\pdfoutput=1
\documentclass[acmtog]{acmart}

\usepackage{url}
\usepackage{siunitx}
\DeclareSIUnit{\px}{px}
\usepackage{physics}
\usepackage{mathtools}
\usepackage{subcaption}
\usepackage[nolist]{acronym}
\usepackage{xspace}
\usepackage{tcolorbox}
\usepackage{enumitem} %
\usepackage{tabularx}
\usepackage{soul}

\newcommand{\couleurorangefonce}[1]{\textcolor[RGB]{233, 155, 116}{#1}}
\newcommand{\couleurorangepale}[1]{\textcolor[RGB]{247, 220, 206}{#1}}
\newcommand{\couleurbleufonce}[1]{\textcolor[RGB]{82, 91, 122}{#1}}
\newcommand{\couleurbleupale}[1]{\textcolor[RGB]{194, 197, 208}{#1}}

\def\SDone{SD 1.0 Inpaint\xspace}
\def\SDtwo{SD 2.0 Inpaint\xspace}
\def\kolors{Kolors\xspace}
\def\SDthree{SD 3.5\xspace}
\def\fluxone{FLUX.1 Fill \scalebox{.7}{[dev]}\xspace}
\def\hidream{HiDream\xspace}
\def\NBone{Nano Banana 1\xspace}
\def\hunyuan{Hunyuan 2.1\xspace}
\def\qwen{Qwen Edit\xspace}
\def\firefly{Firefly 5\xspace}
\def\fluxtwo{FLUX.2 \scalebox{.7}{[dev]}\xspace}
\def\zimageturbo{Z-Image Turbo\xspace}
\def\photoshop{Photoshop\xspace}
\def\fluxtwofour{FLUX.2 \scalebox{.7}{[klein] 4B}\xspace}
\def\fluxtwonine{FLUX.2 \scalebox{.7}{[klein] 9B}\xspace}
\def\NBtwo{Nano Banana 2\xspace}

\AtEndPreamble{
    \usepackage[capitalize]{cleveref}
    \crefname{section}{sec.}{secs.}
    \Crefname{section}{Sec.}{Secs.}
    \crefname{paragraph}{sec.}{secs.}
    \Crefname{paragraph}{Sec.}{Secs.}
    \crefname{table}{tab.}{tabs.}
    \Crefname{table}{Tab.}{Tabs.}
    \crefname{figure}{fig.}{figs.}
    \Crefname{figure}{Fig.}{Figs.}
    \crefname{equation}{eq.}{eqs.}
    \Crefname{equation}{Eq.}{Eqs.}
}

\newtcolorbox{promptbox}{
    colback=gray!5,
    colframe=gray!40,
    boxrule=0.2pt,
    left=5pt, right=5pt, top=2pt, bottom=2pt,
    arc=1mm,
    fontupper=\small,
}

\begin{document}

\title[Shedding Light]{Shedding Light: A Benchmark for Evaluating Lighting Understanding in Generative Image Models}

\author{Justine Giroux}
\authornote{Both authors contributed equally to this research.}
\email{justine.giroux.2@ulaval.ca}
\orcid{0000-0002-2105-8100}

\author{Jack Oliver Hilliard}
\authornotemark[1]
\affiliation{%
  \institution{Université Laval}
  \city{Québec}
  \state{Québec}
  \country{Canada}
}
\email{jack-oliver.hilliard.1@ulaval.ca}
\orcid{0009-0008-0906-1295}

\author{Yannick Hold-Geoffroy}
\affiliation{%
  \institution{Adobe Research}
  \city{San Jose}
  \state{CA}
  \country{USA}}
\email{holdgeof@adobe.com}
\orcid{0000-0002-1060-6941}

\author{Javier Vazquez-Corral}
\affiliation{%
  \institution{Computer Vision Center \&}
  \institution{Universitat Autònoma de Barcelona}
  \city{Barcelona}
  \state{Catalonia}
  \country{Spain}
}
\email{javier.vazquez@cvc.uab.cat}
\orcid{0000-0003-0414-7096}

\author{Jean-Fran\c{c}ois Lalonde}
\affiliation{%
 \institution{Université Laval}
 \city{Québec}
 \state{Québec}
 \country{Canada}
}
\email{jean-francois.lalonde@ulaval.ca}
\orcid{0000-0002-6583-2364}

\renewcommand{\shortauthors}{Giroux et al.}

\begin{abstract}
Accurate modelling of illumination is central to realistic image synthesis and scene understanding. Yet, there is little exploration into whether image generative models are good at this task or whether physical plausibility remains a key challenge for them. Clearly, significant progress has been made in realistic image synthesis, but do models truly understand lighting in a physically accurate manner? To answer this question, this work proposes a benchmark to assess the lighting understanding and harmonisation capabilities of generative models. Our key insight is that evaluating lighting understanding for such models only requires testing how well they insert novel objects into real photographs whilst maintaining consistent illumination. To do so, we use a multi-illumination dataset with images containing simple objects serving as ``light probes'', and prompt models to inpaint the same object onto the original image, then compare the generated results against the ground-truth light probes. We then estimate the lighting direction, colour and radiance distribution from the inpainted probes, providing a quantitative measure of illumination accuracy and photometric realism. Our work establishes a scalable evaluation protocol to systematically assess how well generative models capture and reproduce real-world lighting, offering a foundation for benchmarking the photometric accuracy of any future models. All code and data are available at \url{https://lvsn.github.io/SheddingLight/}.
\end{abstract}

\setcopyright{cc}
\setcctype{by}
\acmJournal{TOG}
\acmYear{2026} \acmVolume{45} \acmNumber{6} \acmArticle{227}
\acmMonth{12} \acmDOI{10.1145/3842579}

\ccsdesc[500]{Computing methodologies~Scene understanding}
\ccsdesc[500]{Computing methodologies~Spatial and physical reasoning}
\ccsdesc[500]{Computing methodologies~Perception}

\keywords{Image Diffusion Models, Lighting, Benchmark, Physical Accuracy}
\begin{teaserfigure}
  \includegraphics[width=\textwidth]{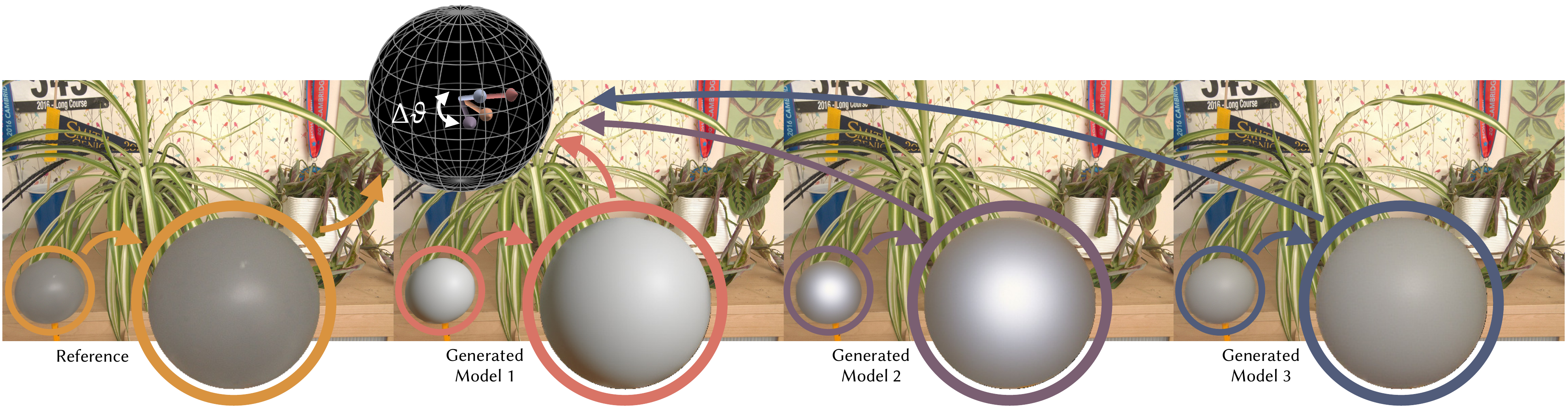}
  \caption{Can generative image models accurately understand the lighting conditions of a scene? To answer this question, we put 16 state-of-the-art models to the test in our novel benchmark, which evaluates their physical lighting accuracy. Our key insight is to rely on inpainting: we prompt models to insert simple objects such as diffuse spheres serving as ``light probes'', which can subsequently be used to robustly estimate illumination properties.
  }
  \Description{Teaser.}
  \label{fig:teaser}
\end{teaserfigure}

\maketitle

\section{Introduction}
\label{sec:intro}

Modern generative image and video models produce visually compelling results with unprecedented quality and realism. Despite this rapid progress, it remains unclear whether these models correctly reason about one of the fundamental components of image formation: illumination. Lighting strongly influences perceived geometry, material appearance, and scene realism, where inconsistencies in illumination often reveal synthetic content even when generated imagery appears otherwise photorealistic.
Recent studies have begun evaluating the physical understanding of generative models in areas such as gravity \cite{thozhiyoor2025objects}, rigid and soft-body dynamics~\cite{wiedemer2025video}, and projective geometry and shadows \cite{Sarkar_2024_CVPR}. In contrast, illumination remains comparatively under-explored despite its central role in visual realism. One reason is that lighting is inherently difficult to evaluate: illumination effects are deeply entangled with scene geometry, material reflectance, exposure, and semantics, making direct photometric comparison challenging. Existing approaches, therefore, remain fragmented, ranging from VLM-based assessments such as PhyBench \cite{meng_phybench_2024} to methods that rely on environment map supervision \cite{giroux2024towards} and manual or domain-specific evaluations \cite{farid_lighting_2022}. As a result, there is currently no standardised framework for quantitatively evaluating illumination consistency in generative models.

In this work, we introduce a physically-grounded benchmark for evaluating illumination consistency in image-editing models by inserting synthetic objects into real scenes. Our insight is to prompt generative models to insert simple objects into images, which can then be used to robustly infer lighting properties. This isolates the model's ability to infer and reproduce scene illumination from other aspects such as geometry, reflectance, and scene semantics.
This enables direct quantitative assessment of photometric consistency in generated imagery.
To this end, our benchmark leverages a multi-illumination dataset with calibrated light probes \citep{murmann_dataset_2019}. Given an image, models are prompted to inpaint a grey diffuse sphere, from which we estimate illumination direction and colour by fitting a physically rendered sphere to the generated result. These estimates are compared against the real-world light probe counterparts to obtain a quantitative measure of photometric consistency. Additionally, this benchmark introduces an optimisation-free measure by directly comparing the ground truth and inpainted spheres based on their radiance distributions.

In short, we propose two main contributions. First, we propose an evaluation protocol based on physically-grounded light estimation to assess lighting accuracy in image synthesis. Second, we leverage this protocol to analyse a comprehensive range of state-of-the-art generative models with varied architectures, revealing their strengths and limitations in reproducing real-world illumination in indoor environments, with a single point light source. Our experiments provide several insights into the behaviour of current generative models. Surprisingly, lighting accuracy has not improved over time, even as models have become more complex (parameter-wise) and trained on more data. In addition, models tend to prefer front-facing lighting directions, and error increases as the azimuth and elevation of the incoming light deviate from this central position. Moreover, most models show understanding of the spatial variation of lighting, which indicates that they are not only doing local harmonisation and have a deeper understanding of the lighting in space. All code and data are available to facilitate reproducibility, support further research, and enable the inclusion of newer models in our evaluation framework.

\section{Related work}
\label{sec:related_work}

\paragraph{Image evaluation} Standard full-reference image quality metrics (e.g. MSE, PSNR~\citep{PSNR}, UQI~\citep{wang_why_2002}, angular error~\citep{barnard_comparison_2002}, VIF~\citep{VIF}, PieAPP~\citep{PIEAPP}, FLIP~\citep{FLIP}, SSIM~\citep{SSIM}, MAD~\citep{larson_most_2010}, LPIPS~\citep{LPIPS}) rely on ground-truth references and therefore cannot be applied to generated images , since this would require generating a new image that is exactly pixel-aligned (identical in viewpoint and content), which current generative imaging models struggle to achieve..
To overcome this limitation, distribution-based metrics such as IS~\citep{IS} and FID~\citep{fid} were introduced.  These metrics compare sets of estimated and reference images, enabling evaluation of generative models even when image content or viewpoint is not perfectly aligned.  However, they are based on InceptionNet~\citep{szegedy_going_2014} pretrained on ImageNet-1k~\citep{deng_imagenet_2009}, which limits their effectiveness in assessing the diversity of the world, a task better captured by newer datasets such as LAION-5B~\citep{schuhmann_laion-5b_2022} and have been heavily critiqued (e.g., \citet{chong2020effectively}).
Finally, no-reference metrics (e.g., BRISQUE~\citep{BRISQUE}, NIQE~\citep{NIQE}, UNIQUE~\citep{zhang2021uncertainty} and HyperIQA~\citep{HyperIQA}) primarily capture perceptual or aesthetic quality rather than physical fidelity, and fail to account for diffusion-specific artefacts.
In contrast, our method directly evaluates the lighting properties of generative models and remains robust to both misalignments and aesthetic variations.

\paragraph{Image editing} Editing images dates back to classical statistical and PDE-based formulations for propagating image structures and textures into missing regions \citep{elharrouss2020image,bertalmio2000image}. The advent of generative modelling transformed this task by introducing learned image priors. Early GAN-based methods \citep{radford2016unsupervised} demonstrated relatively low realism, but this was later refined through architectural improvements such as StyleGAN~\cite{Karras2019stylegan2}, which enabled high-quality semantic completion \citep{richardson2021encoding}. Task-specific designs such as contextual attention \citep{yu2018generative} and gated convolutions \citep{yu2019free} further improved structural consistency and mask handling.

\begin{figure*}[t]
  \centering
  \includegraphics[width=\linewidth]{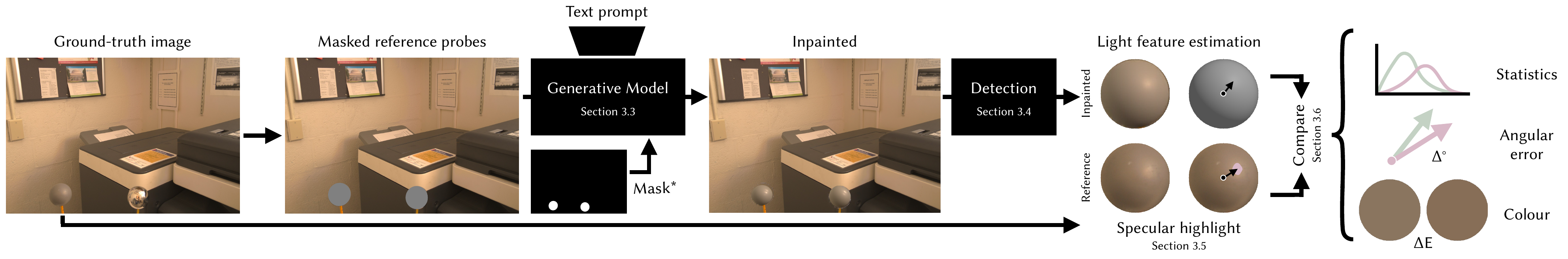}
  \caption{Diagram of the pipeline applied to an image from the Multi-Illumination dataset~\citep{murmann_dataset_2019}. Reference light probes are first masked and the image given to the generative model along with a text prompt. The generated light probe is detected in the image \citep{carion2025sam3segmentconcepts},  cropped (\cref{subsec:methodo_detection}), and the light direction is estimated using inverse rendering (\cref{subsec:methodo_LD}). Finally, we perform various analyses to assess the model's understanding of the scene's lighting.}
  \Description{Pipeline diagram.}
  \label{fig:diagram_pipeline}
\end{figure*}

More recently, diffusion-based models have become the dominant paradigm for image synthesis and editing (see \citep{huang_diffusion_2024} for a survey), demonstrating impressive inpainting performance \citep{lugmayr2022repaint, Avrahami_2022_CVPR, avrahami2023blendedlatent} and attracting interest in fixing artefacts and colour shifts \citep{wang2025towards}. These models also enable richer pipelines for conditional object insertion and scene manipulation, as demonstrated by Paint by Inpaint \citep{wasserman2025paint}. Building on the same principle, \citet{phongthawee_diffusionlight_2024} leveraged the harmonisation properties of diffusion models to inpaint probe regions for environment map estimation. Models like Vision Banana~\citep{visionbanana2026} suggest that generative imaging models have 3D understanding. In this work, we build on these advances to evaluate the physical plausibility of illumination produced during inpainting by state-of-the-art diffusion models.

\paragraph{Evaluation/benchmarking of image editing models}
Most evaluations of recent text-to-image models focus on text-image alignment (e.g., DrawBench~\citep{saharia_photorealistic_2022}, PartiPrompts~\citep{yu_scaling_2022}, CLIPScore~\citep{hessel_clipscore_2022}). Whilst some benchmarks assess multiple aspects of image quality~\citep{cho_davidsonian_2024, lee_holistic_2023, ma2024i2ebench, giakoumoglou_large-scale_2025}, such as colour accuracy (ColorBench~\citep{liang_colorbench_2025}, GenColorBench~\citep{butt_gencolorbench_2025}), few specifically address physical accuracy. PhyBench~\citep{meng_phybench_2024} evaluates physical understanding through prompts related to optical principles, but relies on VLM scoring using GPT-4o~\citep{openai_gpt-4_2024} that only checks for the presence of phenomena rather than their physical correctness. Closer to our work, the metric proposed by \citet{giroux2024towards} targets lighting accuracy but requires ground-truth environment maps, which are typically unavailable for images generated by diffusion-based models.  LumosBench~\citep{liu_unilumos_2025} evaluates relighting methods across six illumination attributes---direction, type, intensity, temperature, dynamics, and optics---but relies on VLM scoring rather than direct measurements, thereby inheriting the ambiguity inherent in natural-language lighting descriptions.

To assess lighting coherence, \citet{farid_lighting_2022} analysed garden spheres generated by DALL$\cdot$E~2~\citep{ramesh_hierarchical_2022} using spherical harmonics~\citep{ramamoorthi_relationship_2001}. He found that the estimated lighting properties were broadly consistent with those in real photographs, but exhibited noticeable deviations in multi-sphere scenes. However, the analysis was restricted to outdoor environments and relied on manual inspection, limiting its scalability and suitability for systematic benchmarking.

\paragraph{Lighting estimation}
Earlier single-image lighting estimation methods relied on specific light cues such as sky, cast shadows, and surface shading to determine the scene's lighting \citep{lalonde2010sun, lalonde2012estimating, barron2015shape}.  Later, deep learning methods were used to replace the handcrafted features with implicitly learnt representations \citep{gardner2017learning, hold2017deep, gardner2019deep, garon2019fast, zhang2019all, legendre2019deeplight, weber2022editable, li2023spatiotemporally, wang2021learning, wang2022stylelight, Dastjerdi_2023_ICCV}.  Nowadays, generative models are used for lighting estimation, with methods such as DiffusionLight~\citep{phongthawee_diffusionlight_2024, Chinchuthakun2025DiffusionLightTurbo} and LuxDiT~\citep{liang_luxdit_2025} leveraging learnt priors to improve their estimations.

These methods are usually evaluated by comparing a sphere rendered with the
predicted lighting against a reference sphere, either photographed in the
real scene \citep{karsch_rendering_2019, legendre2019deeplight} or rendered
with the ground-truth lighting, and computing common Image Quality Assessment (IQA) metrics (e.g., RMSE, PSNR~\citep{PSNR}, LPIPS~\citep{LPIPS}, FID~\citep{fid}, and RGB angular error~\citep{legendre2019deeplight}) to obtain a measure of accuracy. \citet{giroux2024towards} proposed a perceptual metric specifically designed to evaluate the accuracy and plausibility of the lighting predicted by these methods.  We draw inspiration from these established evaluation practices: we ask a generative model to inpaint a diffuse light probe into the images and compare it against a ground-truth reference probe.

Recent work has also used generative models to relight images \citep{zhang_iclight_2025, zeng_dilightnet_2024, jin_neural_2024,
zheng_learning_2026} and videos \citep{he_unirelight_2025,
zhou_lightavideo_2025, gao_flowportal_2025, peng_lightctrl_2026}  directly; whereas these methods make illumination an explicit control signal, we ask whether general-purpose image editing models handle it correctly without being trained for that task specifically.

\section{Methodology}
\label{sec:methodo}

Our objective is to evaluate the physical lighting accuracy of generative inpainting and image-editing models. Our idea relies on inpainting: we ask generative models to render a simple object in an image and compare the generated object to its real counterpart captured in the same scene under the same lighting conditions. This design provides a controlled proxy for illumination evaluation, as it fixes scene geometry and reflectance whilst isolating the model’s ability to infer and reproduce scene lighting.

\subsection{Overview}

We propose the pipeline illustrated in \cref{fig:diagram_pipeline}. First, we require a reference dataset of scenes captured under various illumination conditions (\cref{subsec:methodo_dataset}). We assume images contain a simple object that can be used to robustly estimate lighting properties of the image (a ``light probe''~\citep{debevec_rendering_1998}). Here, we propose leveraging diffuse grey spheres, since all incoming light directions can be recovered by the Lambertian terminator and shading gradient, which are observable within the hemisphere of visible surface normals.

We provide these images (\cref{subsec:methodo_dataset}) to a generative image model and prompt it to insert a diffuse light probe within the image (\cref{subsec:methodo_inpainting}). Next, we detect and extract the generated light probe (\cref{subsec:methodo_detection}). Finally, lighting parameters are acquired from the real and rendered probes using an inverse rendering approach (\cref{subsec:methodo_LD}). We then systematically analyse the model's illumination understanding (\cref{sec:evaluation}) using several measures (\cref{subsec:methodo_measures}).

\subsection{Dataset}
\label{subsec:methodo_dataset}

We leverage the Multi-Illumination dataset~\cite{murmann_dataset_2019}, a collection of \num{1015} indoor scenes, each captured in HDR under \num{25} lighting directions originating from a fixed flash, whose orientation is controlled by a servo motor. This dataset provides a pair of real diffuse and chrome light probes in each image, which serve as measurements of the scene's lighting. Here, only the diffuse light probe is used.
We filter the dataset to remove images where the probes are overexposed, too noisy, or contain multiple dominant specular highlights, resulting in a final ground-truth dataset of \num{23023} images across \num{1006} scenes and \num{24} lighting directions. Since the dataset does not include ground-truth light directions for each light probe, we inferred light direction vectors by detecting the centre of mass of the specular highlight (the real probe is slightly glossy) on the HDR image, and assigned it as the halfway vector between the view and light directions.

\subsection{Inpainting probes}
\label{subsec:methodo_inpainting}

We prompt the generative model to inpaint two grey matte spheres:
\begin{promptbox}
    ``Inpaint two perfectly round light grey matte spheres, fully visible, with smooth, uniform Lambertian shading that accurately responds to the existing scene lighting. The spheres are directly in front of the camera and appear in front of everything. There should be two spheres over the two circular grey masks in the image."
\end{promptbox}
In the image provided to the model, we replace the light probes with two uniform grey circles, preventing the model from ``cheating'' by simply reproducing the ground-truth probes.
For inpainting models that accept a mask alongside the input image, we also provide the locations of the grey circles as binary masks.
Examples of the spheres generated into scenes by various models are shown in \cref{fig:exemples_results} and \cref{fig:qualitative_grid} in the supp. mat.

When specifying the sphere albedo in the prompt, we found that the models did not show significant variations in the generated probe intensity and therefore struggle to overcome the light/albedo ambiguity.

\subsection{Probe detection}
\label{subsec:methodo_detection}

\begin{figure}[t]
\centering
\tiny
\setlength\tabcolsep{0.5pt}
\renewcommand{\arraystretch}{0.1}
\def\colW{0.19}
\def\colw{0.1}
\newlength{\textgap}
\setlength{\textgap}{0.05cm}
\newlength{\imagegap}
\setlength{\imagegap}{0.06cm}
\newlength{\rowgap}
\setlength{\rowgap}{0.2cm}
\begin{tabularx}{\linewidth}{ccccc}

    Reference & \SDone & \SDtwo & \kolors & \SDthree\\[\textgap]
    \raisebox{0pt}{\includegraphics[width=\colW\linewidth]{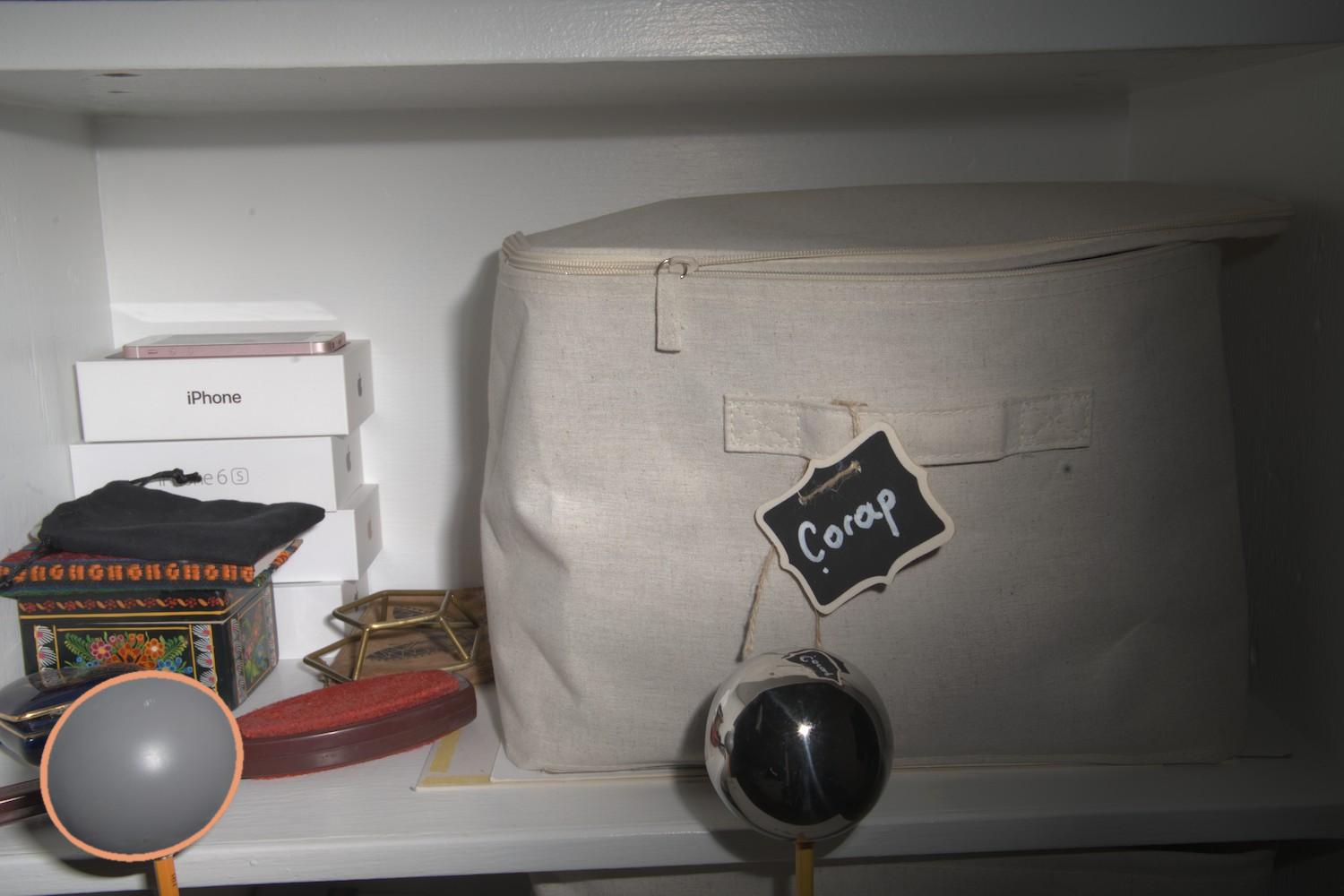}}
    &\raisebox{0pt}{\includegraphics[width=\colW\linewidth]{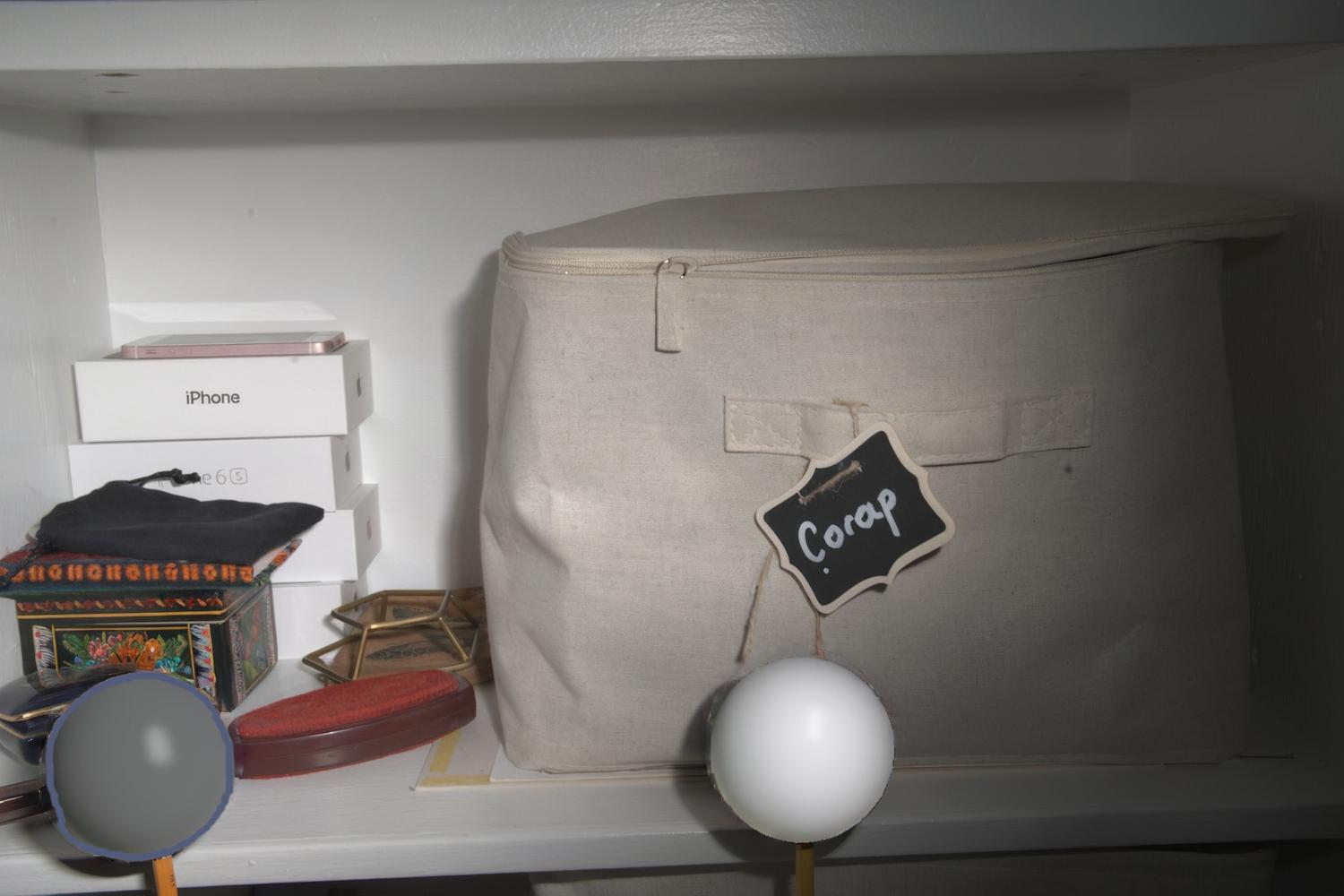}} &\raisebox{0pt}{\includegraphics[width=\colW\linewidth]{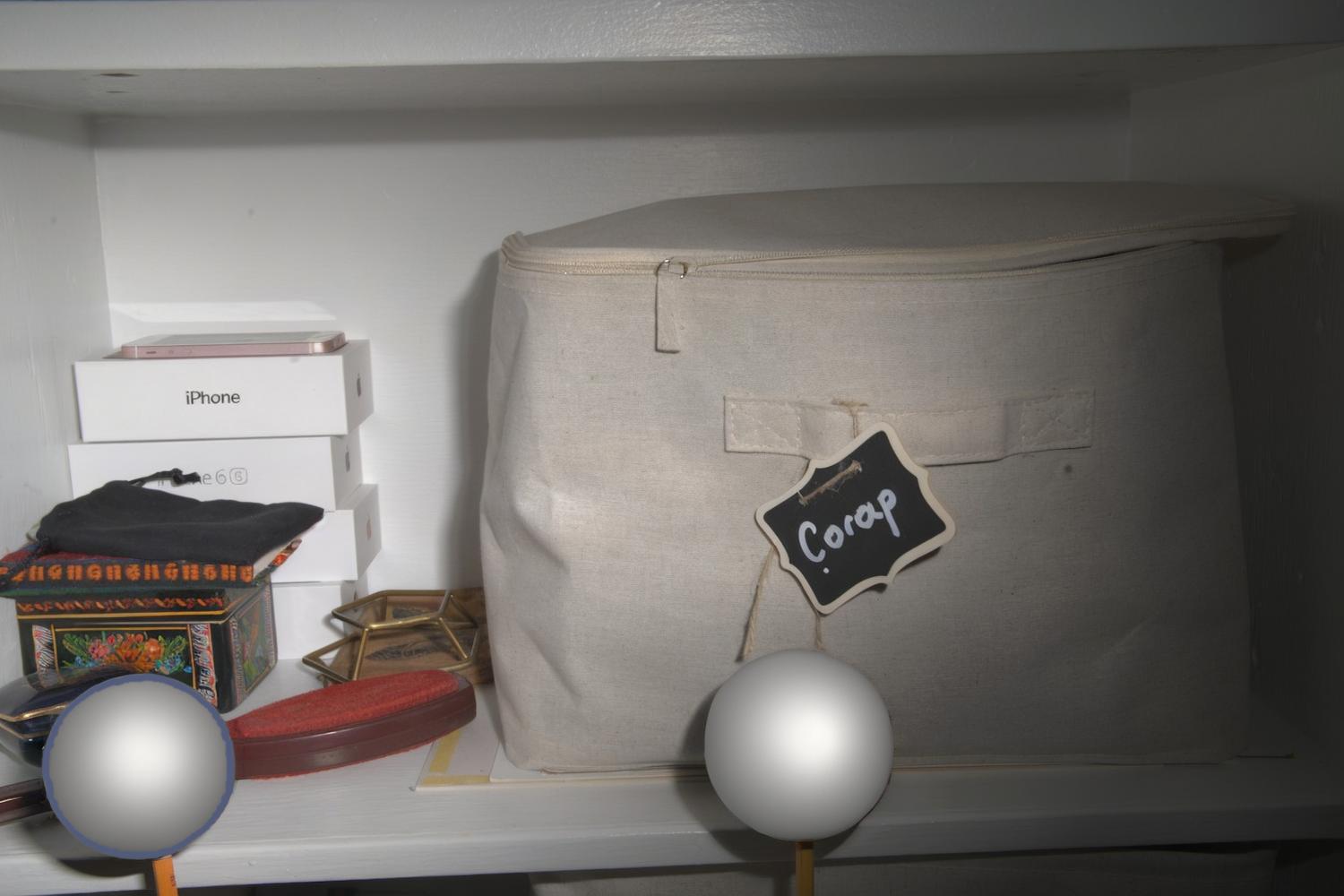}} &\raisebox{0pt}{\includegraphics[width=\colW\linewidth]{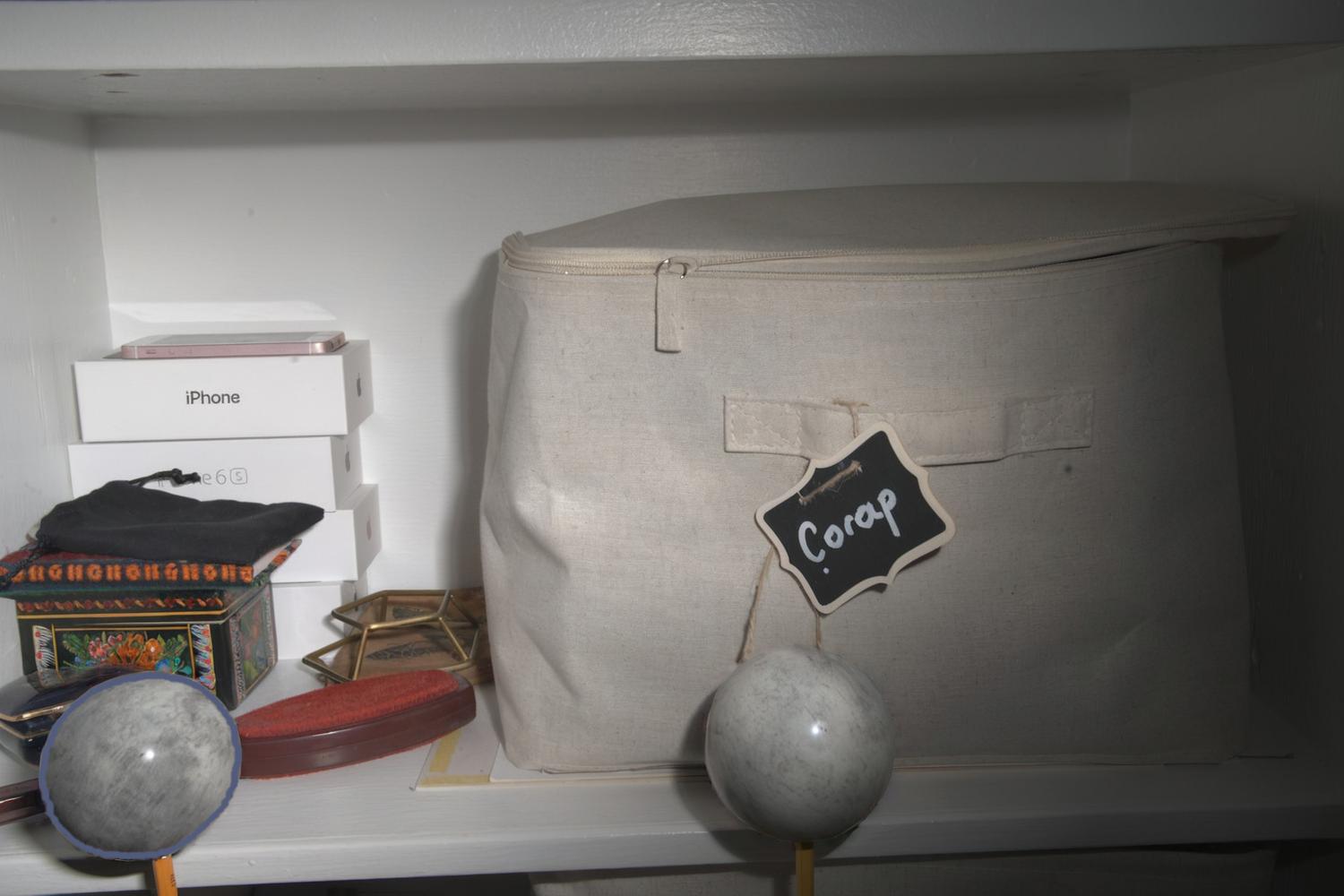}} &\raisebox{0pt}{\includegraphics[width=\colW\linewidth]{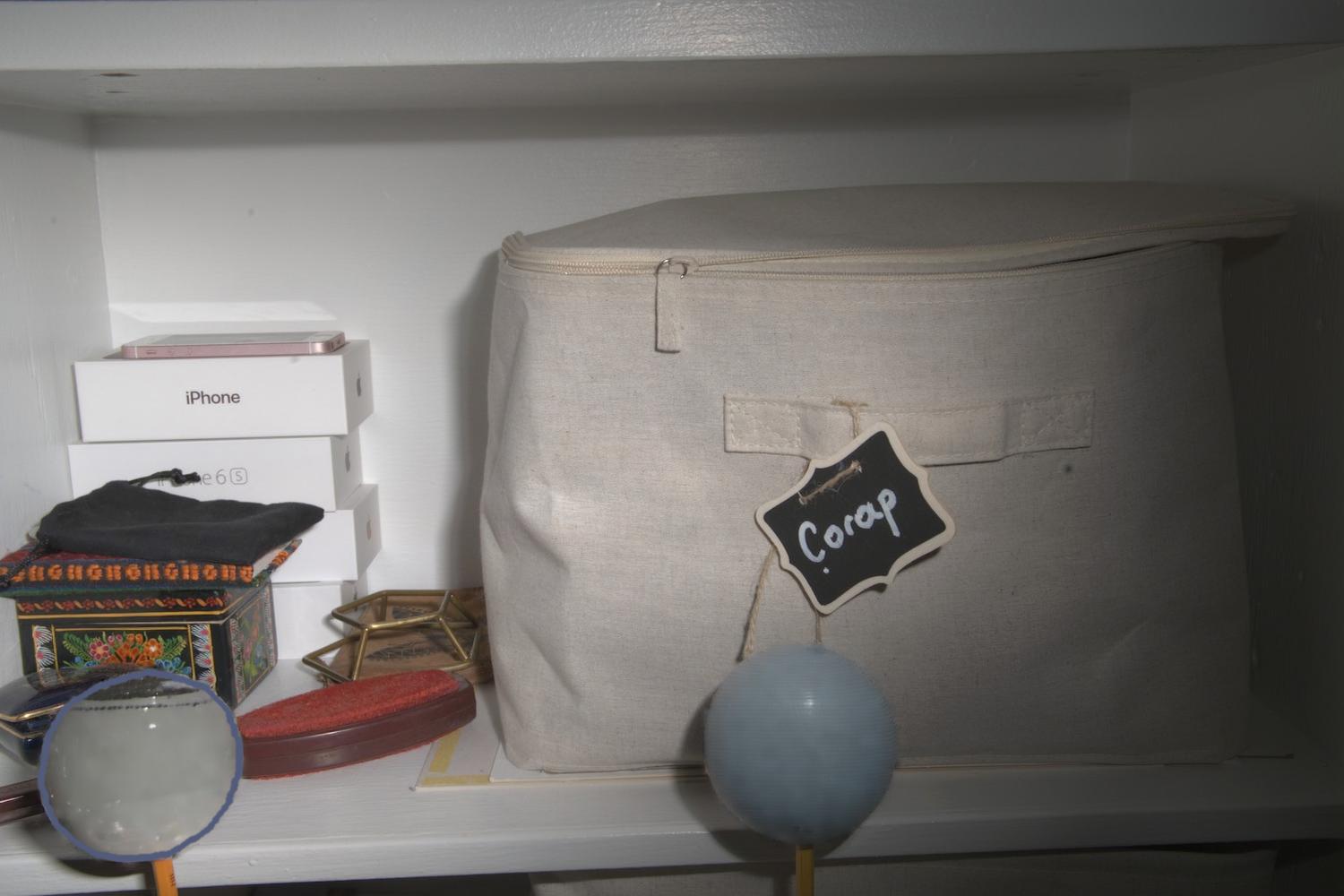}}\\[\imagegap]

    \raisebox{0pt}{\centering\includegraphics[width=\colw\linewidth]{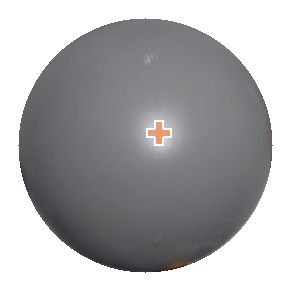}} &\raisebox{0pt}{\centering\includegraphics[width=\colw\linewidth]{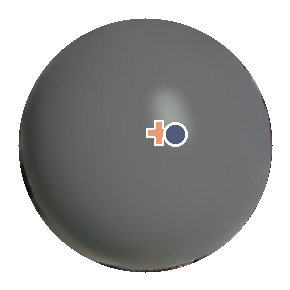}} &\raisebox{0pt}{\centering\includegraphics[width=\colw\linewidth]{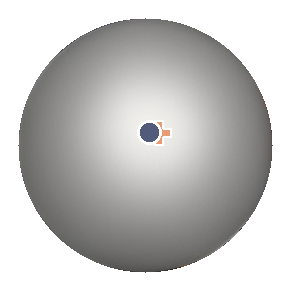}} &\raisebox{0pt}{\centering\includegraphics[width=\colw\linewidth]{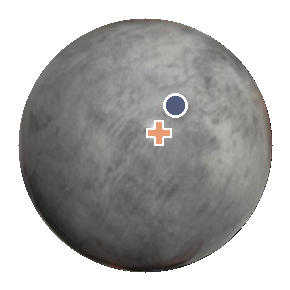}} &\raisebox{0pt}{\centering\includegraphics[width=\colw\linewidth]{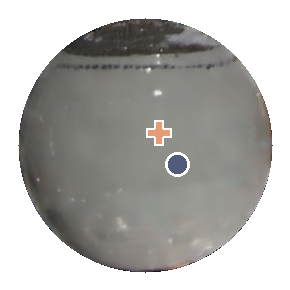}}\\[\rowgap]

    Reference & \fluxone & \hidream & \NBone & \hunyuan\\[\textgap]\
    \raisebox{0pt}{\includegraphics[width=\colW\linewidth]{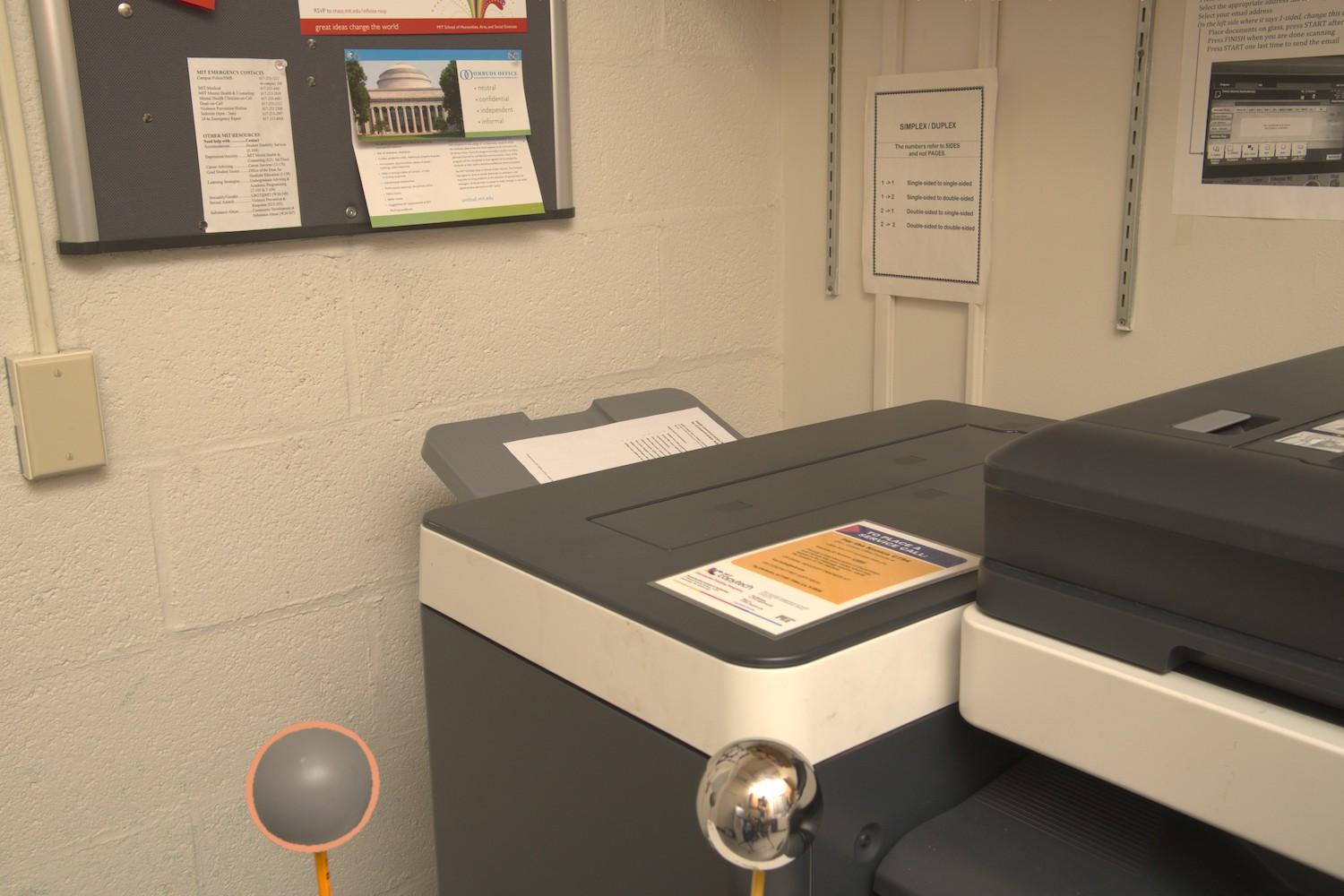}}
    &\raisebox{0pt}{\includegraphics[width=\colW\linewidth]{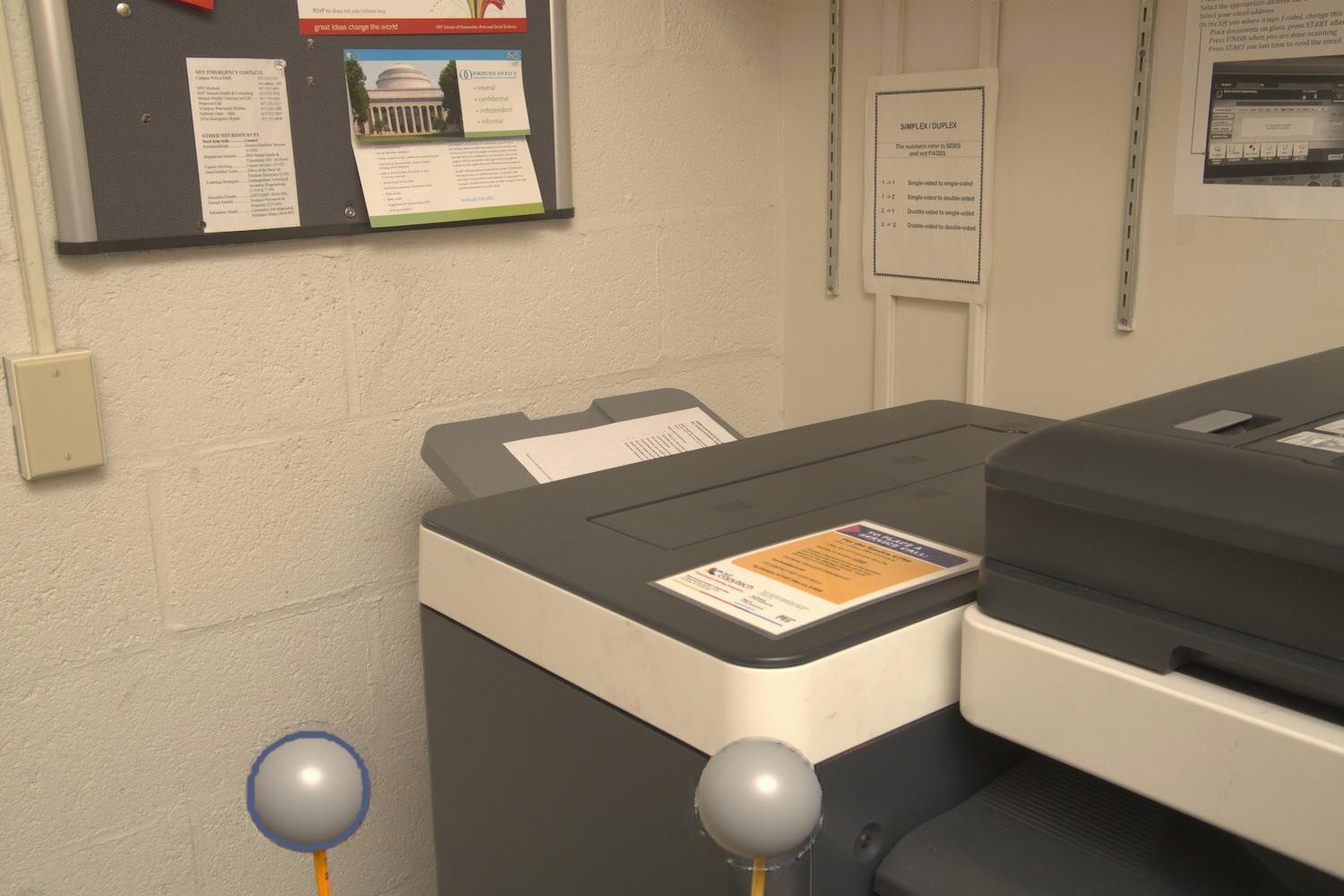}} &\raisebox{0pt}{\includegraphics[width=\colW\linewidth]{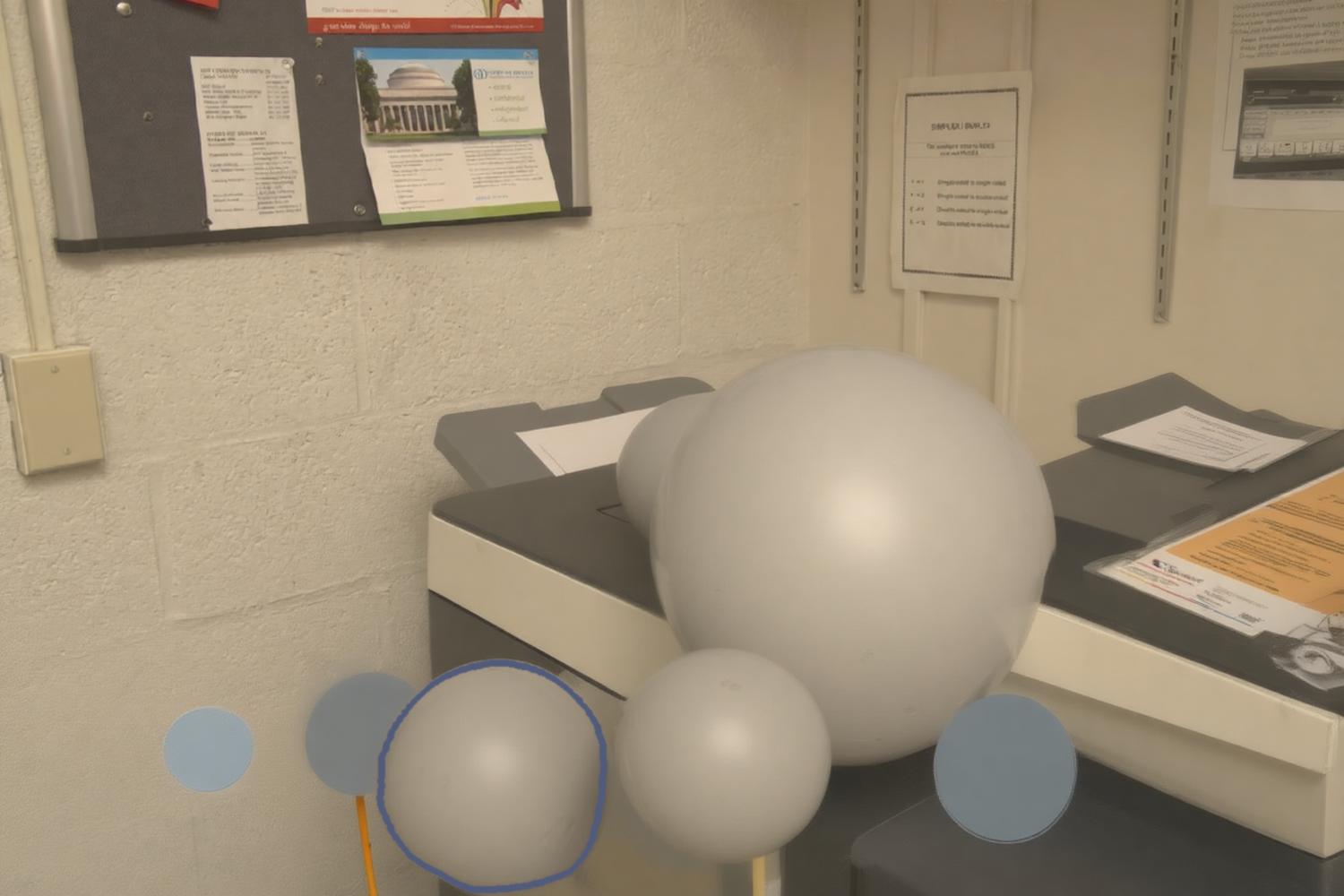}} &\raisebox{0pt}{\includegraphics[width=\colW\linewidth]{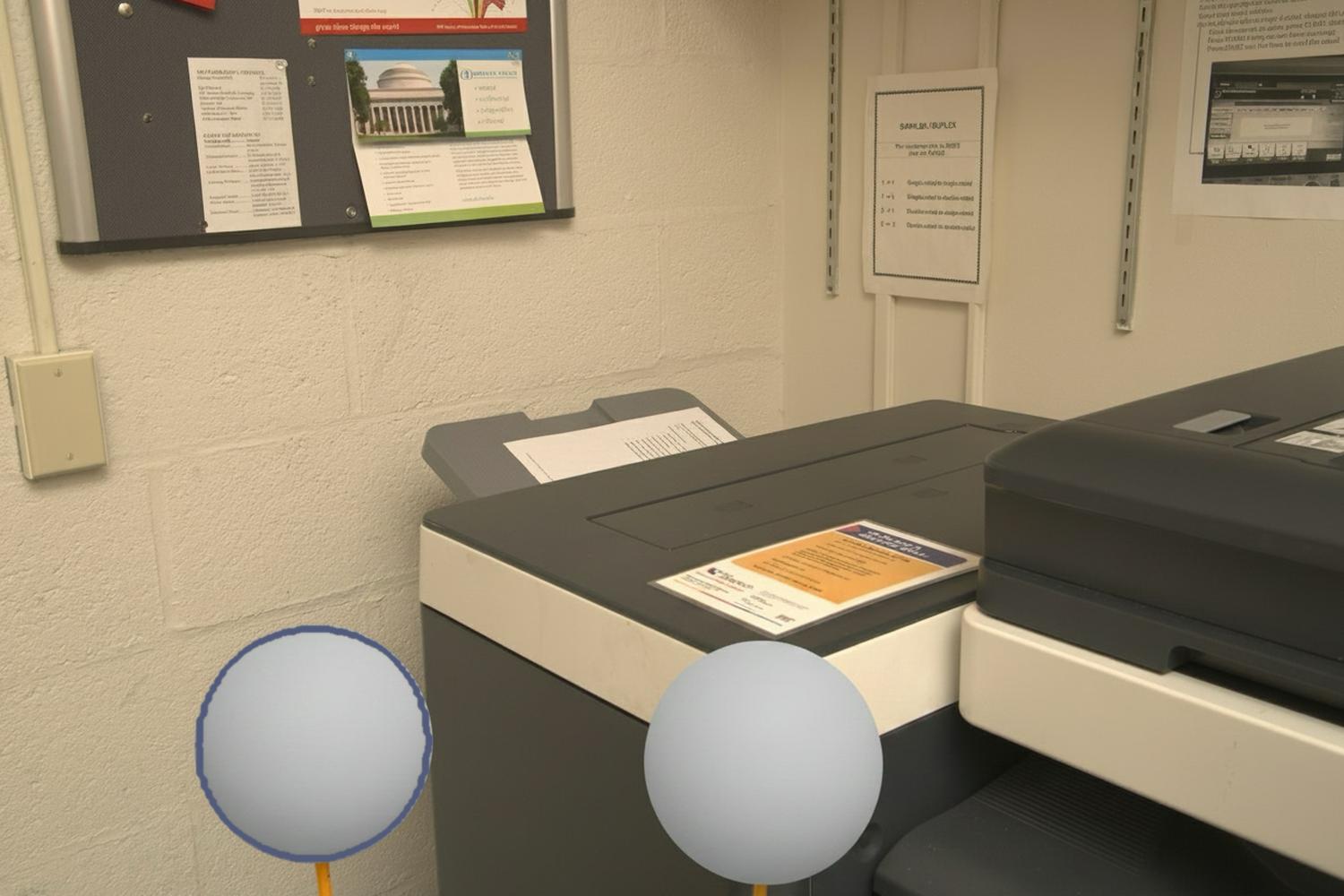}} &\raisebox{0pt}{\includegraphics[width=\colW\linewidth]{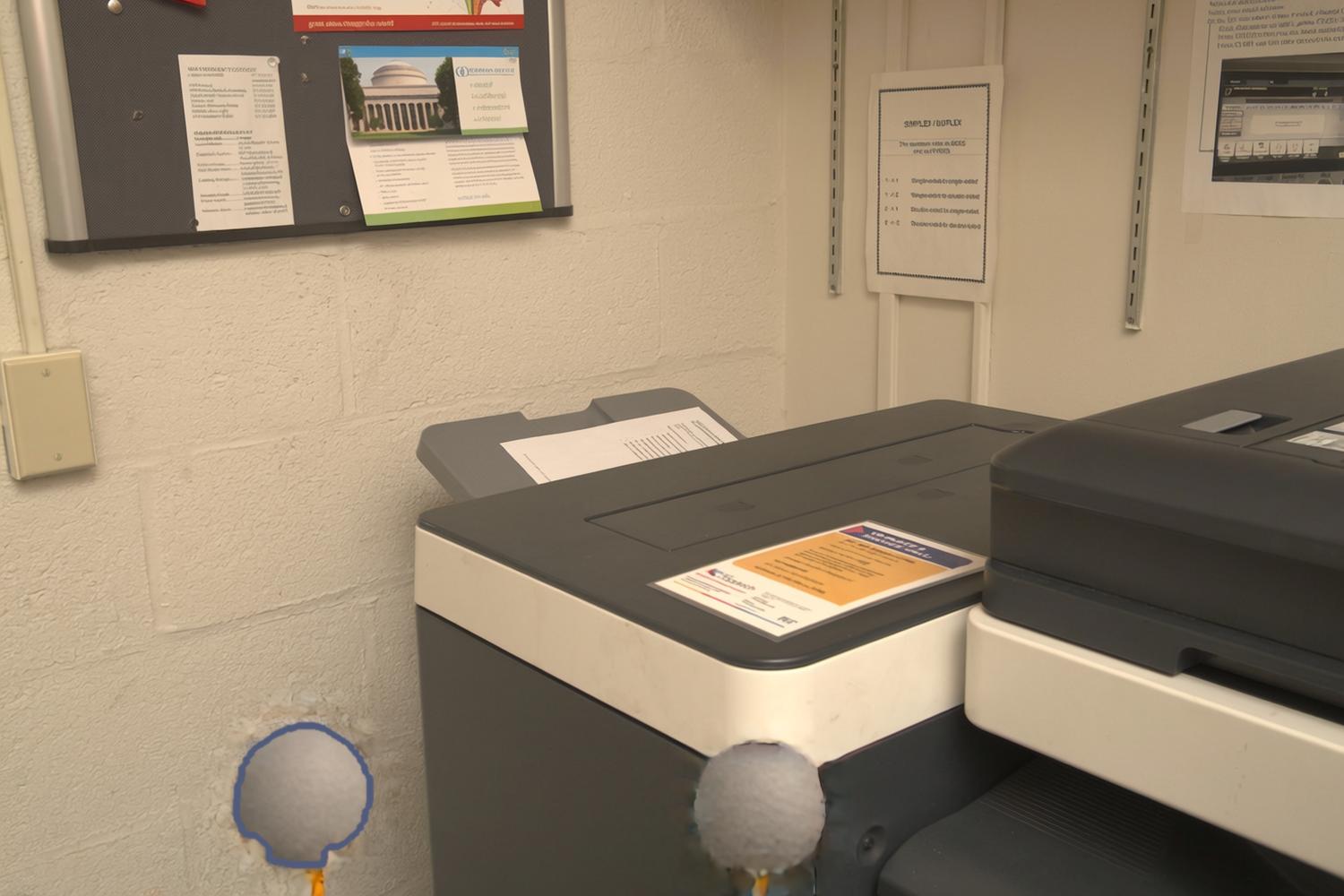}}\\[\imagegap]

    \raisebox{0pt}{\centering\includegraphics[width=\colw\linewidth]{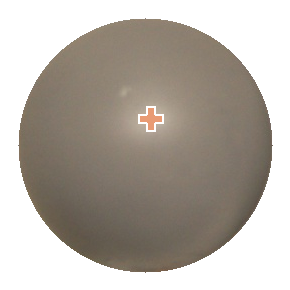}}&\raisebox{0pt}{\centering\includegraphics[width=\colw\linewidth]{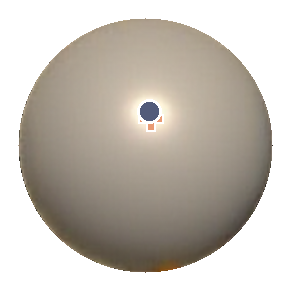}} &\raisebox{0pt}{\centering\includegraphics[width=\colw\linewidth]{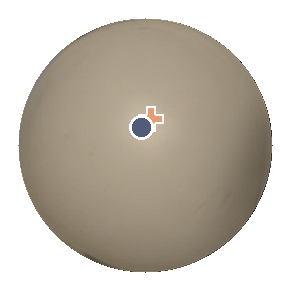}} &\raisebox{0pt}{\centering\includegraphics[width=\colw\linewidth]{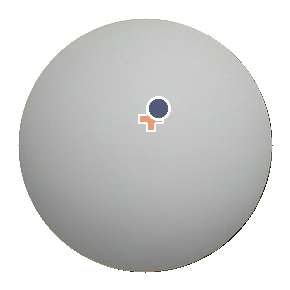}} &\raisebox{0pt}{\centering\includegraphics[width=\colw\linewidth]{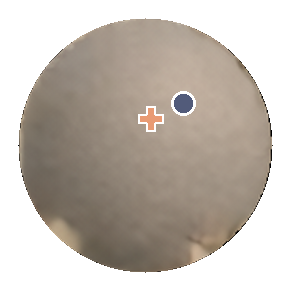}}\\[\rowgap]

    Reference & \qwen & \firefly & \fluxtwo & \zimageturbo\\[\textgap]
    \raisebox{0pt}{\includegraphics[width=\colW\linewidth]{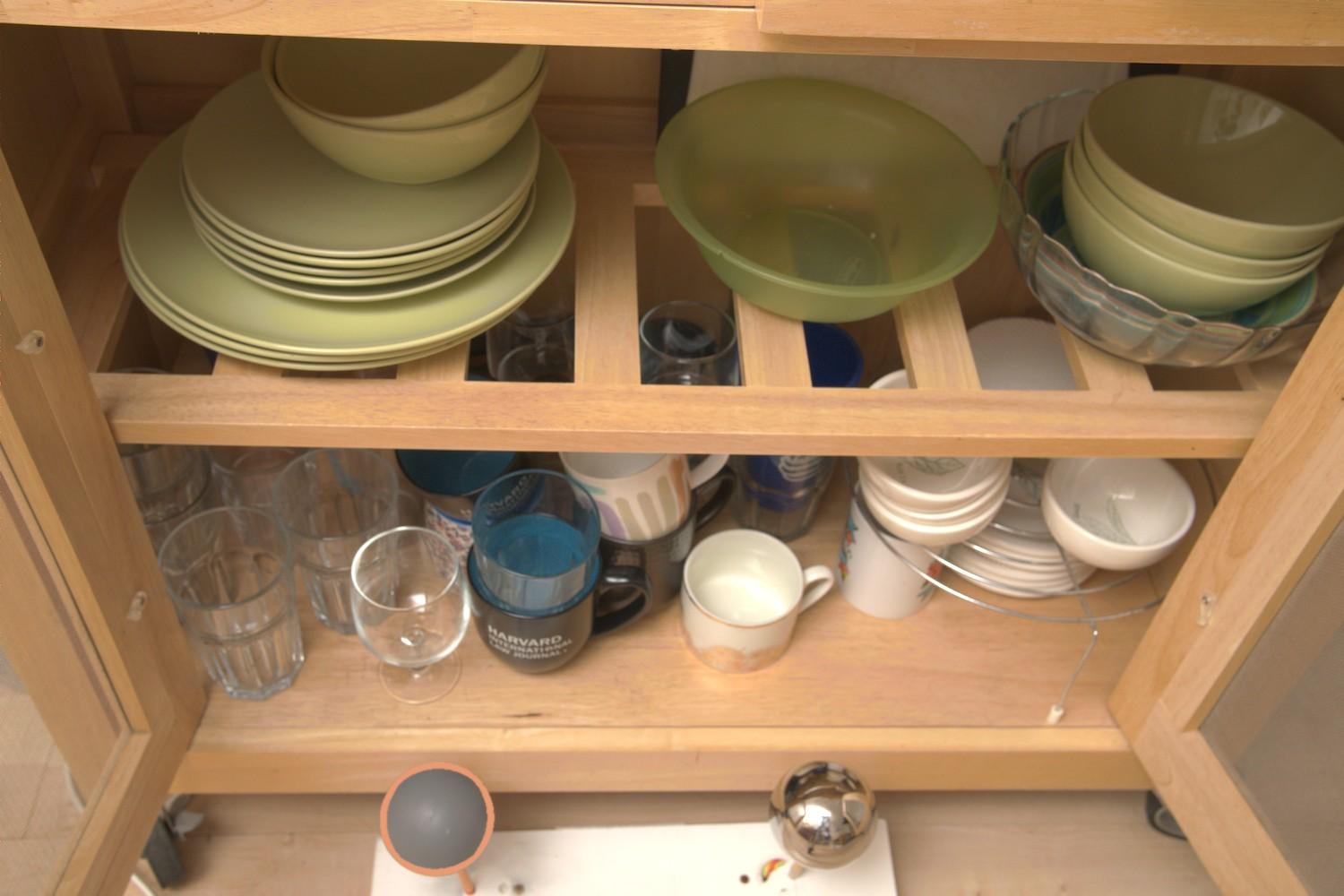}}
    &\raisebox{0pt}{\includegraphics[width=\colW\linewidth]{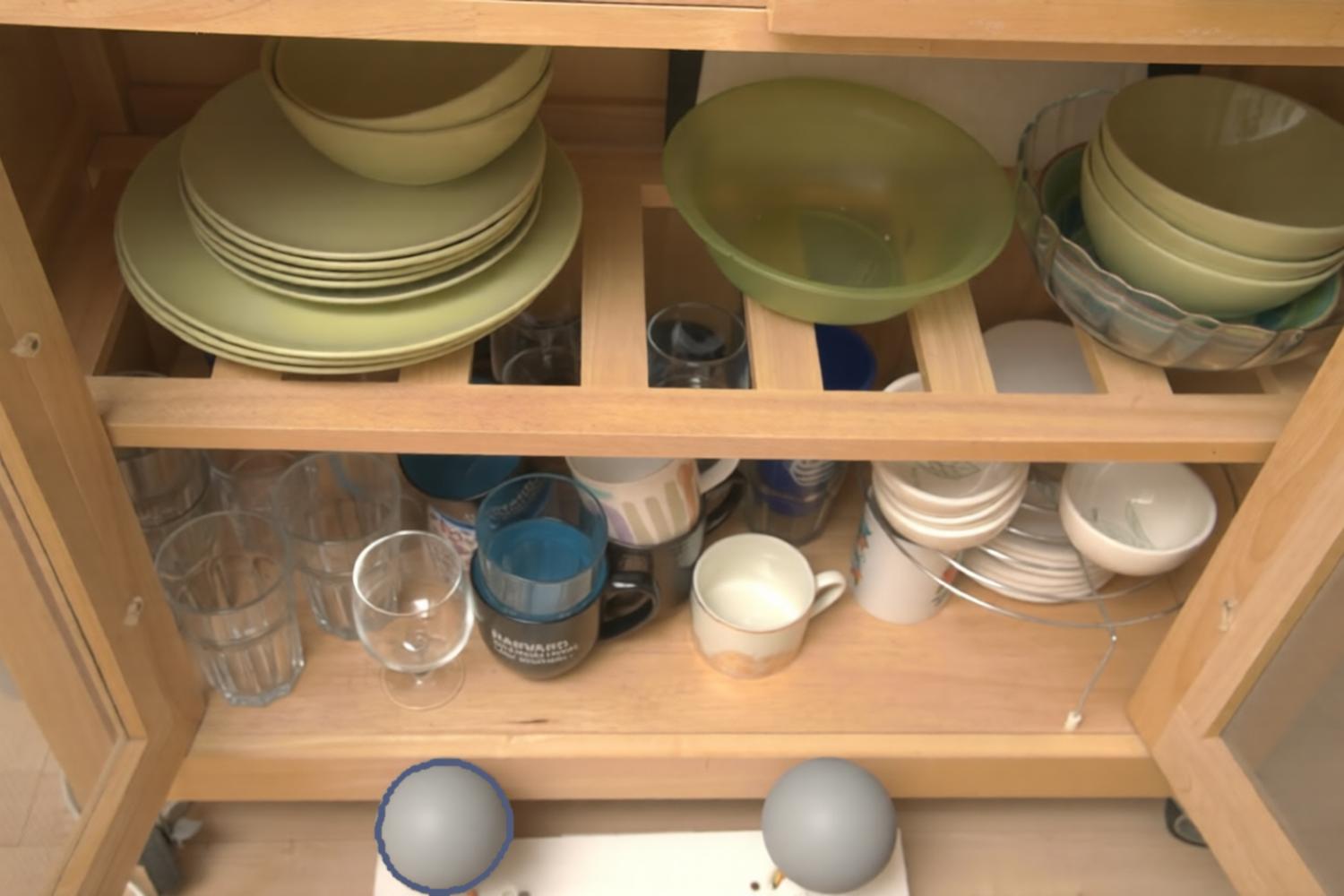}} &\raisebox{0pt}{\includegraphics[width=\colW\linewidth]{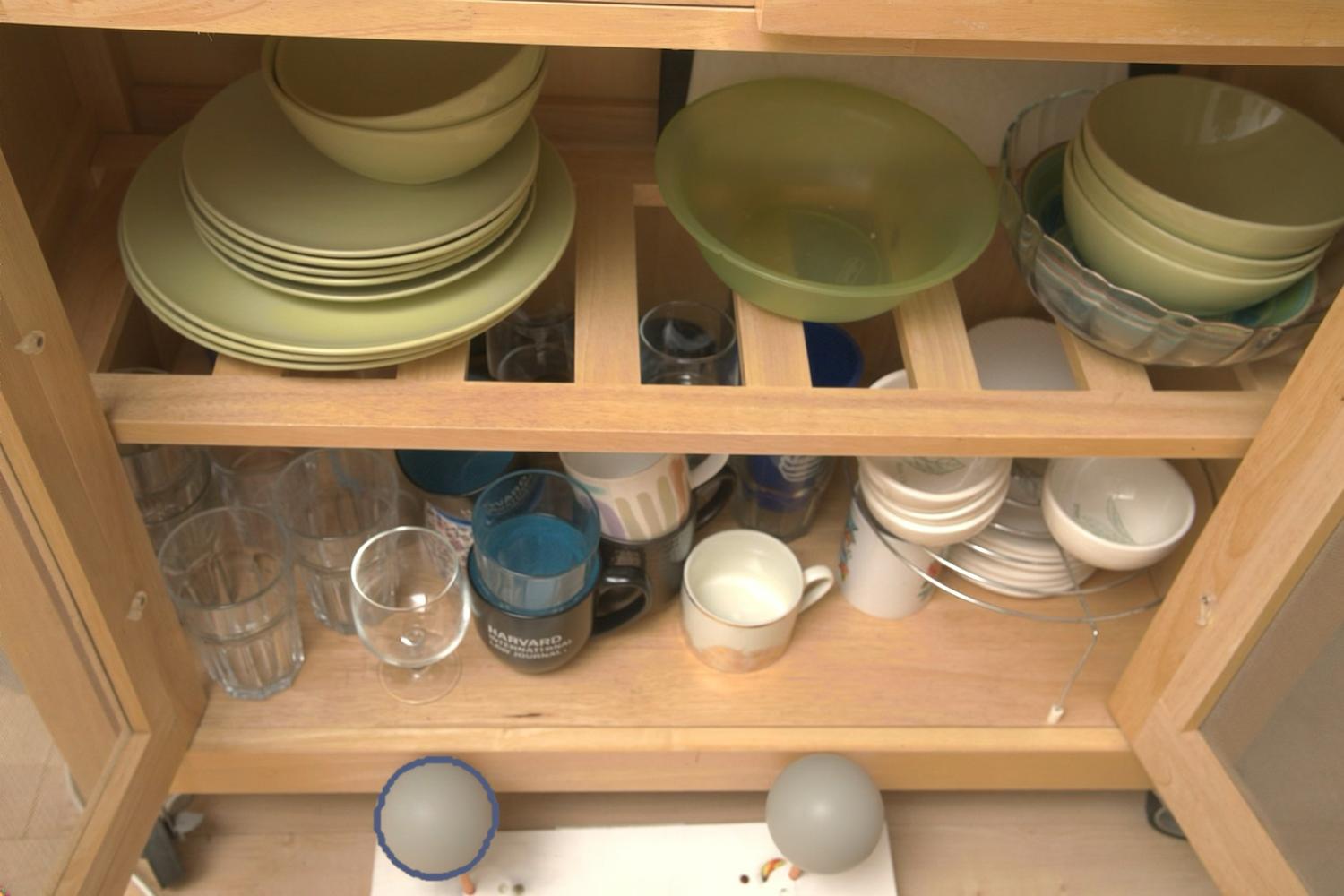}} &\raisebox{0pt}{\includegraphics[width=\colW\linewidth]{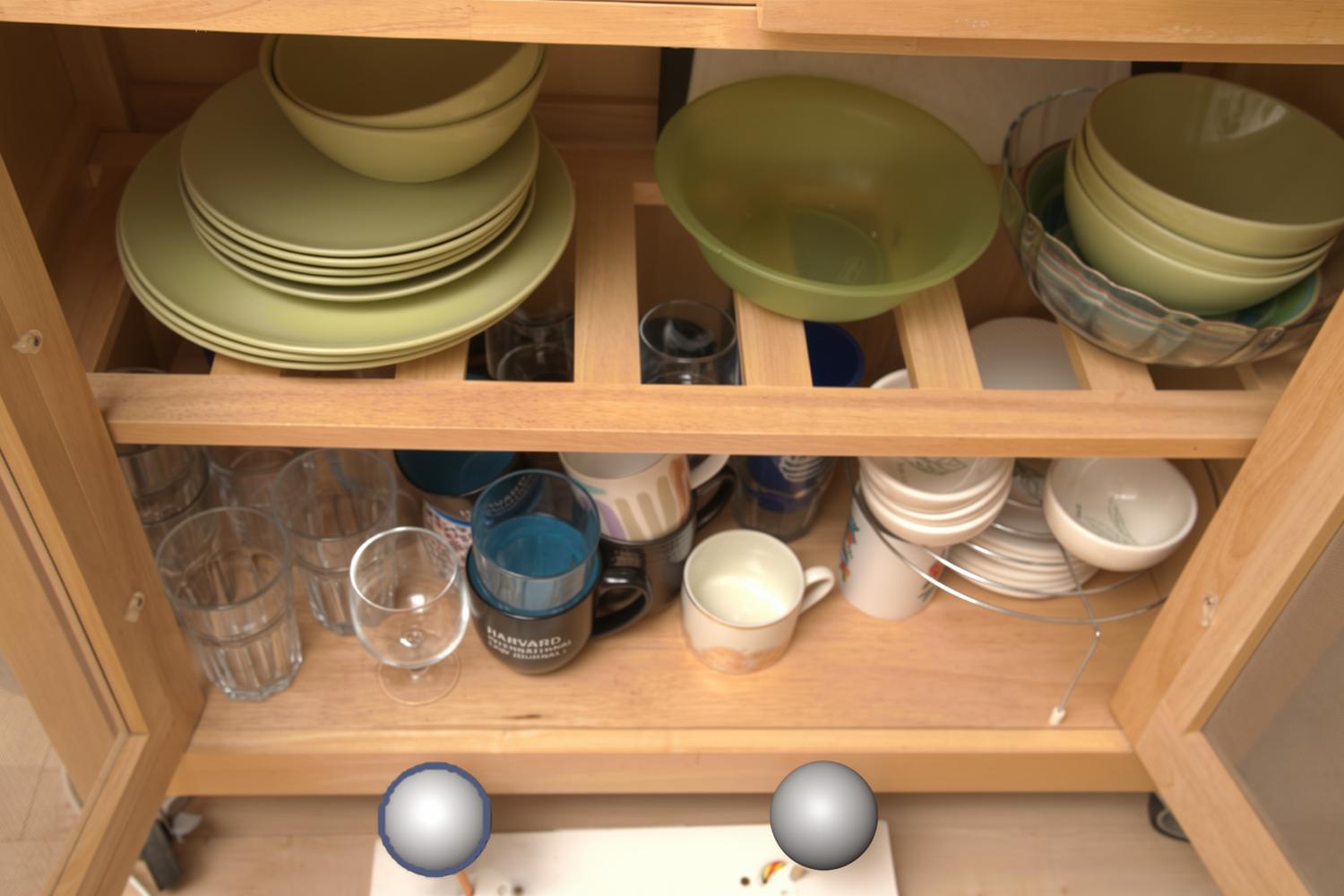}} &\raisebox{0pt}{\includegraphics[width=\colW\linewidth]{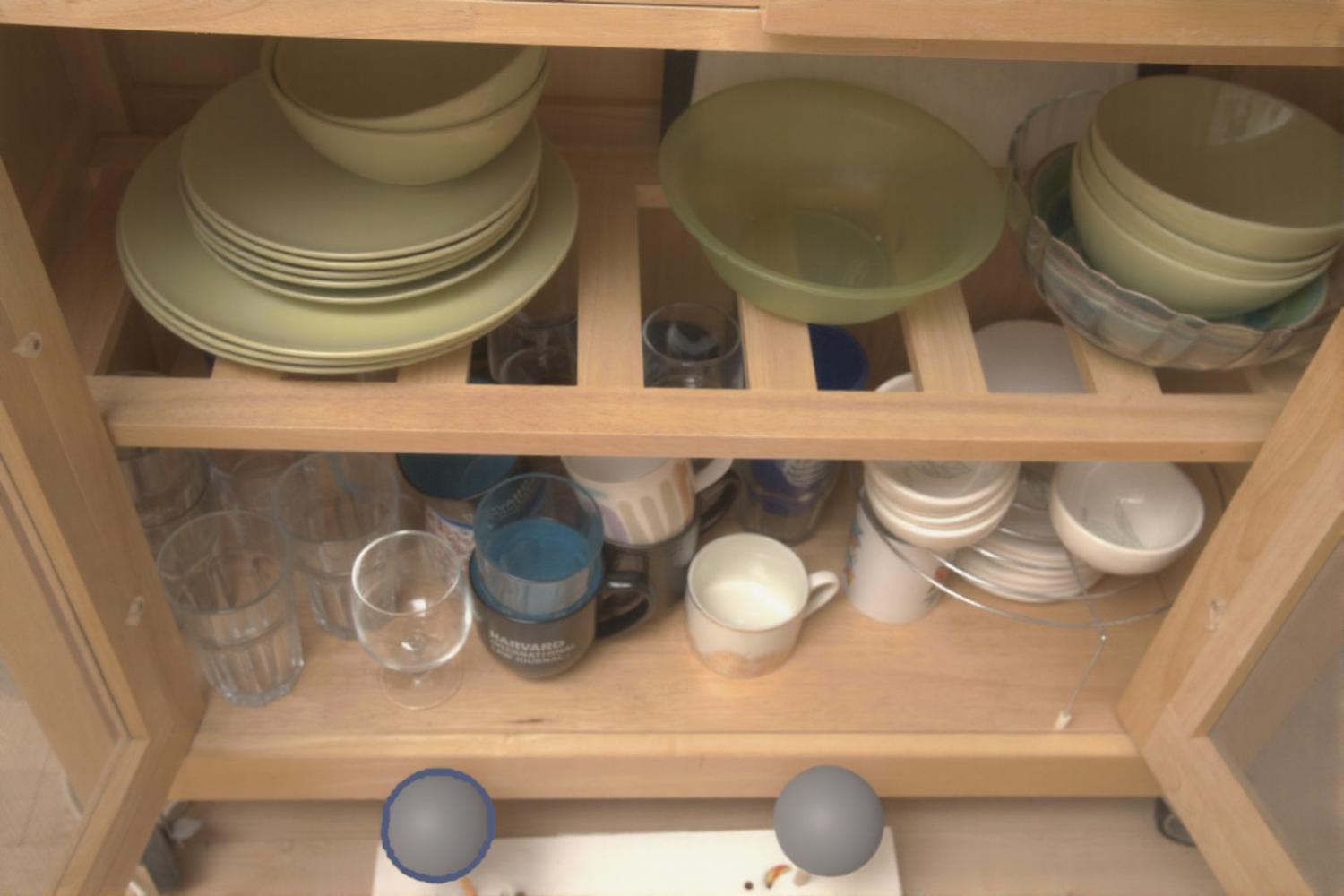}}\\[\imagegap]

    \raisebox{0pt}{\centering\includegraphics[width=\colw\linewidth]{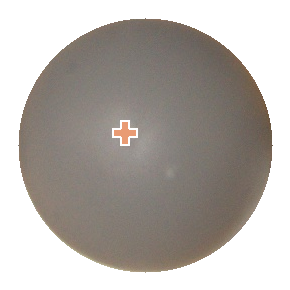}}&\raisebox{0pt}{\centering\includegraphics[width=\colw\linewidth]{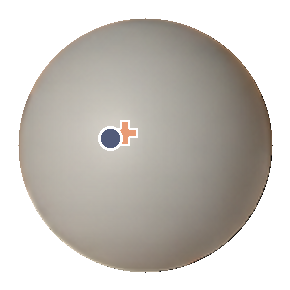}} &\raisebox{0pt}{\centering\includegraphics[width=\colw\linewidth]{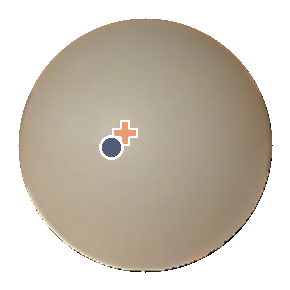}} &\raisebox{0pt}{\centering\includegraphics[width=\colw\linewidth]{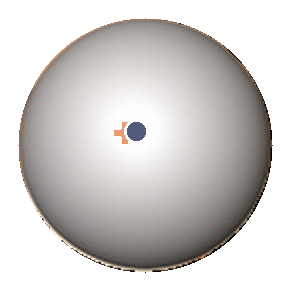}} &\raisebox{0pt}{\centering\includegraphics[width=\colw\linewidth]{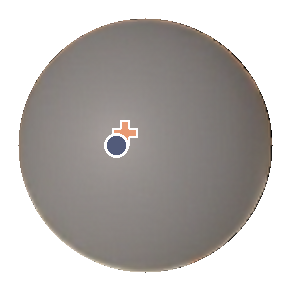}}\\[\rowgap]

    Reference & \photoshop & \fluxtwofour & \fluxtwonine & \NBtwo\\[\textgap]
    \raisebox{0pt}{\includegraphics[width=\colW\linewidth]{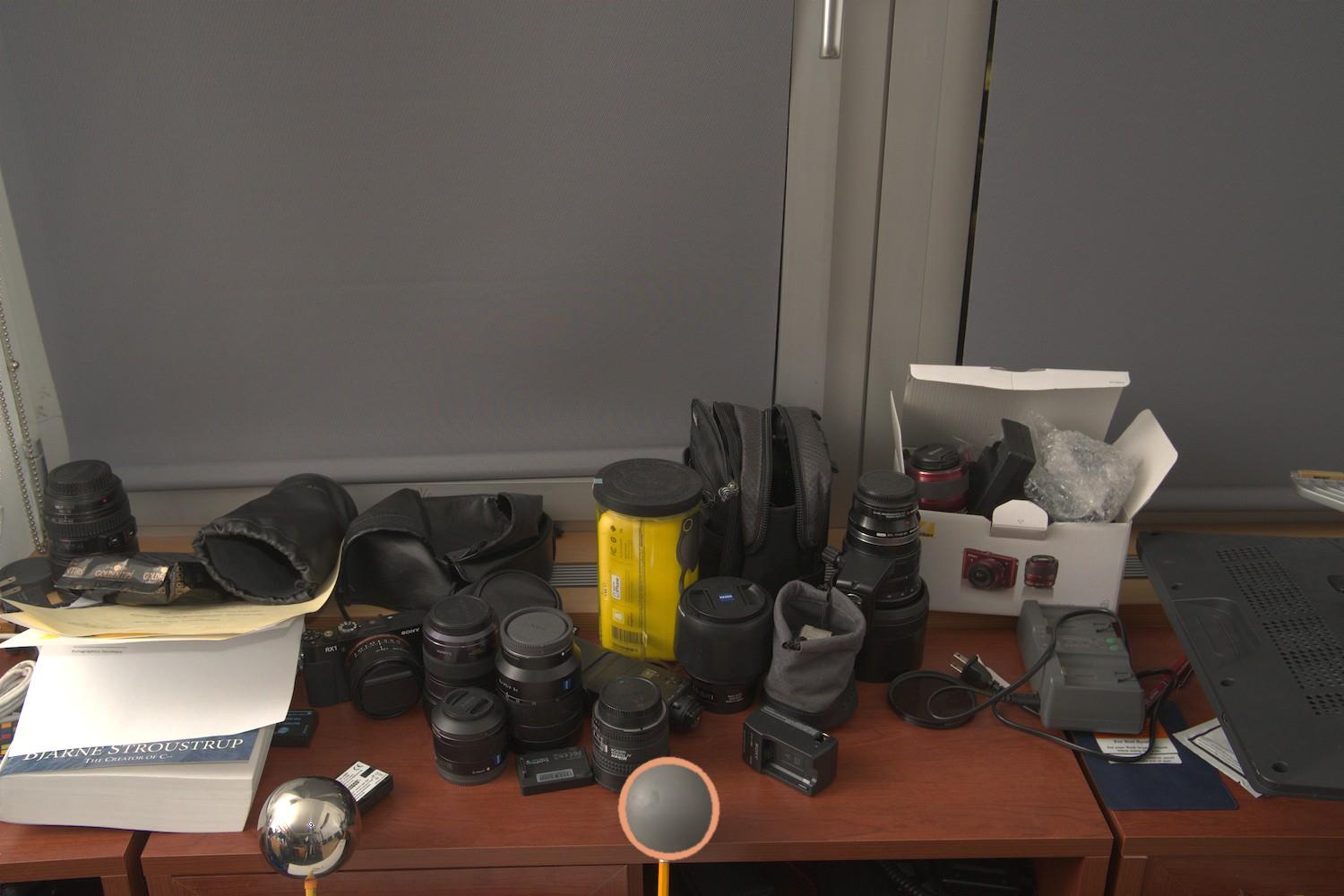}}
    &\raisebox{0pt}{\includegraphics[width=\colW\linewidth]{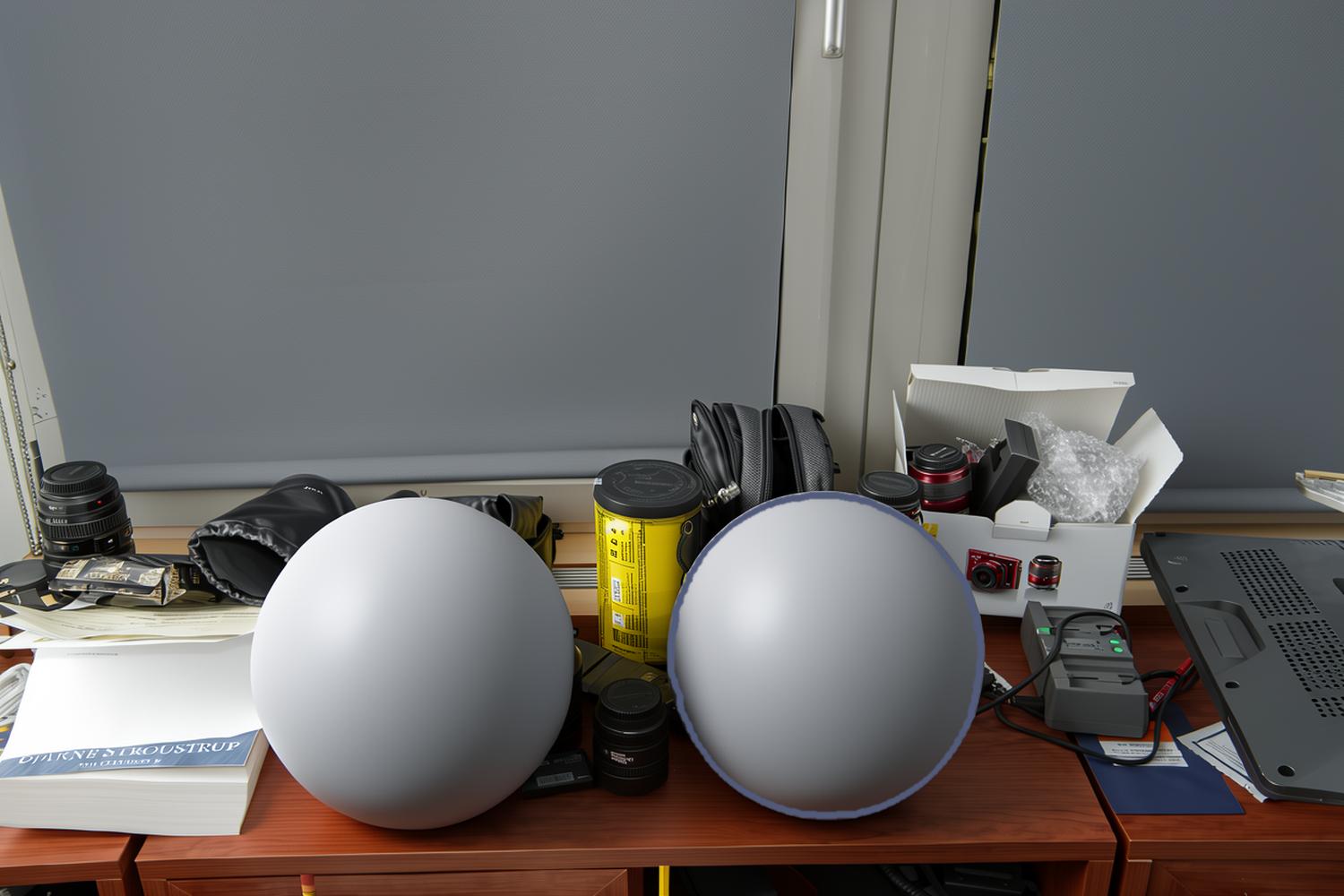}} &\raisebox{0pt}{\includegraphics[width=\colW\linewidth]{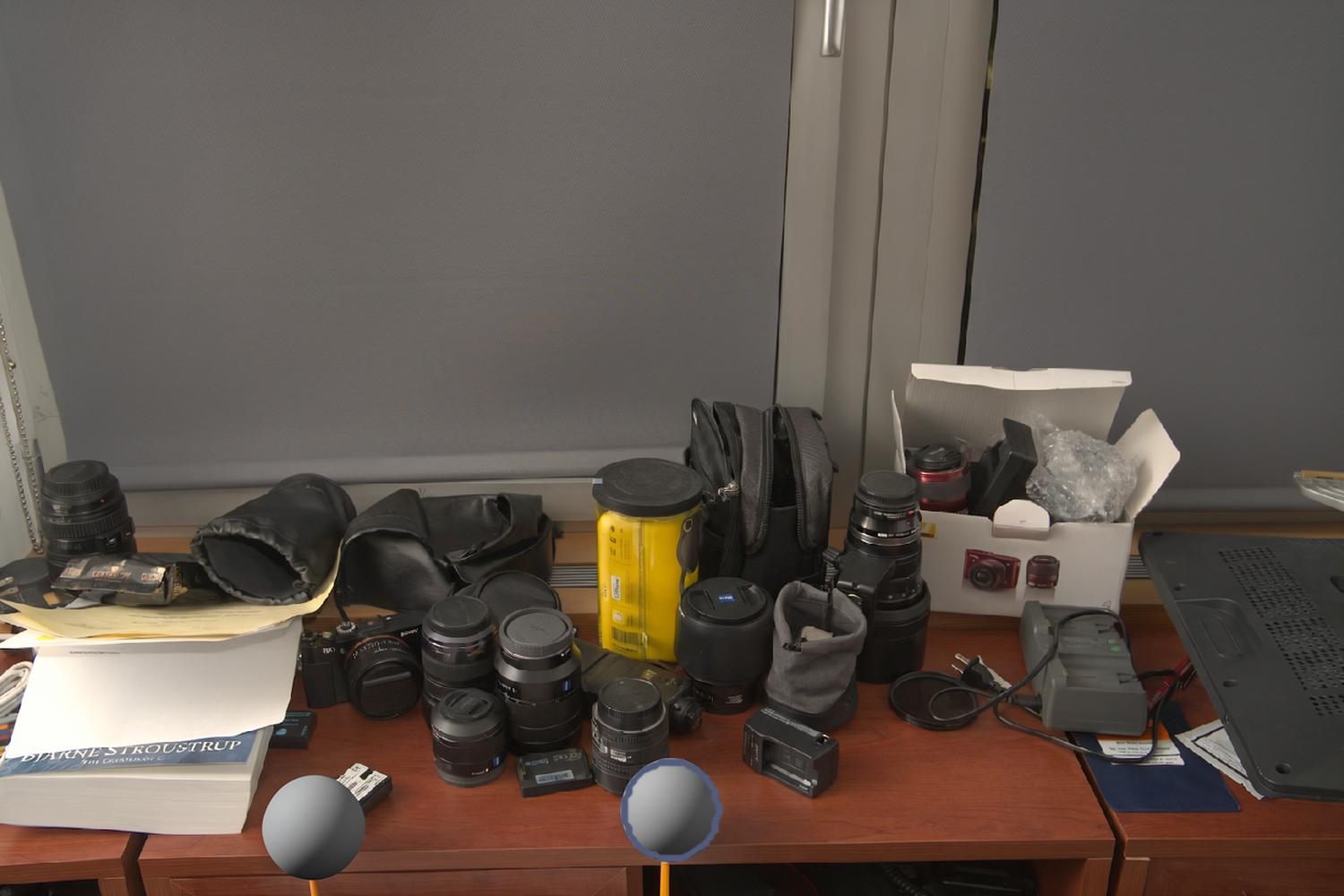}} &\raisebox{0pt}{\includegraphics[width=\colW\linewidth]{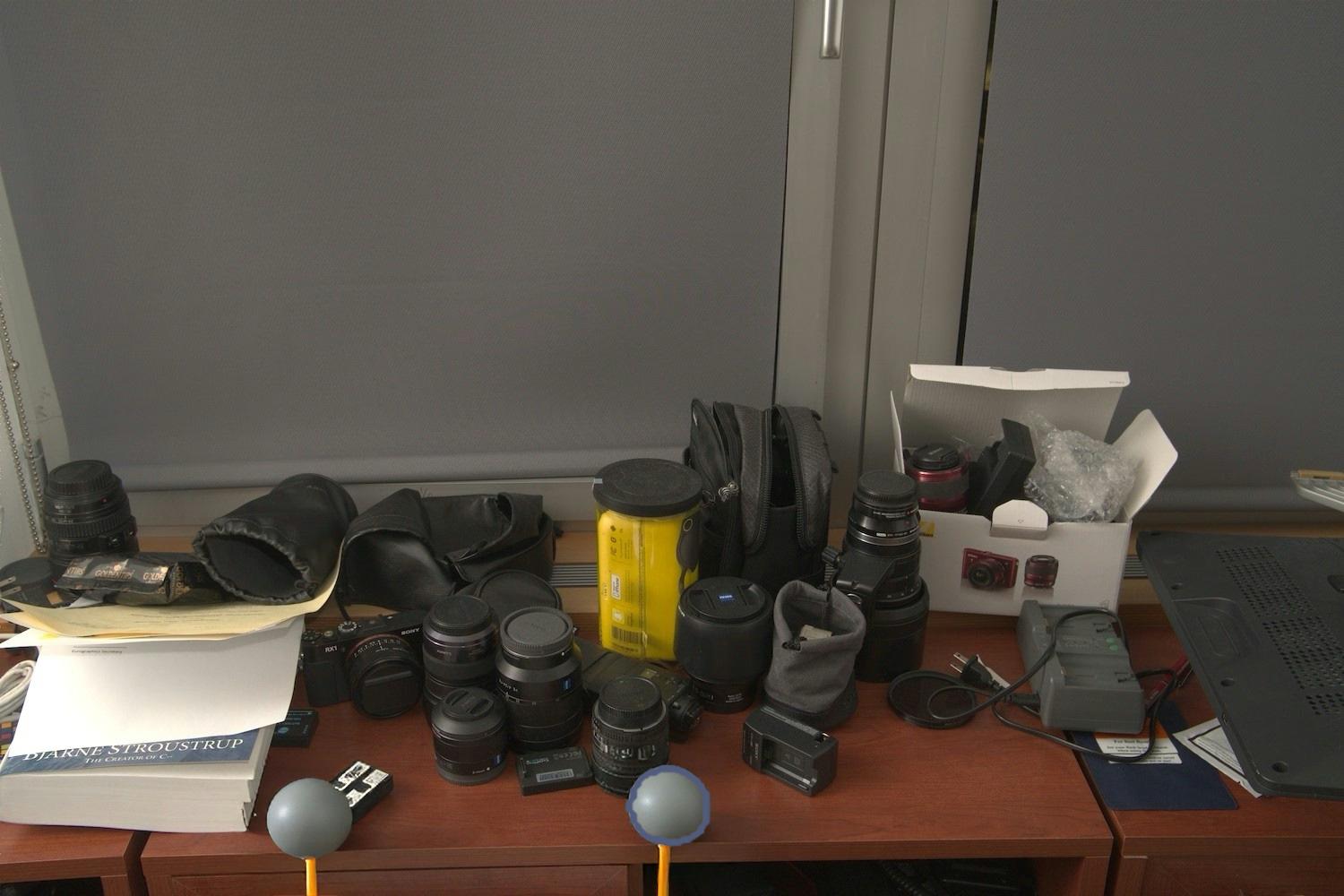}} &\raisebox{0pt}{\includegraphics[width=\colW\linewidth]{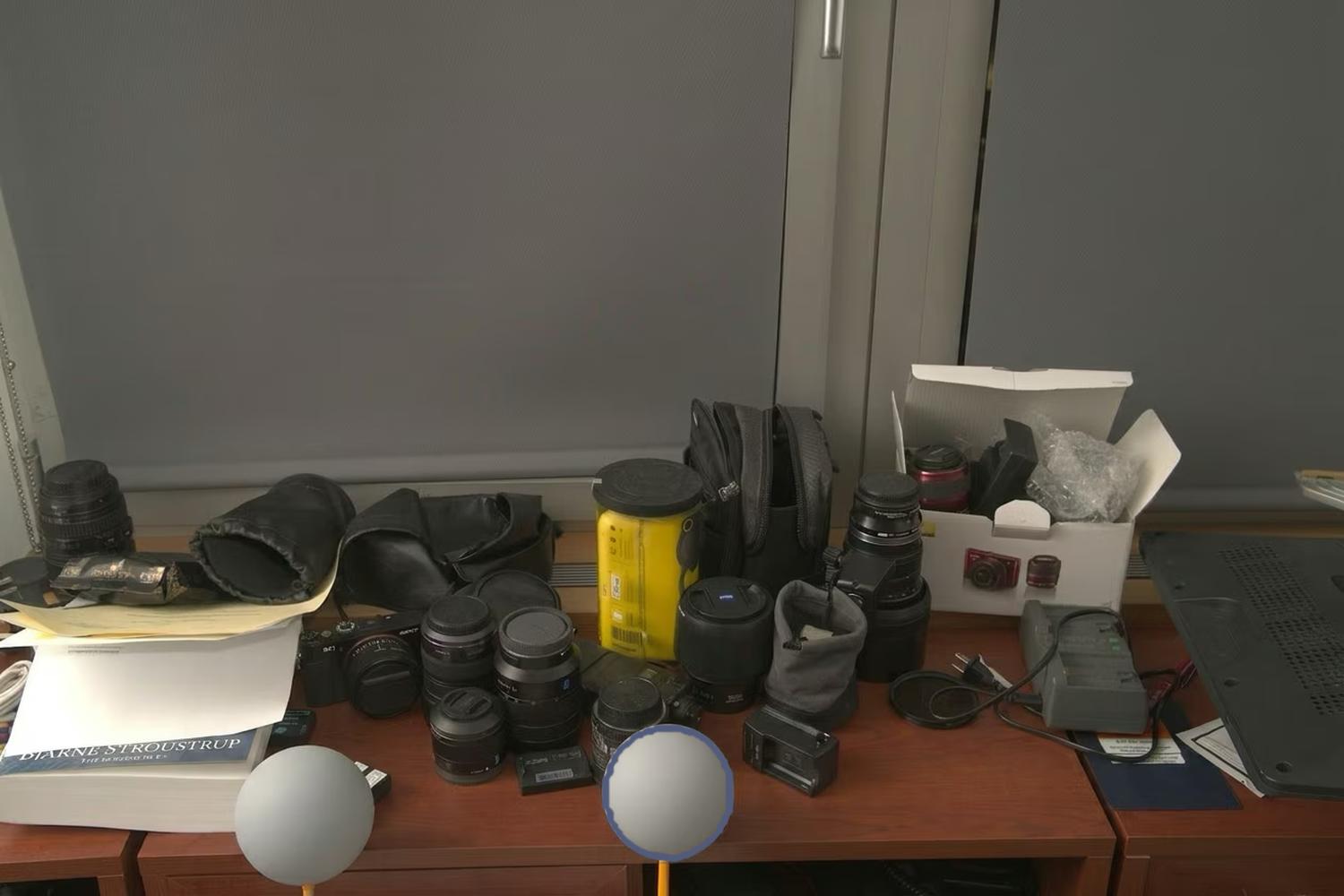}}\\[\imagegap]

    \raisebox{0pt}{\centering\includegraphics[width=\colw\linewidth]{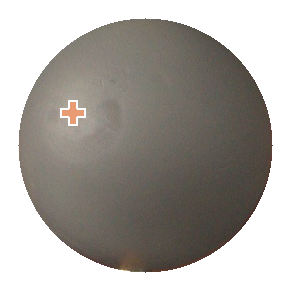}}&\raisebox{0pt}{\centering\includegraphics[width=\colw\linewidth]{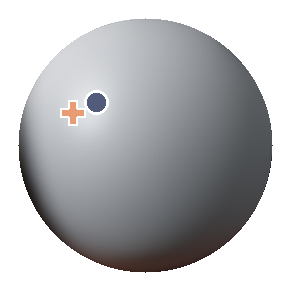}} &\raisebox{0pt}{\centering\includegraphics[width=\colw\linewidth]{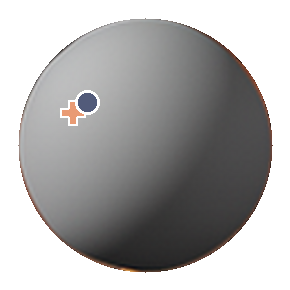}} &\raisebox{0pt}{\centering\includegraphics[width=\colw\linewidth]{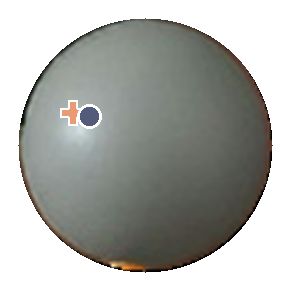}} &\raisebox{0pt}{\centering\includegraphics[width=\colw\linewidth]{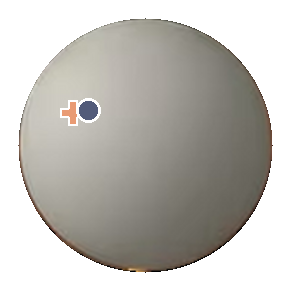}}\\

\end{tabularx}
\vspace{-2mm}
\caption{Examples of generated images (top row) from the Multi-Illumination dataset~\citep{murmann_dataset_2019} for different inpainting and editing models (columns) with the diffuse probe outlined in \couleurbleufonce{blue} and the reference probe in \couleurorangefonce{orange}. The extracted probe (bottom row) is shown with the estimated light direction \couleurbleufonce{blue} and the reference light direction in \couleurorangefonce{orange}. The first column corresponds to the reference image and probe.}
\Description{Examples of the generated images.}
\label{fig:exemples_results}
\end{figure}

There is no guarantee that the generative image models will correctly place the generated probes at the exact same location and size as the masked region. Therefore, the generated probes are first detected and segmented using SAM3~\cite{carion2025sam3segmentconcepts}, and the probe that overlaps with the ground-truth diffuse probe is selected as reference. Probes near the image boundaries often appear elongated due to perspective distortion; we rectify them to a circular shape. To validate that the model has correctly inpainted the required diffuse sphere, the gradient energy score, radial consistency, and relative colour variance of pixels within the segmentation mask are computed. These measures are compared to predefined thresholds (\num{>25}, \num{<0.991}, \num{<0.5} resp.), and detected probes that fail to meet any threshold are discarded.  This procedure ensures that the probes are spherical, fully contained within the image borders, unoccluded by scene elements, and that the model has produced a shaded sphere. Examples of the extracted probes are shown in \cref{fig:exemples_results}. See \cref{suppsubsec:methodo_detection} of the supp. mat. for more details.

\subsection{Light parameters estimation}
\label{subsec:methodo_LD}
Once probes are detected and extracted, we estimate the lighting parameters. Specifically, we focus on the dominant light direction and colour (as light intensity cannot be disentangled from the albedo). Colour is measured directly on the probe by averaging pixel values in CIE Lab space.

To obtain the dominant light direction, we use inverse rendering and fit a rendered sphere using the Cook-Torrance BRDF model~\cite{cook_reflectance_1982}, lit by a combination of directional and ambient lighting. We minimise the $L_1$ pixelwise distance between the rendered and target probe images, similar to  \citet{poirierginter_diffusion_2024}.
The Cook-Torrance model was selected for its ability to model both specular and Lambertian shading. Whilst the generative models are prompted to produce a sphere with Lambertian shading, they occasionally generate spheres exhibiting more complex reflectance properties. This microfacet model demonstrates greater robustness to such variations. The shading model optimises the material (albedo, roughness, metallicity and Fresnel coefficient) and the lighting parameters (light direction and intensity, and an ambient term) for each probe. Note, however, that only the light direction parameter is used for the analysis.

Given that this inverse rendering task is ill-posed, we initialise the shading model parameters to encourage convergence. Specifically, the initial light direction is calculated by assuming the brightest pixel is the centre of the specular highlight, and therefore the half-vector of the light and view directions. The ambient term is calculated from the intensity of the darkest pixel. The remaining parameters are initialised heuristically using median values from optimising on a synthetic test set and are further verified during testing of the generative image models.
The \texttt{Adam} optimiser~\citep{kingma2015adam} is used for \num{1500} steps with an initial learning rate of \num{1e-2}. Additional details of the light parameter optimisation are provided in \cref{suppsubsec:methodo_LD} of the supp. mat.

\subsection{Evaluation measures}
\label{subsec:methodo_measures}

As mentioned in \cref{subsec:methodo_LD}, we cannot assume the generative model has rendered a perfectly diffuse sphere. We therefore focus our evaluation on the estimated dominant light direction, which is robust to the underlying material generated by the evaluated model, and compute the angular error between the ground truth and the inpainted light directions. In addition, we also compute the colour difference $\Delta\textrm{E}_{ab}$ between the ground truth and inpainted probes, computed in the chroma channels of Lab space to mitigate the albedo/intensity ambiguity (\cref{subsec:evaluation_colour}). We cannot evaluate light intensity as it is conflated with the sphere reflectance, which can vary at each model inference (see supp. mat.). Finally, we evaluate the similarity between the radiance distributions of the ground truth and the inpainted spheres using the Kullback-Leibler divergence (\cref{subsec:distribution}).

\section{Benchmarking generative models for lighting accuracy}
\label{sec:evaluation}

We now proceed to analyse the performance of a variety of generative models (\cref{sec:models}) on light direction (\cref{subsec:evaluation_LD}), colour accuracy (\cref{subsec:evaluation_colour}), and radiance distribution (\cref{subsec:distribution}).

\subsection{Generative models}
\label{sec:models}

We evaluate \num{16} generative image models released over the past four years, spanning both inpainting models (\SDone~\cite{rombach2022high}, \SDtwo~\cite{rombach2022high}, \kolors~\cite{kolors}, \SDthree~\cite{pmlr-v235-esser24a}, \fluxone~\cite{blackforestlabs2024flux1fill}, \hunyuan~\cite{HunyuanImage-2.1}, \zimageturbo~\cite{imageteam2025zimageefficientimagegeneration}, \photoshop~\cite{firefly}) and image-editing models (\hidream~\cite{HiDream_2025_MM}, \NBone~\cite{geminiteam2025geminifamilyhighlycapable}, \qwen~\cite{wu2025qwenimagetechnicalreport}, \firefly~\cite{firefly}, \fluxtwo, \fluxtwofour, \fluxtwonine~\cite{flux-2-2025}, \NBtwo~\cite{geminiteam2025geminifamilyhighlycapable}), and including both open-weight and closed (API-only) systems. We selected these models to evaluate a variety of architecture types (U-Net-based diffusion, Diffusion Transformers (DiTs), rectified flow models, scalable single-stream diffusion models, \dots), purposes (inpainting/image-editing, photo-realism, general purpose), and number of parameters (1.3B to 32B for open models) to understand how they affect lighting estimation accuracy. It should be noted that \zimageturbo is a text-to-image model, so a ControlNet is used to allow image-to-image inference. Due to API access limitations for some of the closed models (\NBone, \NBtwo, \firefly), we were unable to test them on the full dataset.
The full list, release dates, parameter counts, and inference settings are summarised in \cref{tab:model_params} (supp. mat.). For each model, we use a fixed seed and the prompt of \cref{subsec:methodo_inpainting}; further per-model hyperparameters are reported in the supp. mat. Default published inference settings are used unless otherwise noted.

\begin{figure*}[t]
  \centering
  \includegraphics[width=\linewidth]{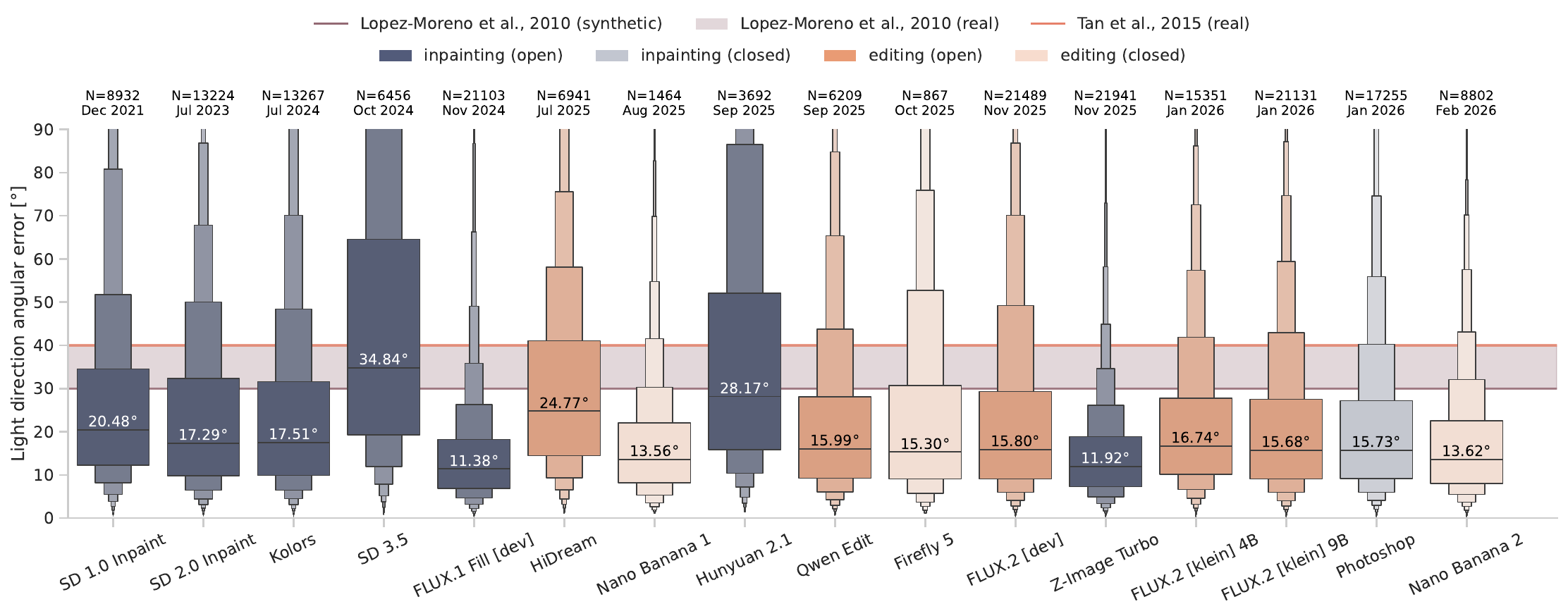}
  \caption{Distributions of the angular error of the light directions for each model, sorted by release date (displayed above each distribution).  The middle line in the box and the value displayed on the distribution correspond to the median.  Each step corresponds to quantiles of the distribution.  The distributions are coloured by type (inpainting or editing) and availability (open or closed).  The number of probes ($N$) used to compute each distribution is displayed at the top of the distribution.  The tails of the distribution, containing the outliers (above \ang{90}), are cropped out of the figure, to better see the distributions. The \citet{lopez2010measuring} (synthetic) line at \ang{30} represents their threshold where observers started being above chance level on their synthetic stimuli, and the \qtyrange[range-phrase=--, range-units=single]{30}{40}{\degree} range was determined by their psychophysical study using real images, identified by the (real) range.  The \citet{tan2015perception} (real) line at \ang{40} is the suggested limit based on their study on complex outdoor images with multiple light features conflated.}
  \Description{Angular error of the light direction for each model.}
  \label{fig:angular_error_direction_lumiere}
\end{figure*}

\subsection{Light direction}
\label{subsec:evaluation_LD}

We analyse how well the dominant light direction is captured by generative models and compare their abilities with their angular error distributions (\cref{subsubsec:evaluation_LD_ang_err}). We then explore model generation bias in terms of distribution per ground-truth light direction (\cref{subsubsec:evaluation_LD_ang_err_cluster}) and in azimuth and elevation (\cref{subsubsec:evaluation_LD_azimuth_elevation}).

\subsubsection{Light direction angular error distribution}
\label{subsubsec:evaluation_LD_ang_err}

The distribution of angular error between the light direction of the ground truth and inpainted probes for each model is shown in \cref{fig:angular_error_direction_lumiere}. The distributions reveal how accurately the different generative models infer and reproduce the correct light direction in a scene using the generated light probe. The following observations are made.

\paragraph{Best and worst models}
Drawing conclusions from the median angular error value for each model, we observe that \fluxone (\ang{11.38}) and \zimageturbo (\ang{11.92}) reproduce the most accurate lighting, compared to SD 3.5 (\ang{34.84}), \hunyuan (\ang{28.17}) and \hidream (\ang{24.77}), which have much higher second moments, as illustrated by the height of the quantiles.

It is important to note that not all distributions contain the same amount of data ($N$, displayed above each model in \cref{fig:angular_error_direction_lumiere}), since generative models generate acceptable light probes with varying degrees of success (\cref{fig:number_data_per_model_per_step} in the supp. mat.). However, a bootstrap equalisation analysis, performed with a sample corresponding to the model with the smallest $N$ (\firefly, $N=867$), indicates that the model with the largest equalised mean variation (\kolors) is \ang{0.049} (computed on \num{1000} bootstrap samples), therefore showing this number is sufficient for robust statistics.

\paragraph{Lighting accuracy has stayed mostly constant with time} Models appear in time-of-publication order in \cref{fig:angular_error_direction_lumiere}, showing that there is no clear improvement in lighting accuracy over time. Even though \fluxone was released in November 2024, it remains the most accurate model for lighting direction.

\paragraph{Later models from the same family do not outperform previous generations} For example, \SDtwo improves over \SDone, but \SDthree is significantly worse. We also observe that \fluxone has considerably different performances than the later versions (\fluxtwo, \fluxtwofour, and \fluxtwonine), which all perform similarly. This seems to indicate that models do not train specifically for lighting accuracy---it rather comes as a by-product of training on their respective tasks, with varying degrees of success.

\paragraph{Closed models do not outperform open models} On average, closed models (identified with a lighter shade in \cref{fig:angular_error_direction_lumiere}) have lower angular error in the light direction than most open models. The best closed models (\NBone and \NBtwo) perform similarly, but are still slightly worse than the best open models (\fluxone and \zimageturbo).

\paragraph{There is no superior image editing approach} The median value of the median angular error of the light direction for the inpainting models is \ang{16.51} (\num{6} models, when excluding the outliers \hunyuan and \SDthree, \ang{17.40} for all \num{8} models) and \ang{15.74} for the editing models (\num{8} models), which results in only a difference of \ang{0.77}, indicating that there is no clear correlation between either task and their light direction accuracy.

\paragraph{Model complexity does not reflect lighting accuracy} Considering all models, there is no correlation between their number of parameters and median angular error.
A good example is the performances of \fluxtwofour (\ang{16.74}) and \fluxtwonine (\ang{15.68}), yet only a difference of \ang{1.06} can be observed.
Furthermore, \fluxtwo (\ang{15.80}) has 32B parameters, yet performs worse than \fluxtwonine by \ang{0.12}.
The number of parameters for each model is listed \cref{tab:model_params} and their relationship in \cref{fig:median_angular_error_vs_parameters}, in the supp. mat.

\paragraph{General observations and human perception}
The lack of global temporal improvement in lighting accuracy and the fact that closed models are not necessarily the most accurate could indicate that the models are being trained to generate aesthetically pleasing rather than physically accurate images.
This is consistent with the fact that images are usually produced to be consumed by humans, who are not particularly good at detecting physical inaccuracies~\cite{cavanagh2005artist}.  As also indicated by \citet{giroux2024towards}, the physical accuracy and perceptual realism of lighting in humans can differ.

To support this, the perceptual sensibility thresholds for humans to lighting incoherences, as determined by \citet{lopez2010measuring} and \citet{tan2015perception}, are shown in \cref{fig:angular_error_direction_lumiere}.  The ``synthetic" line corresponds to rendered stimuli of abstract globally convex shapes, asking observers to select the object lit differently, which resembles the task of computing the angular error.  For the ``real" thresholds, the stimuli consisted of real objects lit incoherently in complex scenes, which closely resembles the task humans would perform when consuming the images generated by these inpainting models.  However, these studies differ---in terms of geometry and materials---from the Multi-Illumination dataset~\cite{murmann_dataset_2019} and were conducted on a limited number of scenes with coarse light direction sampling. Therefore, these lines are a guide to suggest that, for the majority of inpainted probes, the scale of the light direction error produced may not be perceptible to the human visual system, especially given the consensus in the literature that humans are not sensitive to minor lighting incoherencies \citep{ostrovsky_perceiving_2005}.

\subsubsection{Angular error distribution per ground-truth light direction}
\label{subsubsec:evaluation_LD_ang_err_cluster}

We refine the analysis by examining the light direction angular error as a function of the ground-truth light directions. We define coarse light direction bins such as ``front'', ``left'', ``top-left'', etc. (see the supp. mat. for more details) to indicate incoming light directions. Note that there are no ``down'' or ``back'' directions since the Multi-Illumination dataset does not contain such images.

\begin{figure}[t]
  \centering
  \includegraphics[width=\linewidth]{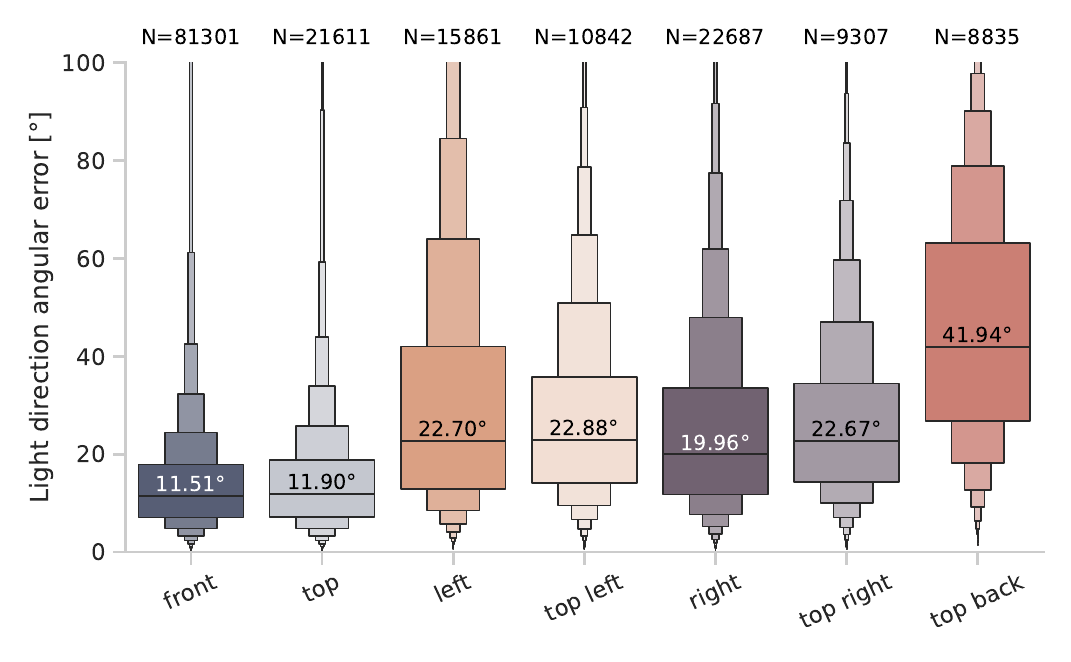}
  \caption{Angular error distribution, per coarse incoming ground-truth light direction, separated in ranges of \ang{90} in azimuth and \ang{45} in elevation. Here, the distributions are computed by aggregating results from all models, except the 3 worst from \cref{fig:angular_error_direction_lumiere} (\SDthree, \hidream, and \hunyuan). The middle line in the box and the value displayed on the distribution correspond to the median. Each step corresponds to quantiles of the distribution. The total number of inpainted probes ($N$) for each cluster is displayed above each distribution.}
  \Description{Angular error distribution.}
  \label{fig:boxplot_gt_clusters_per_direction}
\end{figure}

The angular error distributions for each incoming light direction bin are shown in \cref{fig:boxplot_gt_clusters_per_direction}. Here, results from all but the 3 worst models from \cref{fig:angular_error_direction_lumiere} (namely \SDthree, \hidream and \hunyuan) are aggregated. We observe that generative models tend to struggle when the light comes from the ``top back,'' with an average median value of \ang{41.94}.
Comparatively, the ``front'' and ``top'' directions have average median values of \ang{11.51} and \ang{11.90}, respectively, which indicate that they are the directions the most easily handled by the models. Models handle images with a relative left-right symmetry, as shown by average median values of \ang{22.70} and \ang{19.96} for the ``left'' and ``right'' directions respectively, and similarly for the ``top left'' and ``top right'' (\ang{22.88} and \ang{22.67}).

\subsubsection{Azimuth \& elevation}
\label{subsubsec:evaluation_LD_azimuth_elevation}

\begin{figure*}[h]
  \centering
  \footnotesize
  \begin{tabular}{cc}
  \includegraphics[width=0.49\linewidth]{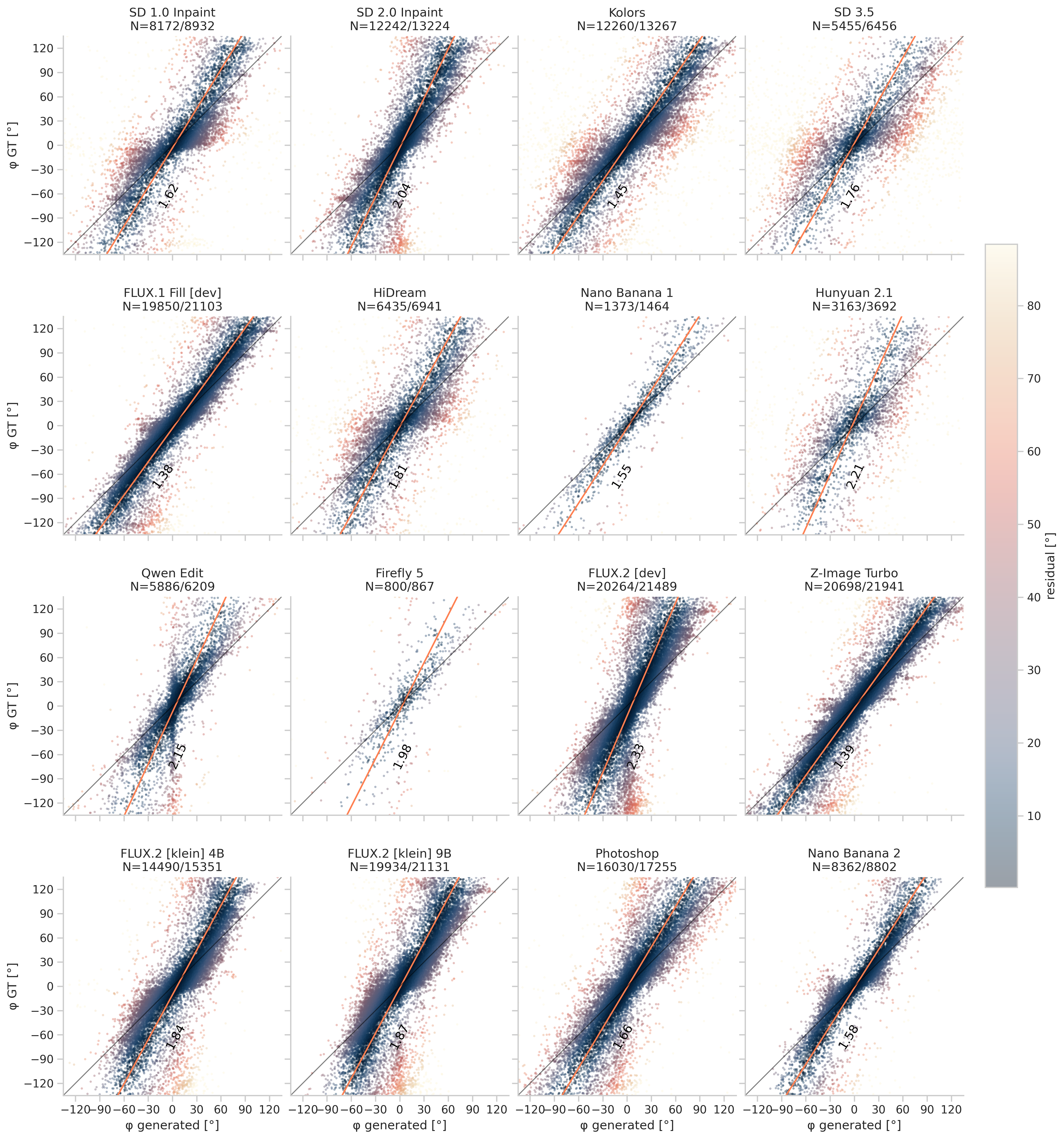} &
  \includegraphics[width=0.49\linewidth]{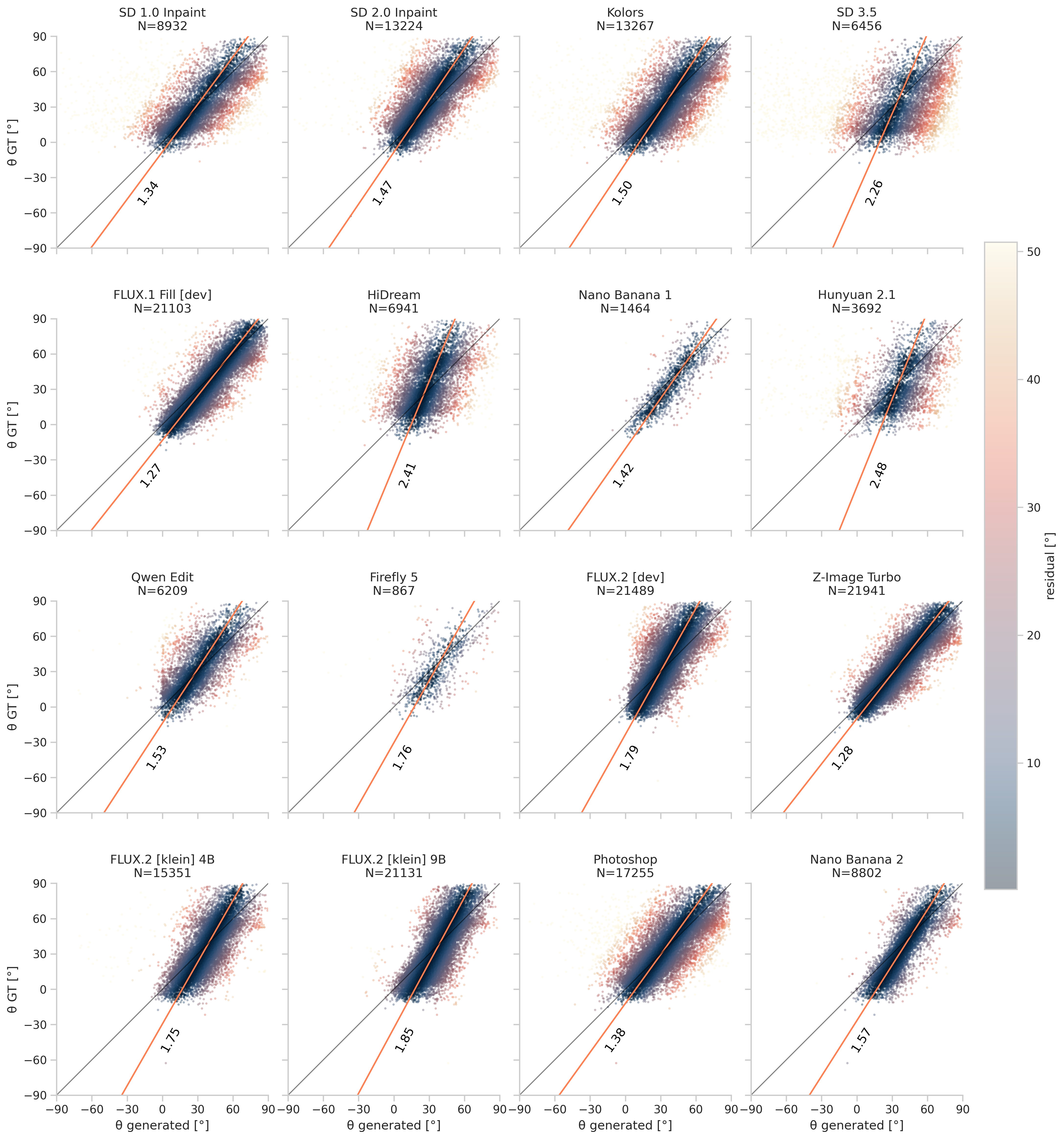} \\
  (a) azimuth & (b) elevation
  \end{tabular}
  \caption{Comparison of the (a) azimuth and (b) elevation errors, where each plot illustrates the angle of the inpainted probe ($x$-axis) against the ground truth ($y$-axis). The theoretical $\text{slope}=1$ (black line) indicates perfect performance. A linear regression (orange line) is fitted on the data, with the slope displayed in black near it. For azimuth, only the [\ang{-135}, \ang{135}] range is considered. Data points are coloured by the residual error from the fit (c.f. \cref{subsec:methodo_LD}).  The number of data points $N$ included for each model is displayed, and for the azimuth, the number of data points included in the [\ang{-135}, \ang{135}] range is also shown.}
  \Description{Comparison of azimuth and elevation errors.}
  \label{fig:comparison_azimuth_elevation_per_model}
\end{figure*}

We further explore potential differences between the azimuth and elevation components of the lighting by computing errors for each of them separately. The comparisons between the ground truth and inpainted azimuth and elevation are shown in \cref{fig:comparison_azimuth_elevation_per_model}.
For the azimuth, only the [-\ang{135}, \ang{135}] interval is kept for analysis.

\paragraph{General observations}
The median slope for all models, obtained by linear regression, is \num{1.79} in azimuth and \num{1.55} in elevation. Globally, all models exhibit a slope $>1$. For the azimuth, this implies that models tend to predict lighting that is biased towards the frontal direction over more extreme left/right directions. For elevation, models struggle to predict lighting from above (consistent with the findings from \cref{subsubsec:evaluation_LD_ang_err_cluster}) and are also biased towards lower elevations.

\paragraph{Some models exhibit an S-shape for azimuth} We observe in \cref{fig:comparison_azimuth_elevation_per_model} that some models (e.g., \SDone, \fluxtwofour, \NBtwo) exhibit an S-shape in their predictions. The S-shape indicate a flatter slope around $\varphi=0$, showing models have lower errors when the light comes straight from the front. However, at the inflexion point (around $\pm$\ang{30}), the models tend to plateau and struggle to match the ground truth. This further illustrates the central bias of models: greater precision in matching the correct azimuth is achieved when $\varphi \in [-30^\circ, 30^\circ]$ approximately. This is in contrast to the top-performing models (such as \fluxone and \zimageturbo), which exhibit a much straighter trend, with the S-shape less noticeable. These models can span a wider range of azimuth angles (as shown by their $\text{slope} \approx 1.4$) and better match the ground truth.

\subsubsection{Spatially varying lighting}
\label{subsubsec:evaluation_LD_spatial_variation}

\begin{figure*}[t]
  \centering
  \includegraphics[width=\linewidth]{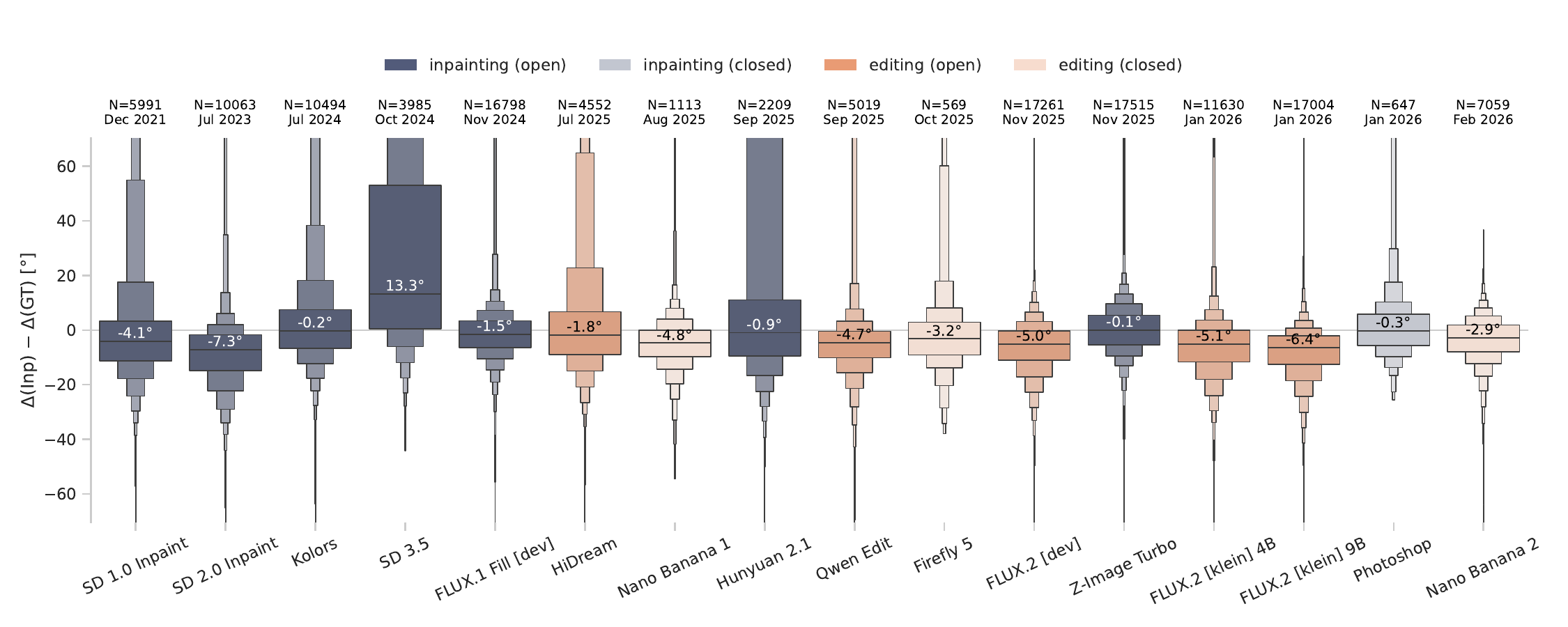}
  \caption{Distributions of the difference in spatial variation of light direction for each model. Models are sorted by release date (above) and coloured by type (inpainting or editing) and availability (open or closed).  The median is shown on middle line, and each step corresponds to distribution quantiles.  The number of probes ($N$) used to compute each distribution is displayed at the top of the distribution.  The tails containing the outliers ($\pm$\ang{60}) are cropped out for legibility.  A value of \num{0} means the generated probes reproduces the ground-truth angles exactly.}
  \Description{Difference between the generated and ground-truth light directions for probes in different locations.}
  \label{fig:direction_lumiere_spatial_variation}
\end{figure*}

To validate that the models understand lighting direction rather than merely performing local harmonisation, we further measure the spatial variation in lighting direction.  To do so, we leverage the two ground-truth probes present in the Multi-Illumination dataset.  The dominant light direction of the second ground-truth probe (chrome) is obtained by detecting its specular highlight (see \cref{subsec:methodo_dataset}).
Since both diffuse and chrome probes are masked when being inpainted, the generative models already produce an image with two inpainted diffuse probes (\cref{fig:qualitative_grid}). We then apply the same estimation algorithm (\cref{subsec:methodo_LD}) to compute the light direction for both inpainted probes. Images lacking two inpainted probes are excluded from evaluation.

Qualitative examples, along with the measured light directions, are shown in \cref{fig:qualitative_grid} (supp. mat.).  We measure the difference between the light directions of the generated probe and the ground-truth, and present the distributions of these spatial variations of the lighting direction for each model in \cref{fig:direction_lumiere_spatial_variation}.  A difference of \num{0} indicates that the model perfectly reproduces the ground-truth variation in spatial lighting direction.  Positive (negative) values indicate that the model has over(under)-estimated how much the lighting direction should vary.  The distributions of the angular differences for each models and the ground truth are shown in the supp. mat. in \cref{fig:spatial_variation_dual_probe_inp_with_gt}.

\paragraph{General observations}
\Cref{fig:direction_lumiere_spatial_variation} shows a median discrepancy below \ang{14} for all models, and at or below \ang{5} for \num{12}/\num{16}, indicating that most evaluated models can successfully modulate lighting spatially in the image, mostly consistent with the ground truth.  \zimageturbo performs remarkably well at understanding spatially varying lighting, with a nearly identical angle between the light directions of its probes and the ground-truth probes.  \SDthree and \SDtwo fail in opposite directions. \SDthree exaggerates lighting variations (\ang{13.3}, a variance failure), whereas \SDtwo has the most negative angle difference (\ang{-7.3}, a bias failure), in which the spatially varying aspect of light is largely absent. We hypothesise that the reason lies in how each architecture propagates spatial context. \SDtwo restricts attention to coarse resolutions and, being otherwise fully convolutional, carries no absolute position signal; long-range cues must pass through a heavily pooled representation, which favours a single scene-level illumination estimate applied regardless of probe position. \SDthree, by contrast, applies full joint attention with explicit positional embeddings, so any patch can attend to any light source in the scene, but nothing constrains those attention weights to be photometrically correct, and spurious long-range associations translate directly into large, confident errors.

\subsubsection{Spherical harmonics coefficients}
\label{subsubsec:evaluation_LD_SH}

\begin{figure}[h]
  \centering
  \includegraphics[width=\linewidth]{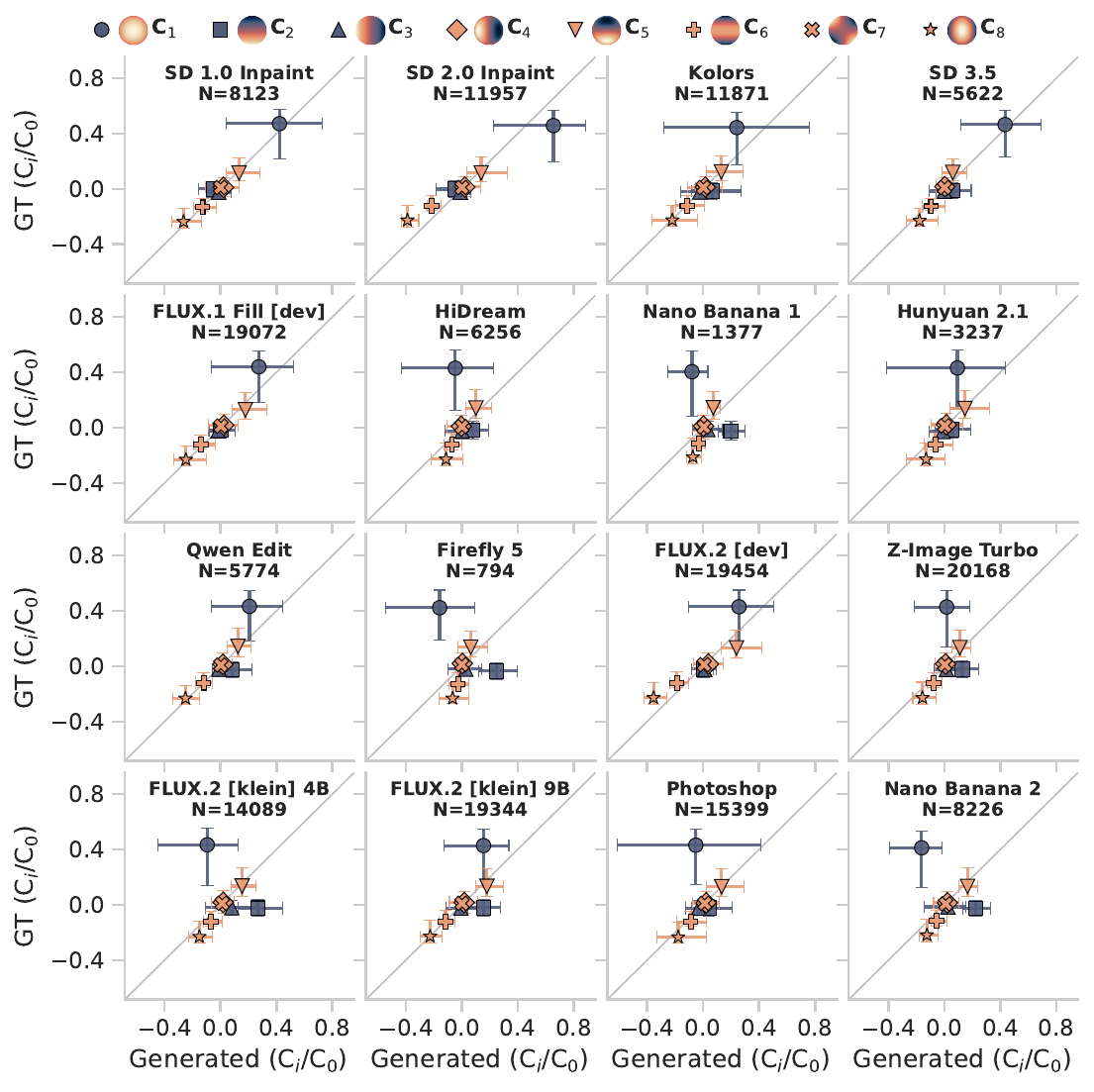}
  \caption{Median relative spherical harmonic coefficients of the generated probes ($x$-axis) compared to the ground truth ($y$-axis) for each model.  The coefficients are divided by the $\textrm{C}_0$ value.  The probes with $\abs{\textrm{C}_0} > 0.1$ are filtered out for stability.  The error bars correspond to the \num{25}th and \num{75}th quantile of the distribution.  The direction each coefficient represents is displayed in the legend, where the \couleurbleufonce{blue} points correspond to the first-order coefficients and \couleurorangefonce{orange} are the second-order coefficients.  The model's names and number of probes used to compute the statistics are displayed at the top of each figure.  The theoretical $\text{slope}=1$ (black line) indicates perfect performance.}
  \Description{Median spherical harmonic coefficients of the ground truth and generated probes for each model.}
  \label{fig:median_SH_coefficients}
\end{figure}

Inspired by \citet{farid_lighting_2022}, we extend our analysis using a spherical harmonics (SH) based comparison.  Since second-order SH capture over 99\% of the energy of the irradiance signal on Lambertian surfaces \citep{ramamoorthi_relationship_2001,ramamoorthi2001efficient}, we compute the second-order SH coefficients for both the generated and ground-truth probes in RGB, then convert the coefficients to greyscale.

The zeroth-order coefficient $\textrm{C}_0$ corresponds to the ambient term of the SH; however, since the probes' albedo is unknown, it conflates irradiance with surface reflectance and cannot be interpreted as a lighting parameter on its own.  We therefore normalise each probe by dividing its coefficients ($\textrm{C}_i$) by the magnitude of the zeroth-order term ($\textrm{C}_0$), which factors out this intensity ambiguity and leaves the lighting shape as the quantity under analysis.  We discard the probes with $\abs{\textrm{C}_0} > 0.1$ for stability, which removes \num{9.23}\% of the data.

\paragraph{General observations}

\Cref{fig:median_SH_coefficients} shows the median relative SH coefficients for each model, compared with the median of the ground-truth probes. The variation among the data points lies mainly along the $x$-axis, showing the various models' biases.  The $\textrm{C}_i$ values of the ground truth ($y$-axis) vary slightly between each models; this is due to the fact that the set of scenes for which the generated probes are valid varies per model, thus each has a different set of associated ground truth probes.  Overall, across all coefficients, the models' medians align well with the ground-truth values, clustering around the unit-slope line (black line).  There does not seem to be any clear trend between the type of model (inpainting or editing) and their availability (open or closed).

\paragraph{First-order coefficients}

The $\textrm{C}_1$ component corresponds to the front/back axis, which is the direction along which our probes are least reliable: small departures from the uniform Lambertian assumption---specular response, spatially varying texture, inter-reflections---redistribute energy between the constant and first-order term $\textrm{C}_1$, so $\textrm{C}_0$ and $\textrm{C}_1$ are readily conflated when observing only the front hemisphere. In addition, for hemispherical observations, there is crosstalk between the coefficients as the SH bases $Y$ are not orthogonal on the hemisphere ($\int_{y \ge 0} Y_{0,0}Y_{1,\text{-}1} \ne 0$). Variation reported along this axis should therefore be interpreted with more caution than the lateral components.

The first-order $\textrm{C}_3$ component (azimuth) shows median values near \num{0}, coherent with the findings from \cref{fig:boxplot_gt_clusters_per_direction} and \cref{fig:comparison_azimuth_elevation_per_model}(a) that the models do not have a left/right directional bias.  The first-order $\textrm{C}_2$ component (elevation) leans towards the positive side, indicating that the models tend to predict that the lighting comes from above more often than the ground truth.  This corroborates the results in \cref{fig:comparison_azimuth_elevation_per_model}(b) and \cref{fig:gt_clusters_spatial_dist}, which show that there are no ground-truth probes lit from below.  The same crosstalk phenomenon between $\textrm{C}_0$ and $\textrm{C}_1$ appears between the second-order $\textrm{C}_5$ and $\textrm{C}_2$ coefficients, and this pair captures the skew towards lighting from above.
This crosstalk is a direct consequence of our partial observation setup: the SH basis is orthonormal over the full sphere, but our probes expose only a hemisphere facing the camera; within that domain, the inner products between basis functions no longer all vanish. The projection is therefore no longer a diagonal system, and coefficients that both encode vertical variation---$\textrm{C}_2$ and $\textrm{C}_5$ in particular---absorb part of each other's energy, with the split between them depending on the estimator rather than on the lighting itself. This ambiguity is an intrinsic weakness of SH coefficients as a lighting representation under hemispherical observation, and motivates our use of the specular highlight to recover light direction on the ground-truth probes (\cref{subsec:methodo_dataset}).

\paragraph{Second-order coefficients}

Several models — \SDone, \kolors, \fluxone, \qwen, \fluxtwonine — show median second-order coefficients (\couleurorangefonce{orange}, $\textrm{C}_4$--$\textrm{C}_8$) close to the ground-truth values, and this group does not coincide perfectly with the models that best recover the dominant lighting direction (\cref{fig:angular_error_direction_lumiere}). The two measurements are not redundant: the first-order band encodes the dominant direction, whereas the second-order band reflects the angular spread of the incident lighting and the presence of secondary sources. A model may therefore reproduce the higher-frequency structure of realistic illumination while incorrectly placing the dominant source, and vice versa. We note two limits on this reading. First, \cref{fig:median_SH_coefficients} reports medians across probes; agreement in aggregate does not imply per-probe agreement, and a model whose errors are symmetric about the ground truth will appear well calibrated here. Second, for a probe lit by a single distant source, the second-order coefficients are largely determined by the first-order ones through the Lambertian transfer kernel, so the two bands are only partially independent.

\subsection{Colour}
\label{subsec:evaluation_colour}

\begin{figure*}[h]
  \centering
  \includegraphics[width=\linewidth]{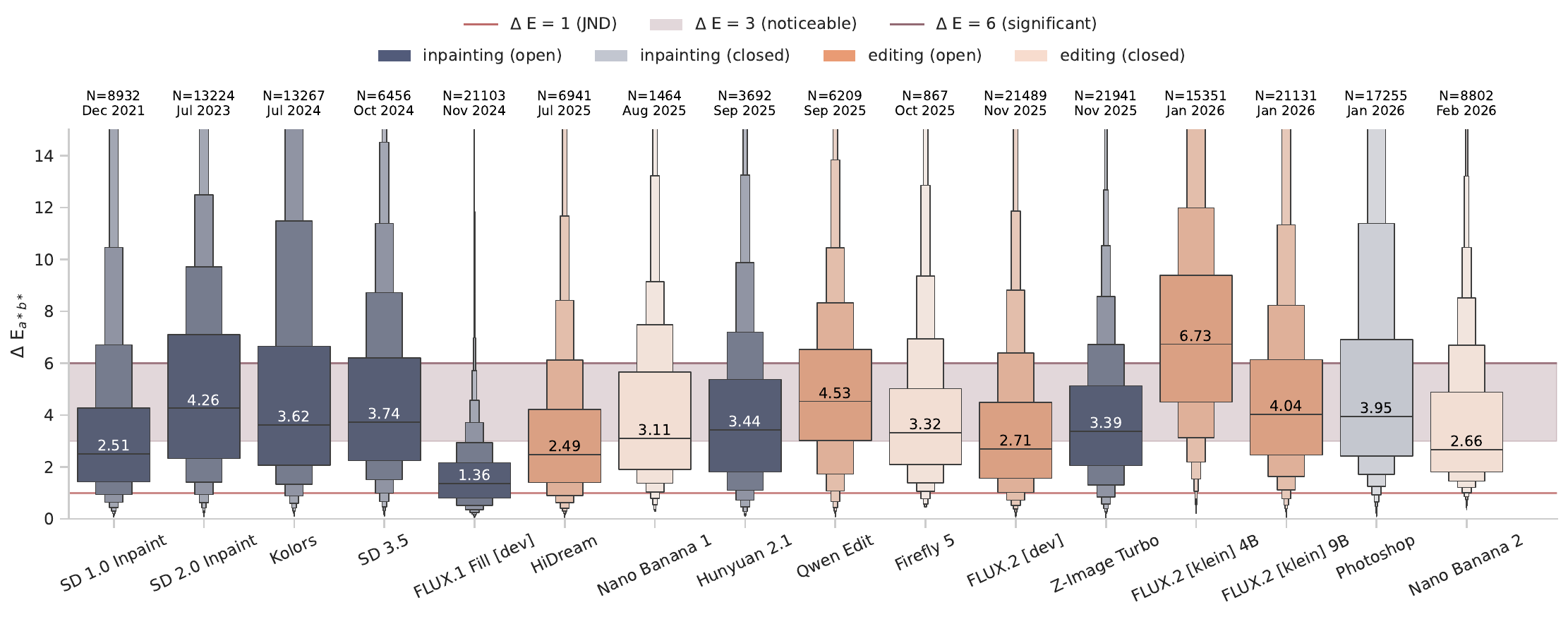}
  \caption{Distributions of the $\Delta\textrm{E}_\textrm{ab}$ for each model, sorted by release date (displayed above each distribution). The middle line and value displayed on the distribution corresponds to the median.  Each step corresponds to quantiles of the distribution.  The distributions are coloured by type (inpainting or editing) and availability (open or closed).  The number of probes ($N$) used to compute each distribution is displayed at the top of the distribution. The $\Delta\textrm{E}_\textrm{ab}=1$ line corresponds to the Just Noticeable Difference (JND) of human perception (when colours are shown next to each other), colour differences start to be noticeable between $\Delta\textrm{E}_\textrm{ab} \in [3, 6]$ and appear as different colours for $\Delta\textrm{E}_\textrm{ab}\geq6$ \citep{fairchild_color_2013}.}
  \Description{Distributions of the $\Delta\textrm{E}_\textrm{ab}$ for each model.}
  \label{fig:delta_E}
\end{figure*}

Our framework further enables evaluation of the colour-accurate illumination capabilities of generative models across the diverse lighting conditions in the Multi-Illumination dataset.
Since we prompt models to generate a ``grey'' sphere, we assume that any resulting colour in the generated probes is due to the estimated lighting. To measure the colour accuracy, we use $\Delta\textrm{E}_{ab}$, which is computed on the $a$ and $b$ (chroma) channels in Lab space. This avoids accounting for luminance, which could introduce uncontrolled factors (e.g., light intensity, albedo).

\paragraph{General observations}
Results from this colour analysis are reported in \cref{fig:delta_E}, which plots the distribution of errors in $\Delta\textrm{E}_{ab}$ and highlights 3 zones: the Just Noticeable Difference (JND) of human perception ($\Delta\textrm{E}_{ab} = 1$), when an error is deemed noticeable ($\Delta\textrm{E}_{ab} = 3$) and when it is significant ($\Delta\textrm{E}_{ab} > 6$). All tested models have median errors less than $\Delta\textrm{E}_{ab} = 6$, with many either very close (\SDone, \hidream, \NBtwo) or less (\fluxone, \hunyuan) than $\Delta\textrm{E}_{ab} = 3$. This strong overall performance could be due to the similarity between light colour estimation and harmonisation: image editing and inpainting models are likely trained to harmonise inserted objects with their backgrounds.  Nevertheless, we note that luminance is not considered in this measure, as it cannot be disentangled from the sphere's albedo; as a result, achieving values closer to non-significant differences may be somewhat easier.

\begin{figure}[t]
  \centering
  \includegraphics[width=\linewidth]{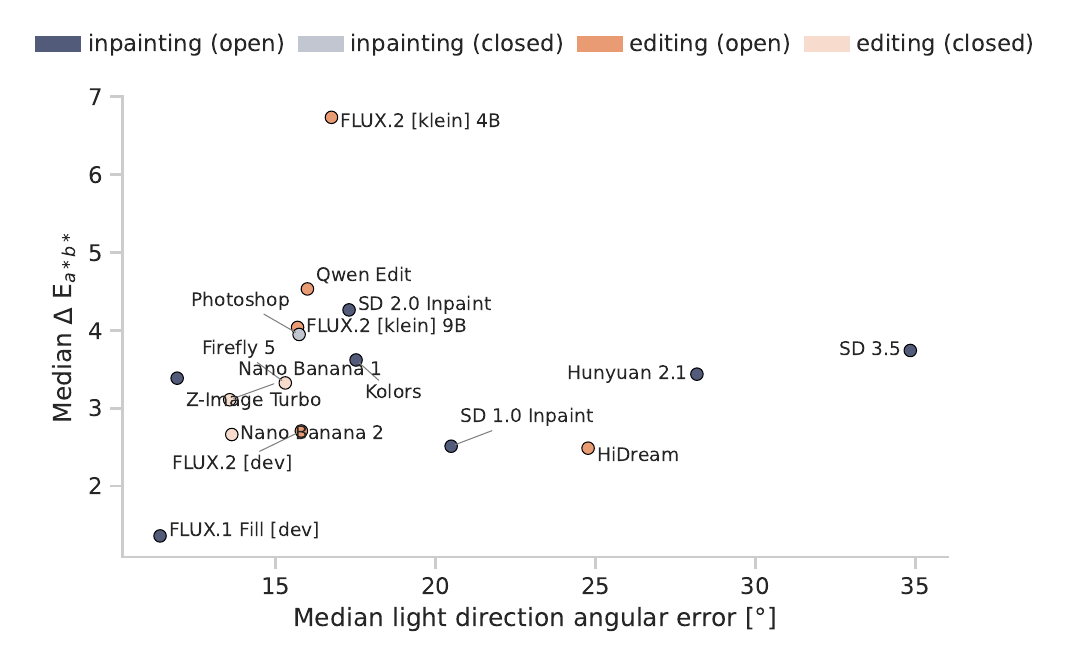}
  \caption{Median colour error as a function of median light direction angular error, for all models evaluated. Models are coloured by type (inpainting or editing) and availability (open or closed), as in \cref{fig:angular_error_direction_lumiere}.}%
  \Description{Light direction error as a function of colour error.}
  \label{fig:correlation_LD_colour}
\end{figure}

\paragraph{Comparing light direction and colour}
When plotting both light direction and colour in
\cref{fig:correlation_LD_colour}, we observe that most models follow the trend that a high error in direction is correlated with a higher error in colour, with \fluxone performing best at both tasks. We note outliers to this trend, however. On the one hand, \hidream and \SDone, which have relatively poor performance in light direction, perform comparatively better in colour. On the other hand, \zimageturbo shows competitive performance in terms of light direction accuracy compared to \fluxone, yet performs considerably worse in terms of colour accuracy.

\subsection{Probe intensity distribution}
\label{subsec:distribution}

\begin{figure*}[t]
  \centering
  \includegraphics[width=\linewidth]{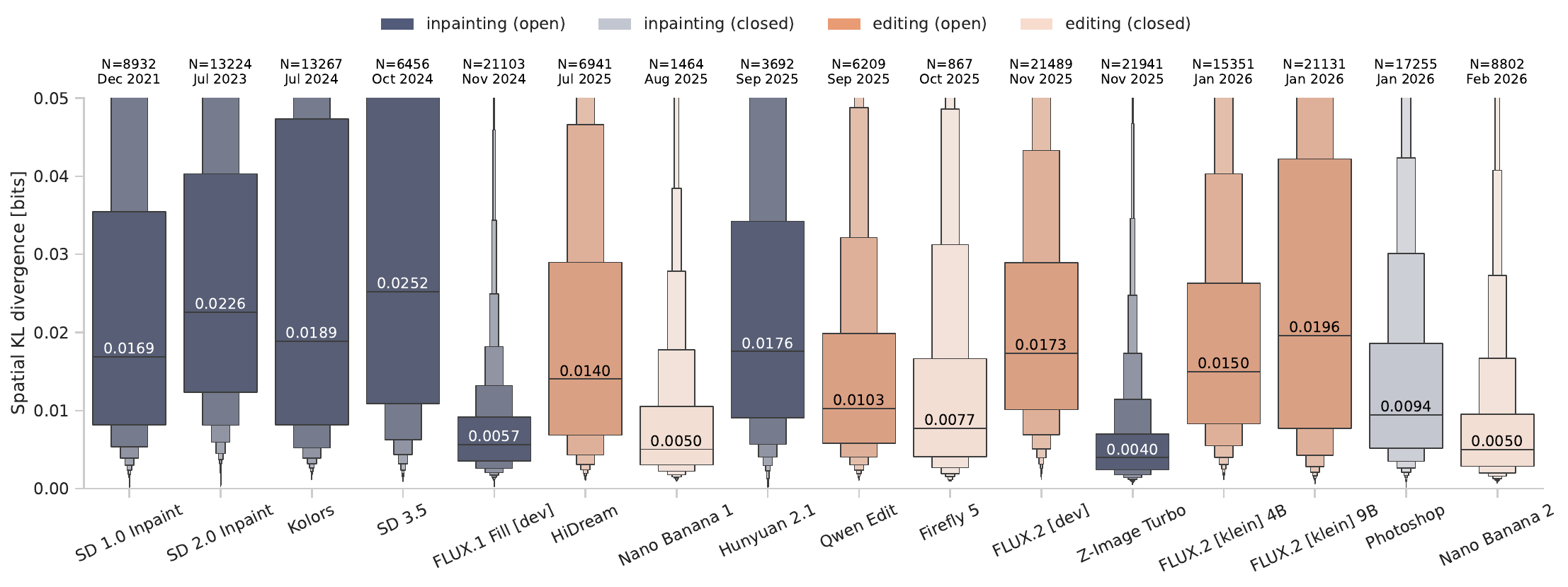}
  \caption{Distributions of Kullback-Leibler divergence between the estimated and ground-truth light probes. Here, the pixel intensities are normalised and treated as 2D distributions $P(\vartheta, \varphi)$ before comparison.  The middle line in the box and the value displayed on the distribution correspond to the median.  Each step corresponds to quantiles of the distribution. The distributions are coloured by type (inpainting or editing) and availability (open or closed).  The number of probes ($N$) used to compute each distribution is displayed at the top of the distribution, as well as the model's release date.  The tails of the distribution, containing the outliers (above \num{0.05} bits) are cropped out of the figure, to better see the distributions.}
  \Description{Spatial Kullback-Leibler divergence between the inpainted and ground-truth probes.}
  \label{fig:spatial_kl_boxenplot}
\end{figure*}

The analyses in \cref{subsec:evaluation_LD,subsec:evaluation_colour} both rely on optimising a shading model (see \cref{subsec:methodo_LD}). Whilst this approach recovers interpretable physical quantities such as dominant light direction, it also introduces uncertainties and potential sources of error. A natural concern is whether the conclusions drawn in \cref{subsec:evaluation_LD,subsec:evaluation_colour} reflect genuine differences in lighting accuracy across models, or instead reflect biases of the inverse rendering procedure itself.

To verify this, we treat the probe region as a two-dimensional distribution of pixel intensities and compare the inpainted and ground-truth probes using a Kullback-Leibler (KL) divergence.
We normalise both the inpainted and ground-truth probes to unit sum, treating each as a discrete probability distribution $P(\vartheta, \varphi)$ over angular position on the visible hemisphere of the sphere.

\paragraph{General observations}
\Cref{fig:spatial_kl_boxenplot} presents the KL divergence analysis for each model. The models with the most accurate light direction and colour estimates in \cref{fig:angular_error_direction_lumiere,fig:delta_E} also have the lowest distributional divergence: \fluxone, \zimageturbo, and the Nano Banana family lead the ranking.
The two measures are obtained independently: one uses inverse rendering, the other a model-free distribution comparison. Yet their rankings closely match, suggesting that our optimisation genuinely captures the illumination conditions.

\section{Discussion}
\label{sec:discussion}

\paragraph{Local harmonisation}
Our analysis investigates whether generative models are merely performing local harmonisation or have a more global understanding of the lighting in the space. Our results reveal that successful generative models achieve both. The colour experiments (\cref{subsec:evaluation_colour}) are closely related to harmonisation: matching scene chromaticity is a simple yet effective prior that does not require reasoning about the full lighting. However, the light direction analysis (\cref{subsec:evaluation_LD}) goes beyond, since matching local colour statistics cannot produce a probe with the correct light direction. Finally, \cref{subsubsec:evaluation_LD_spatial_variation} demonstrates that models reproduce the correct spatially varying light differences across image locations: a behaviour that cannot be explained by local harmonisation since it requires inferring a consistent scene-level lighting.

\paragraph{Approximating the measurement error}
Assuming a probe is correctly generated and segmented, quantifying the exact measurement error of our pipeline is difficult since we do not know the material properties of the generated probes. As a proxy for evaluation, we generated a physics-based rendering (PBR) dataset of spheres of varying material properties and known light directions, resembling those created by the generative models (more details in \cref{suppsubsec:evaluation_LD}). Our inverse rendering algorithm from \cref{subsec:methodo_LD} achieves \ang{2.55} median error on this PBR data. In addition, we do not know the material properties of the real probes either. Real images may also contain other artefacts, such as multiple light sources, noise, or interreflections with nearby surfaces, which may affect the estimate of the dominant light direction. Here, our inverse rendering algorithm obtained a median angular error of \ang{9.35} compared to our specular highlight detection. We consider this to be an upper bound on the measurement error of our pipeline.

\paragraph{Survivorship bias}

\begin{figure}[t]
  \centering
  \includegraphics[width=\linewidth]{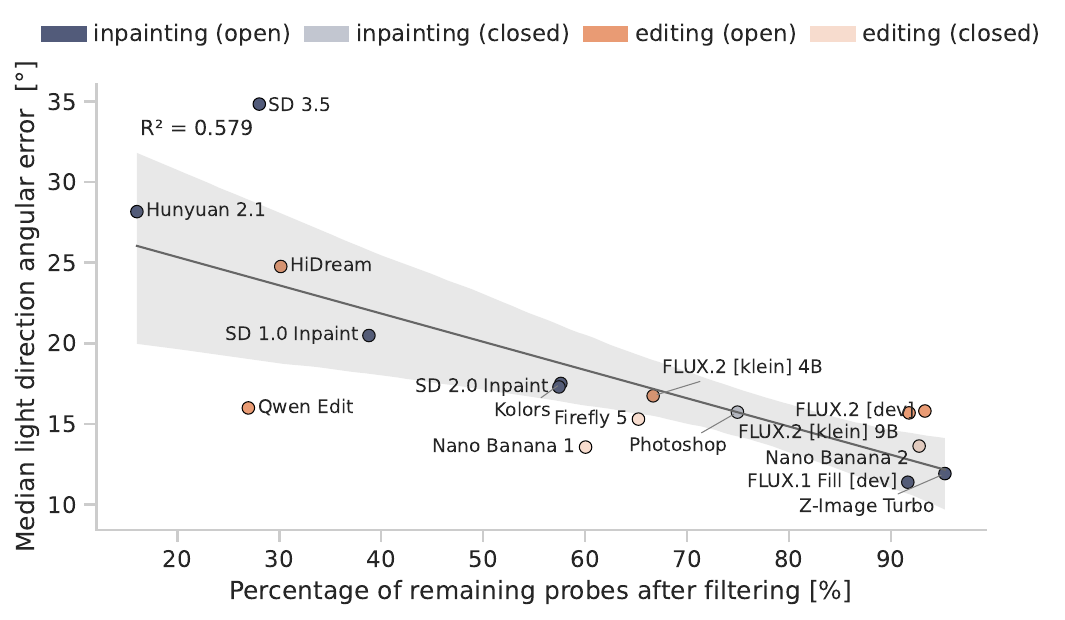}
  \caption{Median light direction angular error as a function of the percentage of quality filtered probes. A linear regression (black line) is fitted to the data, with the $R^2$ displayed at the top, with the fit uncertainty around the line. Models are coloured by type (inpainting or editing) and availability (open or closed), as in \cref{fig:angular_error_direction_lumiere}.}
  \Description{Survivorship bias of the median light direction angular error.}
  \label{fig:discussion_survivorship_bias}
\end{figure}

Correctly generated diffuse spheres are required to obtain robust lighting estimates, but filtering the probes (c.f. \cref{subsec:methodo_detection}) may induce a bias. \Cref{fig:seed_comparison} (see supp. mat.) provides insights into this bias by comparing the same model and prompt across seeds, which results in varying survival rates. We further observe in \cref{fig:number_data_per_model_per_step} (supp. mat.) that lower rates tend to accompany higher median angular error models, including \hunyuan and \SDthree. Therefore, the bias works against these models as they are penalised for producing fewer usable probes, since their surviving probes tend to be worse than those of other models.  Note that this value takes into account the probes that have been filtered because they were of too low quality (see \cref{fig:exemples_badprobes} in the supp. mat.) and the images where the model did not generate a probe.

\Cref{fig:discussion_survivorship_bias} illustrates this relationship more directly by plotting the light direction median angular error for each model as a function of the percentage of generated probes that survive the quality filtering.  The slope of \cref{fig:discussion_survivorship_bias} indicates that the models with the lowest number of valid probes have higher median angular errors.
We therefore observe a bias: models that yield a low number of valid probes tend to have lower accuracy.  This indicates that these models struggle to inpaint a probe (rightmost category in \cref{fig:number_data_per_model_per_step}), and when they succeed, many are low-quality and filtered out.  These models are left with a small yield of probes of lower quality than those from models that perform well on the task.

\paragraph{Limitations}
Our benchmark assumes a single dominant light source, which, along with the ambient term, accounts for about 87\% of the energy on a Lambertian sphere \cite{ramamoorthi2001efficient}. This is a reasonable estimate, but it may overlook energy contributions from more complex illumination conditions.
The Multi-Illumination dataset varies lighting by rotating a flash, which is useful but incomplete: it excludes backlighting, is limited to indoor scenes, and provides a single dominant light direction. We therefore cannot draw conclusions about the capacity of generative models to understand light in outdoor scenes, nor in the presence of multiple lighting sources.
While we tested for the models' capacity to model spatially-varying lighting (\cref{subsubsec:evaluation_LD_spatial_variation}), we only did so at two locations in the scene. A more detailed, 3D analysis is left as future work.
No single prompt is optimal across models; despite our efforts to ensure robustness (\cref{suppsubsubsec:evaluation_probegen_prompt} in the supp. mat.), per-model tuning could improve results.
Additionally, the calibration object used for this analysis was a simple grey sphere, thus it is not possible to conclude on the quality of the lighting produced by these models for objects with more complex geometry and materials.  However, we believe that using a simple sphere as a proxy for more complex objects is a good indication that the model shows understanding of the lighting, especially that the models have demonstrated coherency with the ground-truth lighting (\cref{subsec:evaluation_LD}) and spatial variation (\cref{subsubsec:evaluation_LD_spatial_variation}).

\section{Conclusion}
\label{sec:conc}
This paper introduces a physically grounded benchmark for quantitatively evaluating the accuracy of lighting in image-editing models. By leveraging a dataset containing calibrated light probes in various indoor scenes and exploiting the inpainting capabilities of modern generative models, our benchmark isolates lighting inference from confounding factors such as geometry, material properties, and scene semantics, enabling a direct and principled assessment of photometric consistency.

Our benchmark computes three complementary measures: angular error in light direction, $\Delta\textrm{E}_{ab}$ colour error, and divergence between radiance distributions. Our results reveal several key findings: lighting accuracy has shown little improvement over time despite rapid gains in overall visual quality; closed models do not exhibit a consistent advantage over open ones; explicit image-editing models do not outperform general inpainting models; and model scale alone is not a reliable predictor of accuracy.  Additionally, we study how well models can recreate spatially varying lighting by matching the generated probes to the ground truth at two locations in the images, which shows that most models are not simply doing local harmonisation.

Interestingly, when contrasting the results of our benchmark with findings from the human perception literature, we observe that most predicted errors fall below typical perceptual detection thresholds and are therefore likely to go unnoticed by human observers. This suggests that current generative models prioritise perceptual plausibility over physical correctness, implicitly learning to produce lighting that appears realistic without explicitly reasoning about the underlying physics. In this sense, high-quality image synthesis appears sufficient to achieve visually convincing lighting, even in the absence of physically accurate light estimation. While such perceptual realism may be appropriate for images produced for human consumption, it may be insufficient for other applications where physical accuracy is key. Examples are the generation of training data for autonomous vehicles or other physically grounded simulations.

We hope this benchmark paves the way for a deeper understanding of how generative models represent and reproduce illumination, serving as a diagnostic tool for future research and encouraging further work toward synthesising physically accurate lighting conditions.

\paragraph{Acknowledgements}

This research was supported by NSERC grant RGPIN 2020-04799 and an NSERC PhD scholarship to JG. JVC was supported by Grants PID2024-162555OB-I00, AIA2025-163919-C52 funded by MCIN/AEI/10.13039/
501100011033 and by ERDF “A way of making Europe”, the Generalitat de Catalunya CERCA Program, and the 2025 Leonardo Grant for Scientific Research and Cultural Creation from the BBVA Foundation. The BBVA Foundation accepts no responsibility for the opinions, statements and contents included in the project and/or the results thereof, which are entirely the responsibility of the authors. Compute support was provided by Digital Research Alliance of Canada (RRG 5299).

\bibliographystyle{ACM-Reference-Format}
\bibliography{reference}

\medskip

\clearpage
\newpage
\appendix

\clearpage
\setcounter{page}{1}

\section{Implementation details}
\label{suppsec:methodo}

\subsection{Dataset: filtering and processing}
\label{suppsubsec:methodo_dataset}

The Multi-Illumination dataset does not include ground-truth masks for each probe. We generated the matte and mirror probe masks with SAM3~\citep{carion2025sam3segmentconcepts} and additionally smoothed them by fitting an ellipse.

\begin{figure}[h]
\centering
\setlength\tabcolsep{0.5pt}
\renewcommand{\arraystretch}{0.2}
\def\colW{0.194}
\begin{tabular}{ccccc}
    \includegraphics[width=\colW\linewidth]{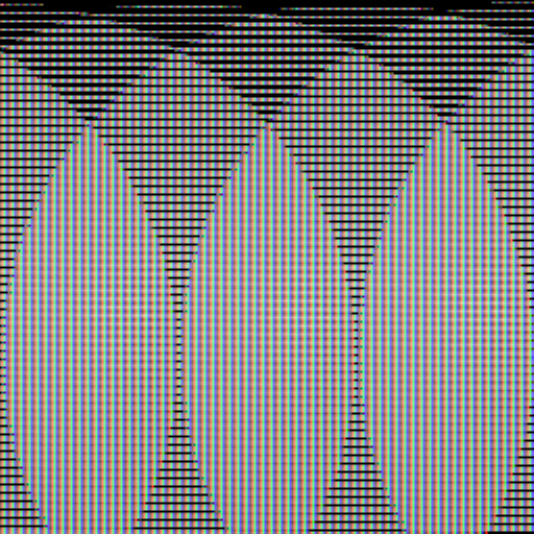} &   \includegraphics[width=\colW\linewidth]{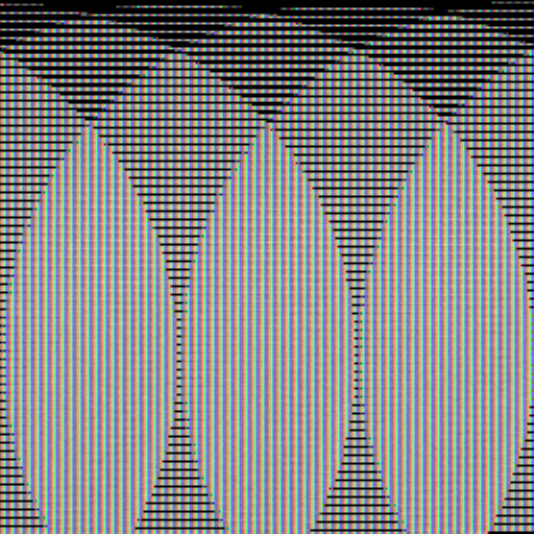} & \includegraphics[width=\colW\linewidth]{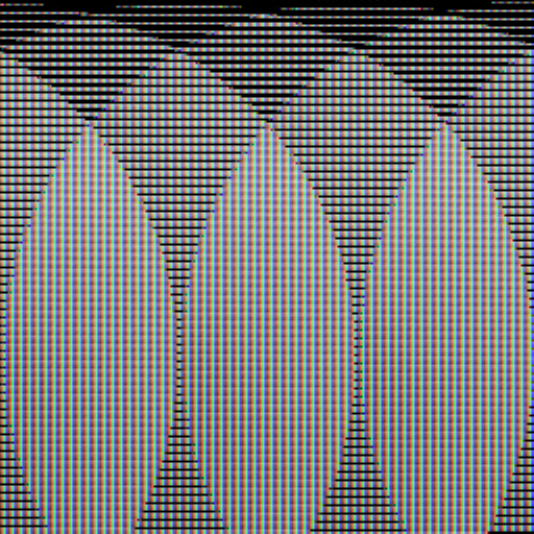} &
    \includegraphics[width=\colW\linewidth]{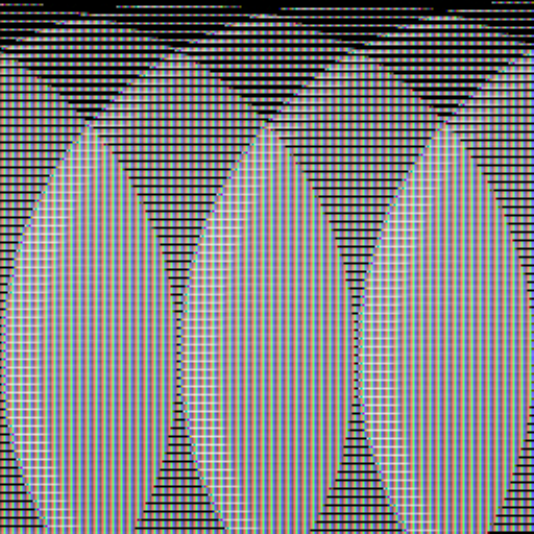} &
    \includegraphics[width=\colW\linewidth]{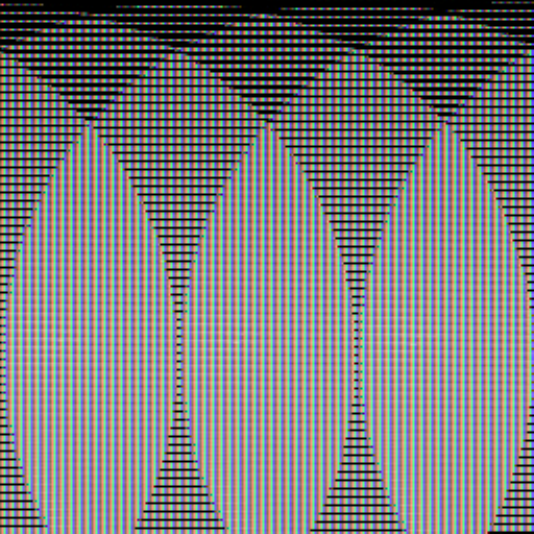} \\
\end{tabular}
\caption{Examples of probes we removed from the Multi-Illumination dataset. \textit{From left to right}: Probe cropped, too noisy, probe stand in view, background in image, probe partially occluded from the light source by an object in scene.}
\label{fig:exemples_badprobes}
\Description{Examples of bad diffuse probes in the Multi-Illumination dataset.}
\end{figure}

We removed \num{2052} images from the dataset that were not suitable for our purposes. Several light probes were illuminated by multiple light sources due to the light directly from the flash or from mirror-reflective objects in the scene reflecting the flash at the probe, in addition to the intended light source from the flash aimed at a point in the scene. A few images in the dataset were overexposed, and their probes were captured with significant noise. Some probes had a significant portion of the background scene within the cropped image provided. Some probes were occluded from the light source by objects in the scene. Finally, some scenes lacked the correct material masks to identify the probes. Examples are shown in \cref{fig:exemples_badprobes}.

We rectify the spheres to be circular by fitting an ellipse to the contours of the SAM3~\cite{carion2025sam3segmentconcepts} mask. We then apply a perspective transform to warp the sphere image so that the fitted ellipse is circular, retaining the spheres' normals.

\subsection{Inpainting: model inference}
\label{suppsubsec:methodo_inpainting}

\begin{figure*}[t]
  \centering
  \includegraphics[width=\linewidth]{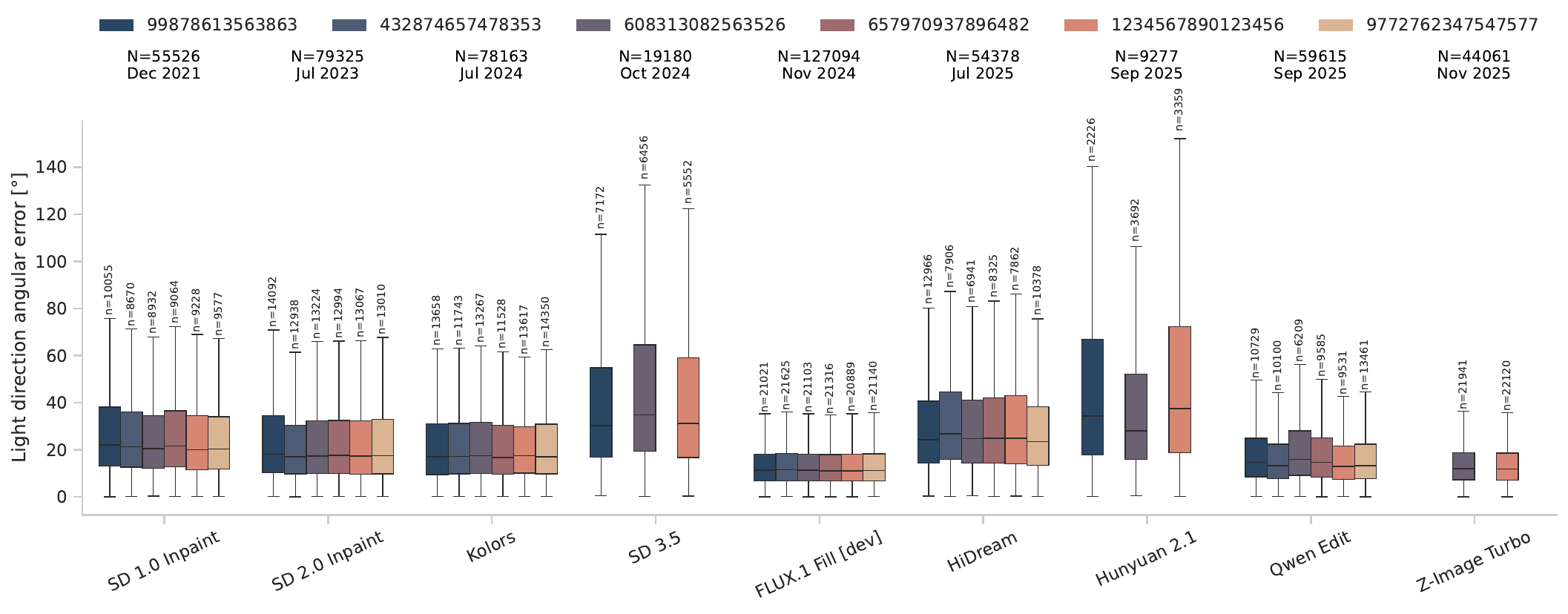}
  \caption{Comparison of \num{6} seeds on \num{10} different models on the whole dataset. For most models, there is no significant difference in which seed is chosen.}
  \Description{Light direction angular error for different seeds.}
  \label{fig:seed_comparison}
\end{figure*}

For each model, we used the default workflows provided by ComfyUI~\cite{comfy}. For open models, we select seed \num{608313082563526}, and show that there is no significant differences between the seeds in \cref{fig:seed_comparison}. The seeds for closed models are listed in \cref{tab:model_params}, and were unable test multiple seeds on these API models.

To ensure we had the correct hyperparameters for the open models, we tested different classifier-free guidance scales on a subset of the dataset \cref{tab:cfg_comparison}. Although \zimageturbo has a lower median angular error with the non-default CFG scale, it is marginal and results in a lower usable probe yield. \fluxone performs marginally better at \num{3} than the default. The remaining models all perform best at their recommended CFG scale. Note that the CFG scale is not consistent across models; some prefer a smaller value (\zimageturbo, \qwen, \fluxone), others prefer a larger value (\SDtwo, \kolors).

We report GPU usage for each model at each pipeline stage in \cref{tab:gpu_hour_breakdown}. The probe detection (\cref{subsec:methodo_detection}) uses a 2g.20gb and the light parameters estimation (\cref{subsec:methodo_LD}) uses a 1g.10gb. Probe analysis is CPU-only and not included in GPU totals.  The total GPU hours for the entire dataset is approximately \num{2395.8} H100-equivalent hours, which corresponds to about \num{100} H100-equivalent days.  These estimates are based on timing measurements from the actual pipeline runs and may vary in practice due to factors like GPU availability, model performance variability, and potential optimisations. For API models, we did not report usage for the inpainting stage. Model checkpoints, hyperparameters, release date and other attributes are listed in \cref{tab:model_params}.

\begin{table}[h]
\caption{Usable probe yield (\% of GT-usable directions) and median angular error per model for varying CFG value. For each model, the default classifier-free guidance (CFG) scale is underlined, and the best median error and yield are highlighted in bold.}
\label{tab:cfg_comparison}
\begin{tabularx}{\linewidth}{lrrr}
\toprule
Model & CFG & Yield [\%]$\uparrow$ & Median Ang Err [\textdegree]$\downarrow$ \\
\midrule
\SDone & 4 & 43.3 & 18.3 \\
\cite{rombach2022high} & \underline{8} & \textbf{47.9} & \textbf{17.4} \\
 & 12 & 47.6 & 18.0 \\
\midrule
\fluxone & \underline{1} & 88.5 & 12.2 \\
\cite{blackforestlabs2024flux1fill} & 3 & \textbf{94.5} & \textbf{11.9} \\
 & 7 & 77.9 & 12.8 \\
\midrule
\qwen & \underline{1} & \textbf{34.3} & \textbf{15.5} \\
\cite{wu2025qwenimagetechnicalreport} & 3 & 27.4 & 22.3 \\
 & 7 & 23.8 & 35.7 \\
\midrule
\zimageturbo & \underline{1} & \textbf{98.9} & 11.8 \\
\cite{imageteam2025zimageefficientimagegeneration} & 3 & 97.0 & \textbf{11.4} \\
& 7 & 6.8 & 49.4 \\
\midrule
\kolors & 1 & 47.0 & 23.2 \\
\cite{kolors} & \underline{3} & \textbf{61.2} & \textbf{19.2} \\
 & 7 & 38.3 & 28.0 \\
\bottomrule
\end{tabularx}
\end{table}

\begin{table}[h]
\vspace{-2mm}
\caption{Estimated GPU hours for dataset generation, by stage, per model. GPU costs are normalised to H100-equivalent hours based on the GPU used per model (`Inpaint GPU`) and the H100 MIG fractions. The last row is the sum of all the models.}
\label{tab:gpu_hour_breakdown}
\centering
\vspace{-2mm}
\scriptsize
\setlength{\tabcolsep}{2pt}
\begin{tabularx}{\linewidth}{lrrr>{\raggedleft\arraybackslash}X >{\raggedleft\arraybackslash}X >{\raggedleft\arraybackslash}X >{\raggedleft\arraybackslash}X}
\toprule
Model & Seeds & Inpainted & Probes & Inpaint GPU & Inpaint H100-eq [h] & Detection H100-eq [h] & Light Estimation H100-eq [h] \\
\midrule
\SDone~\cite{rombach2022high} & 6 & 152100 & 65914 & 2g.20gb & 97.8 & 17.3 & 1.1 \\
\SDtwo~\cite{rombach2022high} & 6 & 152100 & 89975 & 1g.10gb & 38.7 & 26.3 & 1.5 \\
\kolors~\cite{kolors} & 6 & 152100 & 109927 & 3g.40gb & 129.6 & 18.0 & 1.8 \\
\SDthree~\cite{pmlr-v235-esser24a} & 3 & 76050 & 29680 & 3g.40gb & 170.9 & 8.2 & 0.5 \\
\fluxone~\cite{blackforestlabs2024flux1fill} & 6 & 152100 & 142379 & 3g.40gb & 201.2 & 18.2 & 2.4 \\
\hidream~\cite{HiDream_2025_MM} & 6 & 152100 & 76000 & h100 & 910.1 & 26.6 & 1.3 \\
\NBone~\cite{geminiteam2025geminifamilyhighlycapable} & 1 & 2708 & 1775 & API & 0.0 & 0.4 & 0.0 \\
\qwen~\cite{wu2025qwenimagetechnicalreport} & 7 & 153475 & 117372 & 3g.40gb & 119.3 & 37.4 & 2.0 \\
\hunyuan~\cite{HunyuanImage-2.1} & 3 & 76050 & 12055 & 3g.40gb & 61.9 & 5.3 & 0.2 \\
\firefly~\cite{firefly} & 1 & 1484 & 1076 & API & 0.0 & 0.2 & 0.0 \\
\fluxtwo~\cite{flux-2-2025} & 1 & 25350 & 24676 & h100 & 207.2 & 2.4 & 0.4 \\
\zimageturbo~\cite{imageteam2025zimageefficientimagegeneration} & 2 & 50700 & 48962 & 3g.40gb & 98.3 & 13.1 & 0.8 \\
\fluxtwofour~\cite{flux-2-2025} & 1 & 25350 & 18905 & 3g.40gb & 60.1 & 2.9 & 0.3 \\
\fluxtwonine~\cite{flux-2-2025} & 1 & 25350 & 25165 & 3g.40gb & 104.0 & 2.9 & 0.4 \\
\photoshop~\cite{firefly} & 1 & 25350 & 21633 & API & 0.0 & 2.8 & 0.4 \\
\NBtwo~\cite{geminiteam2025geminifamilyhighlycapable} & 1 & 10538 & 9805 & API & 0.0 & 1.4 & 0.2 \\
Total &  & 1232905 & 795299 &  & 2199.1 & 183.4 & 13.3 \\
\bottomrule
\end{tabularx}
\end{table}

\begin{table*}
\centering
\scriptsize
\setlength{\tabcolsep}{2pt}
\caption{Model parameters and details for the models evaluated in this work. The ELO scores are from \url{https://arena.ai/leaderboard/image-edit}.}
\label{tab:model_params}
\begin{tabularx}{\textwidth}{>{\raggedright\arraybackslash}X r r >{\raggedleft\arraybackslash}X r r r r r r r r r}
\toprule
Model & Abbreviation & Release Date & Checkpoint & Architecture & No. Parameters & Precision & Base Resolution & No. Steps & CFG Scale & ELO & Category & Availability \\
\midrule
Stable Diffusion 2 Inpainting~\cite{rombach2022high} & SD 2 Inpaint & 2023-09 & 512-inpainting-ema .safetensors & U-Net & 1303623630 & fp32 & 512 & 20 & 8 & - & Inpainting & Open \\
\hline
SDXL 1.0 Inpainting 0.1~\cite{rombach2022high} & SD 1.0 Inpaint & 2021-12 & sd\_xl\_base\_ 1.0\_inpainting\_ 0.1.safetensors & U-Net & 3468852345 & fp16 & 1024 & 20 & 7 & - & Inpainting & Open \\
\hline
Kolors Inpainting~\cite{kolors} & Kolors & 2024-07 & kolors-inpainting .safetensors & U-Net & 2579473220 & fp16 & 1024 & 20 & 3 & - & Inpainting & Open \\
\hline
Stable Diffusion 3.5~\cite{pmlr-v235-esser24a} & SD 3.5 & 2024-10 & sd3.5\_large\_fp8\_scaled .safetensors & mm-DiT & 8134104590 & fp8 & 1024 & 32 & 8 & - & Inpainting & Open \\
\hline
FLUX.1 Fill [dev]~\cite{flux2024} & FLUX.1 Fill [dev] & 2024-11 & flux1-fill-dev.safetensors & mm-DiT & 11902391360 & bf16 & 1024 & 20 & 1 & - & Inpainting & Open \\
\hline
HiDream-E1.1~\cite{HiDream_2025_MM} & HiDream & 2025-07 & hidream\_e1\_ 1\_bf16 .safetensors & SDiT & 17105733184 & bf16 & 1024 & 20 & 3 & - & Editing & Open \\
\hline
Gemini 2.5 Flash Image Preview~\cite{geminiteam2025geminifamilyhighlycapable} &  Nano Banana 1 & 2025-08 & - & MoE + Diffusion & - & - & 1024 & - & - & 1300 & Editing & Closed \\
\hline
HunyuanImage 2.1~\cite{HunyuanImage-2.1} & Hunyuan & 2025-09 & hunyuan image2.1\_ fp8\_e4m3fn.safetensors & DiT & 17425795520 & fp8 & 2048 & 20 & 3.5 & - & Inpainting & Open \\
\hline
Qwen-Image-Edit-2509~\cite{wu2025qwenimagetechnicalreport} & Qwen Edit & 2025-09 & qwen\_image\_ edit\_2509\_ fp8\_e4m3fn.safetensors & mm-DiT & 20430401088 & fp8 & 1024 & 4 & 1 & 1234 & Editing & Open \\
\hline
Adobe Firefly Image Model 5~\cite{firefly} &  Firefly 5 & 2025-10 & - & \texttt{-} & \texttt{-} & bf16 & 2240 & - & - & - & Editing & Closed \\
\hline
FLUX.2 [dev]~\cite{flux-2-2025} & FLUX.2 [dev] & 2025-11 & flux2\_dev\_ fp8mixed .safetensors & mm-DiT & 32223281408 & fp8 & 1024 & 20 & 5 & 1225 & Inpainting & Open \\
\hline
Z-Image-Turbo~\cite{imageteam2025zimageefficientimagegeneration} & Z-Image-Turbo & 2025-11 & z\_image\_ turbo\_bf16 .safetensors & S3-DiT & 6154908736 & bf16 & 1024 & 20 & 1 & - & Editing & Open \\
\hline
FLUX.2 [klein] 4B Base~\cite{flux-2-2025} & FLUX.2 [klein] 4B & 2026-01 & flux-2-klein-base-4b-fp8 .safetensors & mm-DiT & 3875544732 & fp8 & 1024 & 20 & 5 & 1188 & Inpainting & Open \\
\hline
FLUX.2 [klein] 9B Base~\cite{flux-2-2025} & FLUX.2 [klein] 9B & 2026-01 & flux-2-klein-base-9b-fp8 .safetensors & mm-DiT & 9078581464 & fp8 & 1024 & 20 & 5 & 1224 & Inpainting & Open \\
\hline
Photoshop Generative Fill (Firefly Fill \& Expand)~\cite{firefly} &  Photoshop & 2026-01 & - & - & - & bf16 & 2048 & - & - & - & Inpainting & Closed \\
\hline
Gemini 3.1 Flash Image Preview~\cite{geminiteam2025geminifamilyhighlycapable} &  Nano Banana 2 & 2026-02 & - & MoE + Diffusion & - & - & 1024 & - & - & 1387 & Editing & Closed \\
\bottomrule
\end{tabularx}
\end{table*}

\subsection{Probe detection: SAM3 inference and probe filtering}
\label{suppsubsec:methodo_detection}

We use SAM3~\citep{carion2025sam3segmentconcepts} to detect and segment the generated probes with the prompt:
\begin{promptbox}
    \texttt{matte sphere not mirror reflective sphere}
\end{promptbox}
\begin{figure}[t]
\centering
\setlength\tabcolsep{0.5pt}
\renewcommand{\arraystretch}{0.2}
\def\colW{0.194}
\begin{tabularx}{\linewidth}{ccccc}

    \raisebox{0pt}{\includegraphics[width=\colW\linewidth]{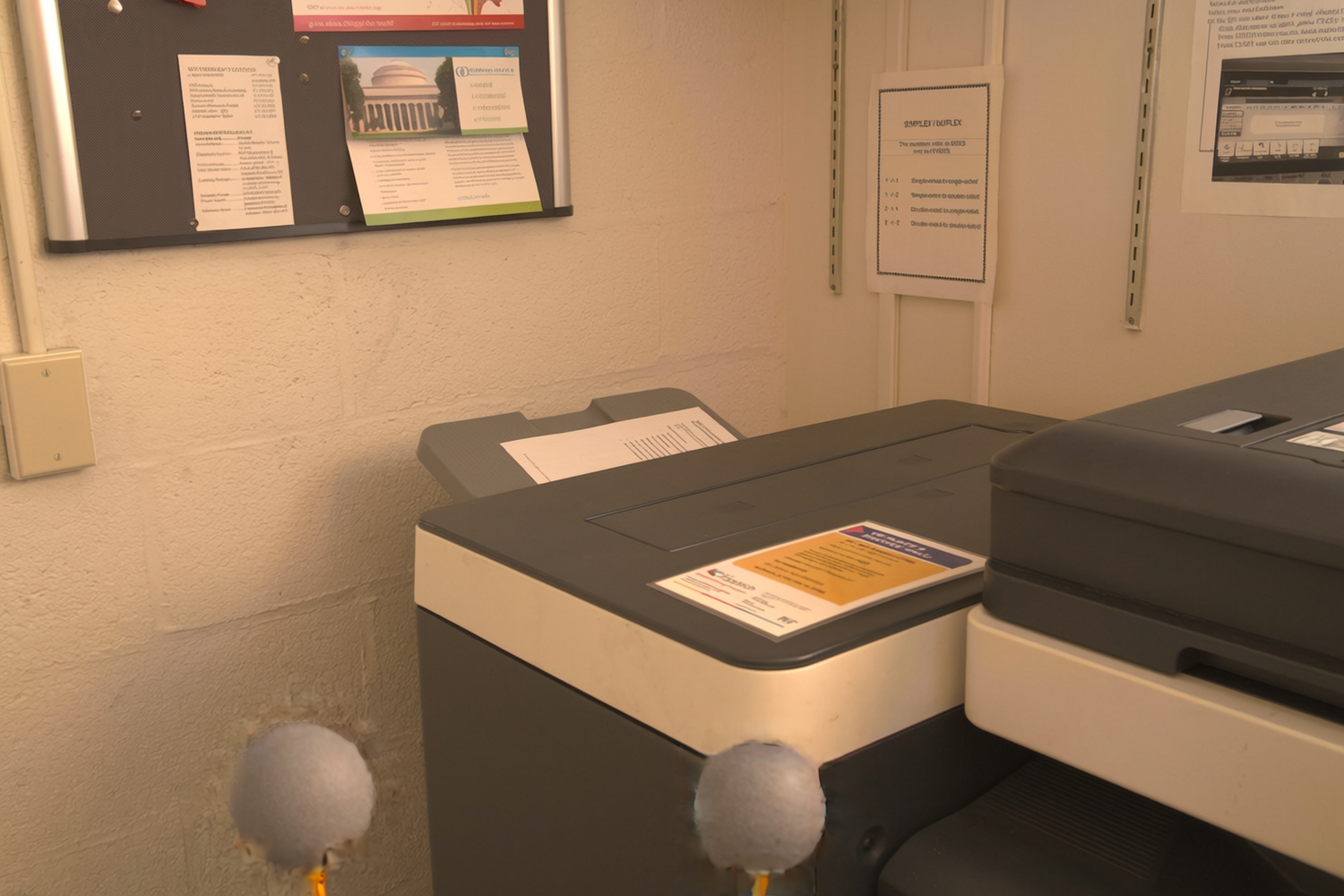}} &\raisebox{0pt}{\includegraphics[width=\colW\linewidth]{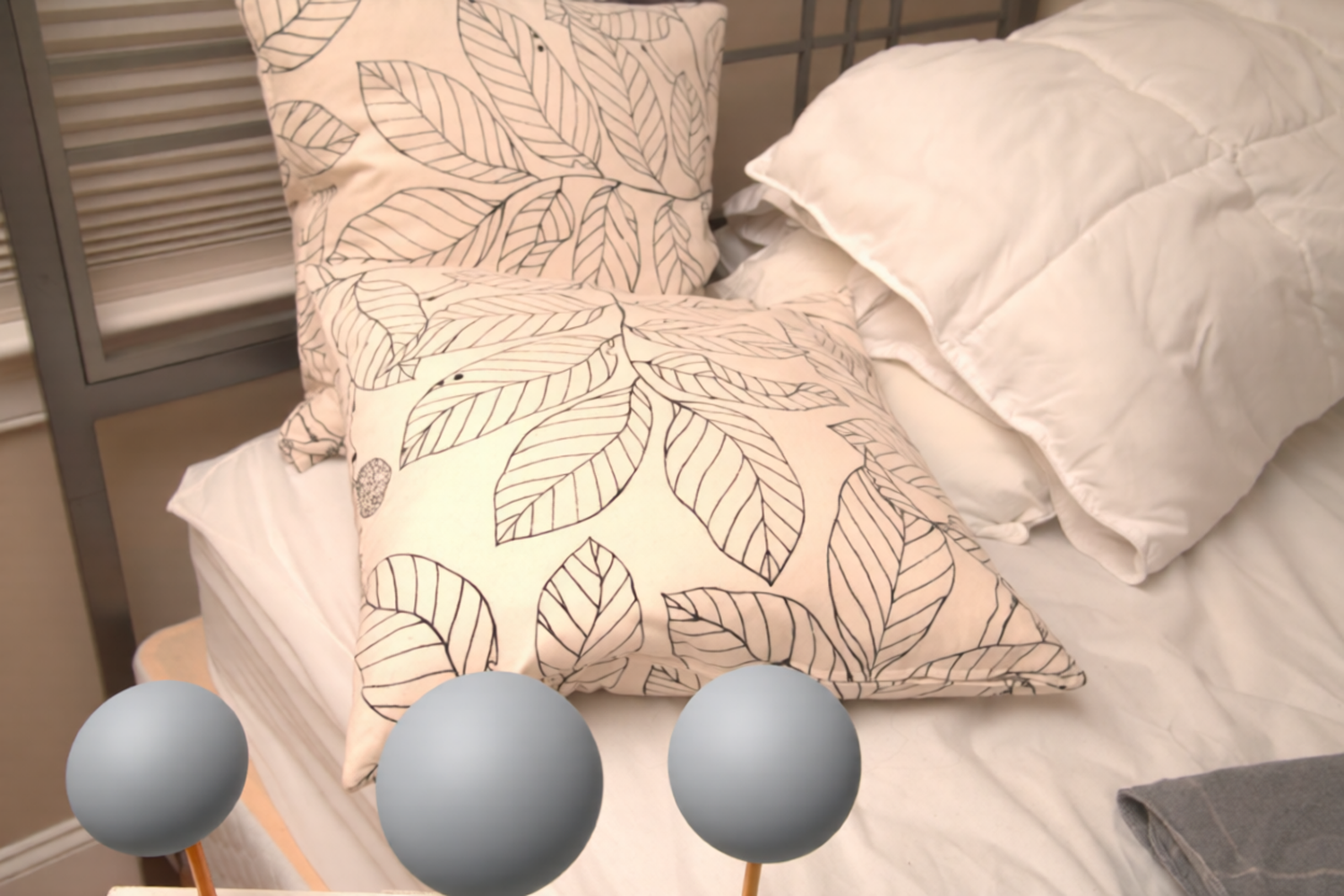}} &\raisebox{0pt}{\includegraphics[width=\colW\linewidth]{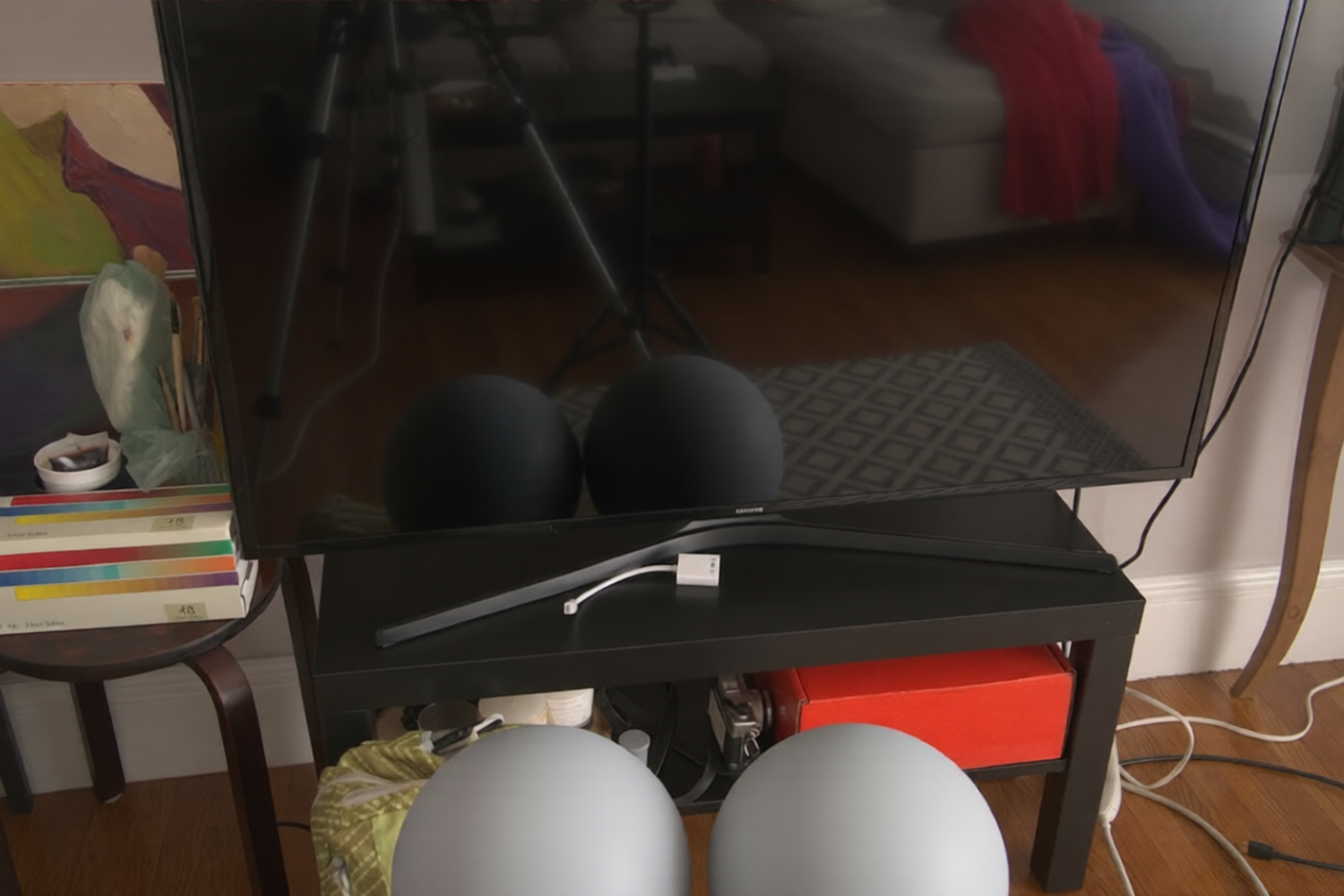}} &\raisebox{0pt}{\includegraphics[width=\colW\linewidth]{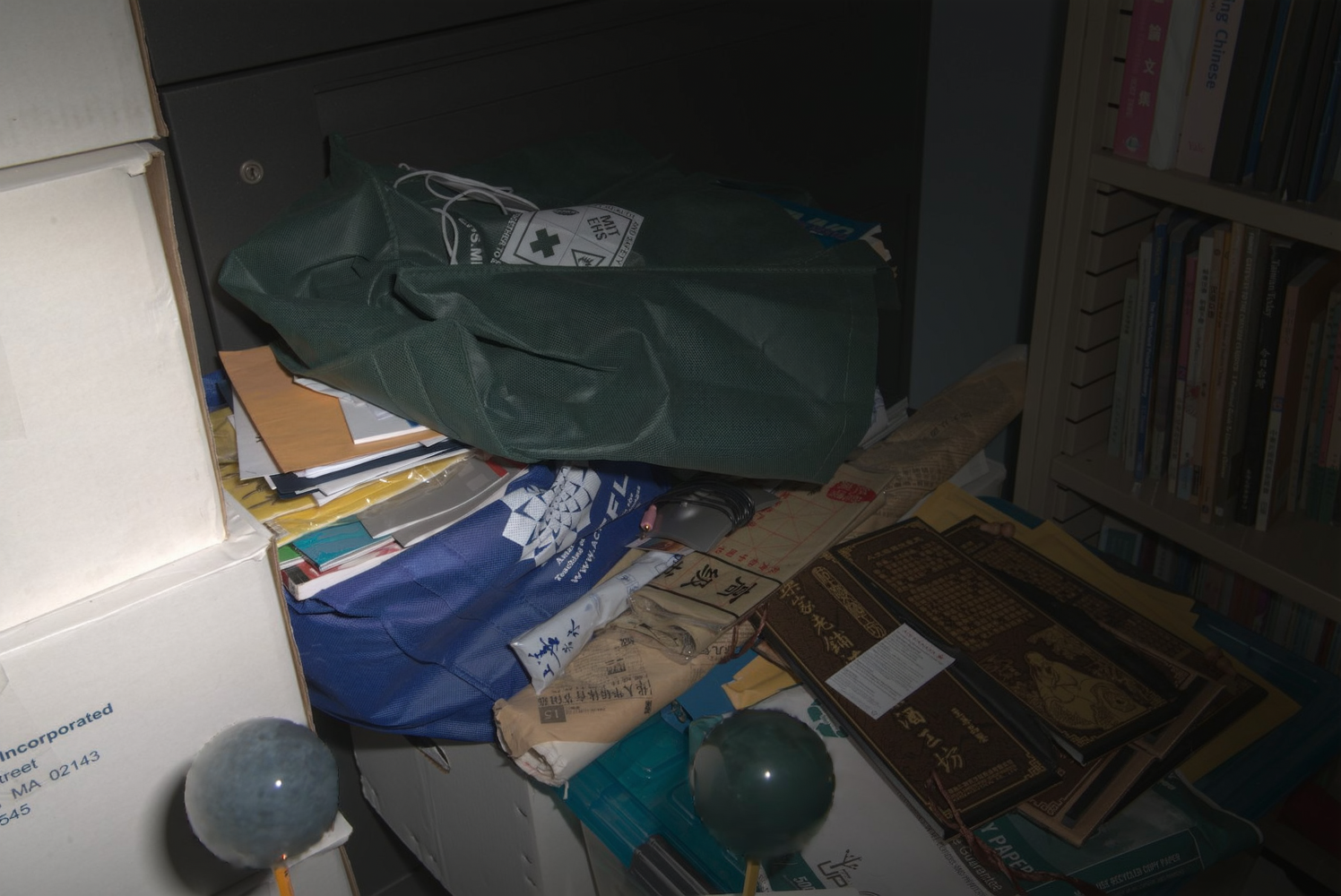}} &\raisebox{0pt}{\includegraphics[width=\colW\linewidth]{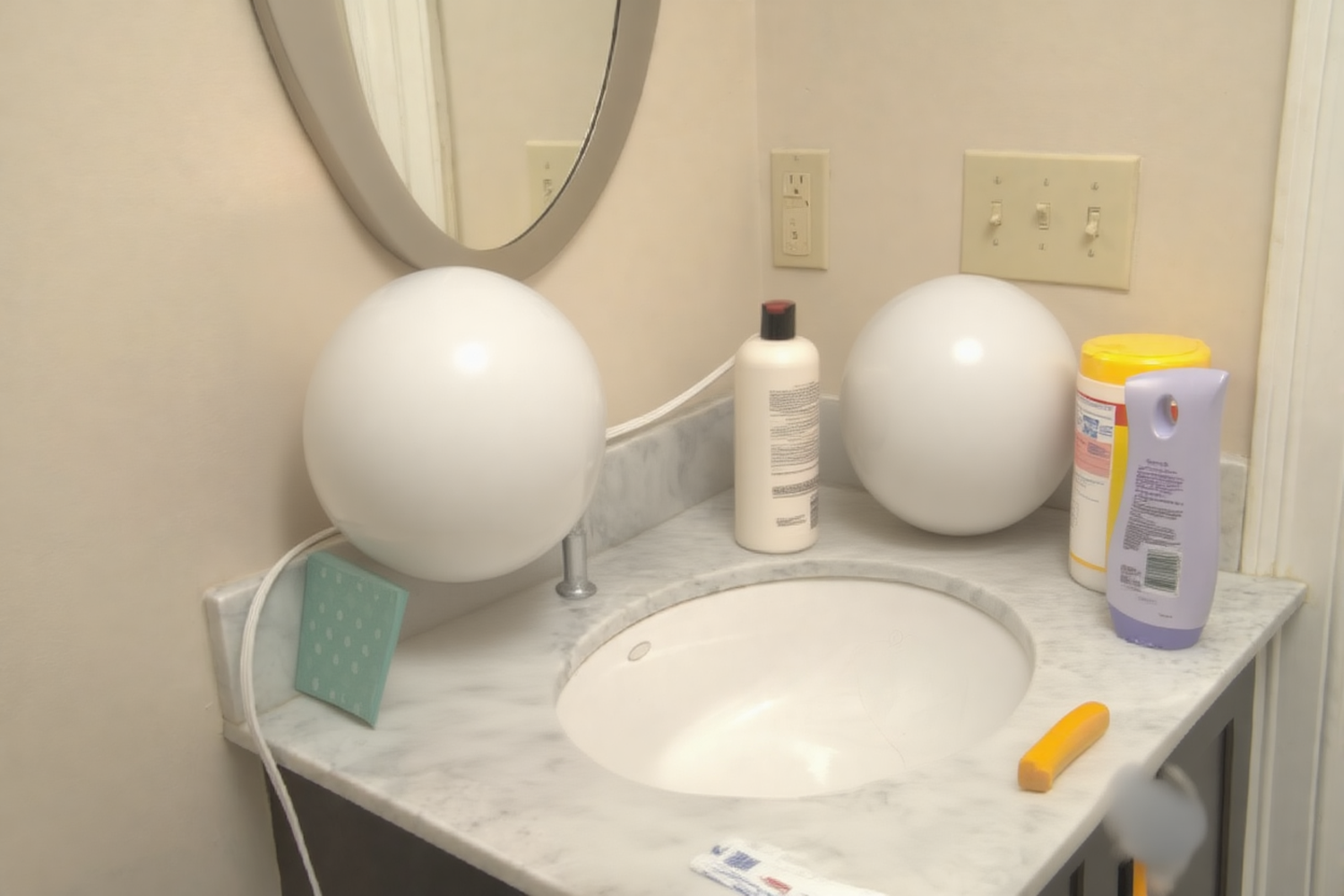}}
    \\

\end{tabularx}
\caption{Examples where the generative image models generated unusable probes.}
\Description{Examples of unusable probes generated by the models.}
\label{fig:exemples_badinpaints}
\end{figure}
Generative image models do not strictly adhere to the prompt's instructions; we present a few common examples in \cref{fig:exemples_badinpaints}. \hunyuan generated noisy probes regardless of the number of time steps used at inference. \qwen and \hidream often generated multiple probes, with \hidream struggling to place them in front of the camera and often embedding them into the scene behind objects. Some models, such as \SDthree, would try to blend the probe's texture and colour with the surrounding scene.

\subsection{Light features estimation: shading model details}
\label{suppsubsec:methodo_LD}

\begin{figure}[t]
\centering
\tiny
\setlength\tabcolsep{0.5pt}
\renewcommand{\arraystretch}{0.2}
\def\colW{0.19}
\def\colw{0.09}
\setlength{\textgap}{0.05cm}
\setlength{\imagegap}{0.08cm}
\setlength{\rowgap}{0.23cm}
\begin{tabularx}{\linewidth}{cccccccccc}

    \multicolumn{2}{c}{\raisebox{0pt}{\includegraphics[width=\colW\linewidth]{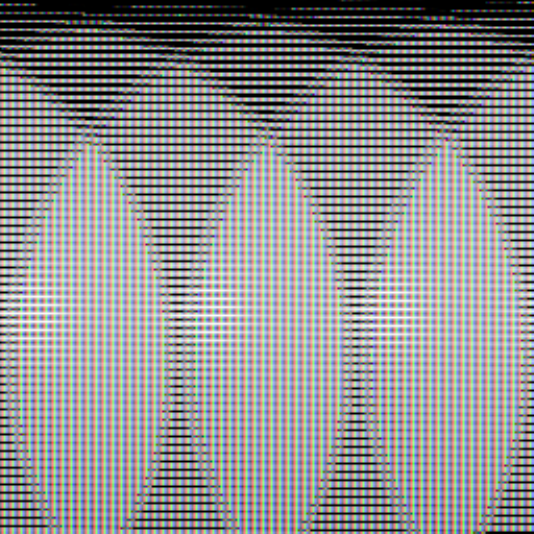}}} &&&&&&&&\\

    \raisebox{0pt}{\includegraphics[width=\colw\linewidth]{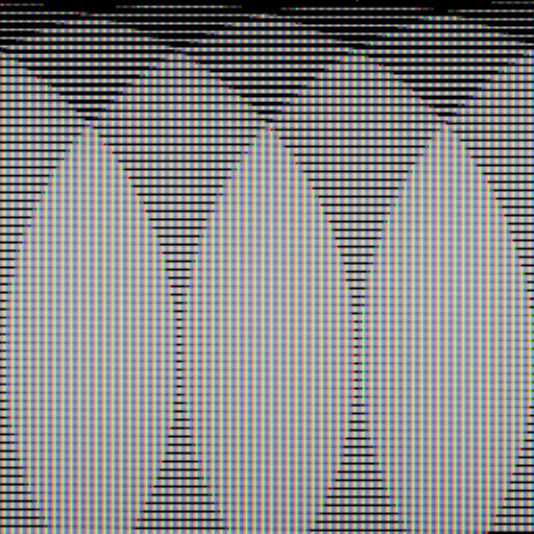}} &\raisebox{0pt}{\includegraphics[width=\colw\linewidth]{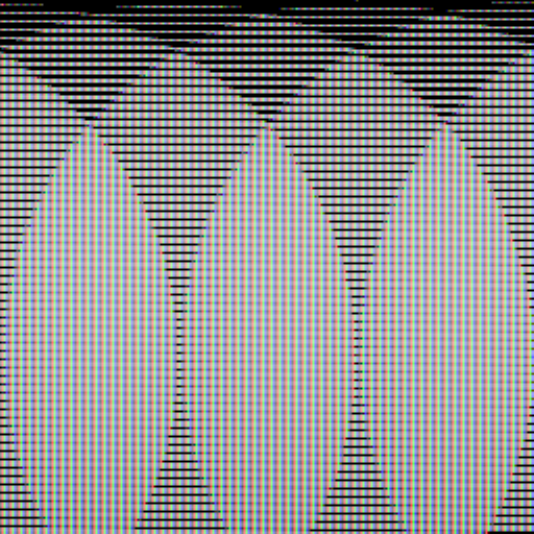}} &\raisebox{0pt}{\includegraphics[width=\colw\linewidth]{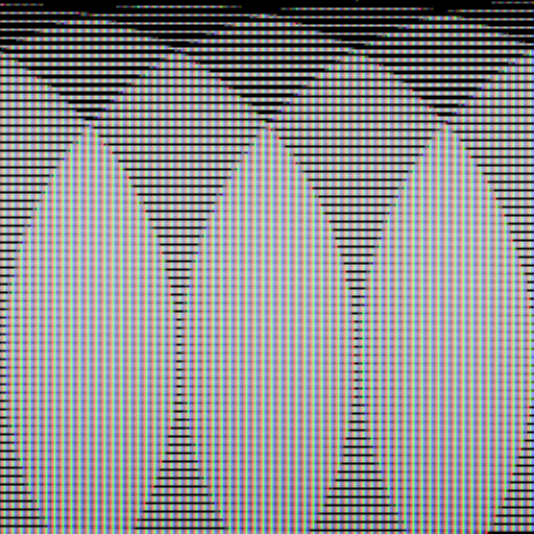}}
    &\raisebox{0pt}{\includegraphics[width=\colw\linewidth]{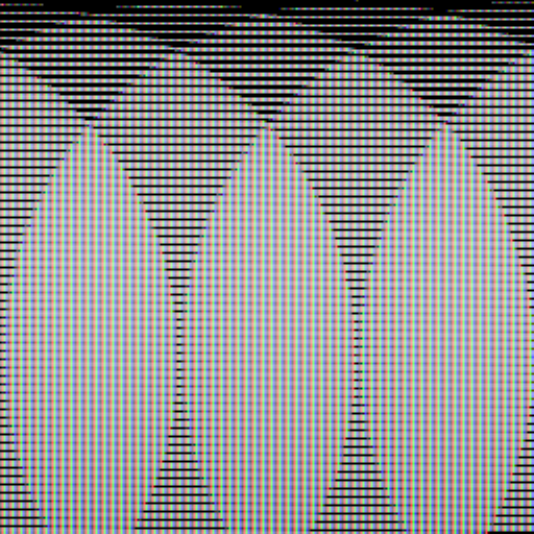}}
    &\raisebox{0pt}{\includegraphics[width=\colw\linewidth]{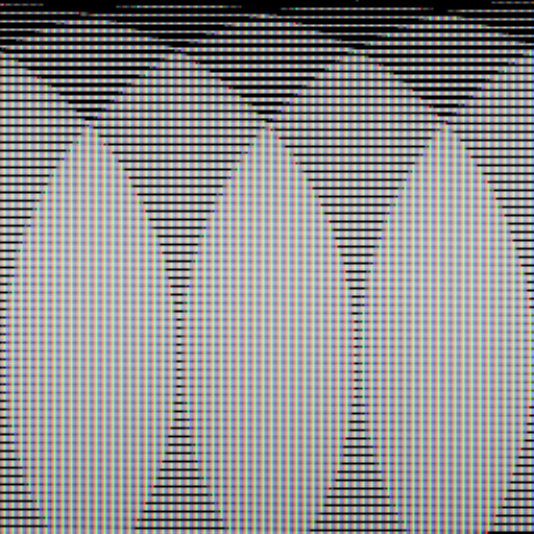}}
    &\raisebox{0pt}{\includegraphics[width=\colw\linewidth]{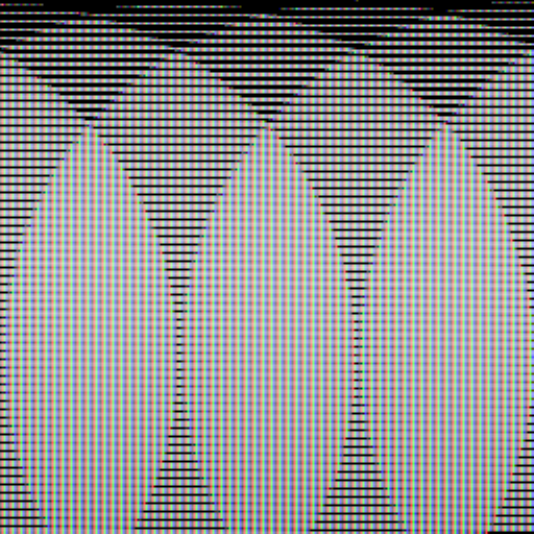}} &\raisebox{0pt}{\includegraphics[width=\colw\linewidth]{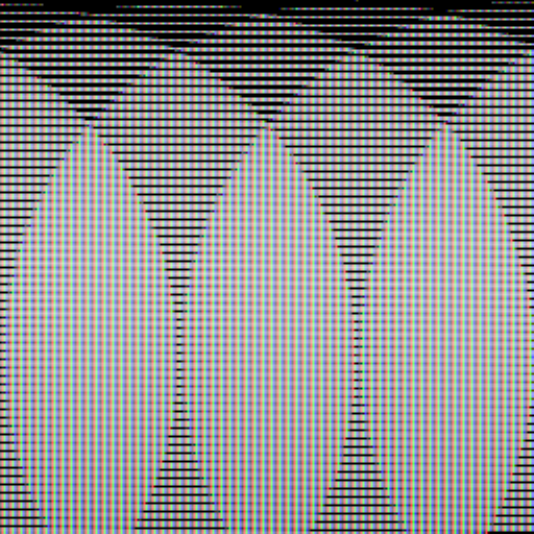}} &\raisebox{0pt}{\includegraphics[width=\colw\linewidth]{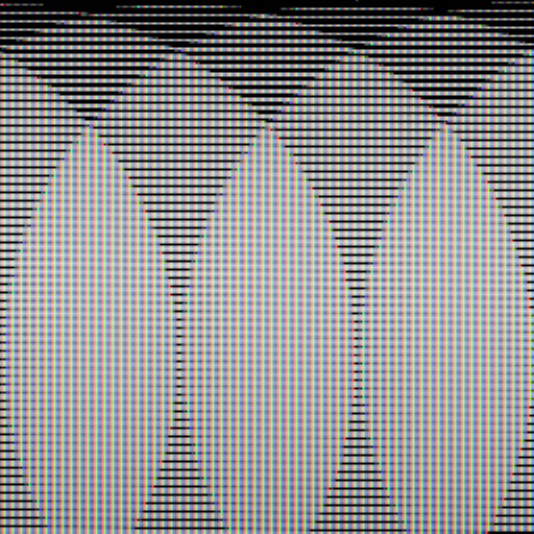}}
    &\raisebox{0pt}{\includegraphics[width=\colw\linewidth]{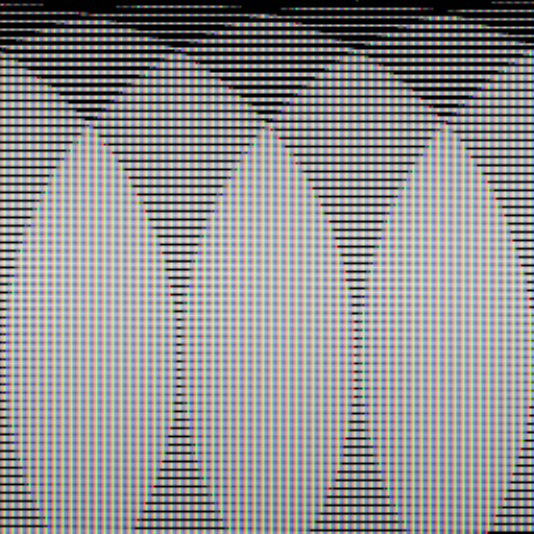}}
    &\raisebox{0pt}{\includegraphics[width=\colw\linewidth]{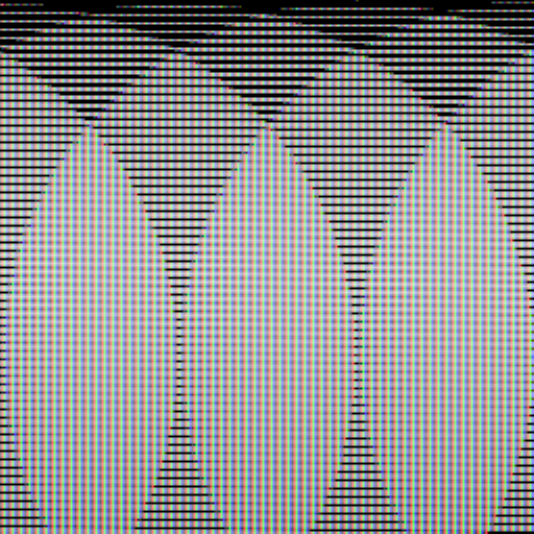}}
    \\[\rowgap]

    \multicolumn{2}{c}{\raisebox{0pt}{\includegraphics[width=\colW\linewidth]{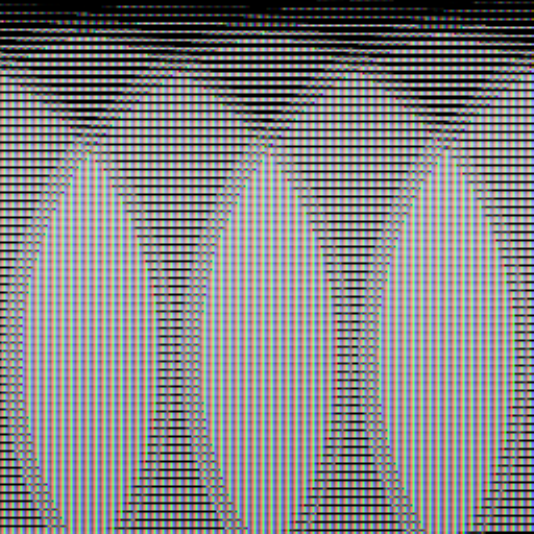}}} &&&&&&&&\\

    \raisebox{0pt}{\includegraphics[width=\colw\linewidth]{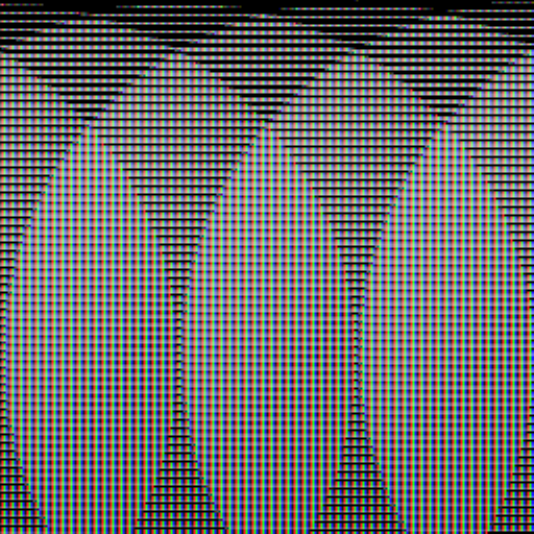}} &\raisebox{0pt}{\includegraphics[width=\colw\linewidth]{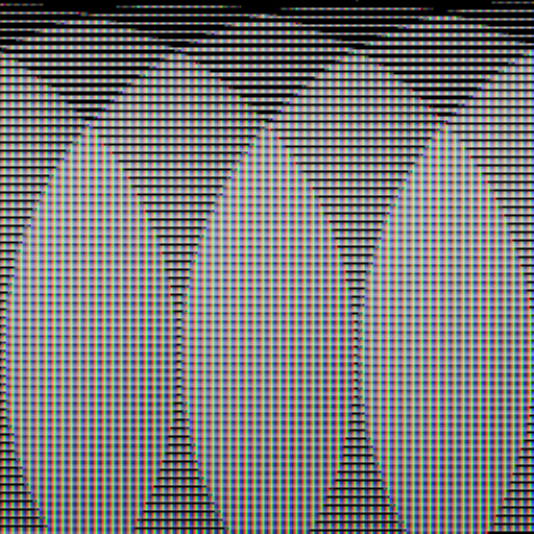}} &\raisebox{0pt}{\includegraphics[width=\colw\linewidth]{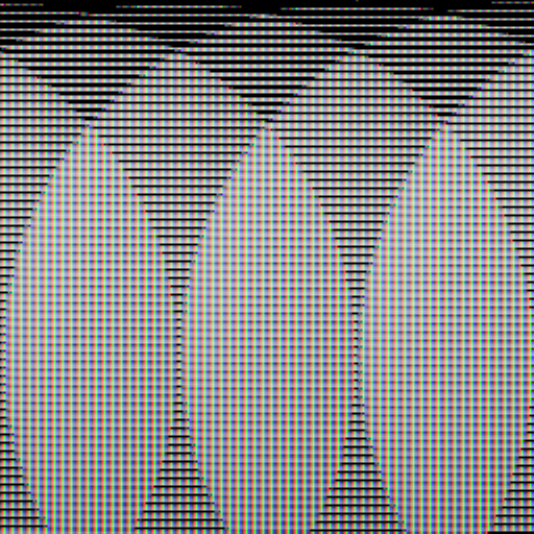}}
    &\raisebox{0pt}{\includegraphics[width=\colw\linewidth]{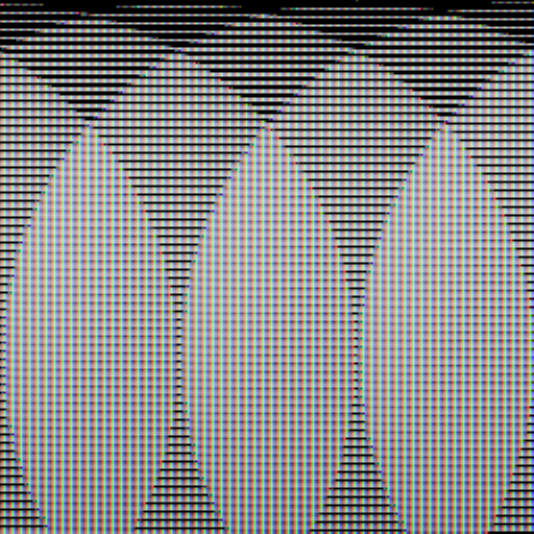}}
    &\raisebox{0pt}{\includegraphics[width=\colw\linewidth]{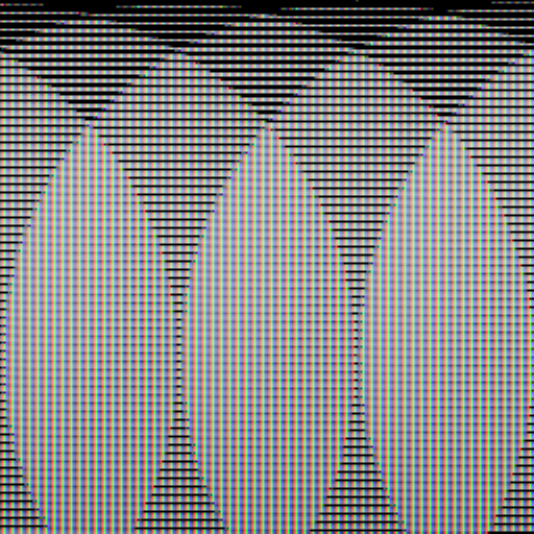}}
    &\raisebox{0pt}{\includegraphics[width=\colw\linewidth]{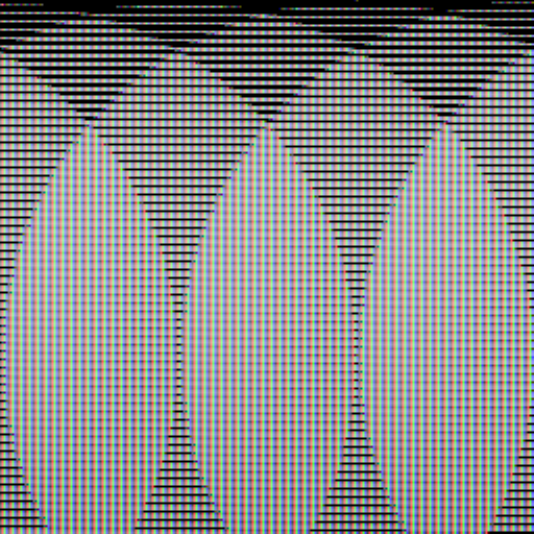}} &\raisebox{0pt}{\includegraphics[width=\colw\linewidth]{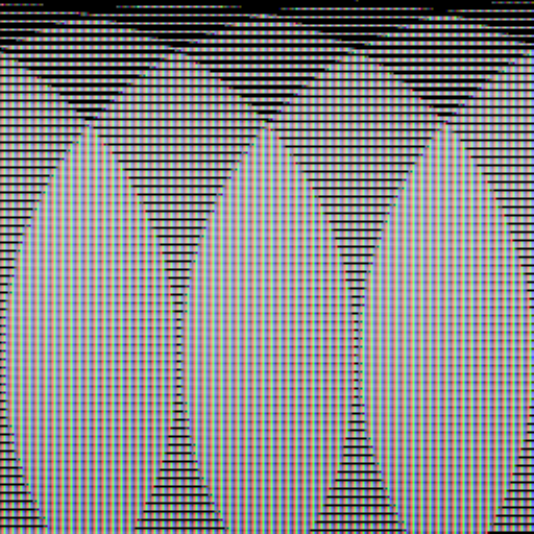}} &\raisebox{0pt}{\includegraphics[width=\colw\linewidth]{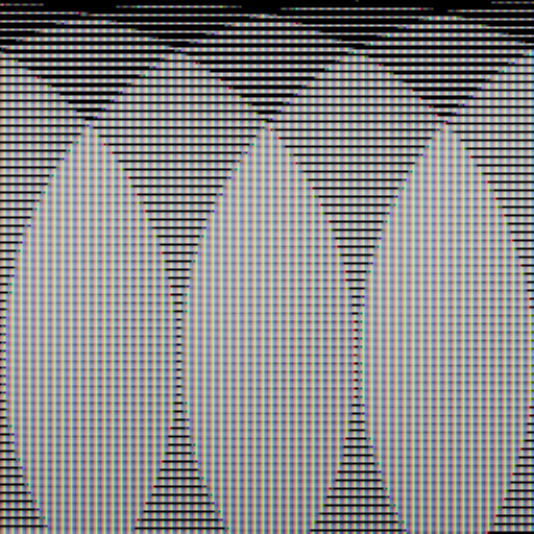}}
    &\raisebox{0pt}{\includegraphics[width=\colw\linewidth]{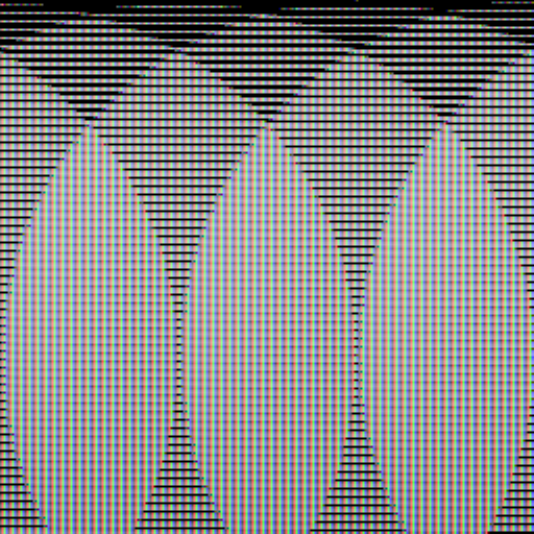}}
    &\raisebox{0pt}{\includegraphics[width=\colw\linewidth]{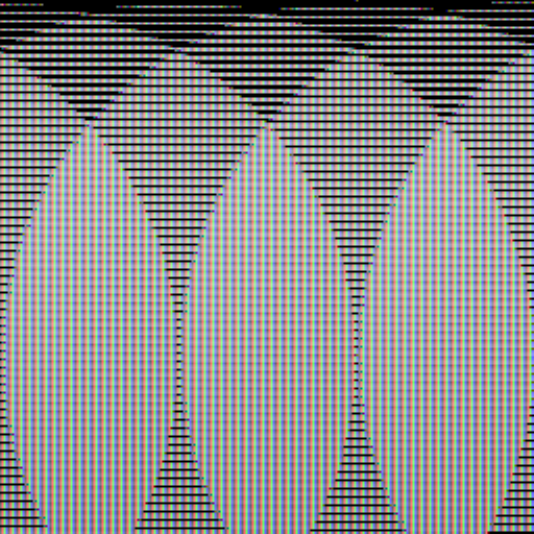}}
    \\

\end{tabularx}
\caption{Two examples of the estimated probe optimising to a target light probe from \fluxone~\cite{blackforestlabs2024flux1fill} in our inverse rendering stage. \textit{Top}: A light probe with a bright specular component. \textit{Bottom}: A fully diffuse light probe. Our light feature estimation stage can accommodate a range of material types.}
\Description{Examples of estimated probes during the inverse rendering optimisation steps.}
\label{fig:exemples_probegif}
\end{figure}

The Cook-Torrence shading model we use is composed of the following approximations. The specular component of our shading model is modelled by the microfacet BRDF, with Trowbridge-Reitz GGX approximation of the normal distribution function~\citep{walter_microfacet_2007}, Schlick's approximation of the GGX Smith geometric shadowing function~\citep{walter_microfacet_2007, Karis2013RealSI} and Schlick's approximation of the Fresnel term~\citep{schlick_brdf_1994}. The diffuse component is modelled using Lambertian lighting with an energy-conservation factor and Fresnel attenuation. The shading model uses material parameters (albedo, roughness, metallicity and Fresnel coefficient) and lighting parameters (light direction and intensity, and an ambient term). Examples of glossy and matte surface probes at different iterations of our optimisation are presented in \cref{fig:exemples_probegif}.

\begin{figure*}[t]
    \centering
    \tiny
    \setlength\tabcolsep{0.5pt}
    \renewcommand{\arraystretch}{0}
    \def\colW{0.095}
    \begin{tabular}{@{}cccccccccc@{}}
    \textbf{GT} & \textbf{SD 1.0 Inpaint} & \textbf{SD 2.0 Inpaint} & \textbf{Kolors} & \textbf{SD 3.5} & \textbf{GT} & \textbf{SD 1.0 Inpaint} & \textbf{SD 2.0 Inpaint} & \textbf{Kolors} & \textbf{SD 3.5} \\
    \includegraphics[width=0.095\textwidth]{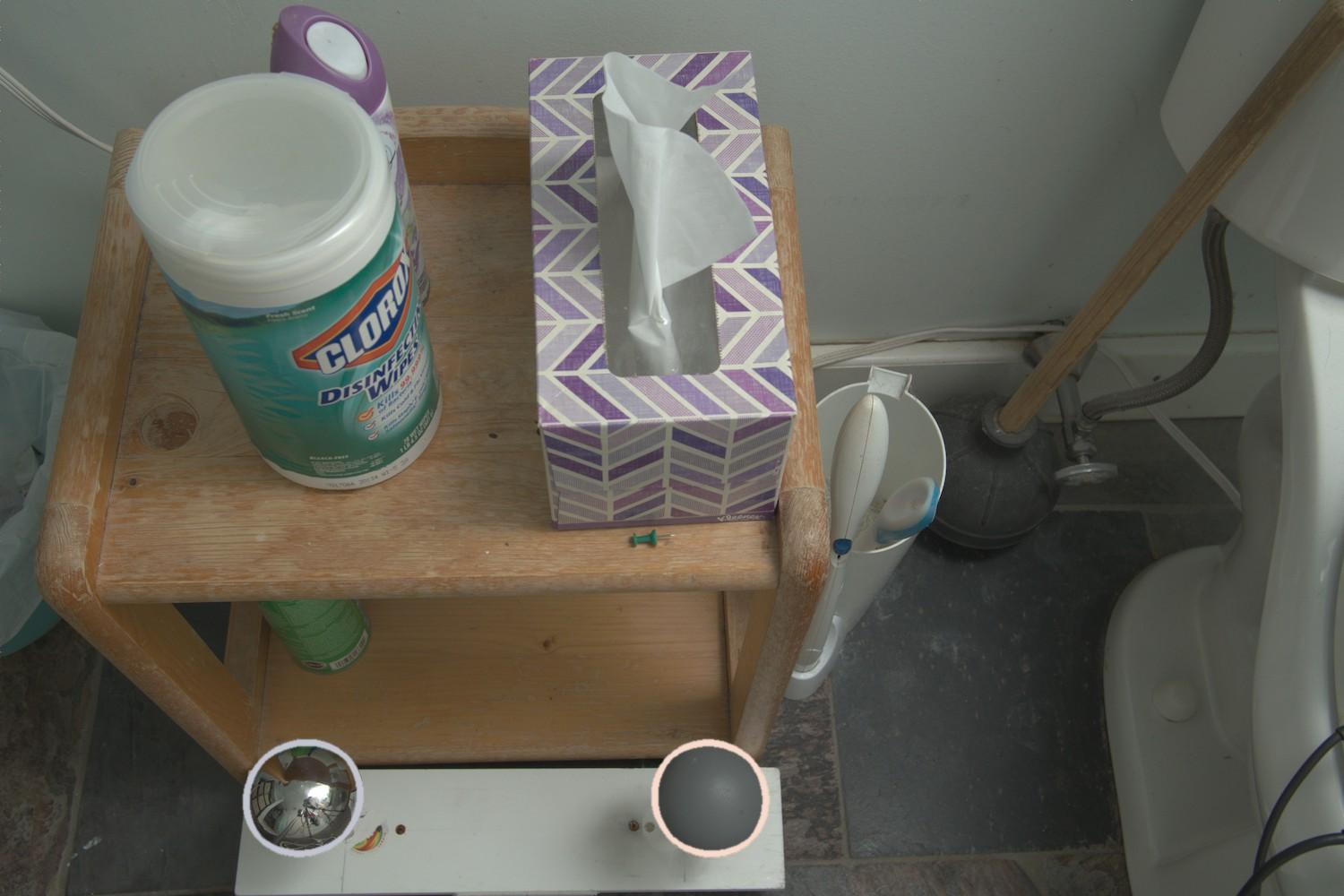} & \includegraphics[width=0.095\textwidth]{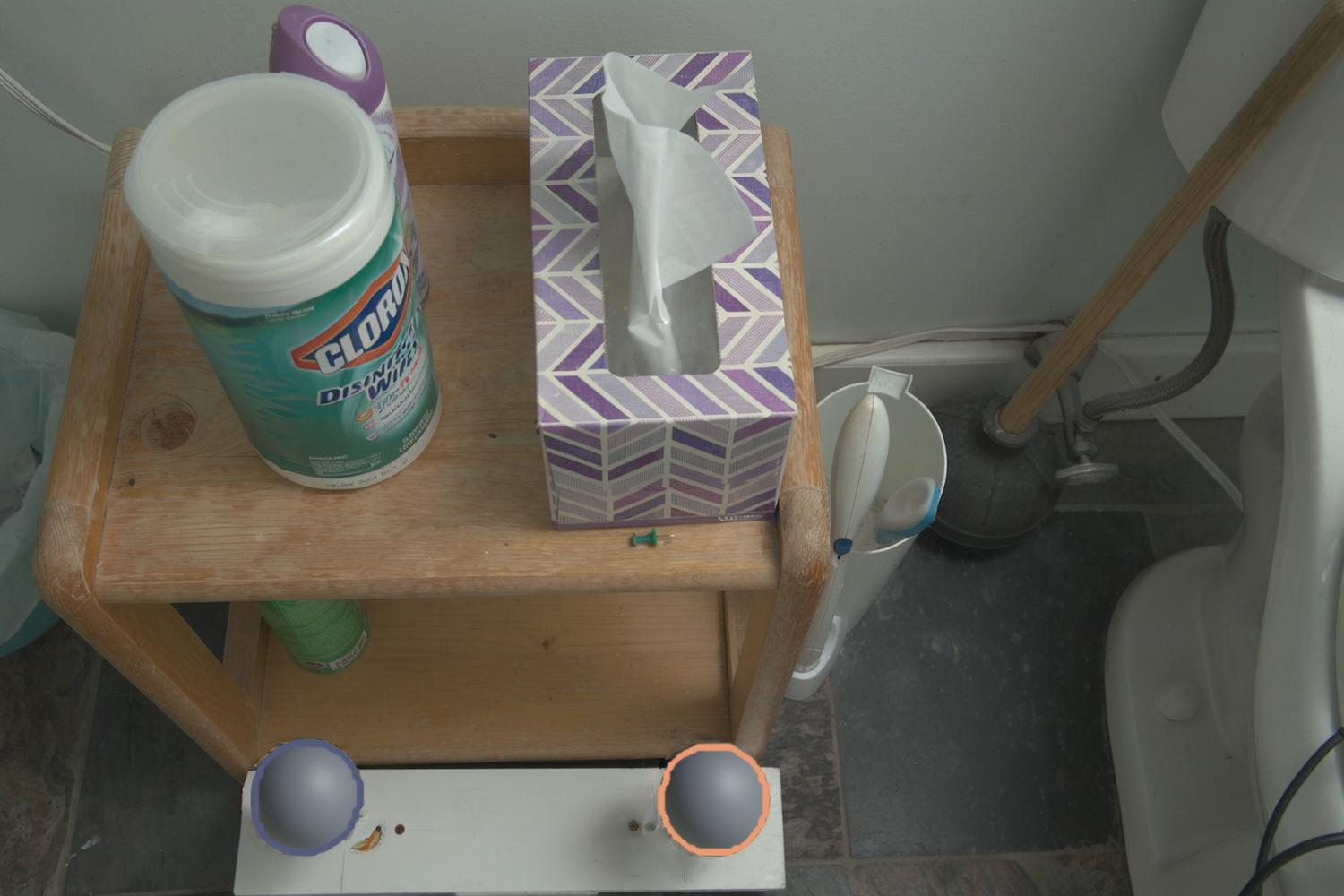} & \includegraphics[width=0.095\textwidth]{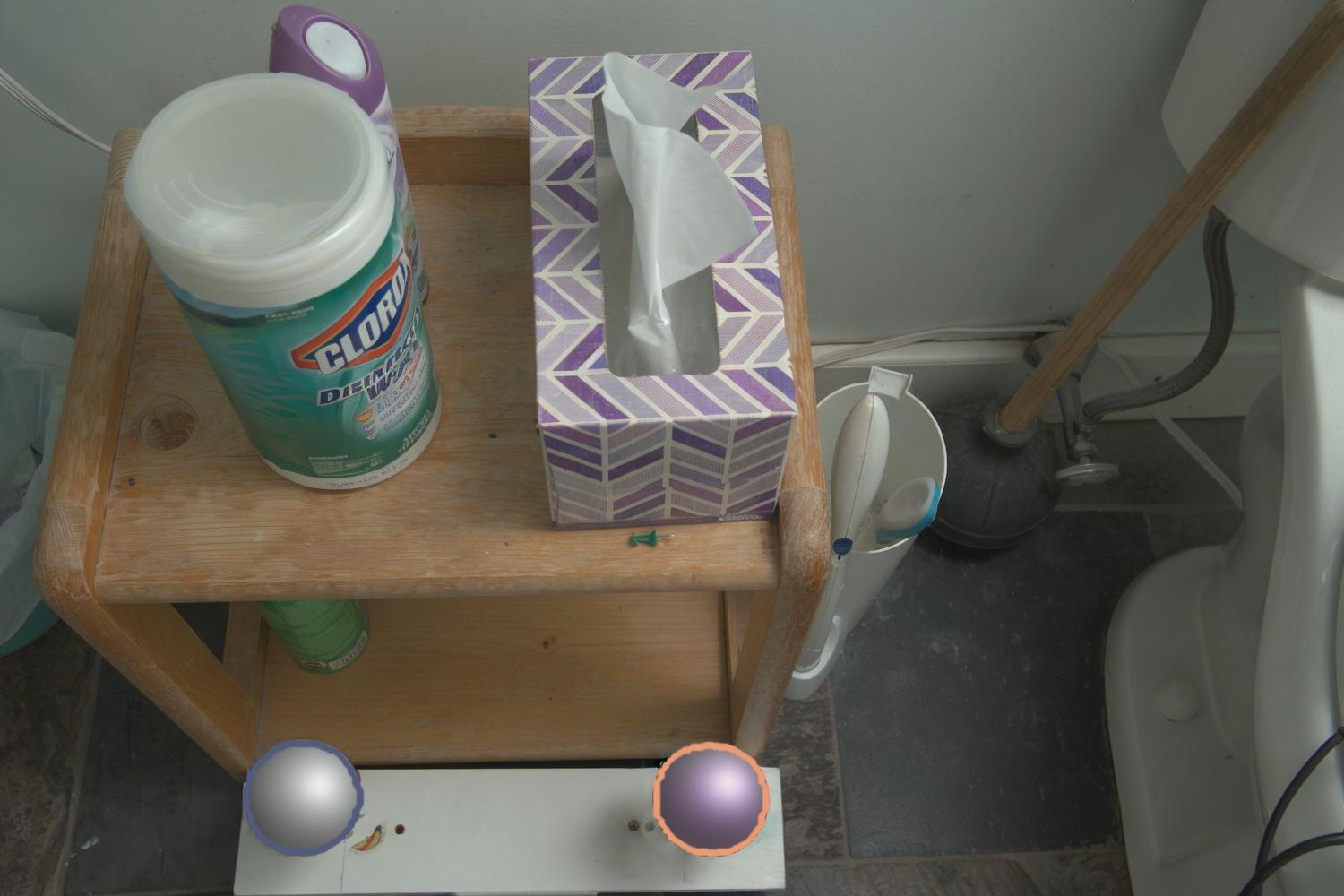} & \includegraphics[width=0.095\textwidth]{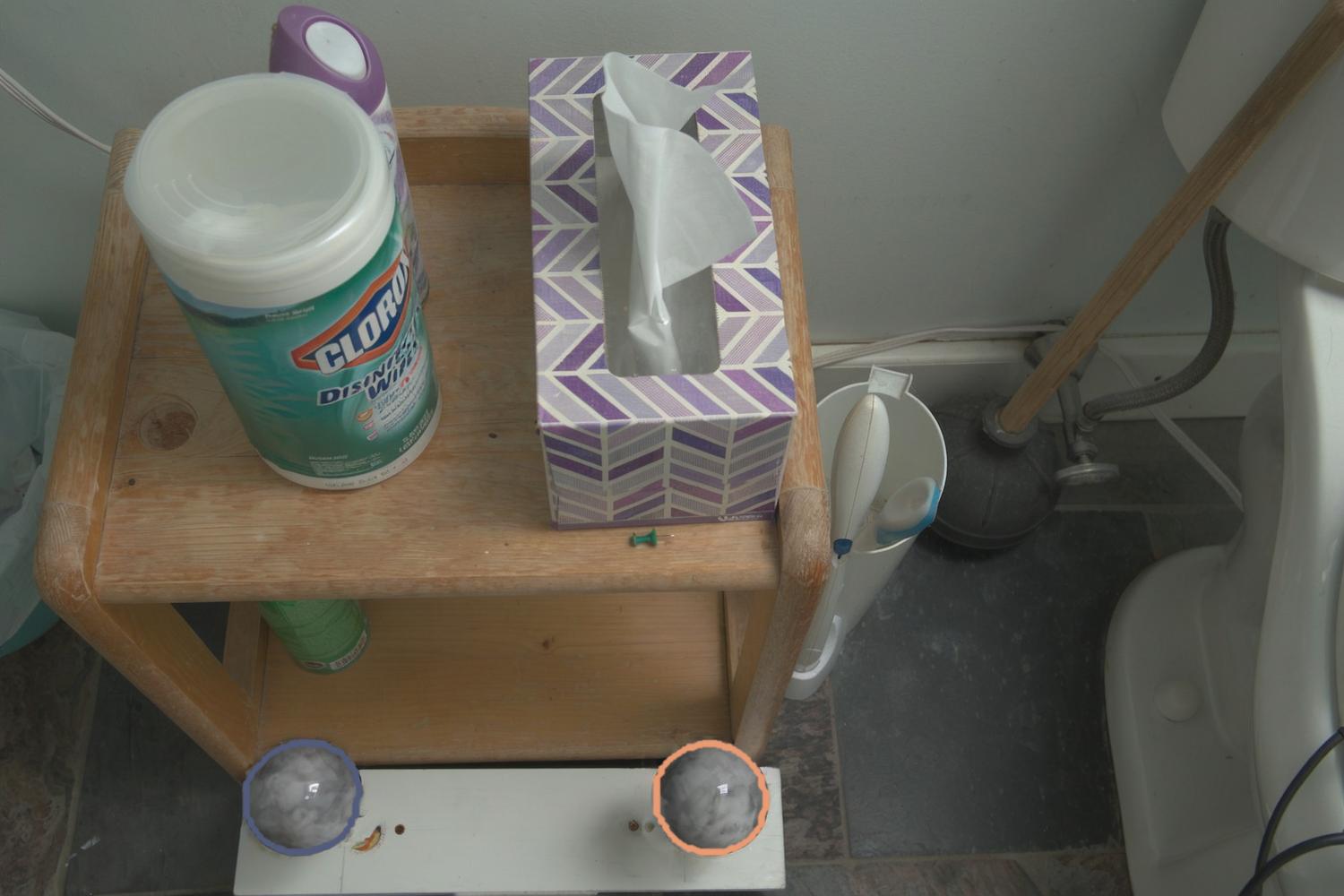} & \includegraphics[width=0.095\textwidth]{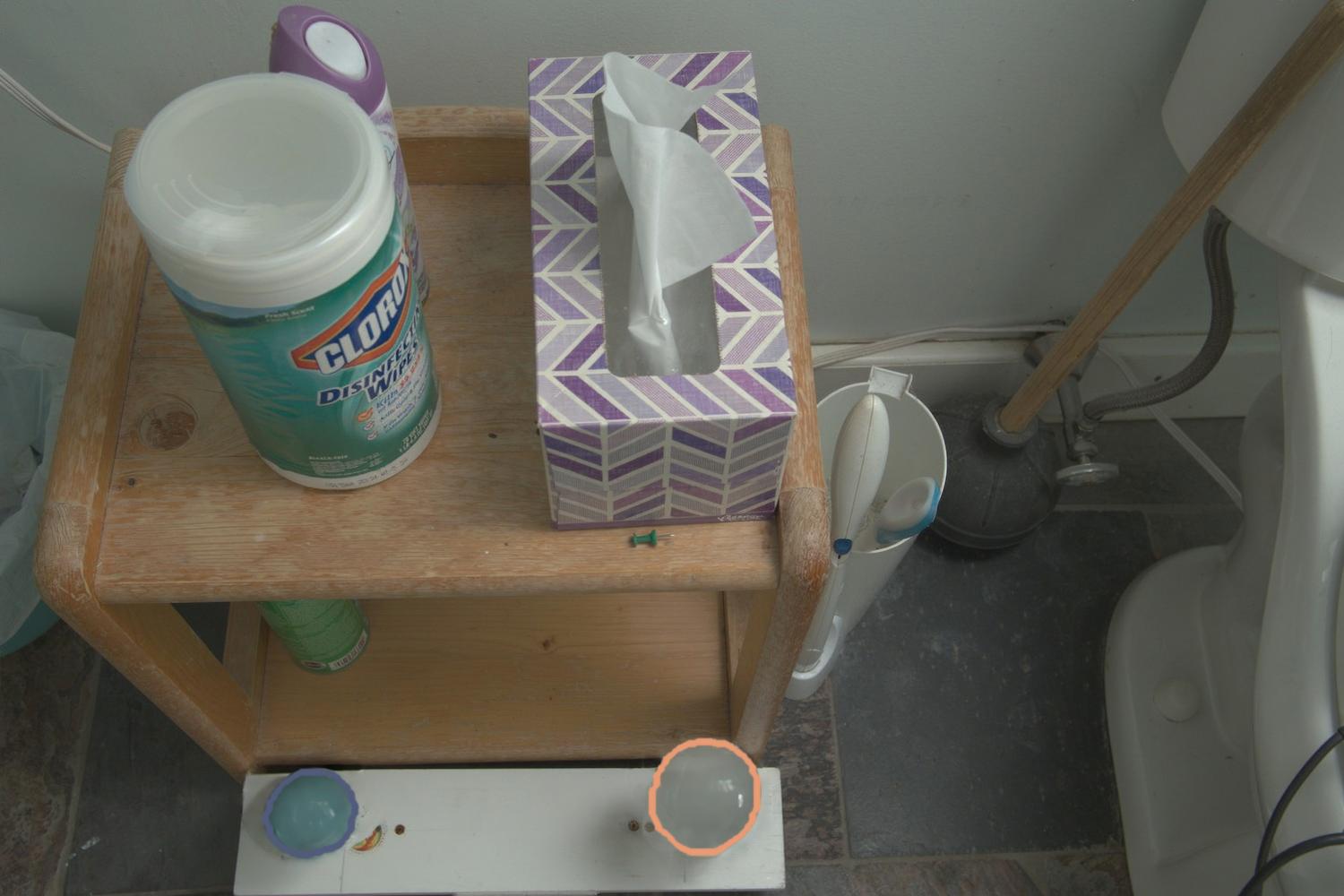} & \includegraphics[width=0.095\textwidth]{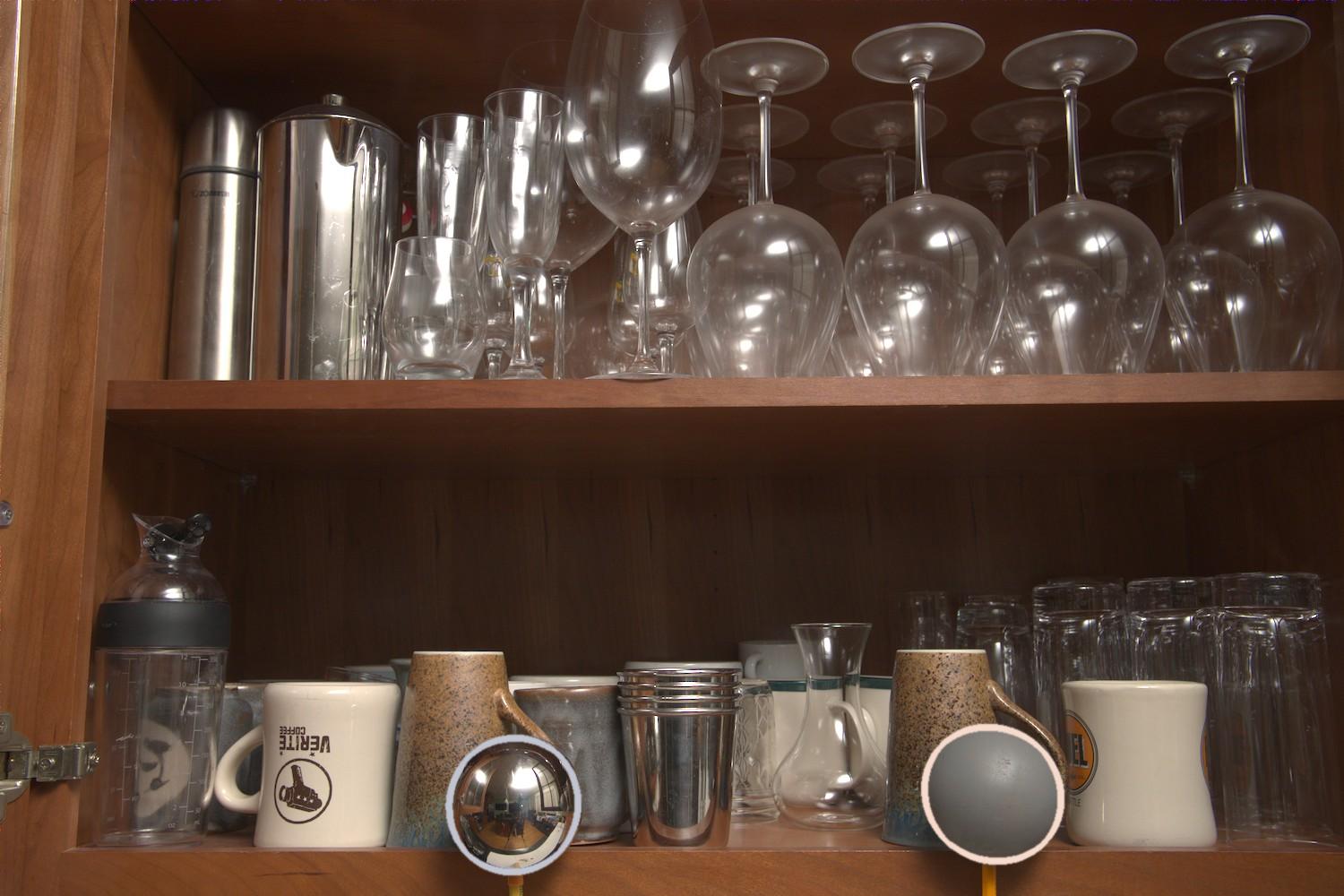} & \includegraphics[width=0.095\textwidth]{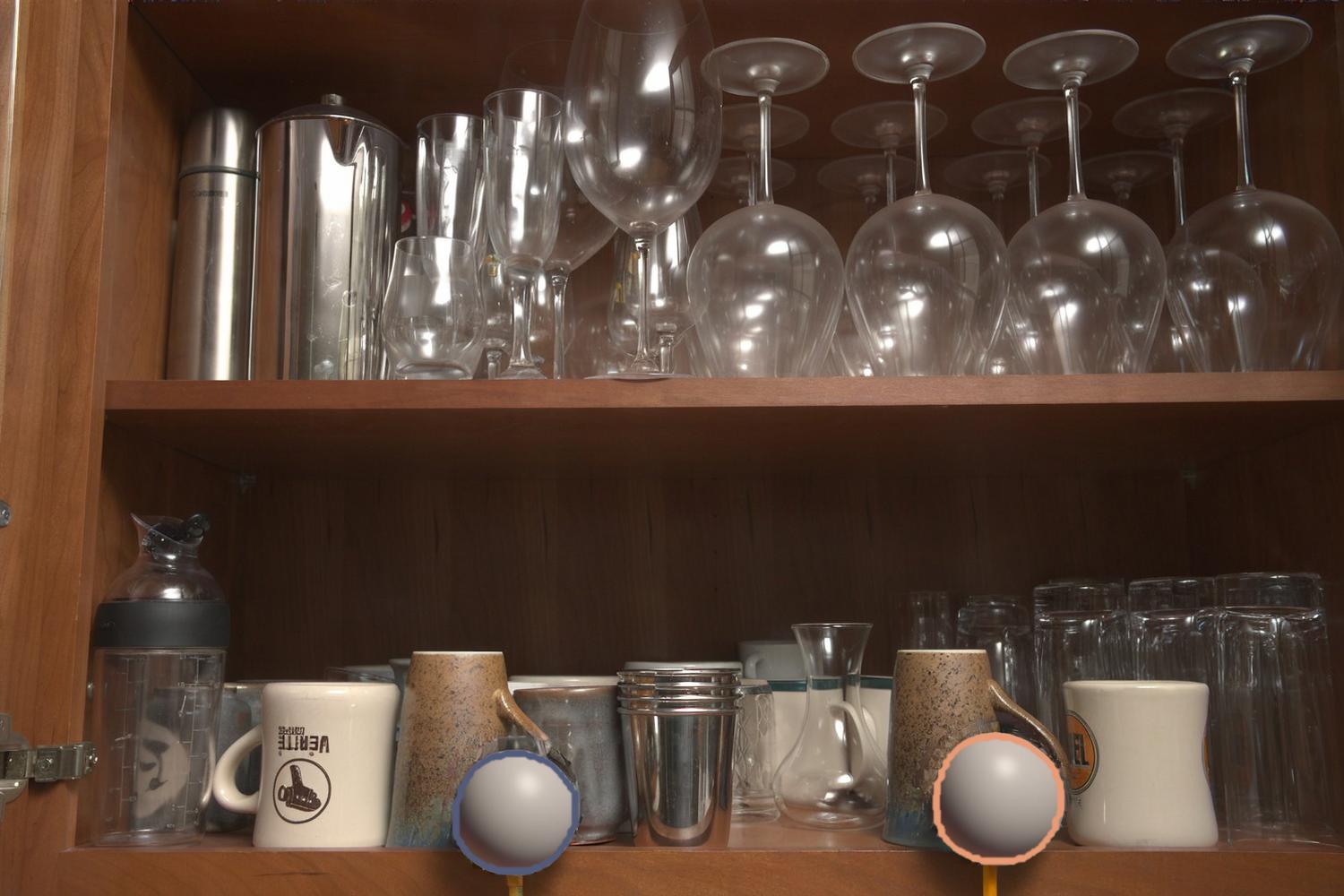} & \includegraphics[width=0.095\textwidth]{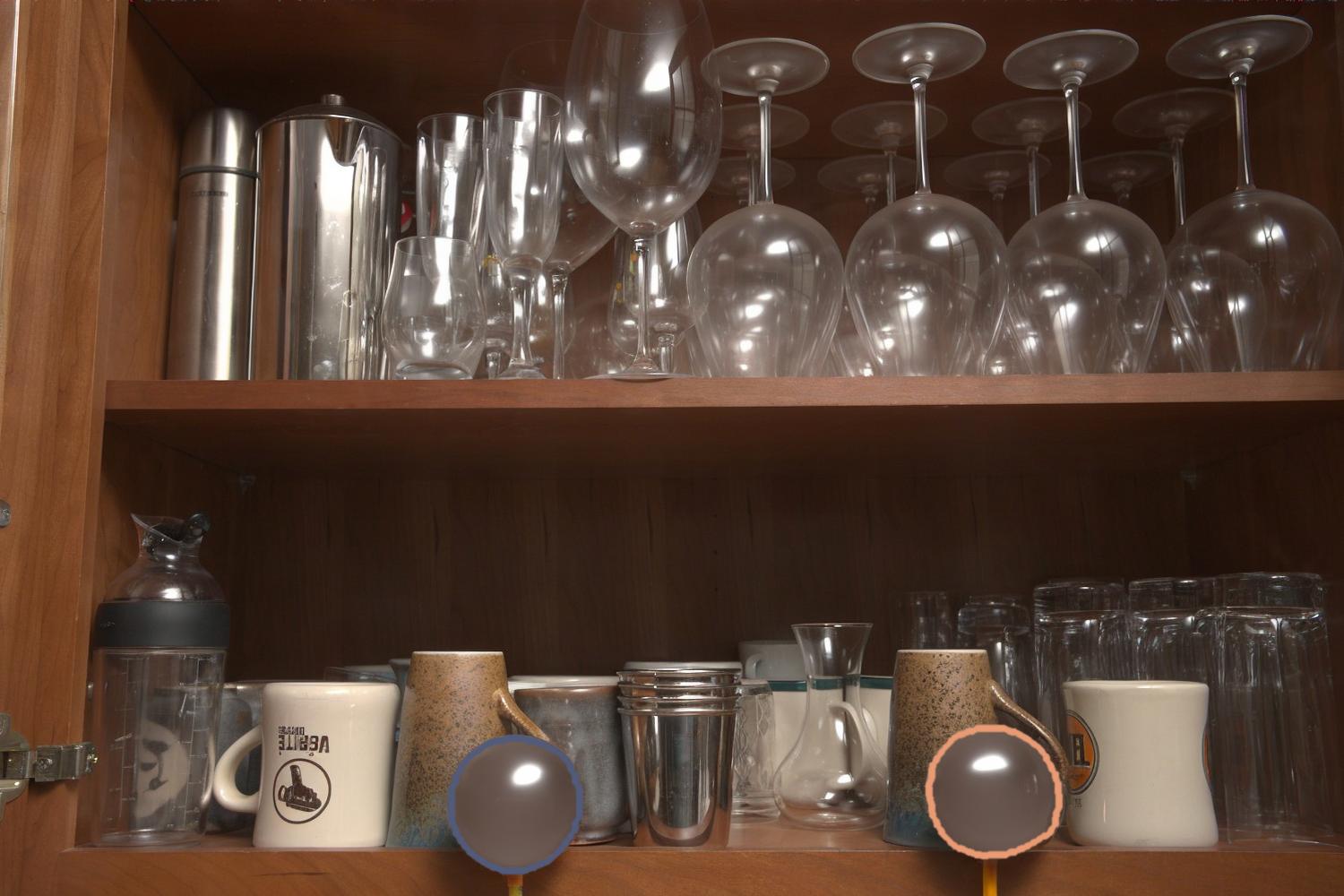} & \includegraphics[width=0.095\textwidth]{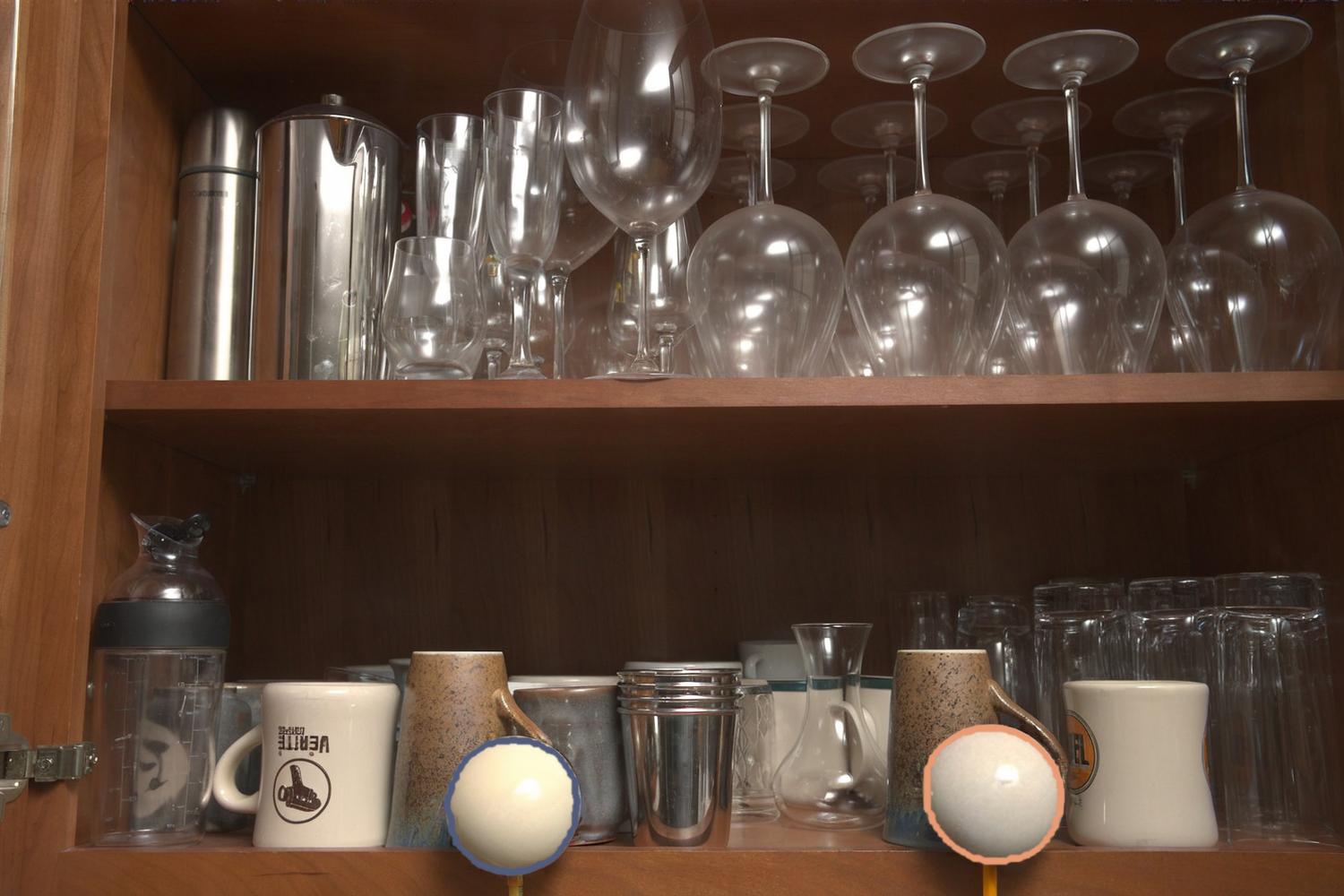} & \includegraphics[width=0.095\textwidth]{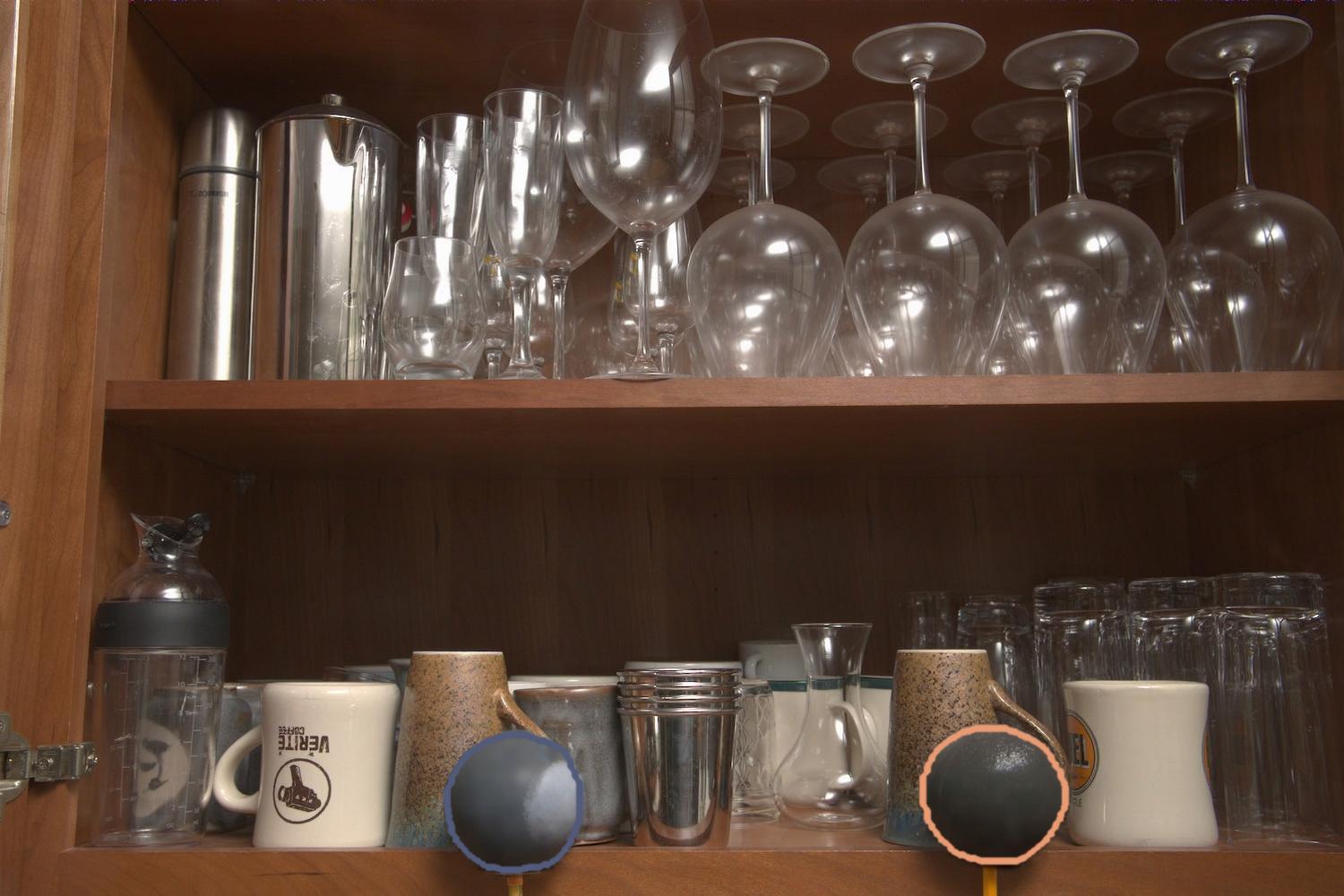} \\[1pt]
    \includegraphics[width=0.070\textwidth]{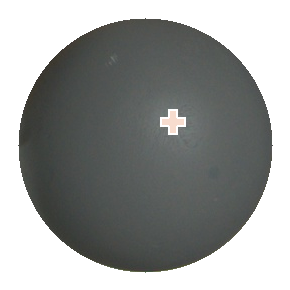} & \includegraphics[width=0.070\textwidth]{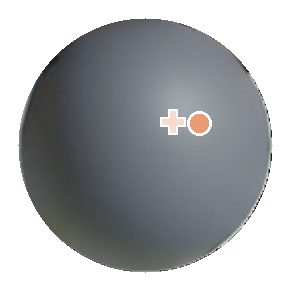} & \includegraphics[width=0.070\textwidth]{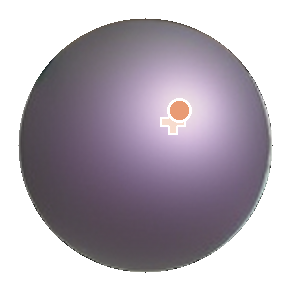} & \includegraphics[width=0.070\textwidth]{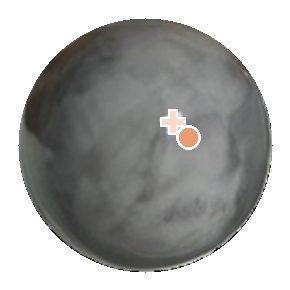} & \includegraphics[width=0.070\textwidth]{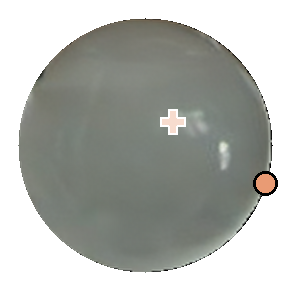} & \includegraphics[width=0.070\textwidth]{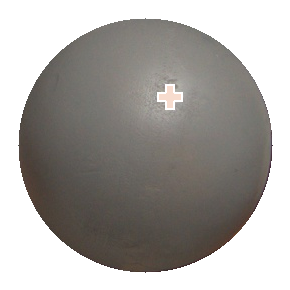} & \includegraphics[width=0.070\textwidth]{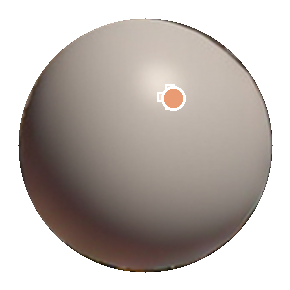} & \includegraphics[width=0.070\textwidth]{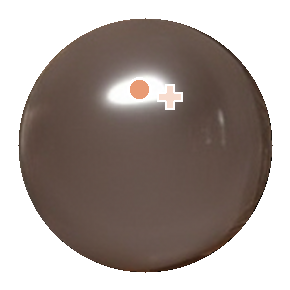} & \includegraphics[width=0.070\textwidth]{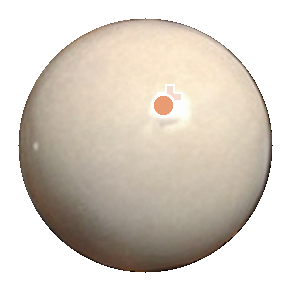} & \includegraphics[width=0.070\textwidth]{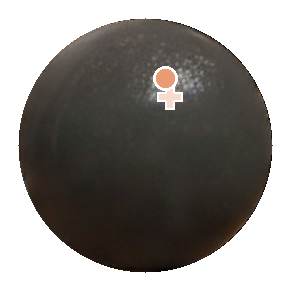} \\
    \includegraphics[width=0.070\textwidth]{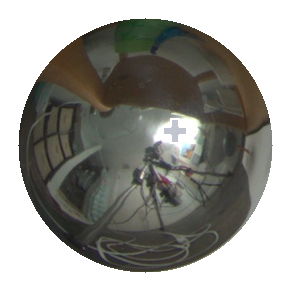} & \includegraphics[width=0.070\textwidth]{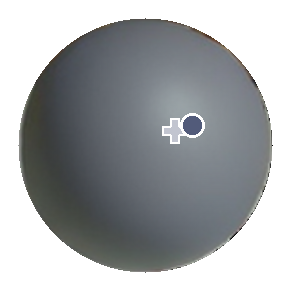} & \includegraphics[width=0.070\textwidth]{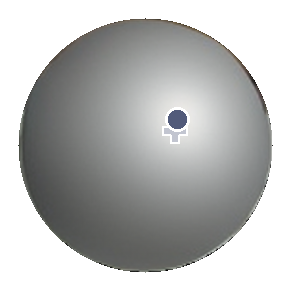} & \includegraphics[width=0.070\textwidth]{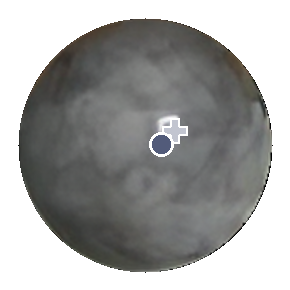} & \includegraphics[width=0.070\textwidth]{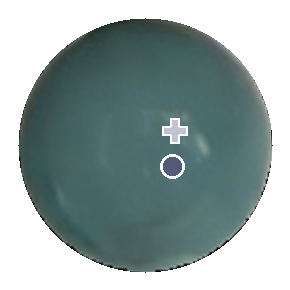} & \includegraphics[width=0.070\textwidth]{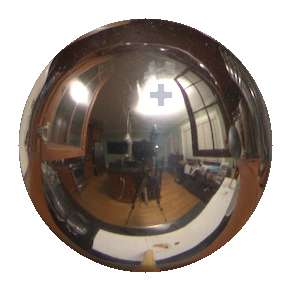} & \includegraphics[width=0.070\textwidth]{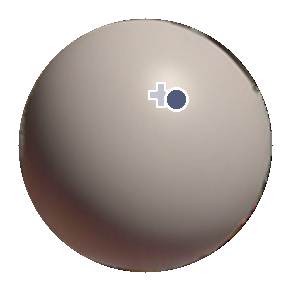} & \includegraphics[width=0.070\textwidth]{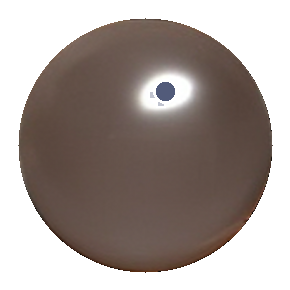} & \includegraphics[width=0.070\textwidth]{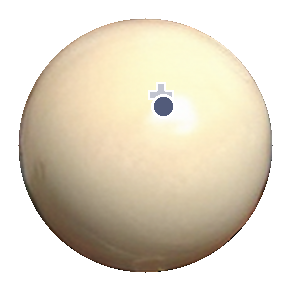} & \includegraphics[width=0.070\textwidth]{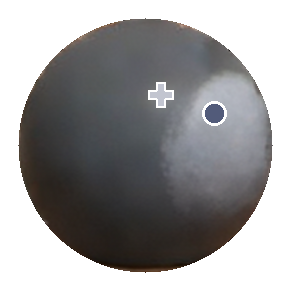} \\
  \end{tabular}
  \\[4pt]
  \begin{tabular}{@{}cccccccccc@{}}
    \textbf{GT} & \textbf{FLUX.1 Fill [dev]} & \textbf{HiDream} & \textbf{Nano Banana 1} & \textbf{Hunyuan 2.1} & \textbf{GT} & \textbf{FLUX.1 Fill [dev]} & \textbf{HiDream} & \textbf{Nano Banana 1} & \textbf{Hunyuan 2.1} \\
    \includegraphics[width=0.095\textwidth]{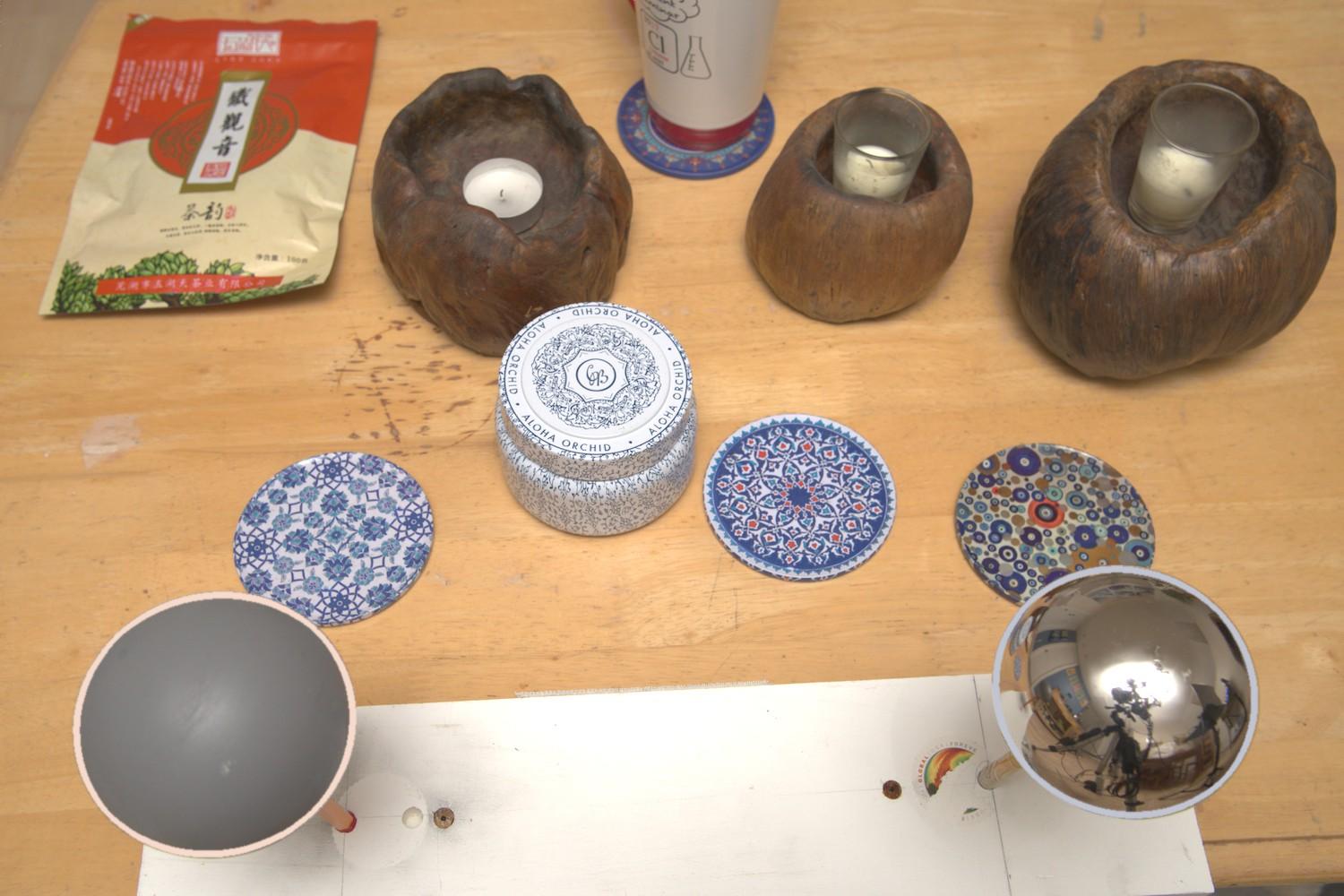} & \includegraphics[width=0.095\textwidth]{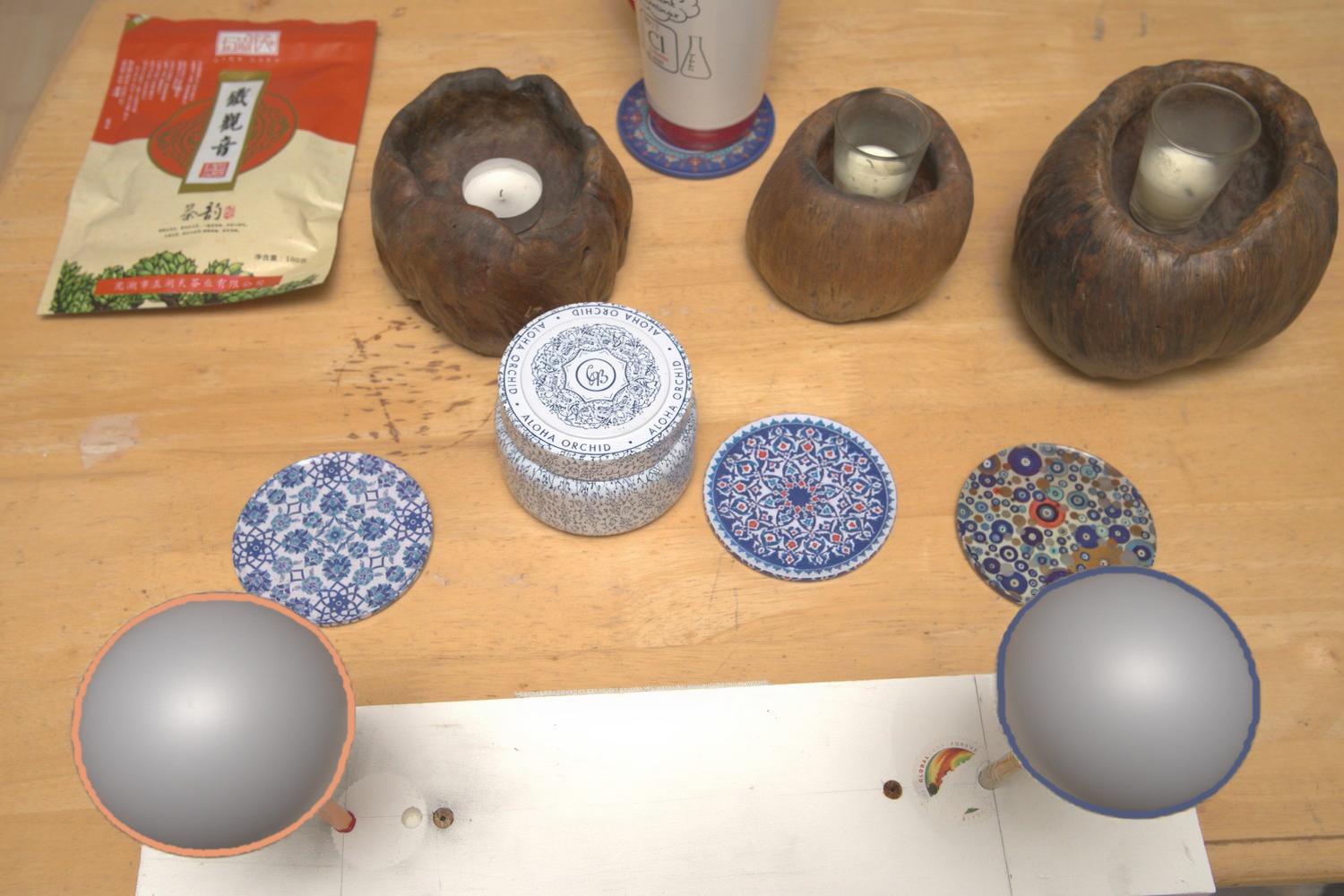} & \includegraphics[width=0.095\textwidth]{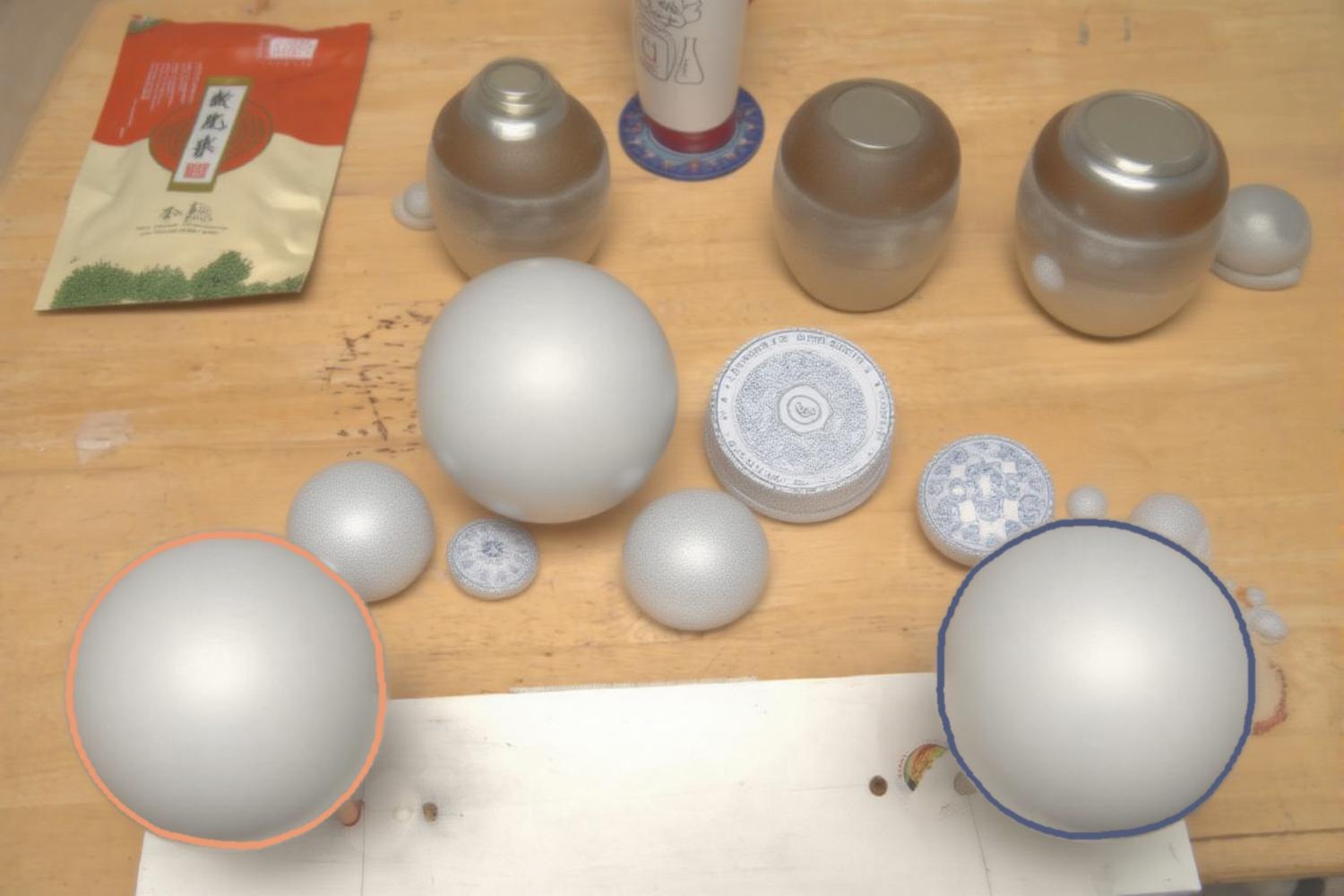} & \includegraphics[width=0.095\textwidth]{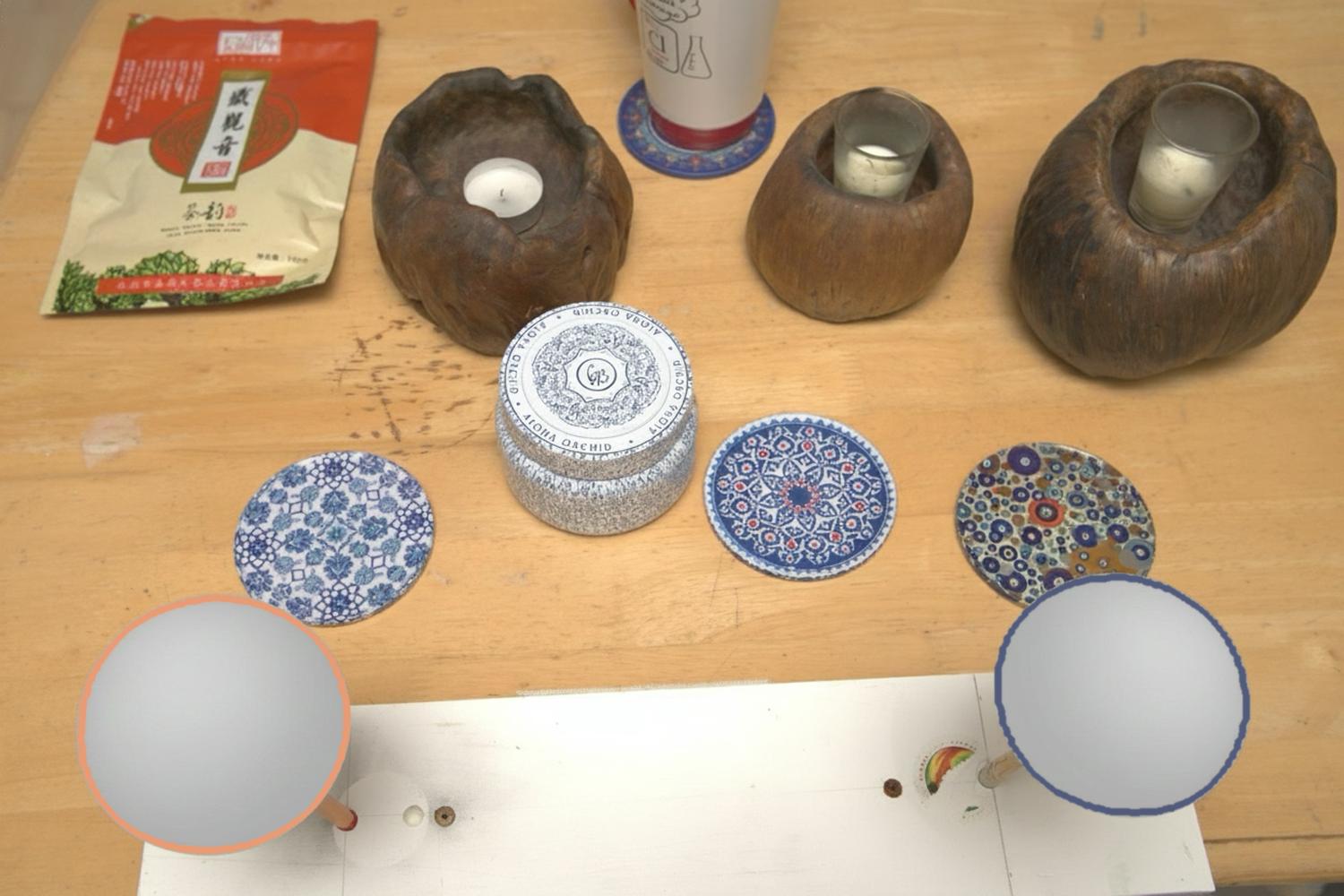} & \includegraphics[width=0.095\textwidth]{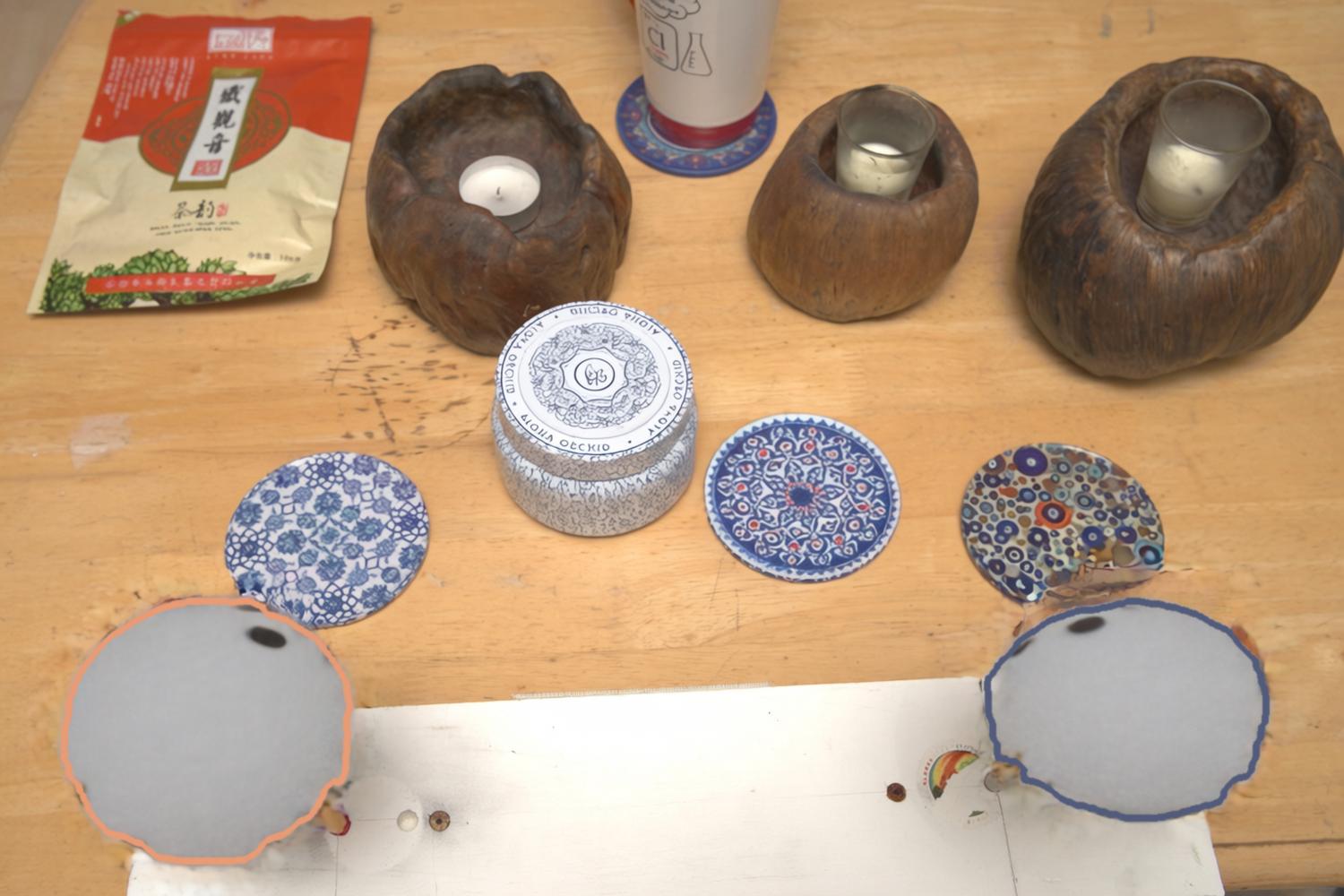} & \includegraphics[width=0.095\textwidth]{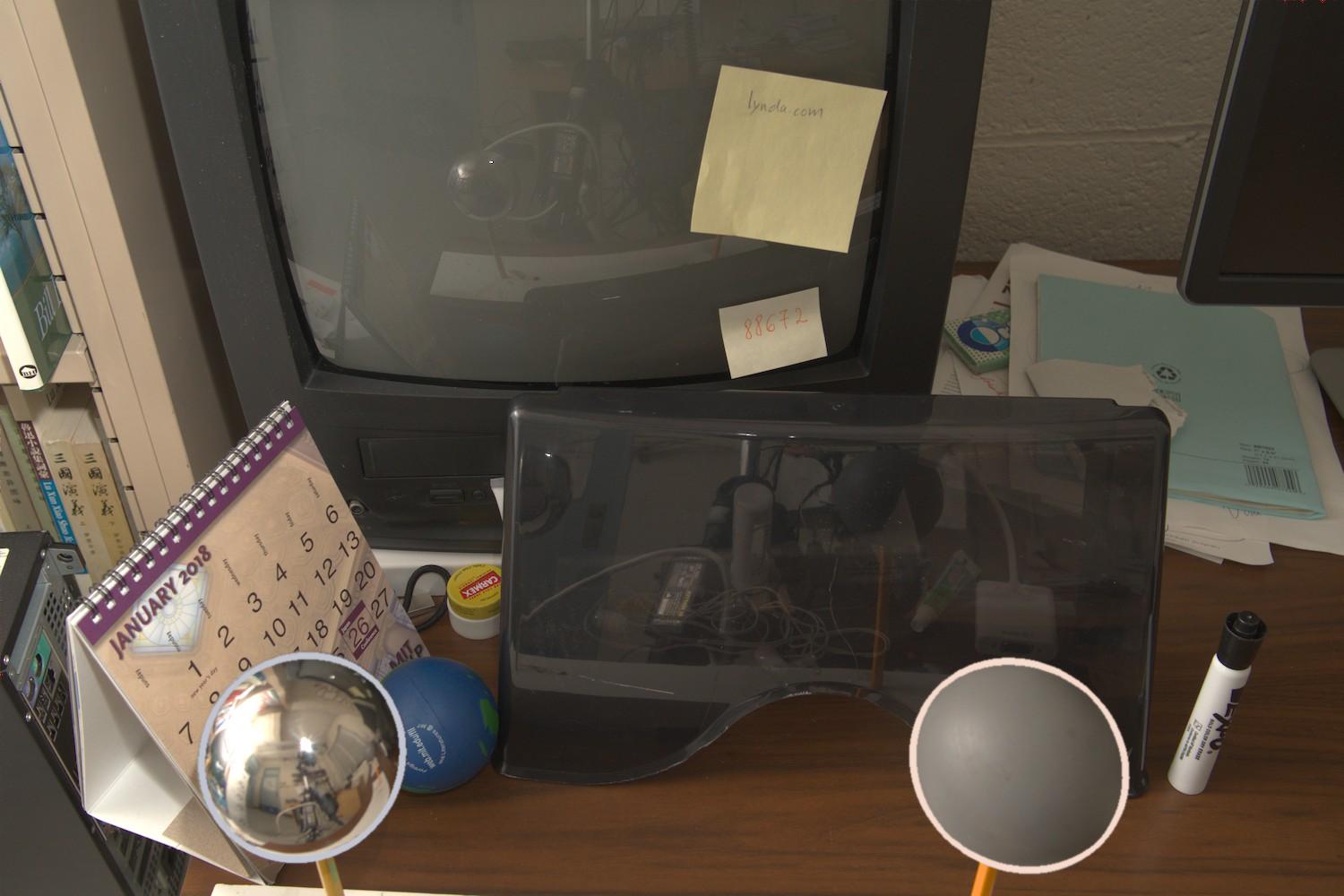} & \includegraphics[width=0.095\textwidth]{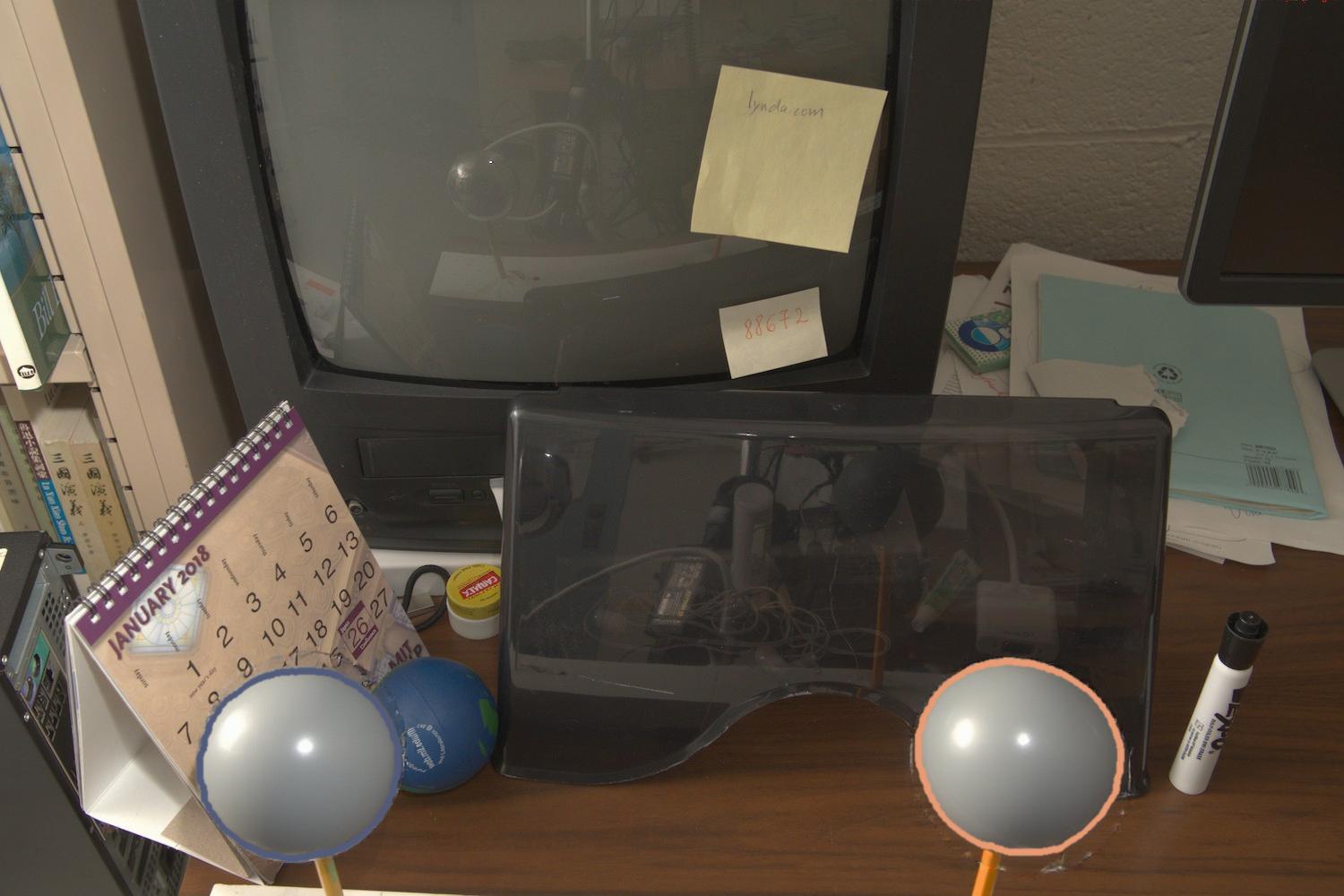} & \includegraphics[width=0.095\textwidth]{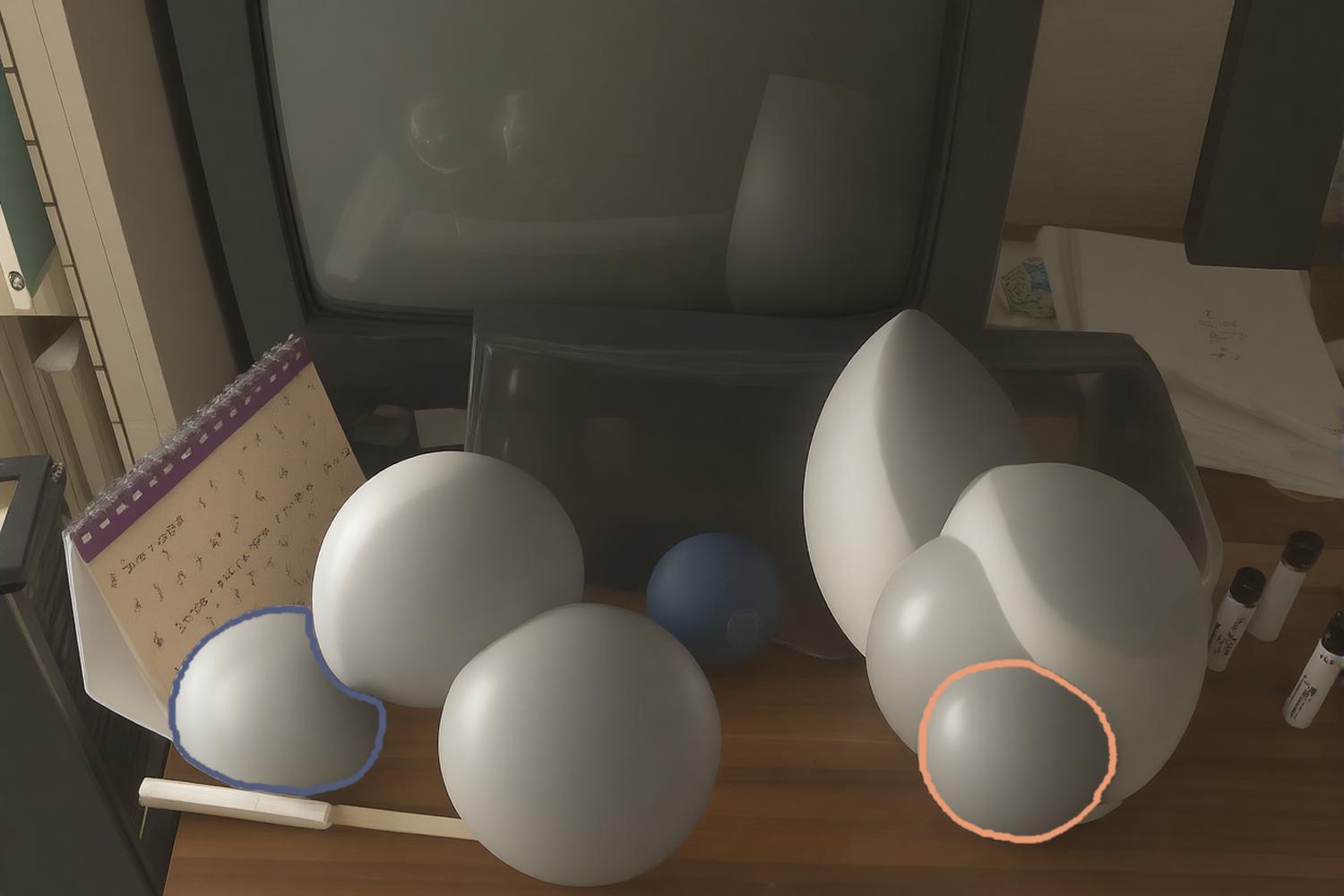} & \includegraphics[width=0.095\textwidth]{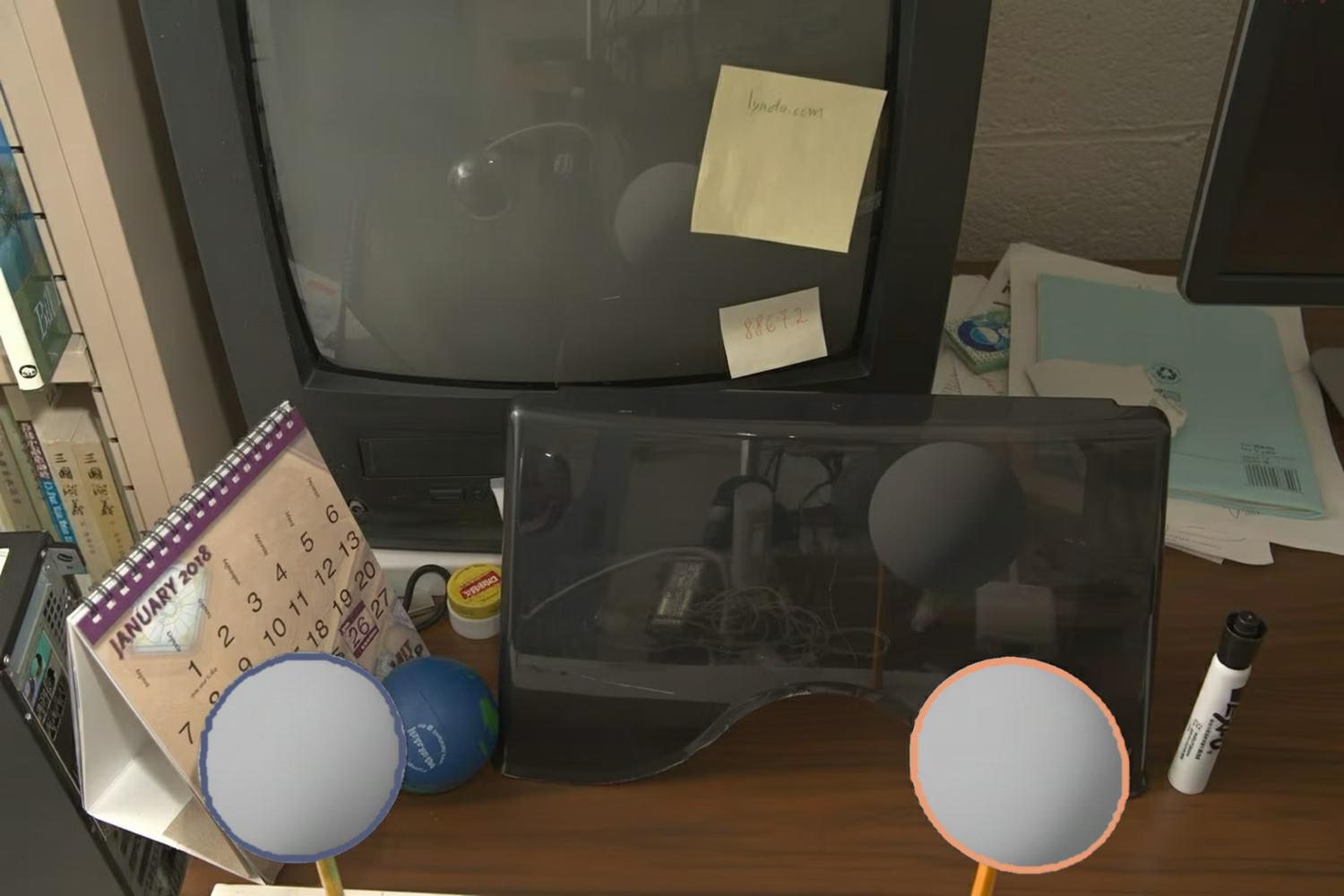} & \includegraphics[width=0.095\textwidth]{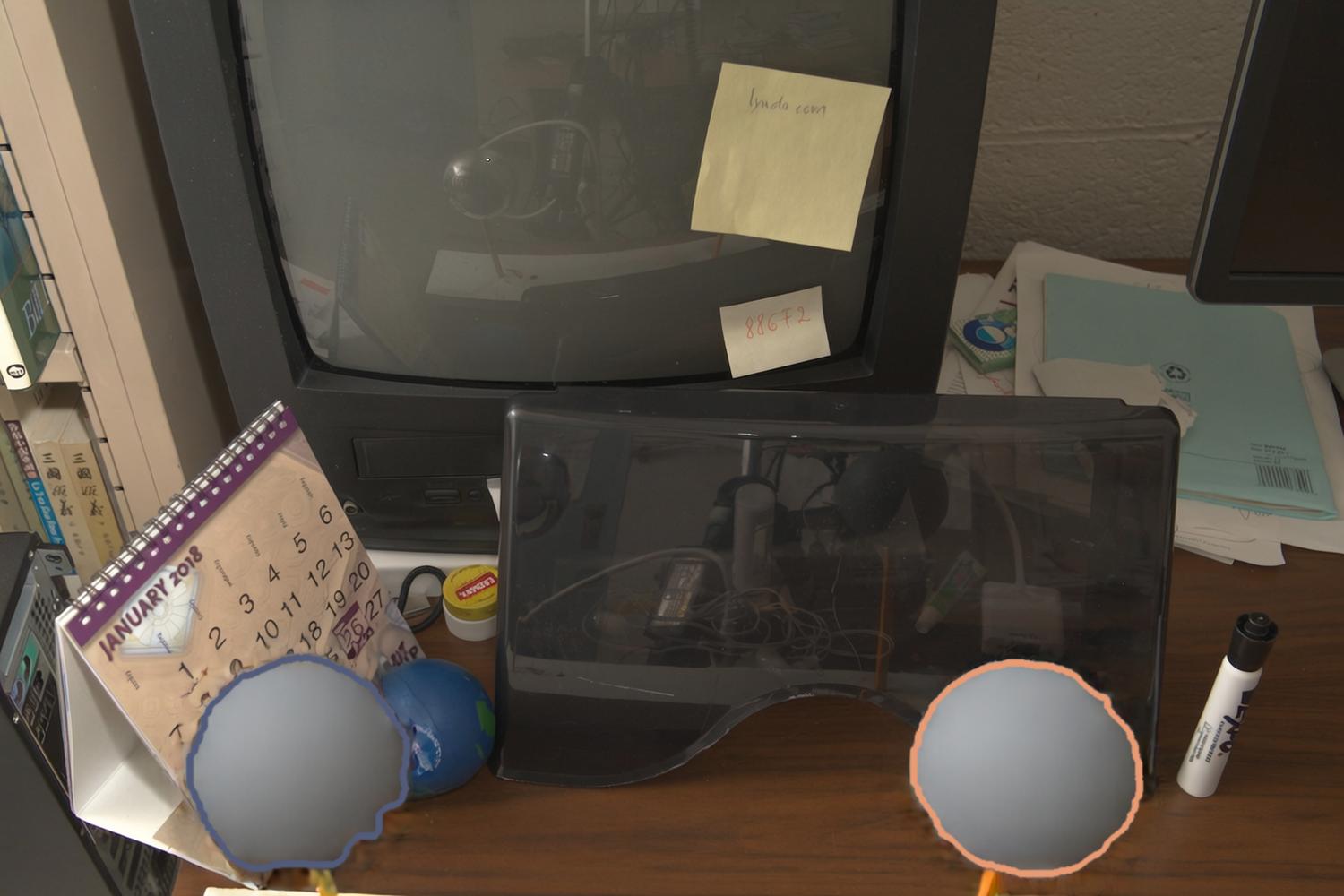} \\[1pt]
    \includegraphics[width=0.070\textwidth]{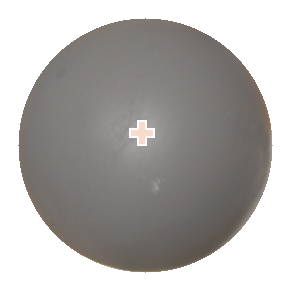} & \includegraphics[width=0.070\textwidth]{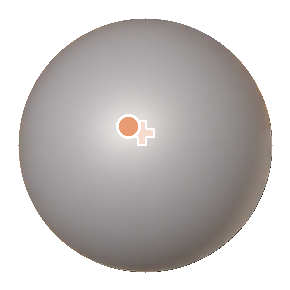} & \includegraphics[width=0.070\textwidth]{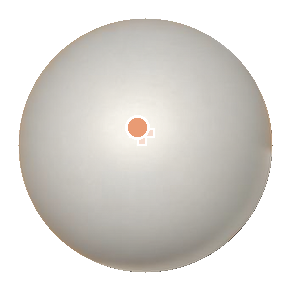} & \includegraphics[width=0.070\textwidth]{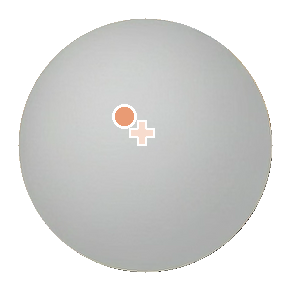} & \includegraphics[width=0.070\textwidth]{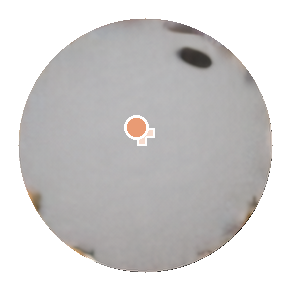} & \includegraphics[width=0.070\textwidth]{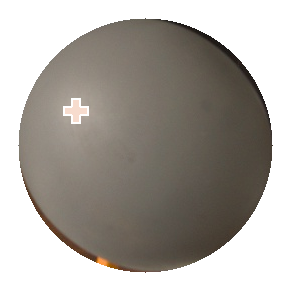} & \includegraphics[width=0.070\textwidth]{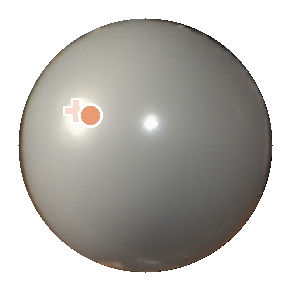} & \includegraphics[width=0.070\textwidth]{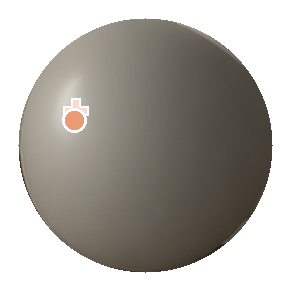} & \includegraphics[width=0.070\textwidth]{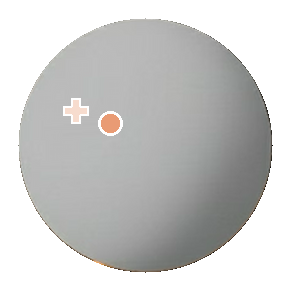} & \includegraphics[width=0.070\textwidth]{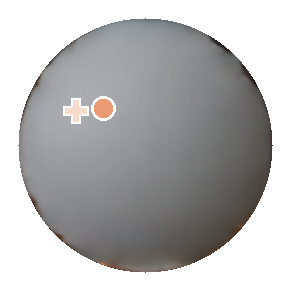} \\
    \includegraphics[width=0.070\textwidth]{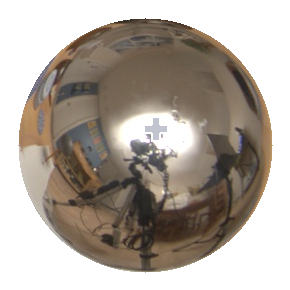} & \includegraphics[width=0.070\textwidth]{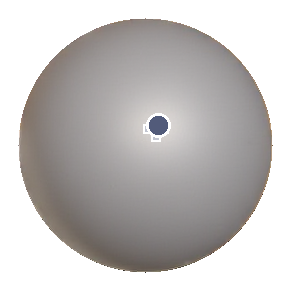} & \includegraphics[width=0.070\textwidth]{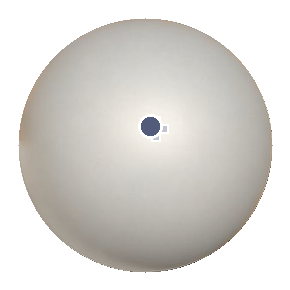} & \includegraphics[width=0.070\textwidth]{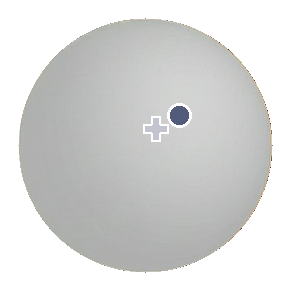} & \includegraphics[width=0.070\textwidth]{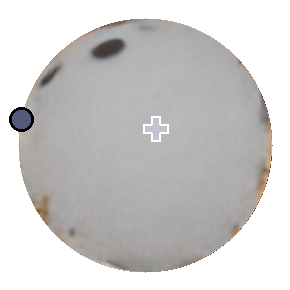} & \includegraphics[width=0.070\textwidth]{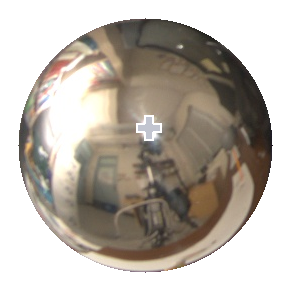} & \includegraphics[width=0.070\textwidth]{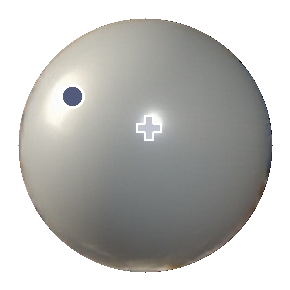} & \includegraphics[width=0.070\textwidth]{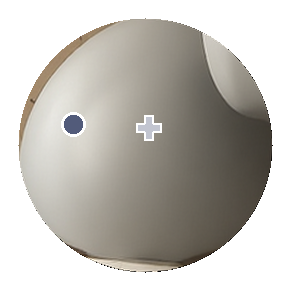} & \includegraphics[width=0.070\textwidth]{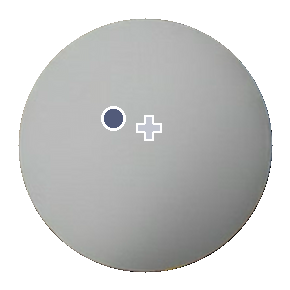} & \includegraphics[width=0.070\textwidth]{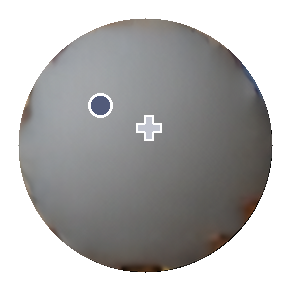} \\
  \end{tabular}
  \\[4pt]
  \begin{tabular}{@{}cccccccccc@{}}
    \textbf{GT} & \textbf{Qwen Edit} & \textbf{Firefly 5} & \textbf{FLUX.2 [dev]} & \textbf{Z-Image Turbo} & \textbf{GT} & \textbf{Qwen Edit} & \textbf{Firefly 5} & \textbf{FLUX.2 [dev]} & \textbf{Z-Image Turbo} \\
    \includegraphics[width=0.095\textwidth]{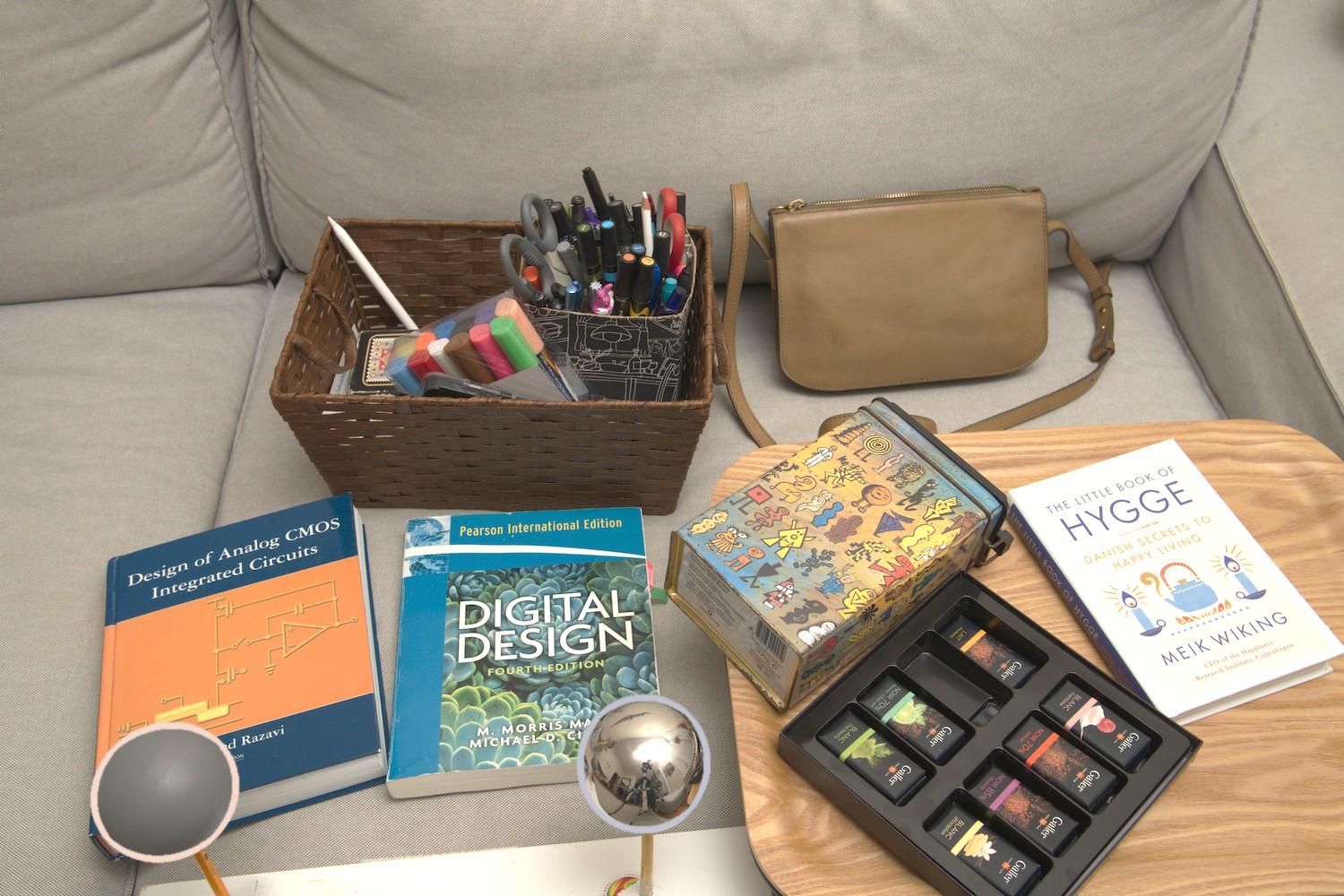} & \includegraphics[width=0.095\textwidth]{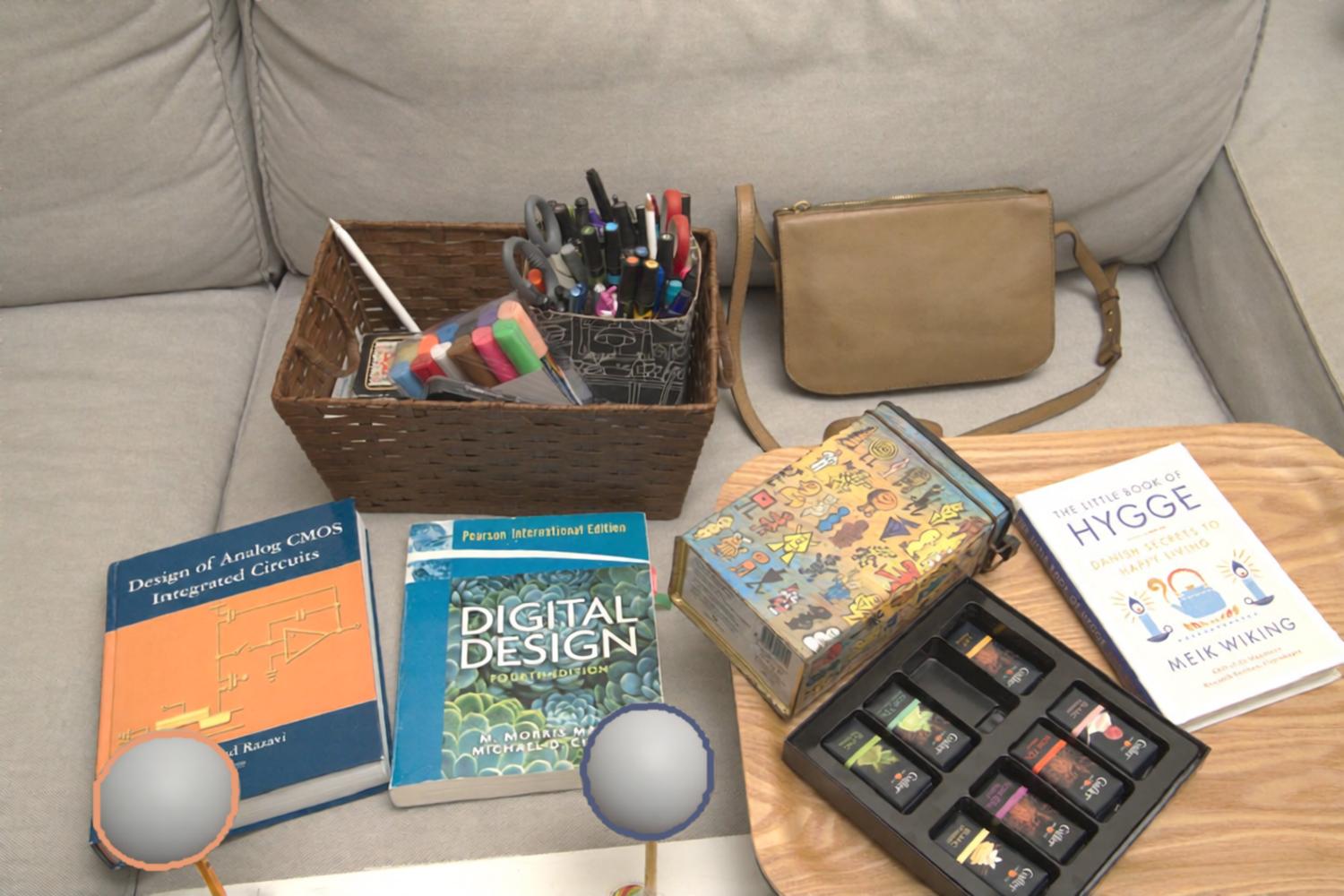} & \includegraphics[width=0.095\textwidth]{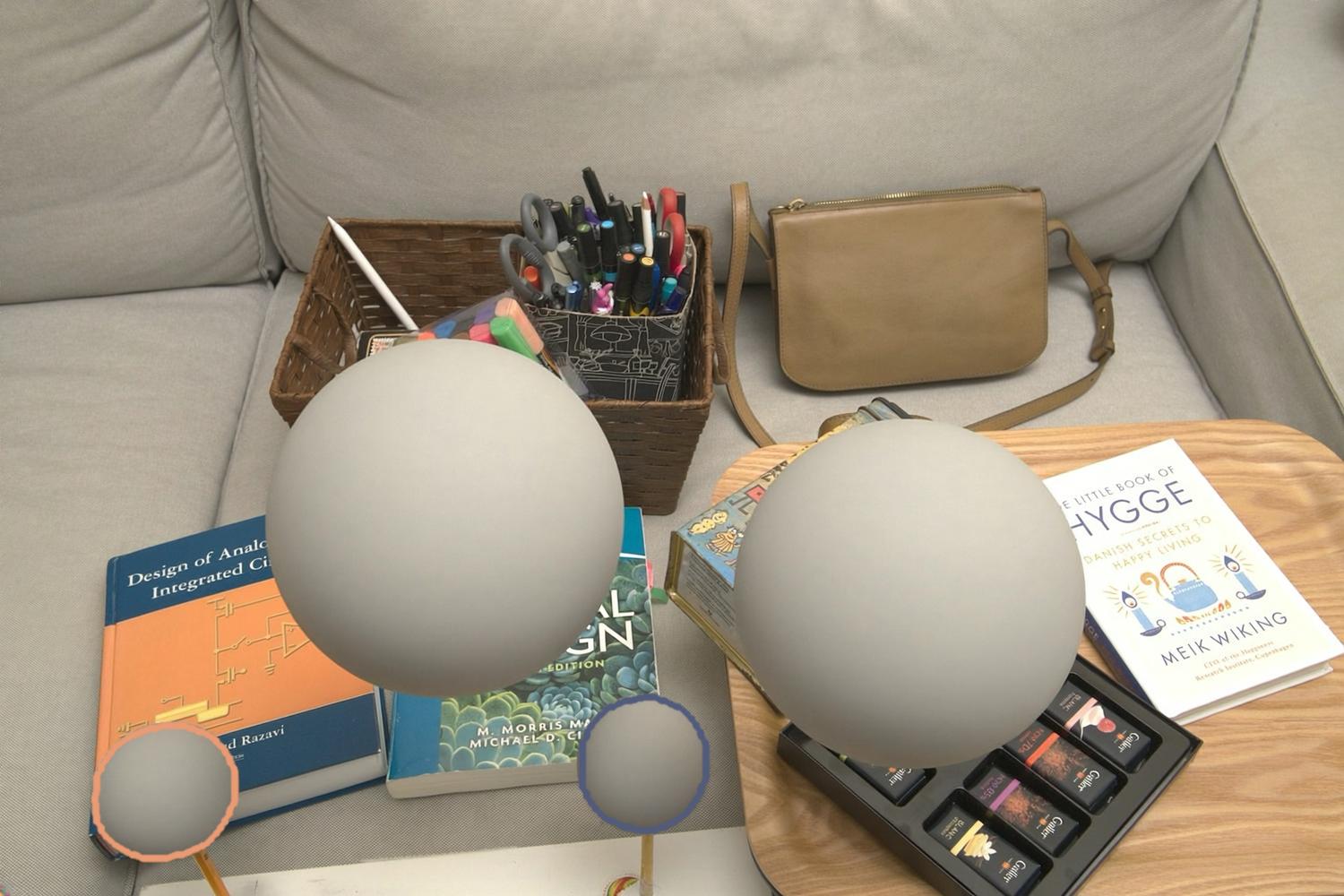} & \includegraphics[width=0.095\textwidth]{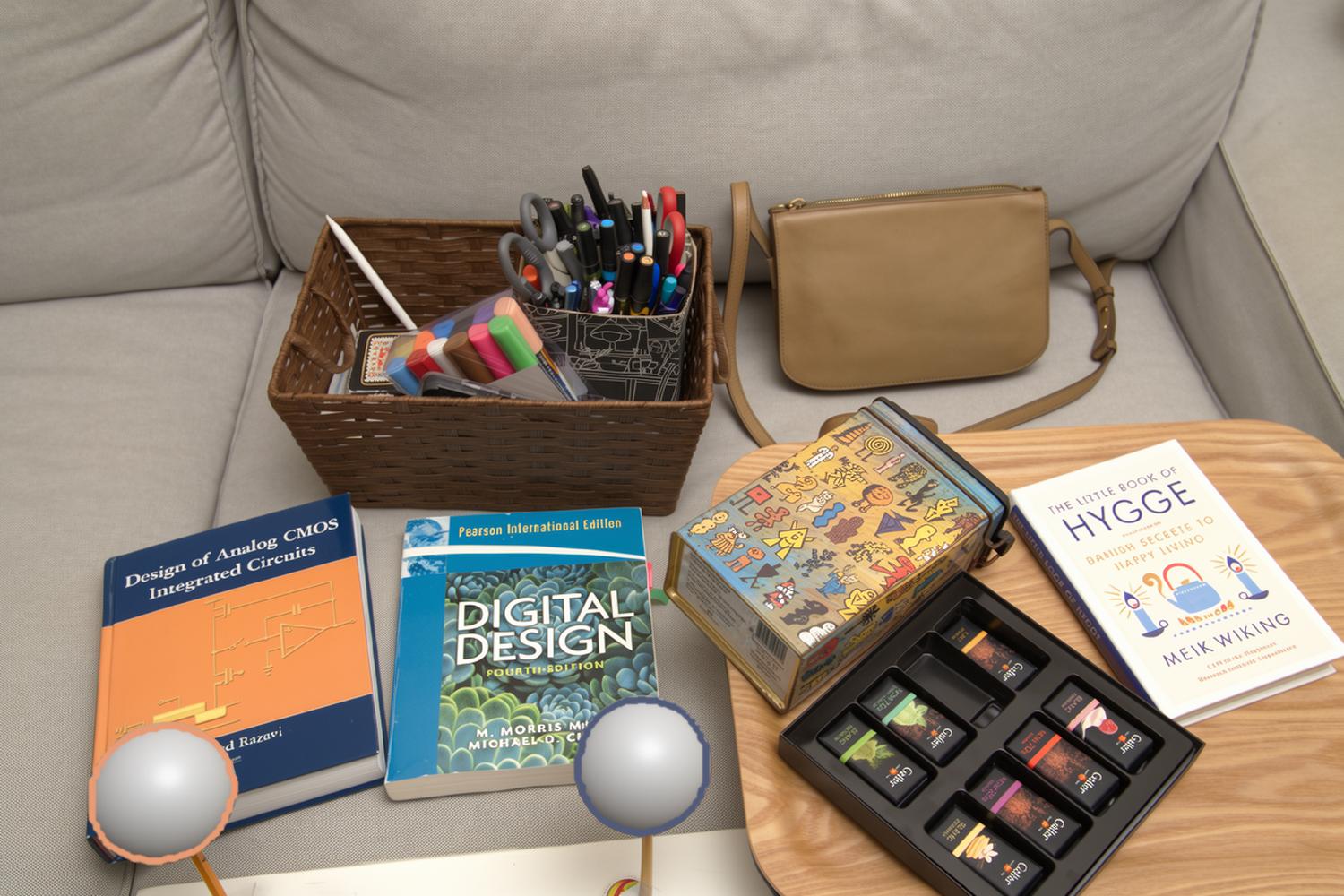} & \includegraphics[width=0.095\textwidth]{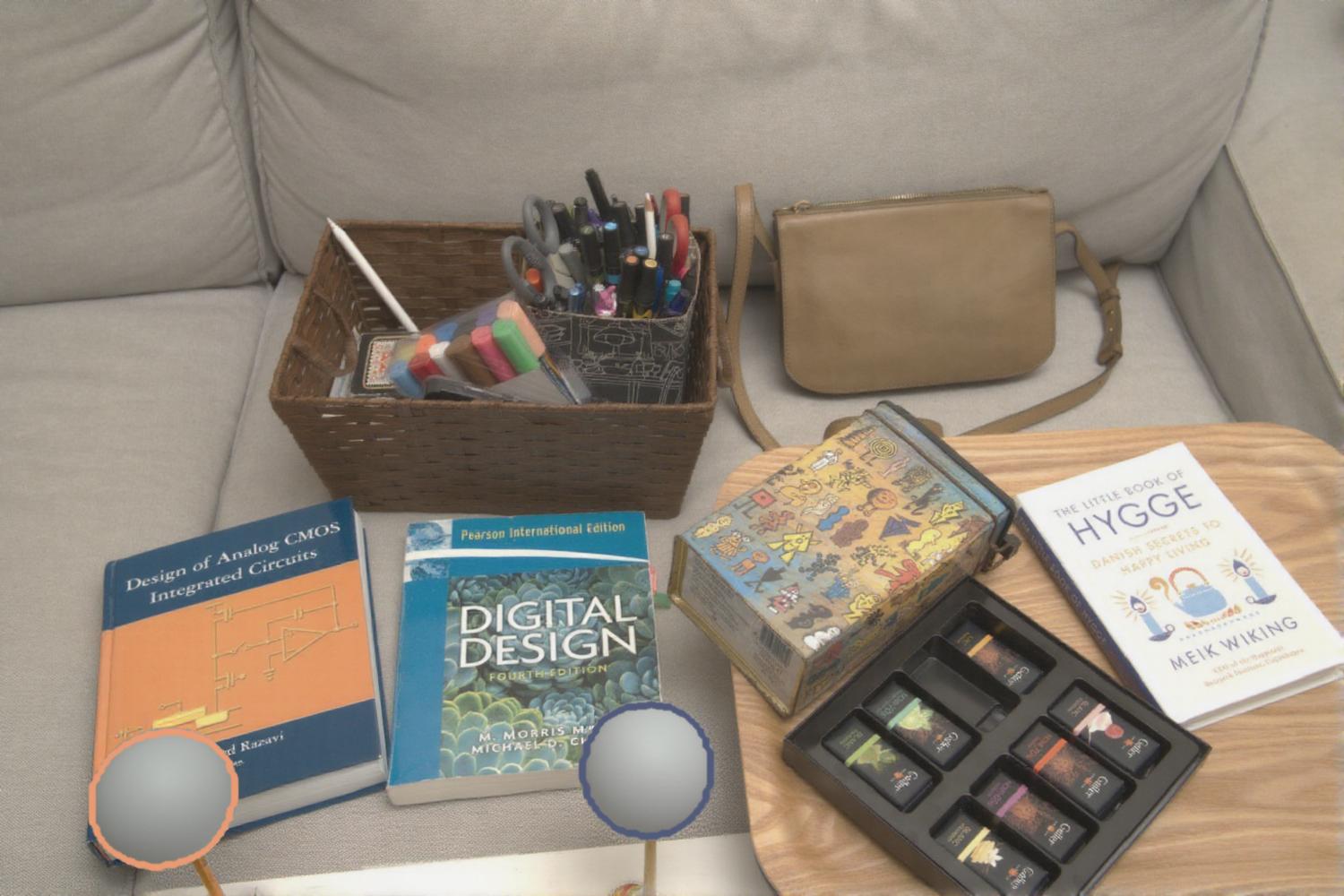} & \includegraphics[width=0.095\textwidth]{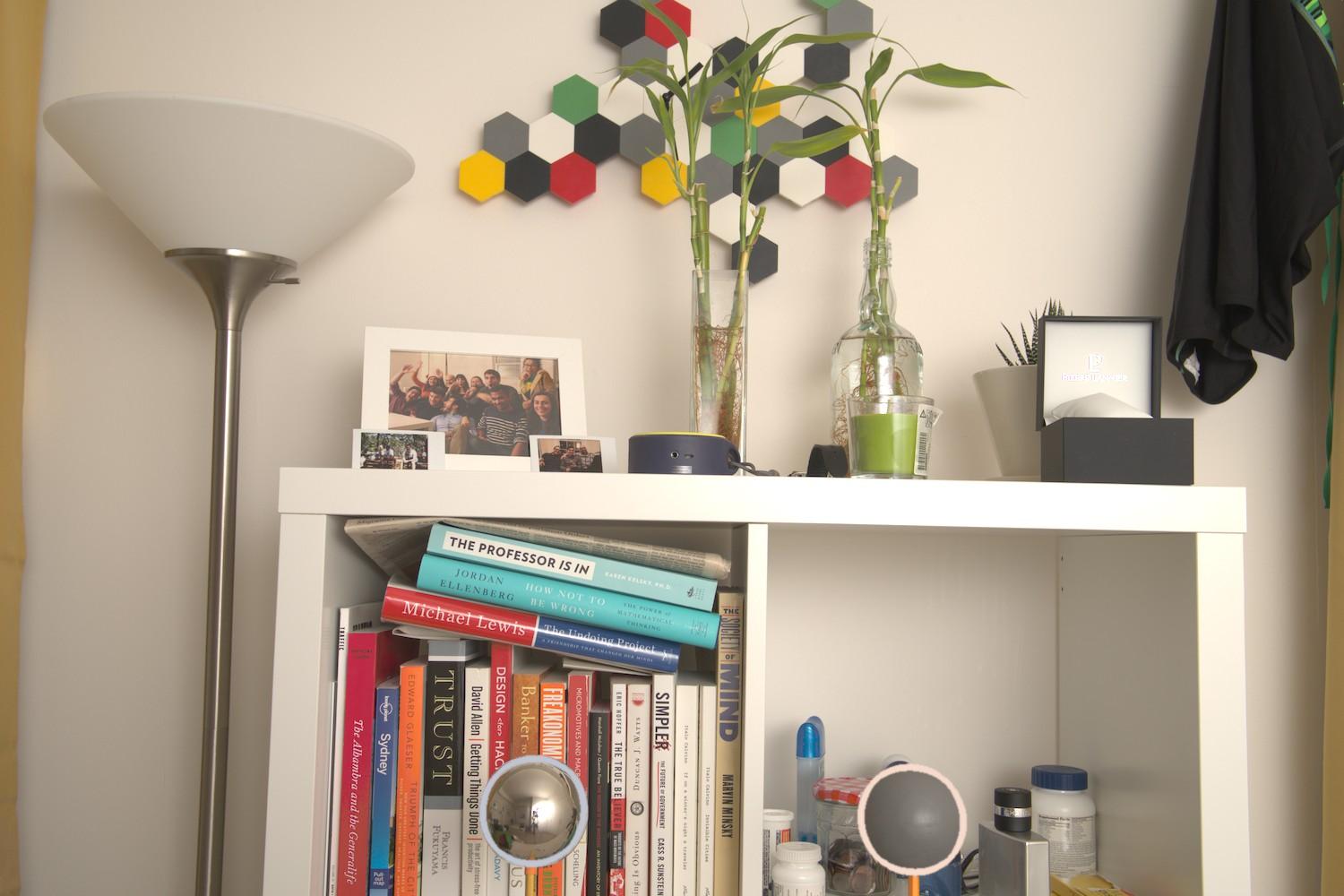} & \includegraphics[width=0.095\textwidth]{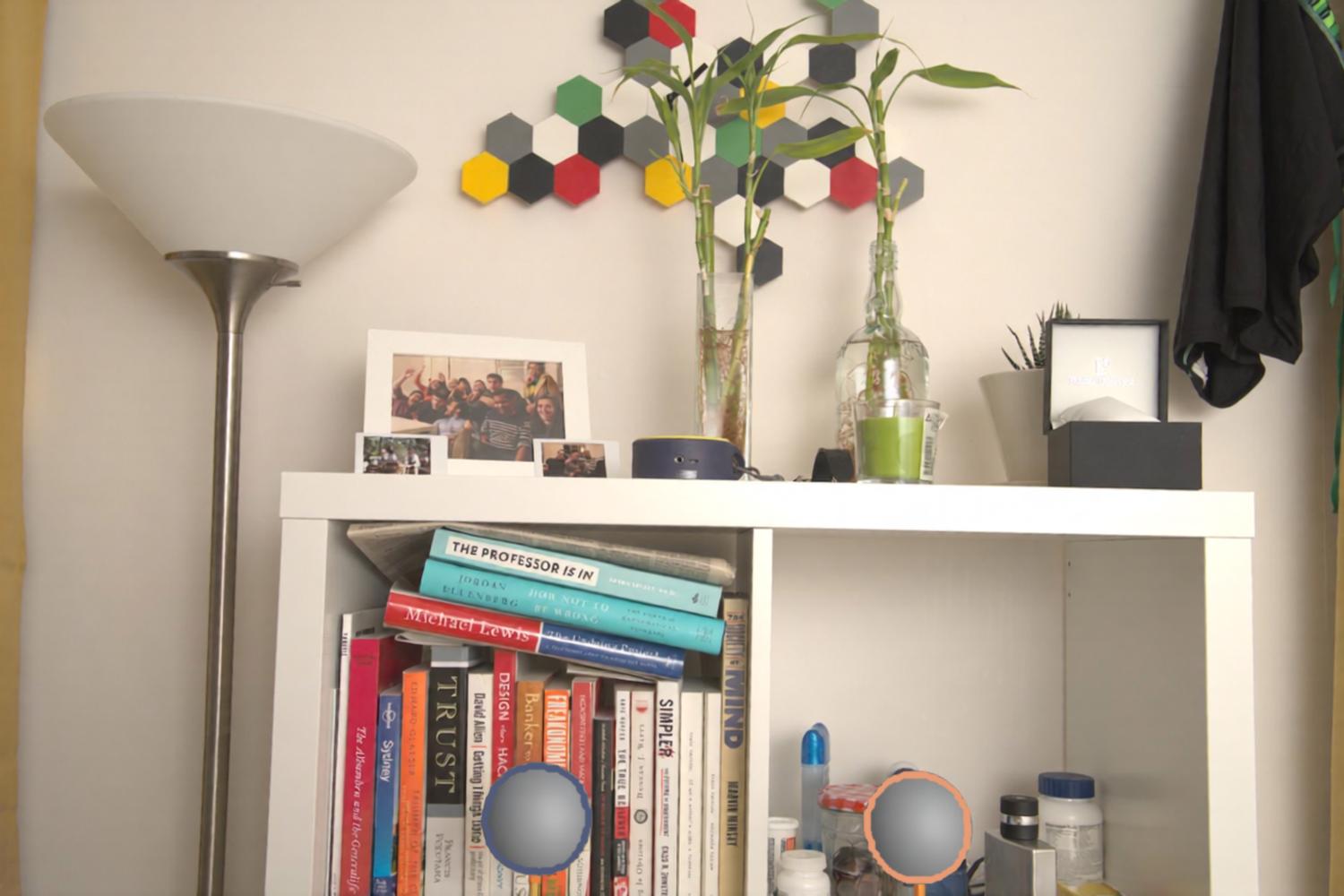} & \includegraphics[width=0.095\textwidth]{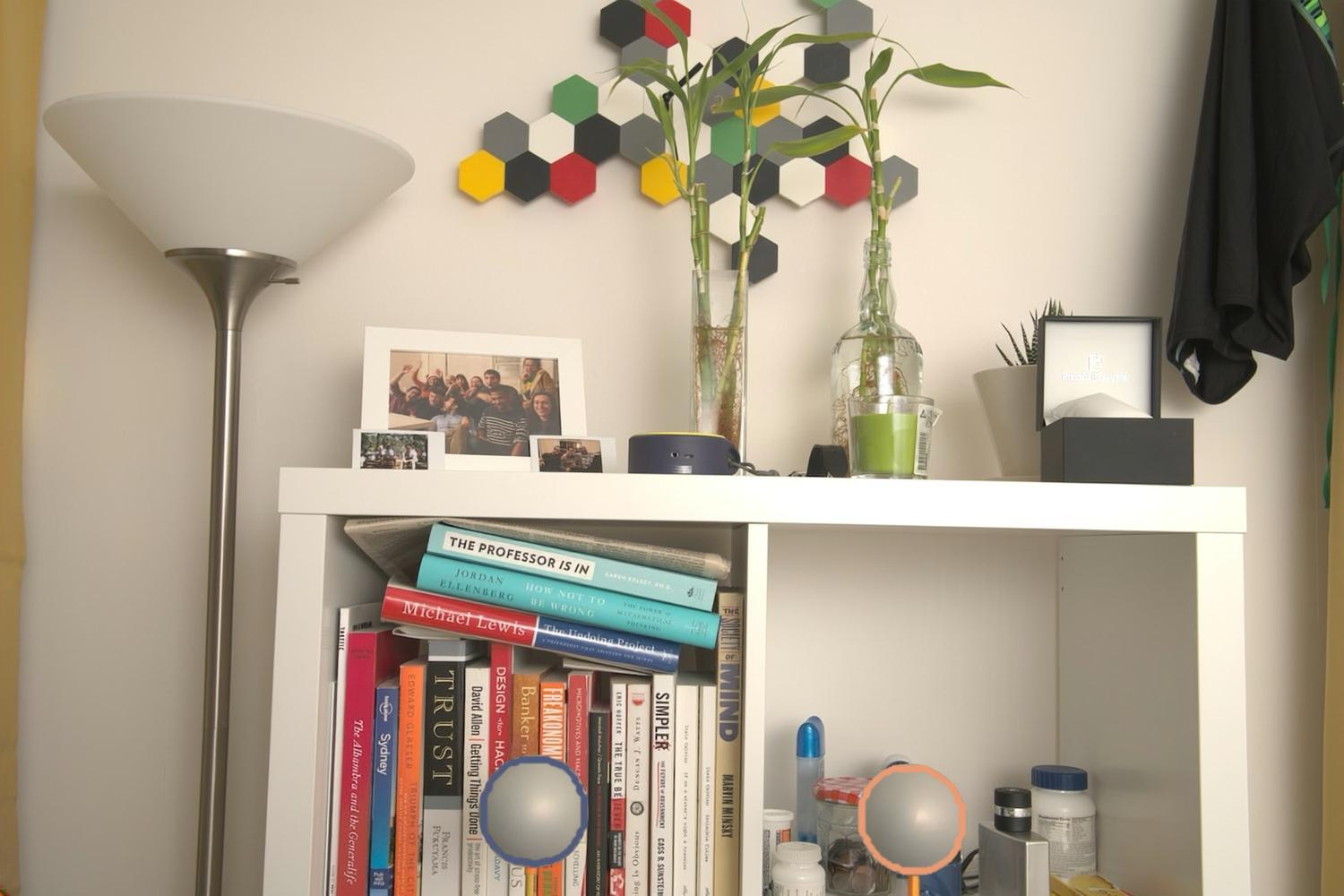} & \includegraphics[width=0.095\textwidth]{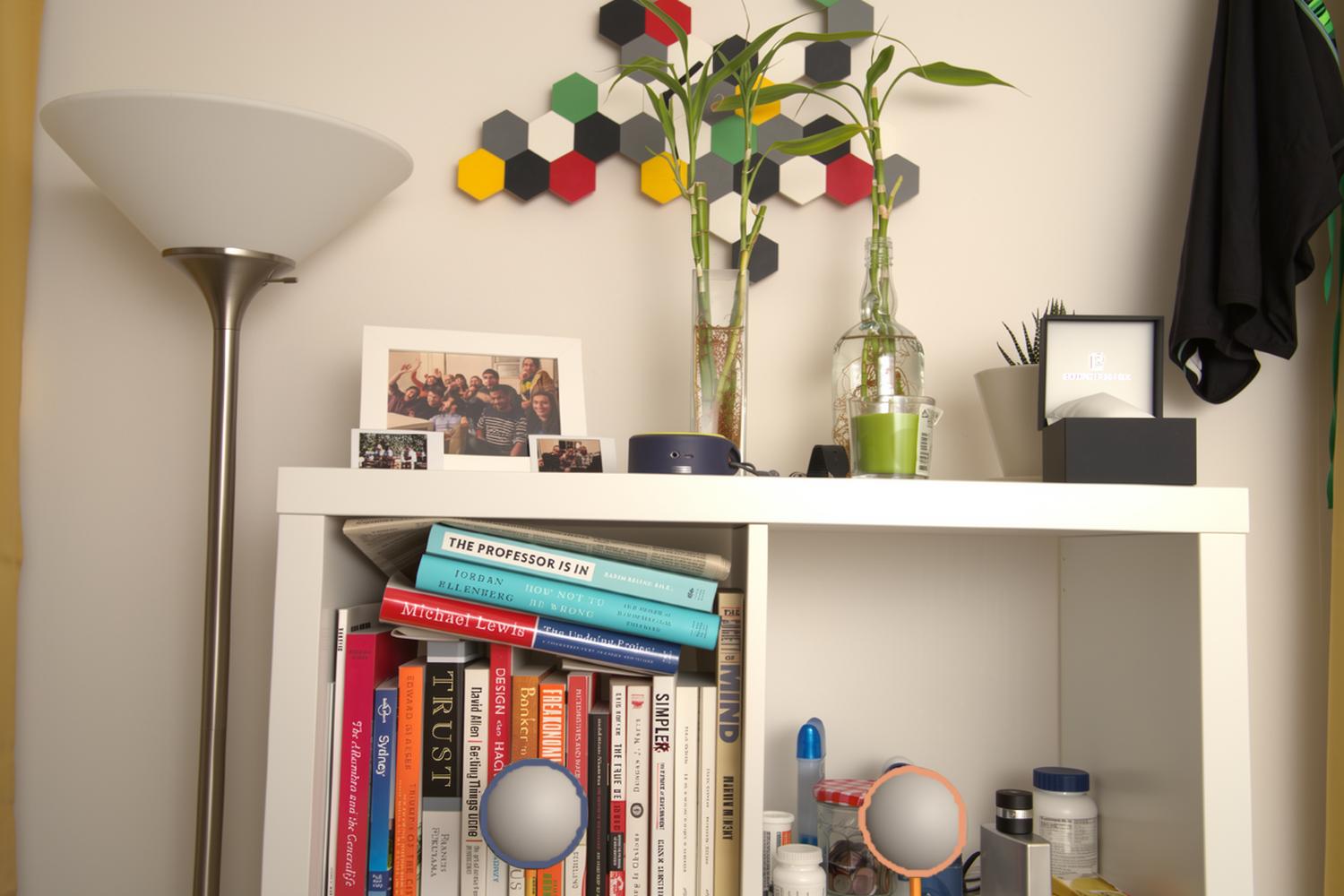} & \includegraphics[width=0.095\textwidth]{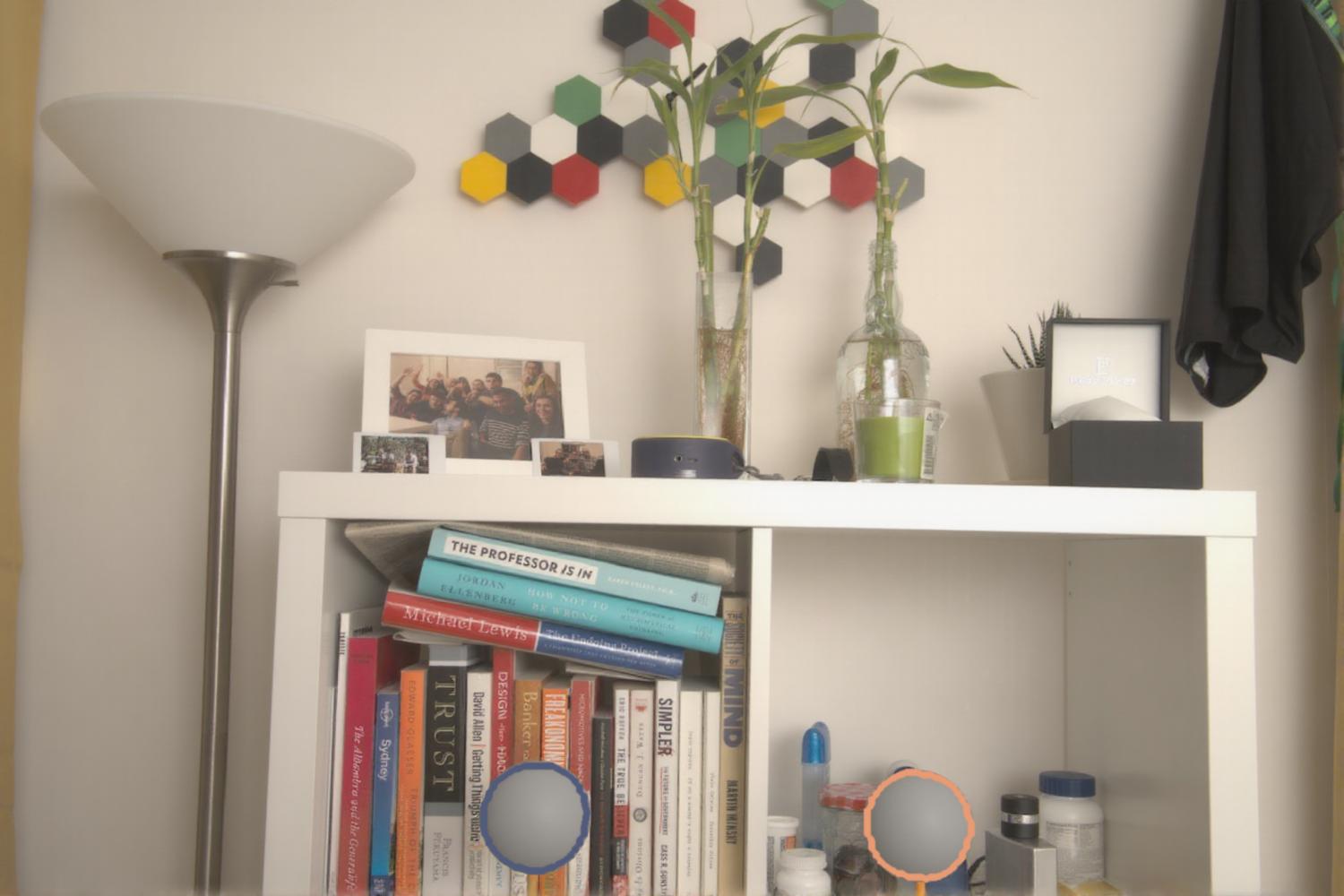} \\[1pt]
    \includegraphics[width=0.070\textwidth]{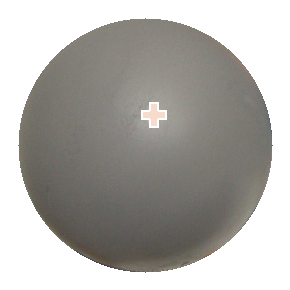} & \includegraphics[width=0.070\textwidth]{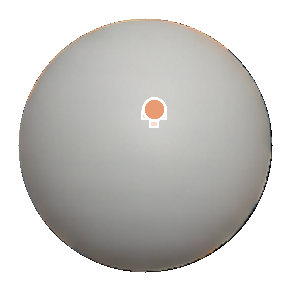} & \includegraphics[width=0.070\textwidth]{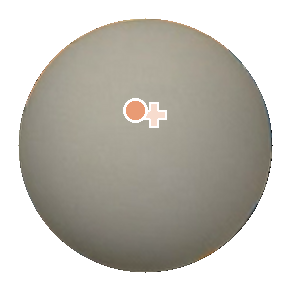} & \includegraphics[width=0.070\textwidth]{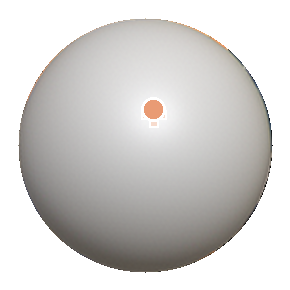} & \includegraphics[width=0.070\textwidth]{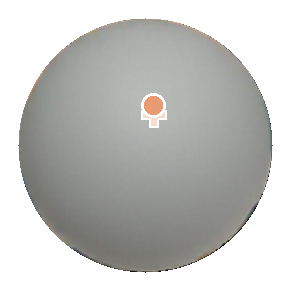} & \includegraphics[width=0.070\textwidth]{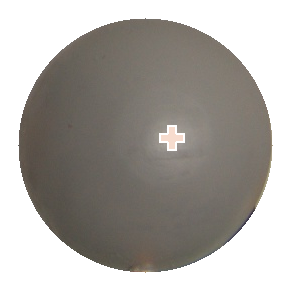} & \includegraphics[width=0.070\textwidth]{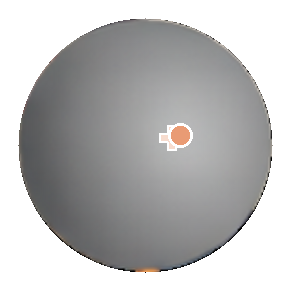} & \includegraphics[width=0.070\textwidth]{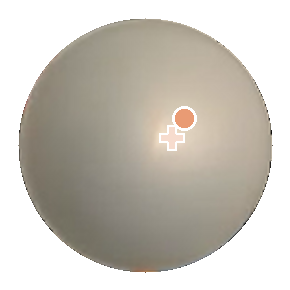} & \includegraphics[width=0.070\textwidth]{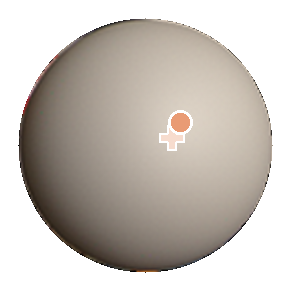} & \includegraphics[width=0.070\textwidth]{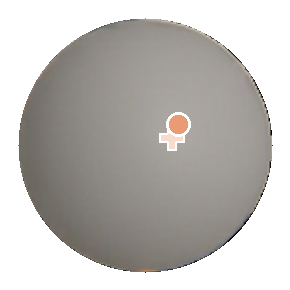} \\
    \includegraphics[width=0.070\textwidth]{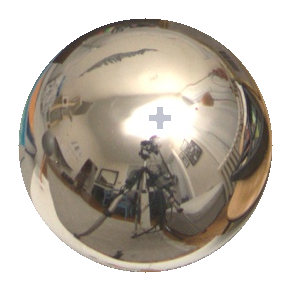} & \includegraphics[width=0.070\textwidth]{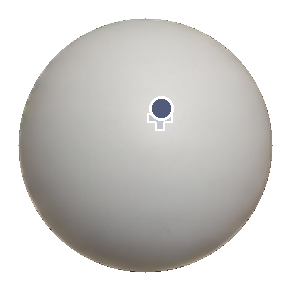} & \includegraphics[width=0.070\textwidth]{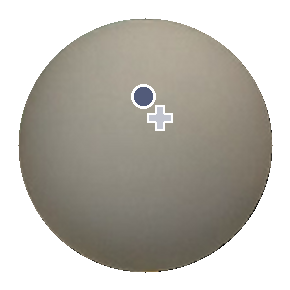} & \includegraphics[width=0.070\textwidth]{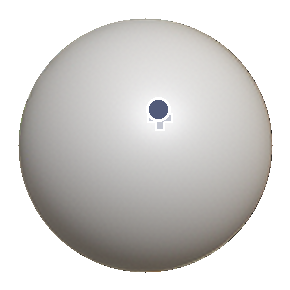} & \includegraphics[width=0.070\textwidth]{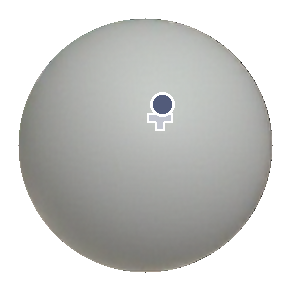} & \includegraphics[width=0.070\textwidth]{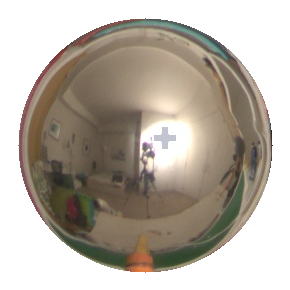} & \includegraphics[width=0.070\textwidth]{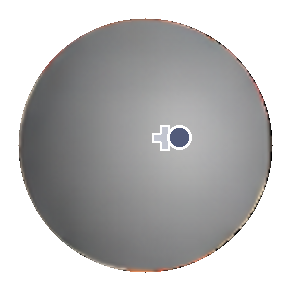} & \includegraphics[width=0.070\textwidth]{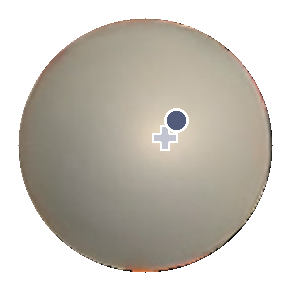} & \includegraphics[width=0.070\textwidth]{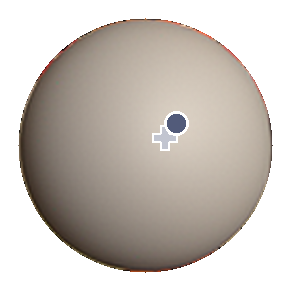} & \includegraphics[width=0.070\textwidth]{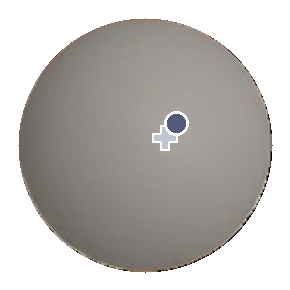} \\
  \end{tabular}
  \\[4pt]
  \begin{tabular}{@{}cccccccccc@{}}
    \textbf{GT} & \textbf{FLUX.2 [klein] 4B} & \textbf{FLUX.2 [klein] 9B} & \textbf{Photoshop} & \textbf{Nano Banana 2} & \textbf{GT} & \textbf{FLUX.2 [klein] 4B} & \textbf{FLUX.2 [klein] 9B} & \textbf{Photoshop} & \textbf{Nano Banana 2} \\
    \includegraphics[width=0.095\textwidth]{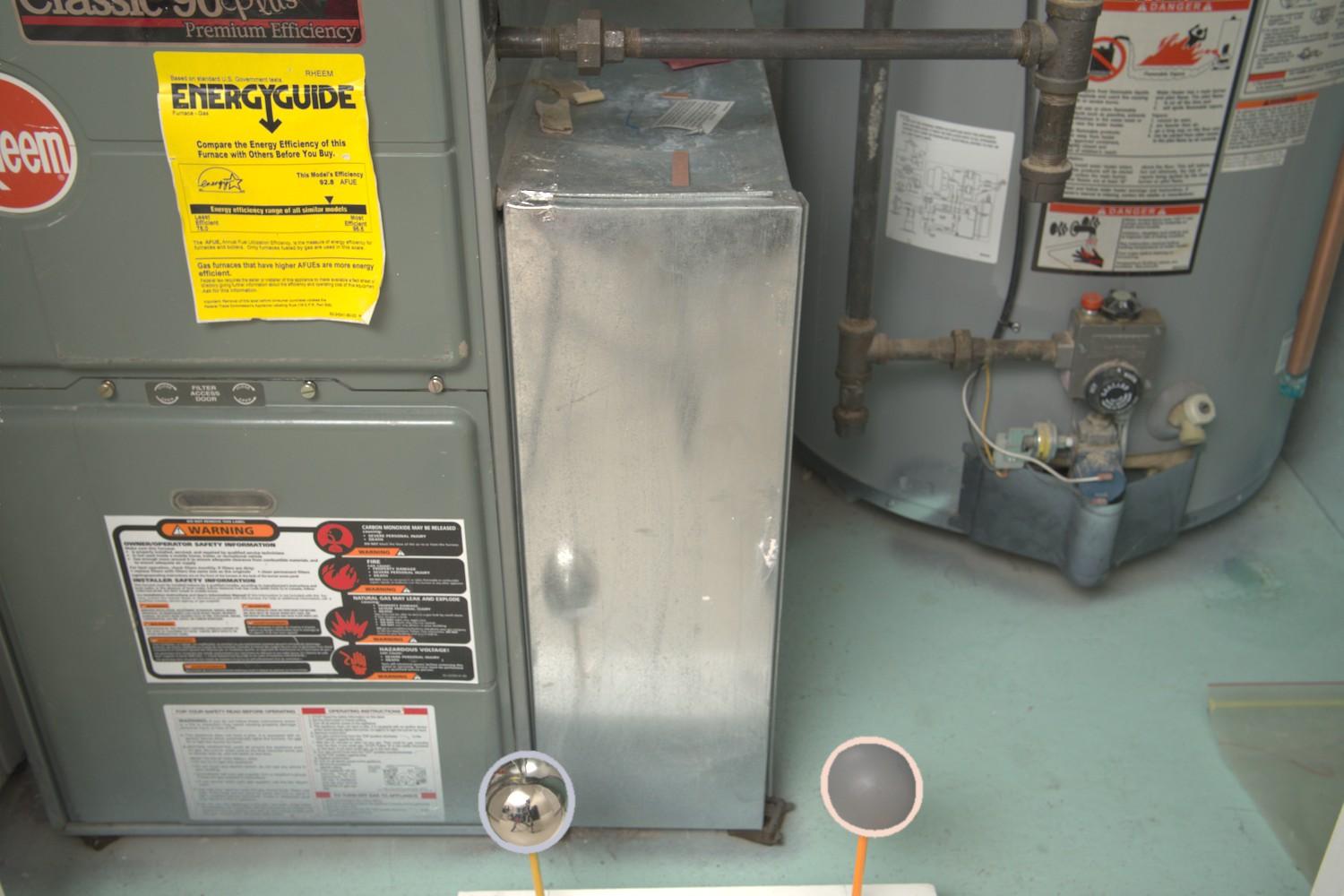} & \includegraphics[width=0.095\textwidth]{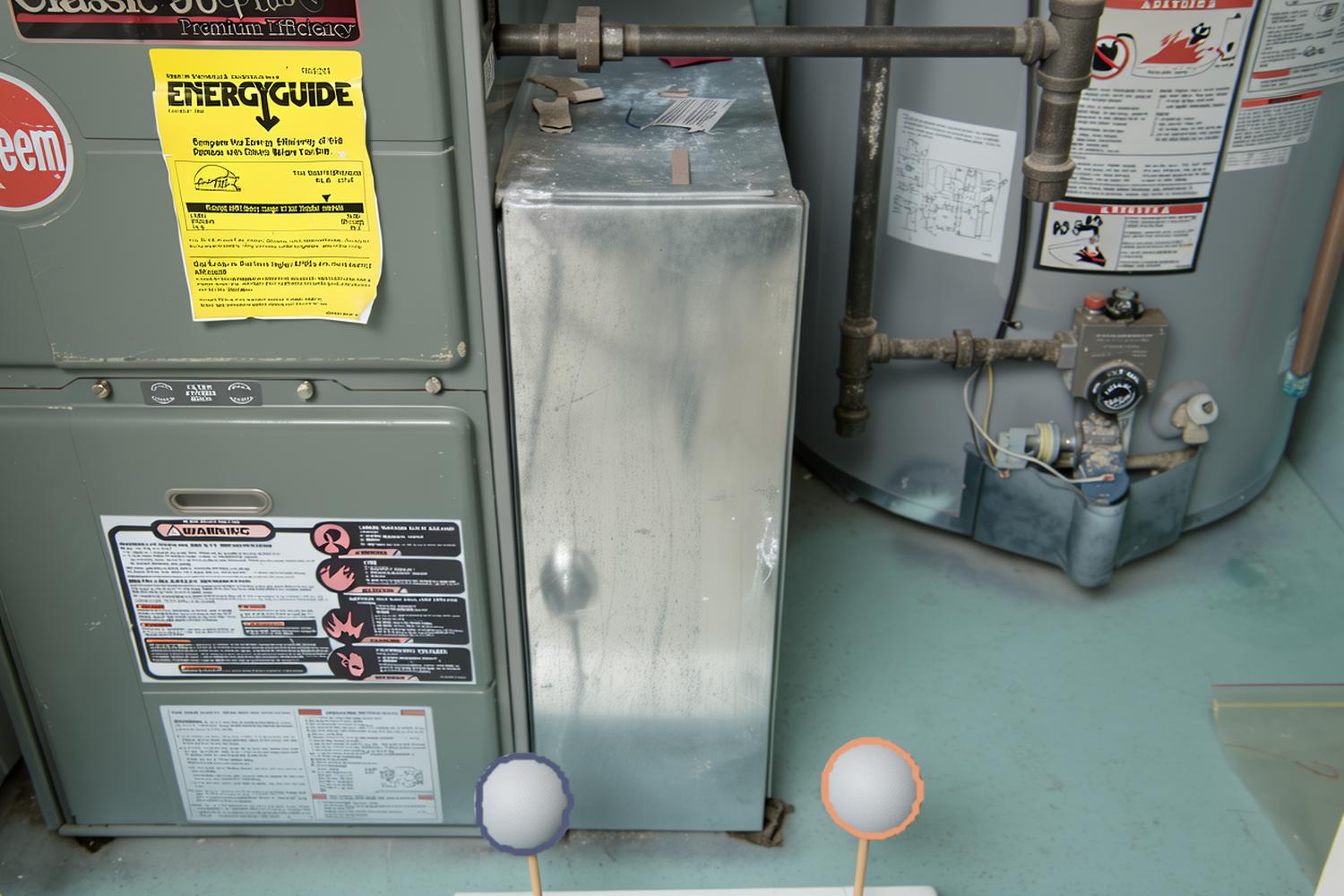} & \includegraphics[width=0.095\textwidth]{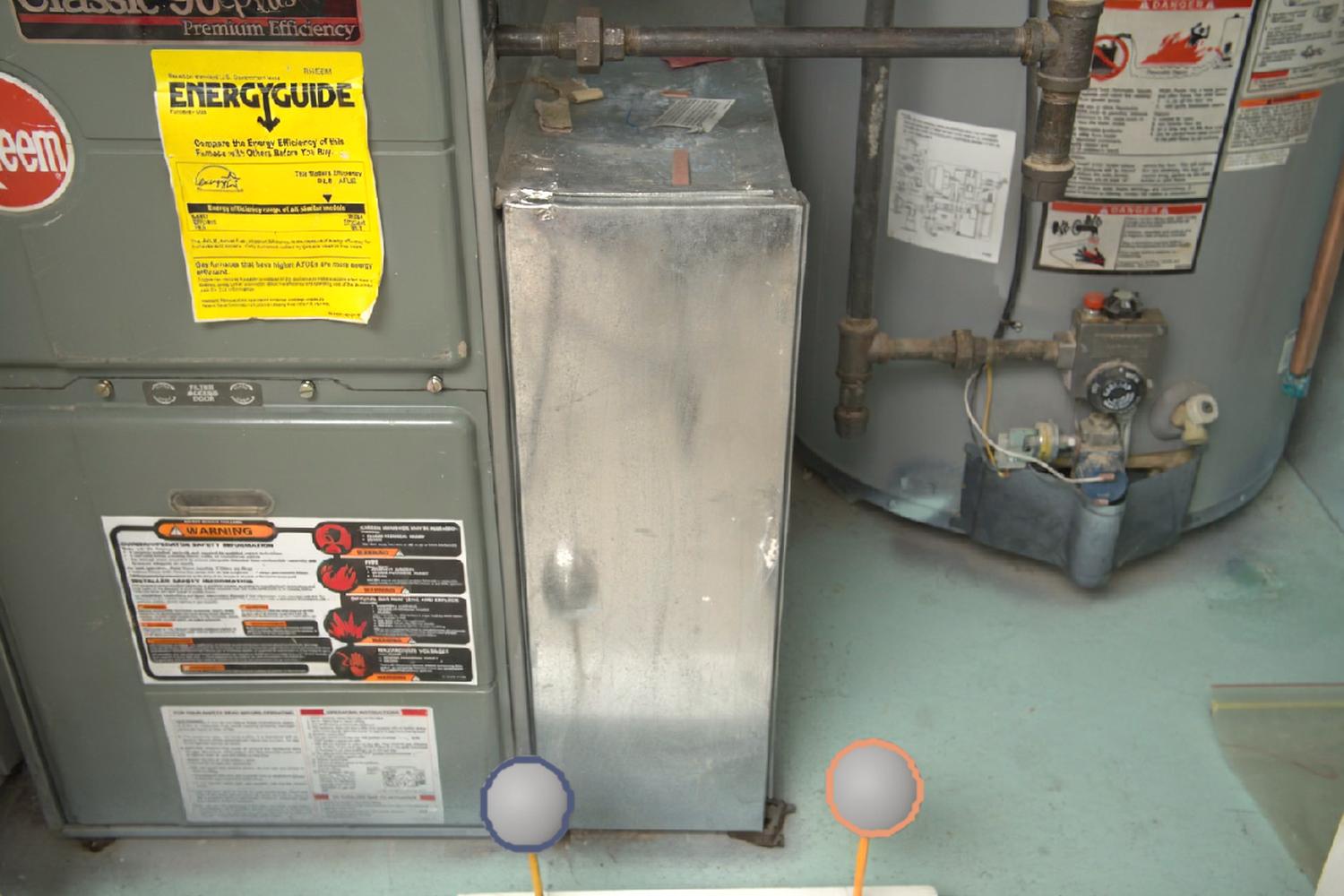} & \includegraphics[width=0.095\textwidth]{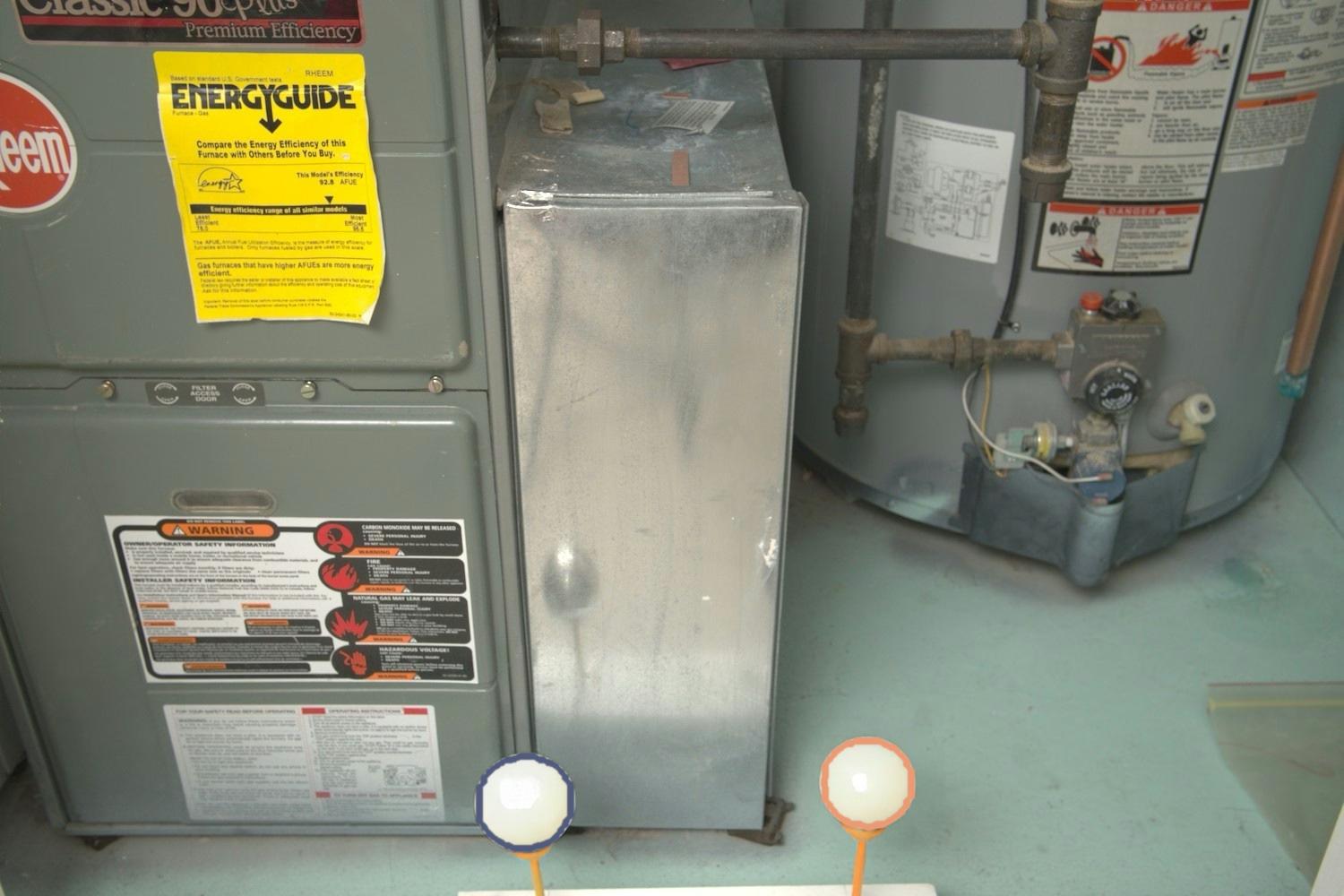} & \includegraphics[width=0.095\textwidth]{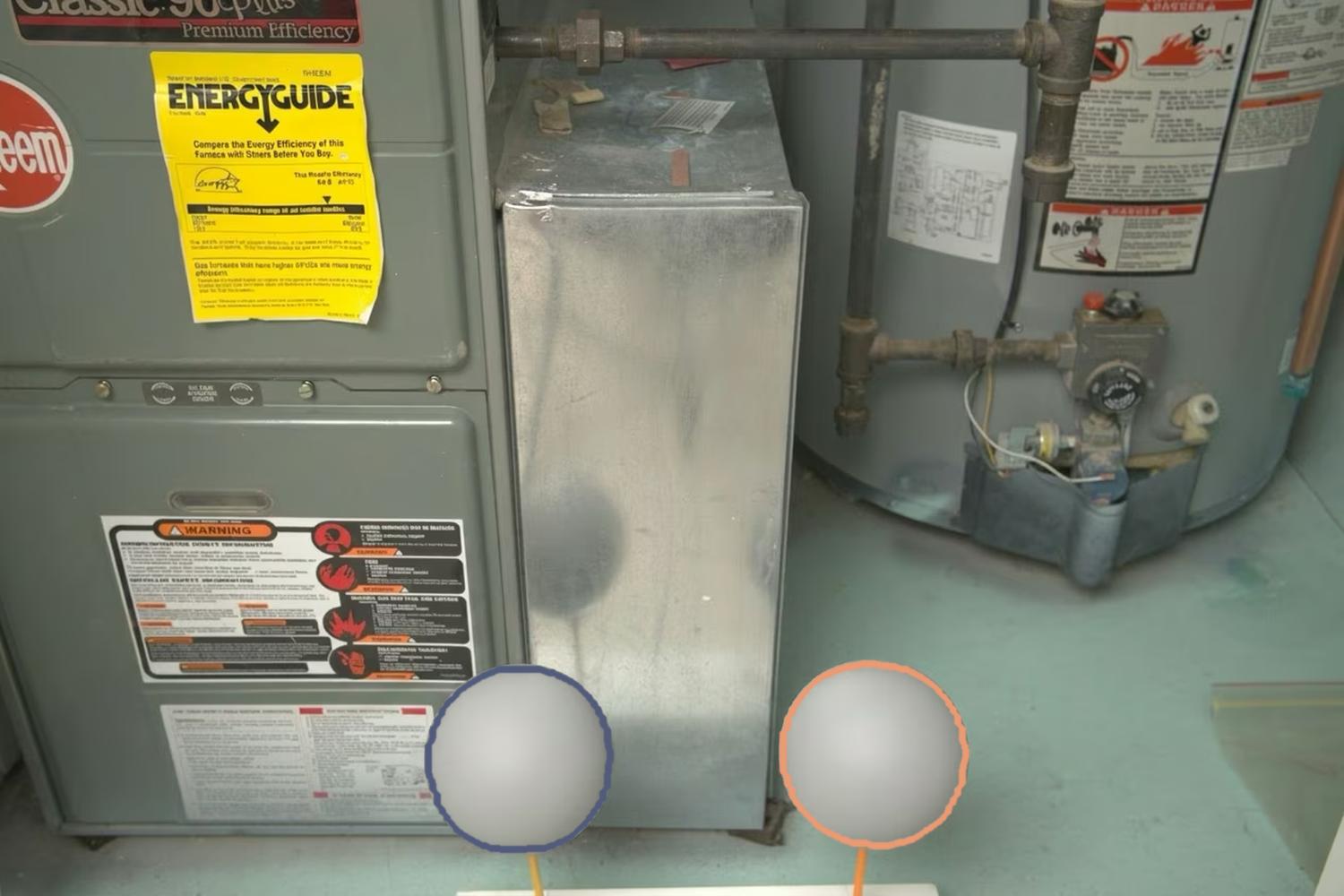} & \includegraphics[width=0.095\textwidth]{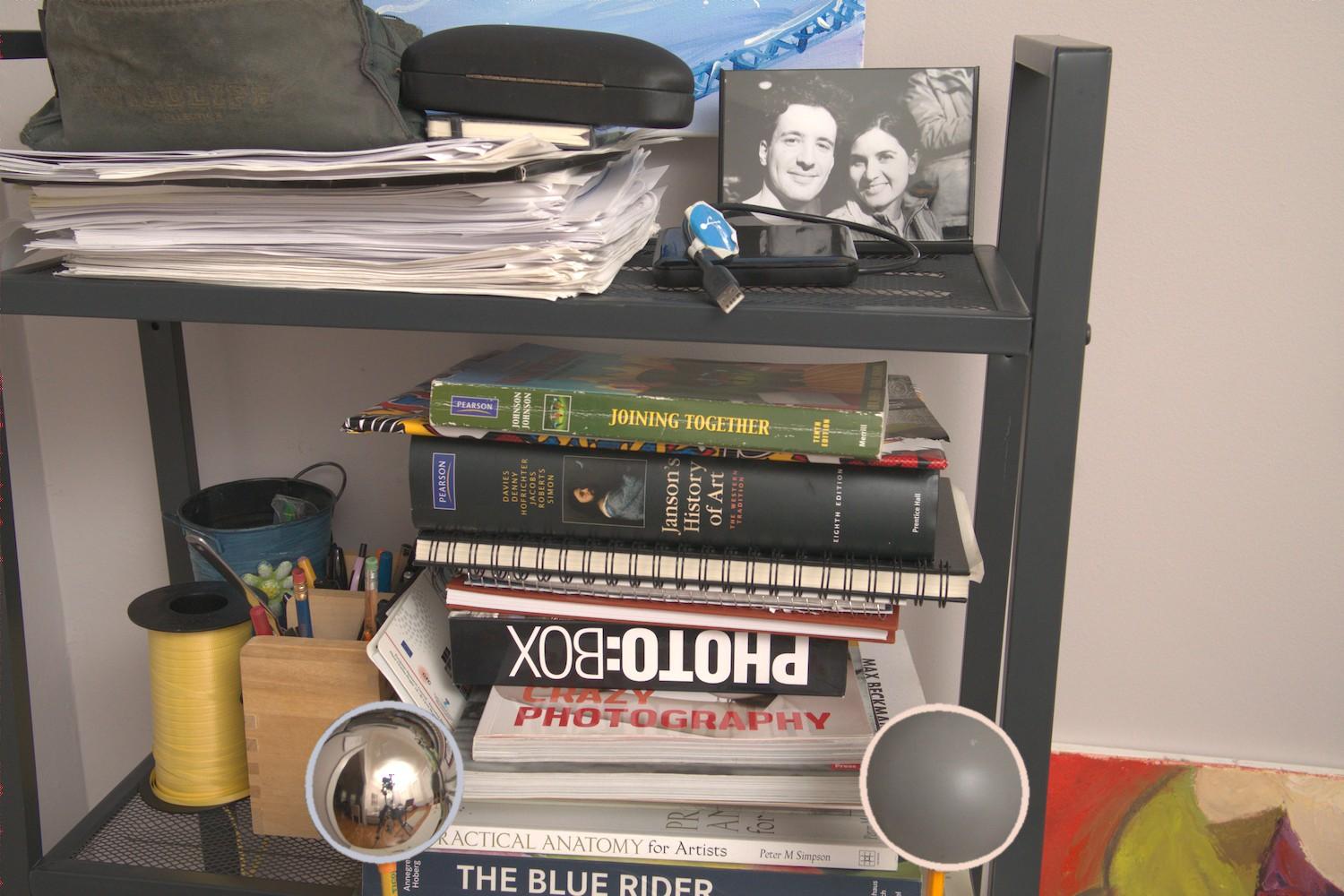} & \includegraphics[width=0.095\textwidth]{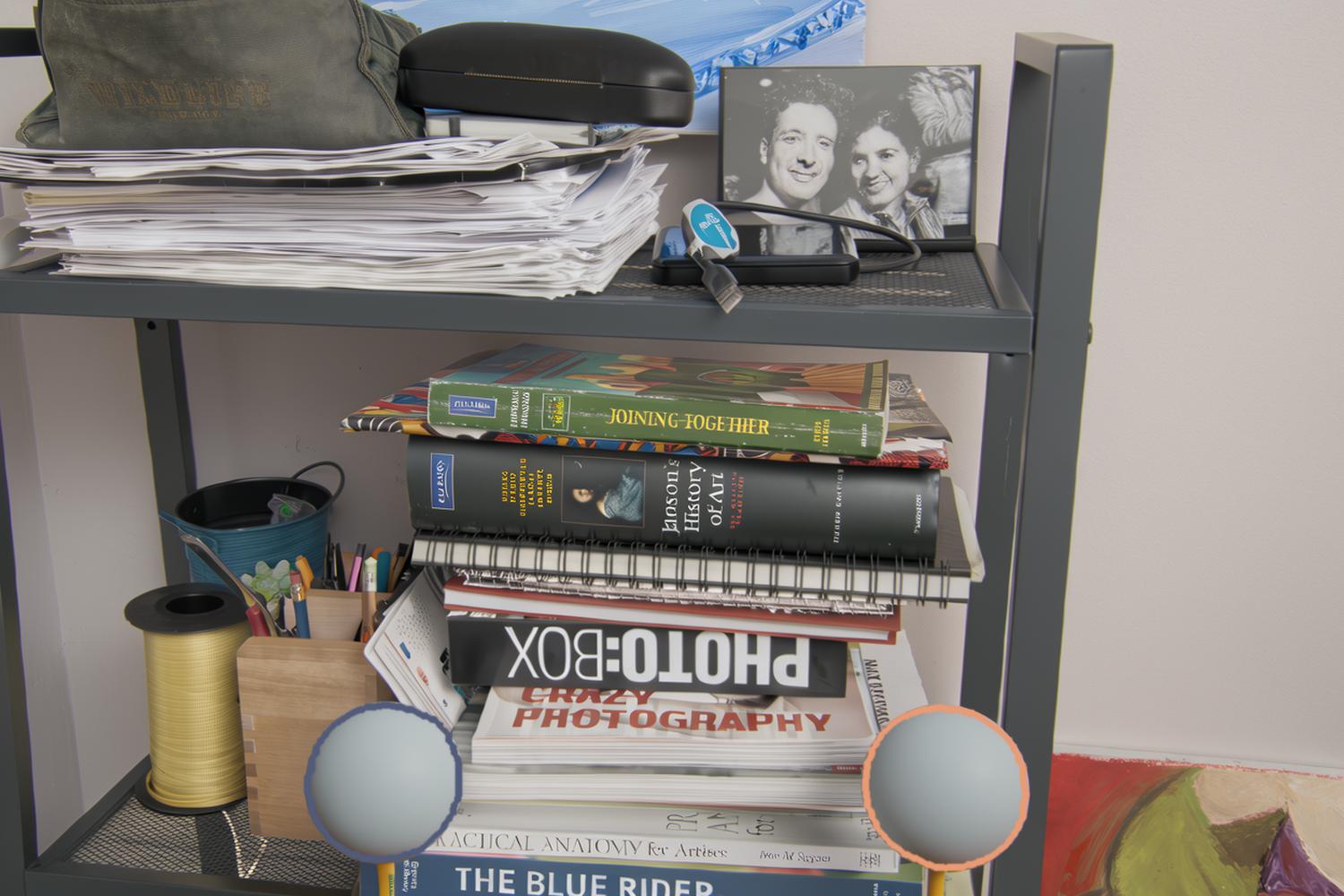} & \includegraphics[width=0.095\textwidth]{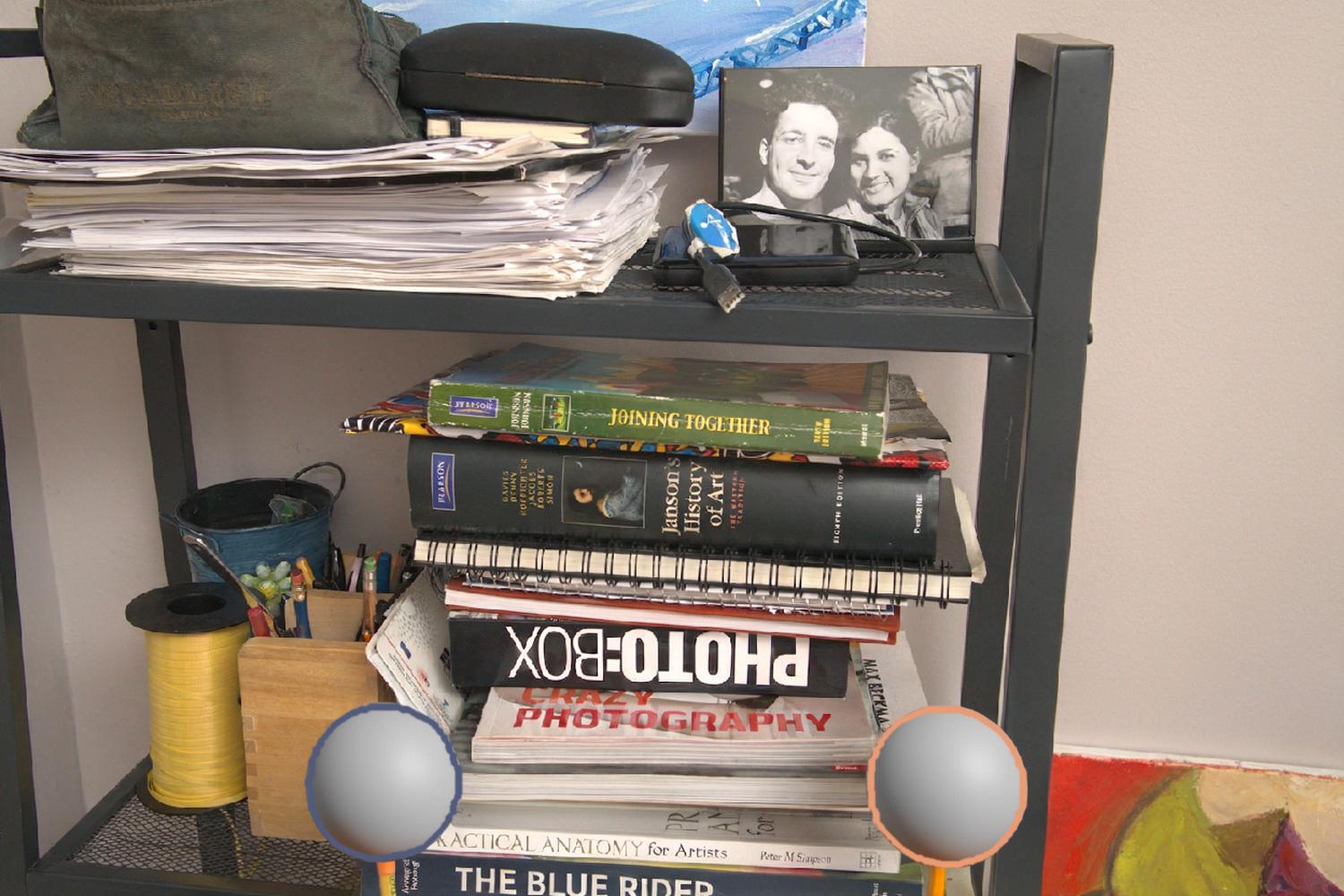} & \includegraphics[width=0.095\textwidth]{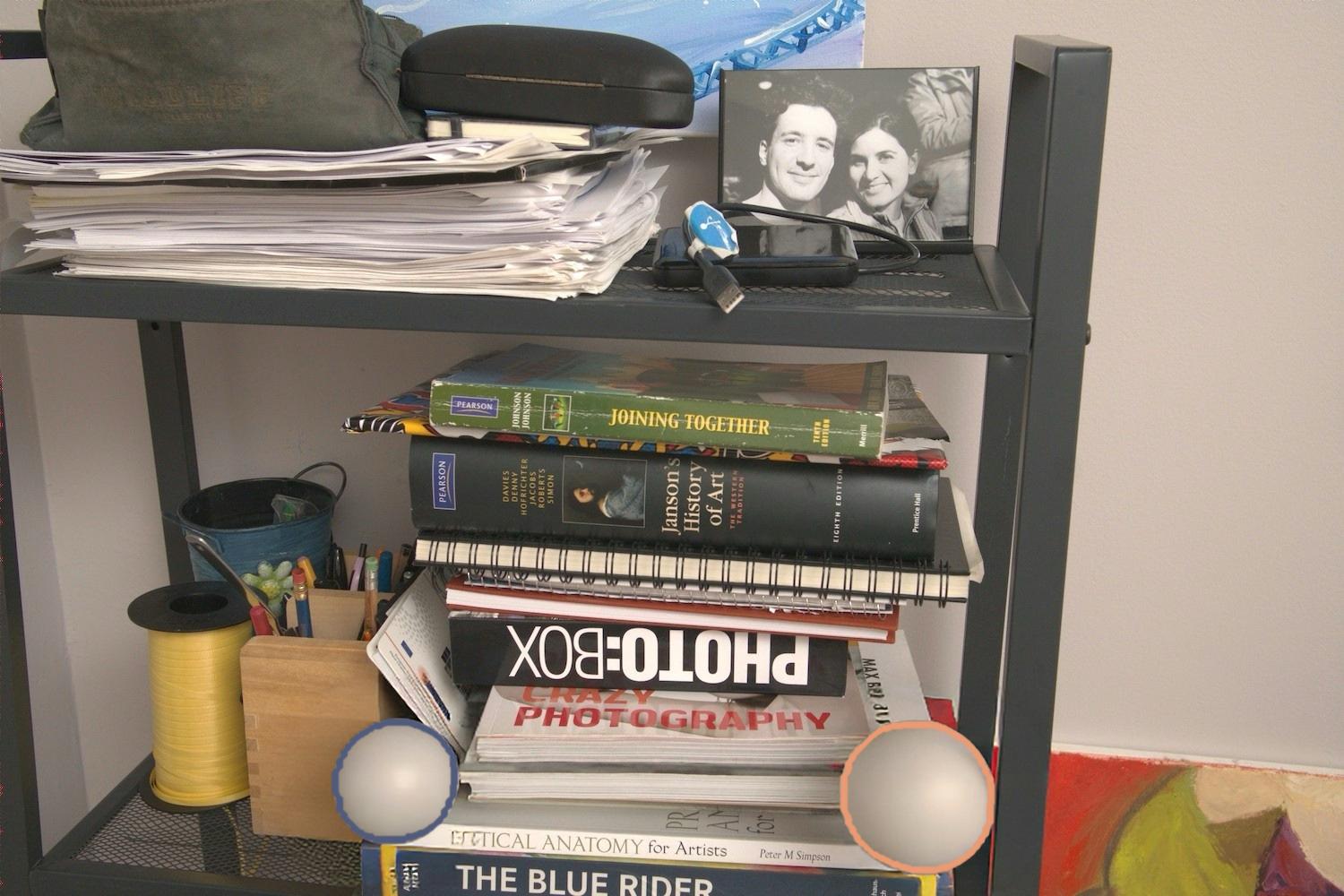} & \includegraphics[width=0.095\textwidth]{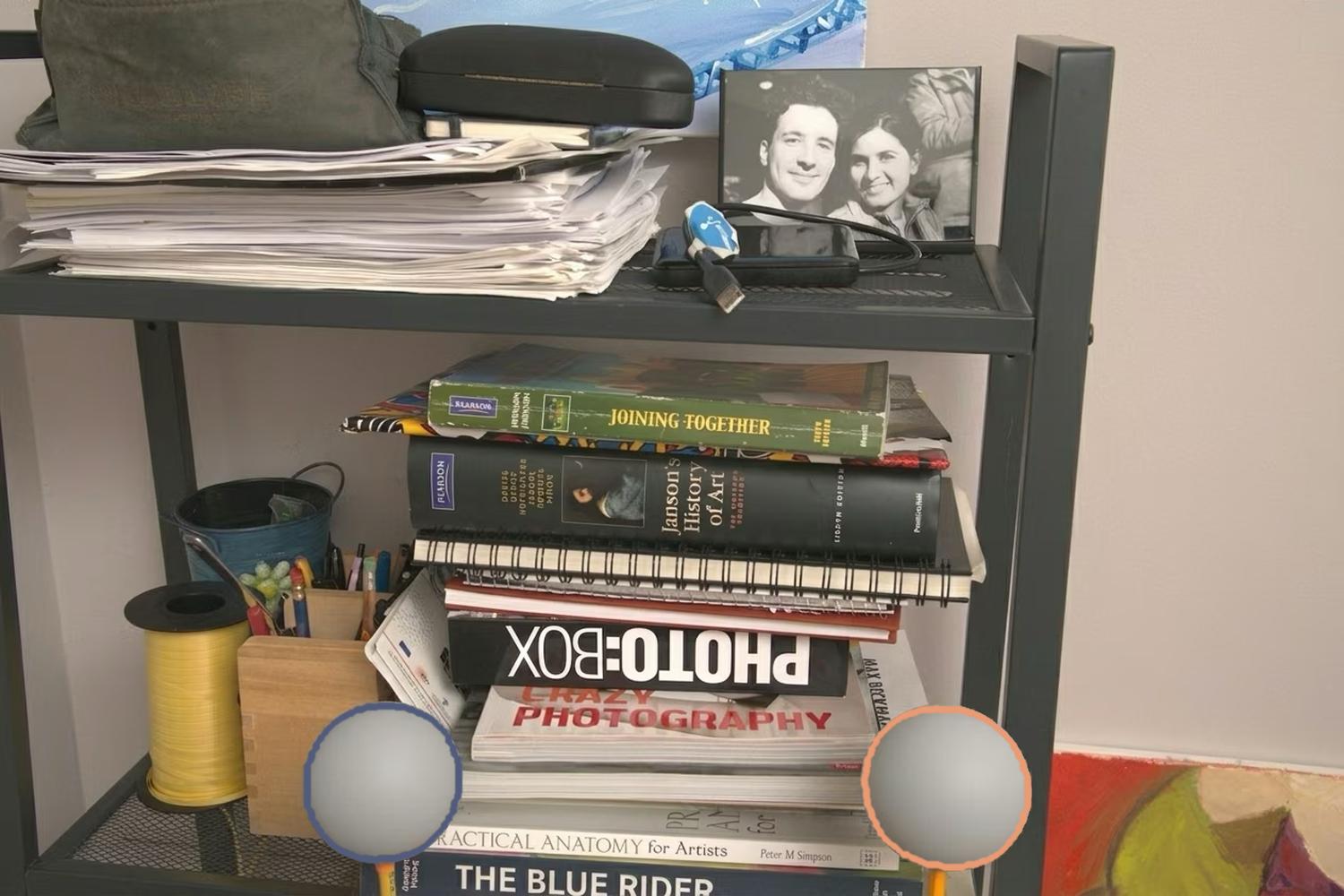} \\[1pt]
    \includegraphics[width=0.070\textwidth]{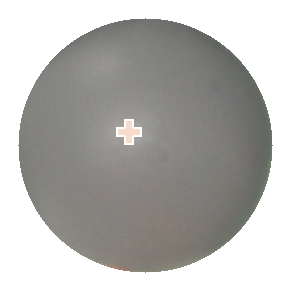} & \includegraphics[width=0.070\textwidth]{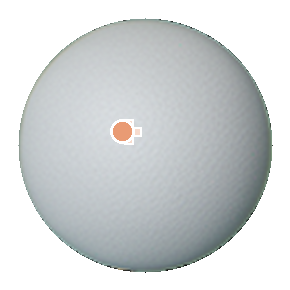} & \includegraphics[width=0.070\textwidth]{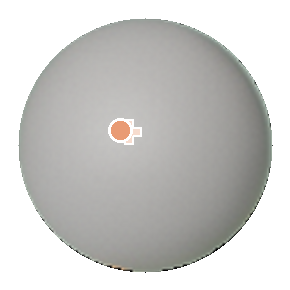} & \includegraphics[width=0.070\textwidth]{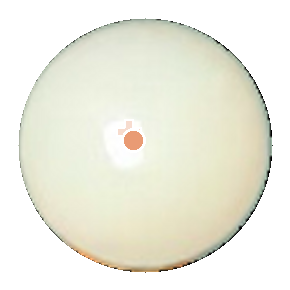} & \includegraphics[width=0.070\textwidth]{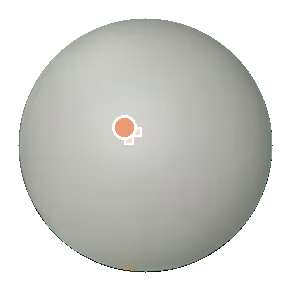} & \includegraphics[width=0.070\textwidth]{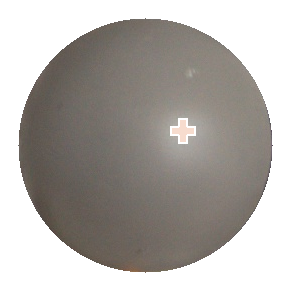} & \includegraphics[width=0.070\textwidth]{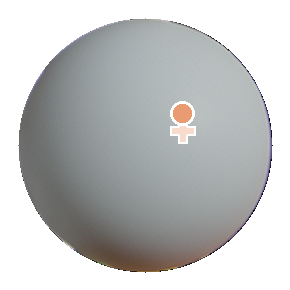} & \includegraphics[width=0.070\textwidth]{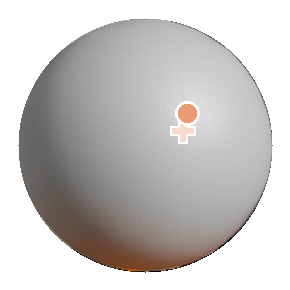} & \includegraphics[width=0.070\textwidth]{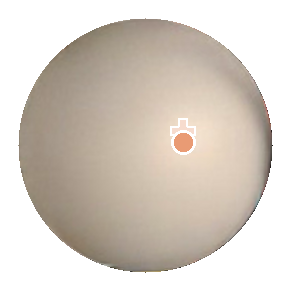} & \includegraphics[width=0.070\textwidth]{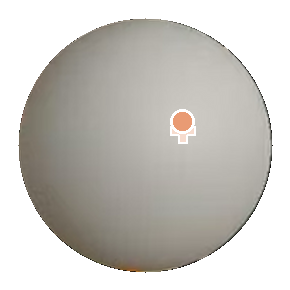} \\
    \includegraphics[width=0.070\textwidth]{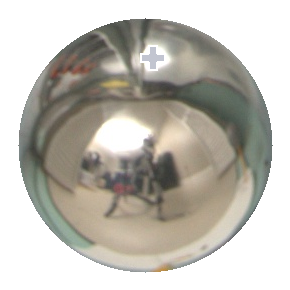} & \includegraphics[width=0.070\textwidth]{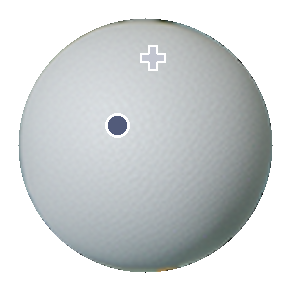} & \includegraphics[width=0.070\textwidth]{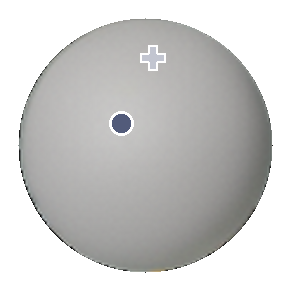} & \includegraphics[width=0.070\textwidth]{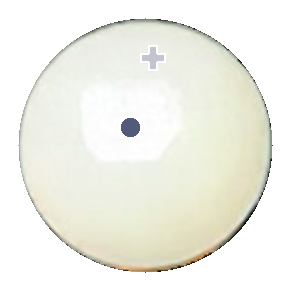} & \includegraphics[width=0.070\textwidth]{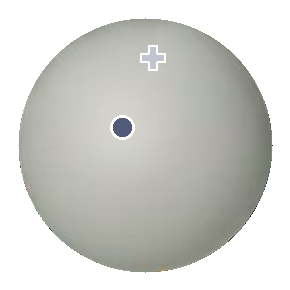} & \includegraphics[width=0.070\textwidth]{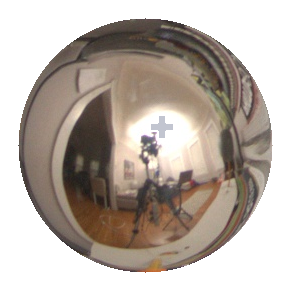} & \includegraphics[width=0.070\textwidth]{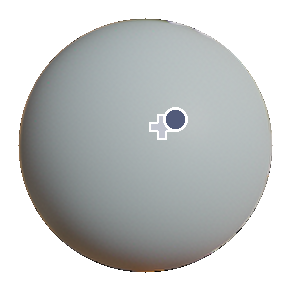} & \includegraphics[width=0.070\textwidth]{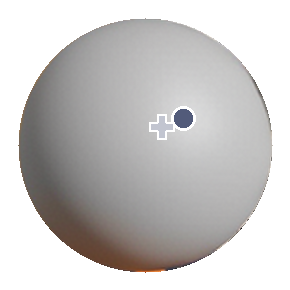} & \includegraphics[width=0.070\textwidth]{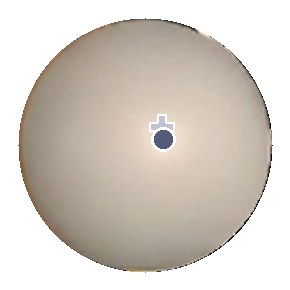} & \includegraphics[width=0.070\textwidth]{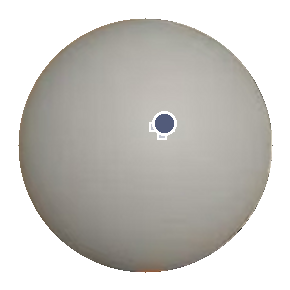} \\
  \end{tabular}
  \caption{Examples inpainted images and their light probes across \num{16} models, sorted by release date.  The first column is the reference image, with the diffuse probe circled in \couleurorangepale{light orange} and the chrome probe in \couleurbleupale{light purple}.  The generated probes are circled in \couleurorangefonce{dark orange} and \couleurbleufonce{dark blue} (inpainted over the reference probes) in the generated images from different models (all other columns).  The light direction is shown on the diffuse reference probe (second row) by the \couleurorangepale{light orange} cross and in the \couleurorangefonce{dark orange} circle for the generated and for the chrome reference probe (third row), the reference light direction is indicated by a \couleurbleupale{light purple} cross and the generated light direction by a \couleurbleufonce{dark blue} circle.  The marker contour is white if the direction is towards the camera and black if it is away from the camera.}
  \Description{Examples of generated images, their probes, and the measured light directions for different models and different scenes.}
  \label{fig:qualitative_grid}
\end{figure*}

Additional examples of probes, for different models and scenes, are shown in \cref{fig:qualitative_grid}, showing the generated image, and both probes extracted, with their estimated light direction, as well as the reference image and probes to compare.

\section{Further evaluation}
\label{suppsec:evaluation}

\subsection{Light parameter optimisation evaluation}
\label{suppsubsec:evaluation_LD}

\begin{figure}[t]
\centering
\setlength\tabcolsep{7pt}
\def\colW{0.19}
\begin{tabularx}{\linewidth}{cccc}

    \raisebox{0pt}{\includegraphics[width=\colW\linewidth]{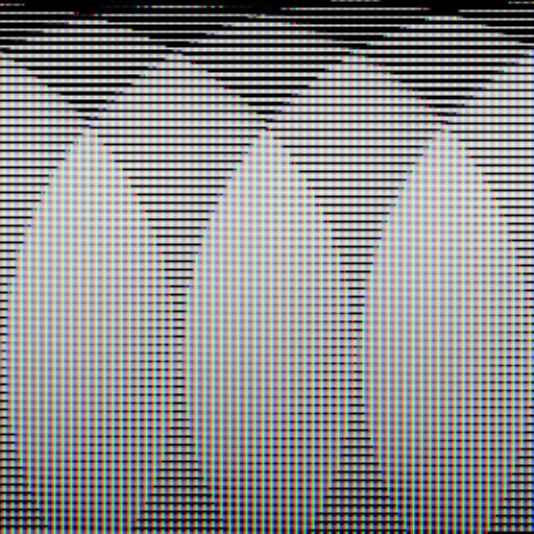}} &\raisebox{0pt}{\includegraphics[width=\colW\linewidth]{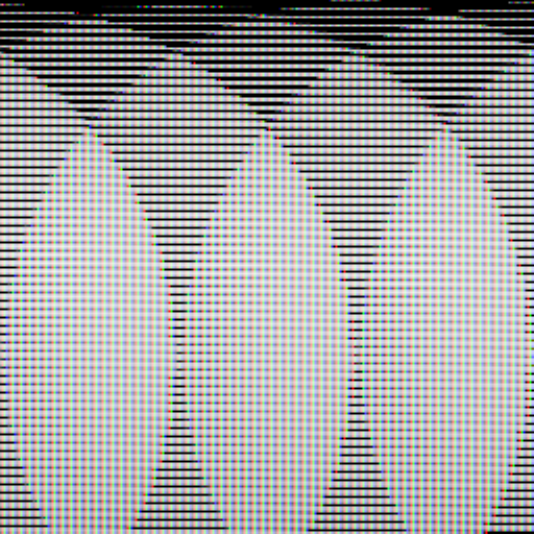}} &\raisebox{0pt}{\includegraphics[width=\colW\linewidth]{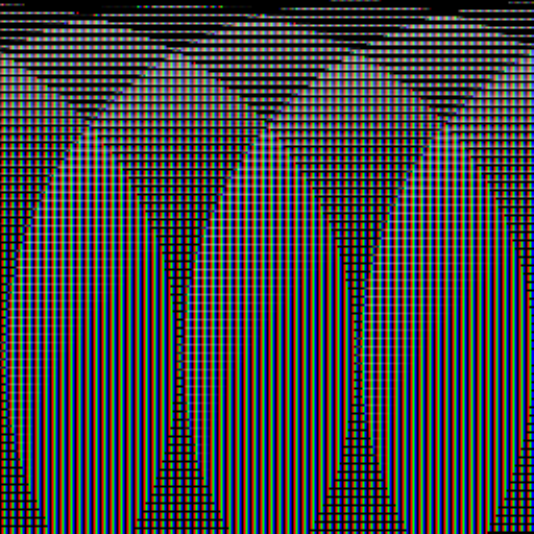}} &\raisebox{0pt}{\includegraphics[width=\colW\linewidth]{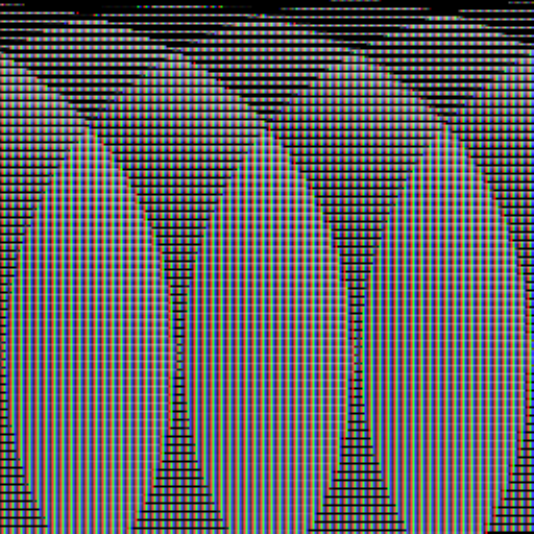}}
    \\

\end{tabularx}
\caption{Examples of probes from our PBR test set used to find the uncertainty of our inverse rendering stage. The test set was created in Blender~\citep{blender} and contains a range of probe material types, light directions and intensities.}
\Description{Examples of our synthetic probe dataset.}
\label{fig:exemples_synthexamples}
\end{figure}

For the proxy evaluation of the estimated light direction from our light parameter optimisation stage, we created a physics-based rendering (PBR) dataset, shown in \cref{fig:exemples_synthexamples}, in Blender~\cite{blender} of a sphere with varying material properties, illuminated by a single light source from different directions. By creating a synthetic test set, we can use ground-truth labels to evaluate our light direction estimation. We test our inverse rendering optimiser on the synthetic dataset in HDR and low-dynamic range (LDR), as the Multi-Illumination dataset provides probes in HDR and the generative image models output LDR. We also evaluate on the Multi-Illumination dataset, comparing against both the specular highlight labels and human-annotated labels on a small split of \num{575} images. Our results in \cref{tab:ldo_median_ae} show that our method is more accurate on HDR data than LDR, and that it outperforms human annotations on a sample of the real dataset.
\begin{table}[h]
\caption{Performance of our light parameter optimisation stage on a synthetic PBR dataset and the Multi-Illumination dataset~\citep{murmann_dataset_2019}. Additionally, we provided results from a human annotator for comparison. We report the median and standard deviation of the angular error.}
\label{tab:ldo_median_ae}
\setlength{\tabcolsep}{5pt}
\footnotesize
\begin{tabularx}{\linewidth}{l >{\raggedright\arraybackslash}X l r r}
\toprule
LD Estimator & Dataset & LD Labels & Median Ang Err [\textdegree]$\downarrow$ & $\sigma$ [\textdegree]$\downarrow$ \\
\midrule
Vanilla & Synthetic HDR & Actual & \textbf{2.43} & 6.59 \\
 & Synthetic LDR & Actual & 2.55 & 6.72\\
\midrule
 & Multi-Illumination HDR & Specular & \textbf{9.35} & 24.80 \\
\midrule
 & Multi-Illumination  [just annotated] & Specular  & \textbf{8.70} & \\
Annotated & Multi-Illumination  [just annotated] & Specular  & 9.40 & \\

\bottomrule
\end{tabularx}
\end{table}

\subsection{Probe generation evaluation}
\label{suppsubsec:evaluation_probegen}

\subsubsection{Probe generation yield}
\label{suppsubsubsec:evaluation_probegen_probesuc}

\begin{figure}[h]
  \centering
  \includegraphics[width=\linewidth]{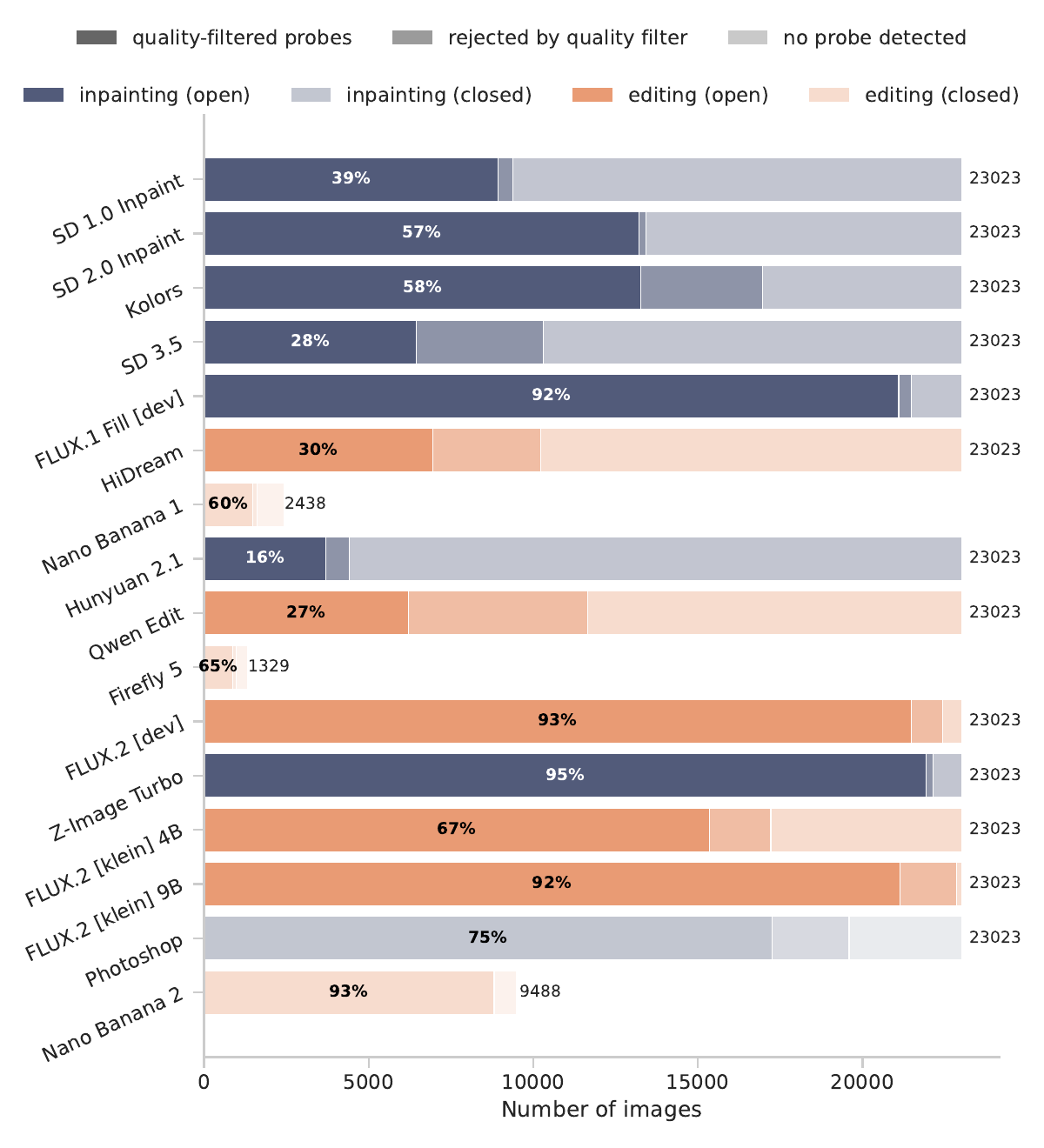}
  \caption{Total number of images generated, probes generated, and usable probes generated by each model. This demonstrates each model's ability to adhere to the prompt by generating a probe into the scene.}
  \Description{Total number of images in the dataset per model.}
  \label{fig:number_data_per_model_per_step}
\end{figure}

At various stages in our pipeline, we evaluate whether a generated light probe was correctly placed and the quality of these probes. We report the number of usable probes generated by each model as a percentage of the images they were inferred on  \cref{fig:number_data_per_model_per_step}. Of note, \qwen generated high-quality probes that ranked 6th in median angular error; however, it struggles to generate probes at the requested position and often embeds them within the scene or partially outside the field-of-view of the image. The stable diffusion family of models, \hunyuan and \hidream, had a low success rate of generating a probe in the correct location, which correlates with the lowest median angular errors. Examples of unsuitable probes are featured in \cref{fig:exemples_badinpaints}. There was no correlation between the remaining model's ability to generate a usable probe and to generate the correct lighting. Neither was there any correlation between inpainting and image-editing models and the percentage of usable probes generated.

\subsubsection{Prompt adherence}
\label{suppsubsubsec:evaluation_probegen_prompt}

To investigate how the text prompt influences the ability of generative image models to accurately recreate a scene's lighting, we test three additional prompts on a selection of models (\fluxone, \qwen, \zimageturbo, and \kolors) using a subset of \num{55} scenes from the dataset.
The prompts were designed with varying levels of complexity, from no discussion of the lighting task or the spheres' material properties to explicitly stating the task, geometric and material properties of the spheres, and the location to generate them:
\begin{promptbox}
\paragraph{\textbf{Simple}}: \texttt{"two grey spheres"}
\end{promptbox}
\begin{promptbox}
\paragraph{\textbf{Medium}}: \texttt{"inpaint two round light grey matte spheres with shading matching the scene lighting"}
\end{promptbox}
\begin{promptbox}
\paragraph{\textbf{Complex}}: \texttt{"Place two light grey Lambertian spheres over the circular grey mask regions. Each sphere is perfectly round, fully in frame, and shaded consistently with the existing scene lighting."}
\end{promptbox}
\begin{promptbox}
\paragraph{\textbf{Ours}}: \texttt{"Inpaint two perfectly round light grey matte spheres, fully visible, with smooth, uniform Lambertian shading that accurately responds to the existing scene lighting. The spheres are directly in front of the camera and appear in front of everything. There should be two spheres over the two circular grey masks in the image."}
\end{promptbox}
No statistically significant differences in light direction accuracy were found across the models when using different prompts. For some models, usable probe yield varied across prompts. Therefore, it was not possible to determine whether the prompt affected the probe generation or the model's understanding of lighting. We found our prompt to be sufficiently robust across all models for the task and included the necessary probe location information that image-editing models require.

\subsection{Model size analysis}
\begin{figure}[t!]
  \centering
  \includegraphics[width=\linewidth]{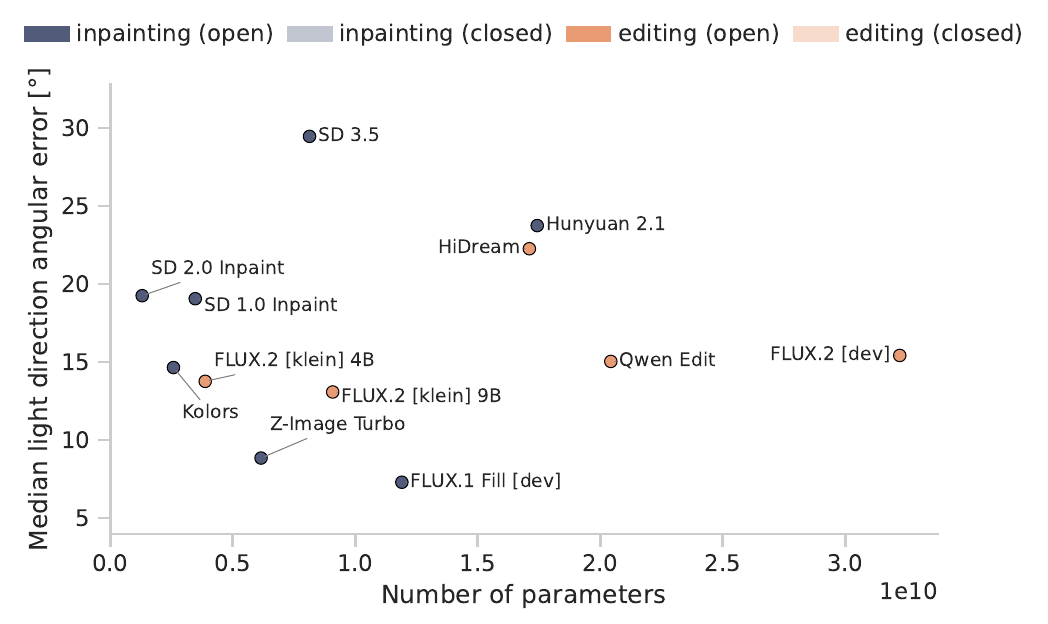}
  \caption{Median angular error of the light direction as a function of the number of parameters for each model.  The markers are coloured by type (inpainting or editing) and availability (open or closed).}
  \Description{Median angular error of the light direction as a function of the number of parameters for each model.}
  \label{fig:median_angular_error_vs_parameters}
\end{figure}

\Cref{fig:median_angular_error_vs_parameters} shows the relation between the number of parameters of the model and the median angular error in light direction estimation. As shown in the figure, no clear correlation emerges, with the best (\fluxone) and worst (\SDthree) models having similar numbers of parameters.

\begin{figure}[t!]
  \centering
  \includegraphics[width=\linewidth]{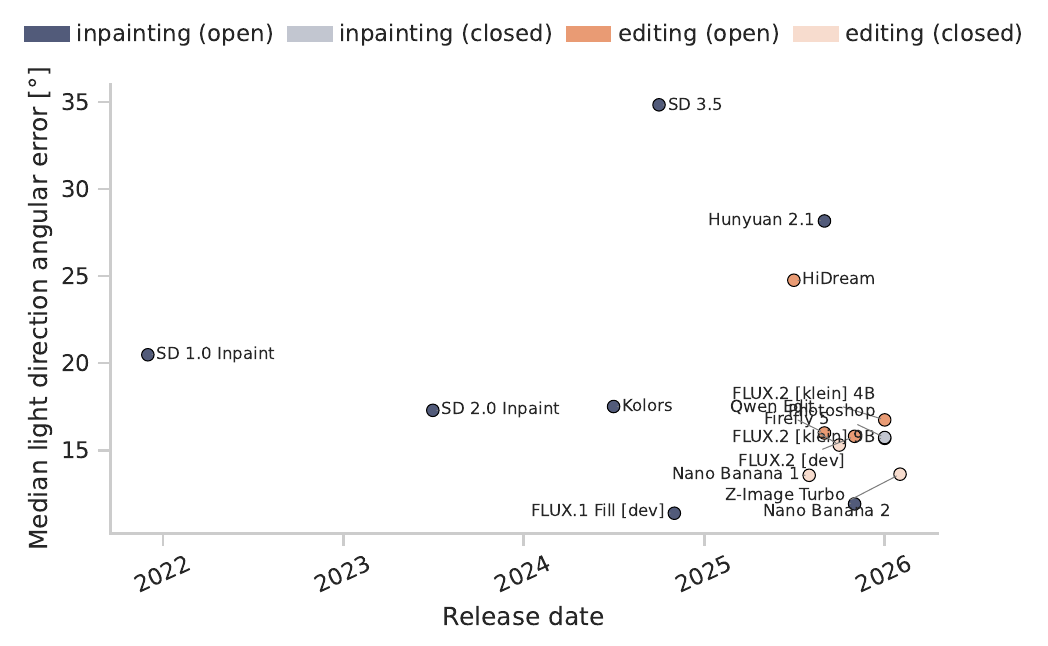}
  \caption{Median angular error of the light direction as a function of the release date of each model.  The markers are coloured by type (inpainting or editing) and availability (open or closed).}
  \Description{Median angular error of the light direction as a function of the release date of each model.}
  \label{fig:median_angular_error_vs_release_date}
\end{figure}

An alternative visualisation of \cref{fig:angular_error_direction_lumiere} to highlight the accuracy of the lighting direction generated by the models through the years is shown in \cref{fig:median_angular_error_vs_release_date}.  The median angular error in light direction for all (quality-filtered) probes generated by the model is compared with the model release date.  The lack of a discernible trend shows that improvements in inpainting and editing generative models over time (and increased parameter counts, c.f. \cref{fig:median_angular_error_vs_parameters}) do not predict light direction accuracy.

\subsection{Spatially varying lighting analysis}
\label{suppsubsubsec:evaluation_spatially_varying}

\begin{figure*}[t]
  \centering
  \includegraphics[width=\linewidth]{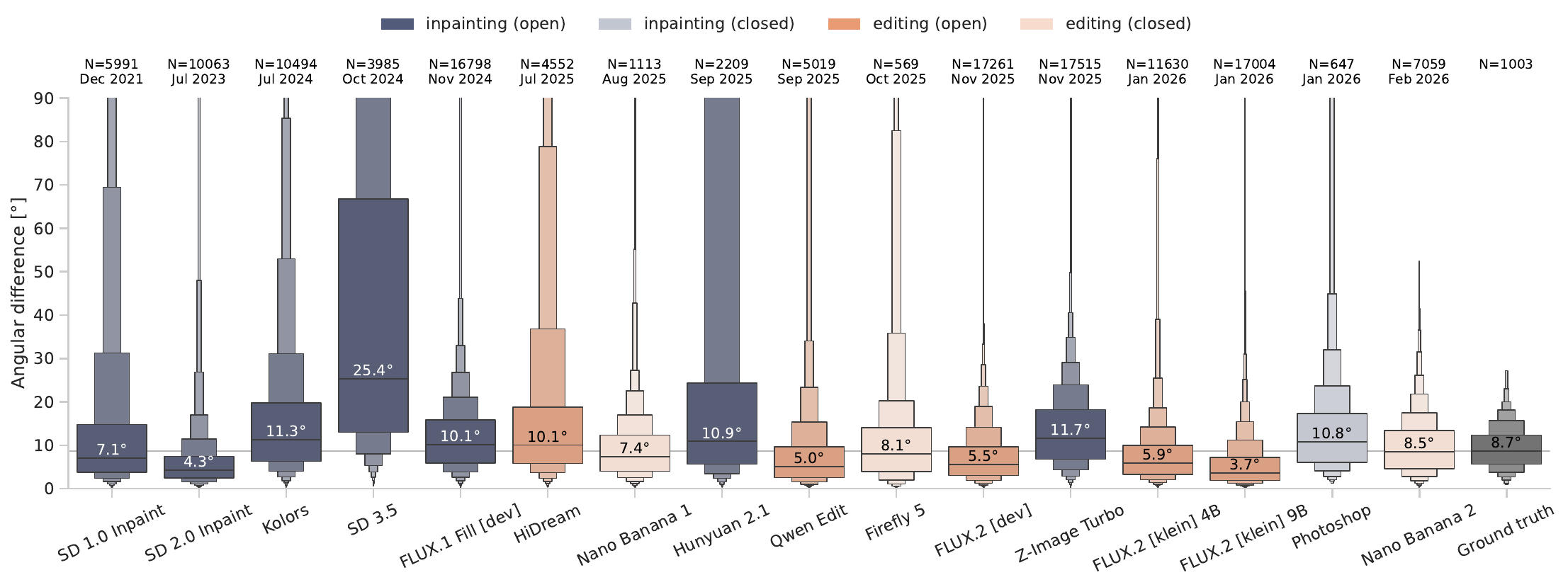}
  \caption{Distributions of the angular difference of the light directions measured in both probes (different locations) for each model and the ground-truth probes (last column). Models are sorted by release date (above) and coloured by type (inpainting or editing) and availability (open or closed).  The median is shown on middle line, and each step corresponds to distribution quantiles.  The horizontal line corresponds to the ground-truth median angular difference.  The number of probes ($N$) used to compute each distribution is displayed at the top of the distribution.  The tails containing the outliers ($\pm$\ang{90}) are cropped out for legibility.}
  \Description{Difference between the light directions for probes in different locations.}
  \label{fig:spatial_variation_dual_probe_inp_with_gt}
\end{figure*}

\Cref{fig:spatial_variation_dual_probe_inp_with_gt} shows the distribution of the angle difference between two probes in different locations within the same image.  Some models tend to predict less variation in their spatial light directions within a single image (i.e., they make the light direction similar in both probes), such as \SDtwo and \fluxtwonine.  In contrast, some models vary the light direction more than expected, such as \SDthree.

Note that this figure only shows the angle difference and does not take into account the direction of the dominant light vector, so the light direction could be wrong on the generated probes, but still have the same difference between both probes as the ground-truth probes.  Also note that this measure does not compare the angular difference between a specific scene and its ground truth (unlike \cref{subsubsec:evaluation_LD_spatial_variation}) and only shows statistical trends, not direct accuracy for a given scene.  However, these trends are similar to the ones observed in the comparison per scene, as in \cref{fig:direction_lumiere_spatial_variation}.

\subsection{Probe size analysis}
\label{suppsubsubsec:evaluation_probe_size}

\begin{figure*}[h]
  \centering
  \includegraphics[width=\linewidth]{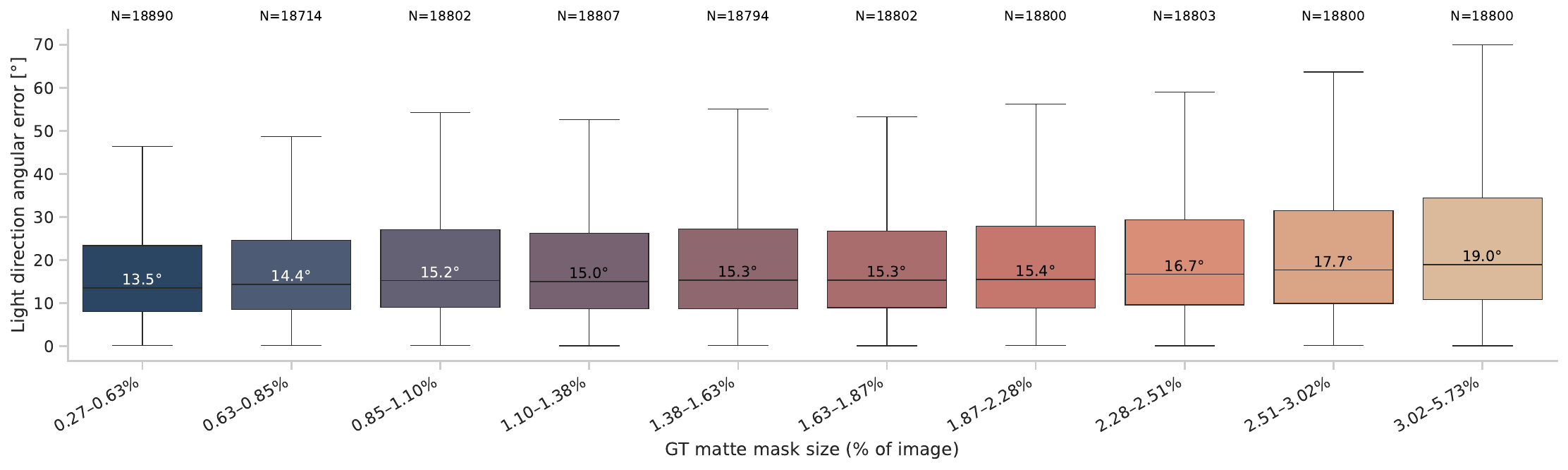}
  \caption{Distributions of the light direction angular error as a function of the percentage of the image area the ground-truth matte probe occupies, for all models.  The mask size ranges have been selected to be balanced, the number of probes ($N$) in each is displayed at the top of each distribution.}
  \Description{Distributions of the light direction angular error as a function of the percentage of the image area the ground-truth probes.}
  \label{fig:mask_size_vs_angular_error}
\end{figure*}

In the Multi-Illumination dataset, the distances of the ground-truth probes from the camera vary across scenes. To analyse the impact of the relative size of the probe in the image to the surrounding objects and the potential artefacts caused by the lower resolution of smaller inpainted probes, we compare the distributions of light direction angular error of all models for each scene in \cref{fig:mask_size_vs_angular_error}.  The relative size of the ground-truth matte probe within the image is computed and clustered into balanced bins for each scene. This shows that smaller probes with fewer pixels do not hinder light-feature estimation, whereas larger probes closer to the camera do.

\subsection{Angular error distribution}

\begin{figure}[t]
  \centering
  \includegraphics[width=\linewidth]{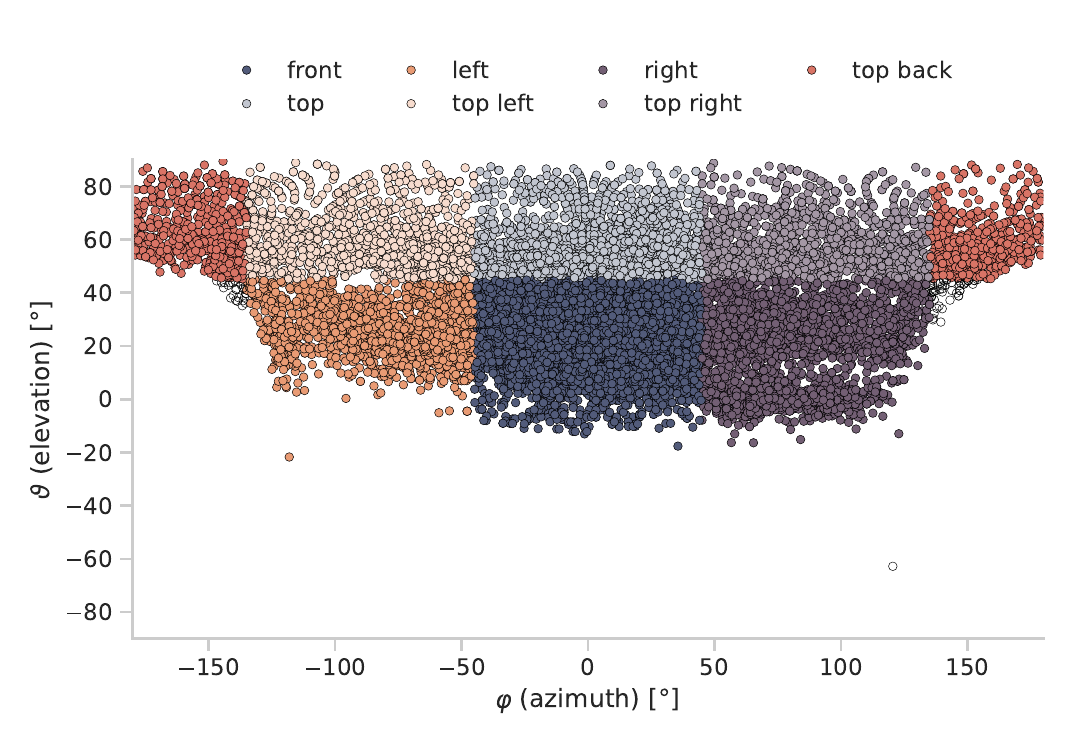}
  \caption{Spatial distribution of the ground-truth light direction clusters, divided equally in space, in intervals of \ang{90} in azimuth and \ang{45} in elevation. Data in white is not considered because there are too few to be a statistically representative group.}
  \Description{Spatial distribution of the ground truth light direction clusters.}
  \label{fig:gt_clusters_spatial_dist}
\end{figure}

\begin{figure*}[t!]
  \centering
  \includegraphics[width=\linewidth]{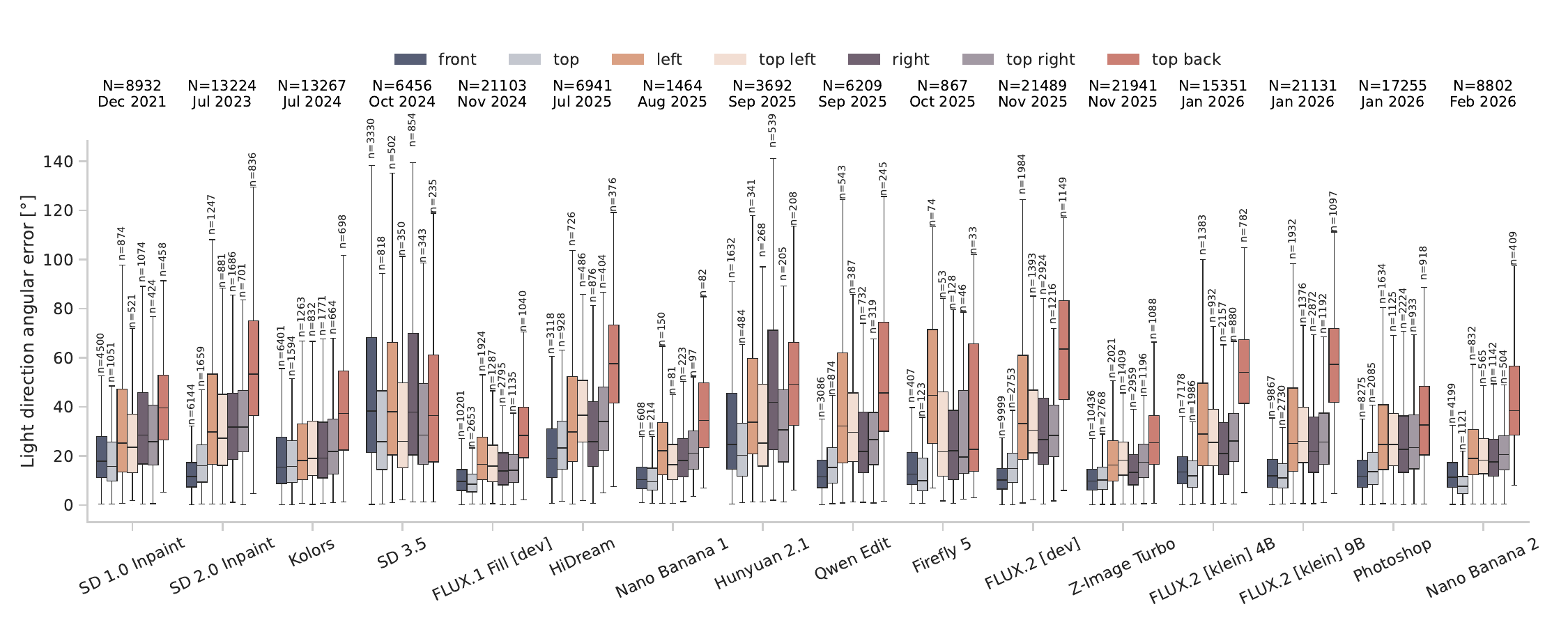}
  \caption{Angular error distribution for each model, per coarse incoming ground-truth light direction, based on \cref{fig:gt_clusters_spatial_dist}. The middle line in the box corresponds to the median, and the edges of the box to the quartiles.  The whiskers represent the confidence intervals.  The number of inpainted probes ($n$) for each model in each ground-truth light direction cluster is displayed above the whisker.  The total number of inpainted probes ($N$) and the year of model introduction are also shown above.}
  \Description{Angular error distribution for each model.}
  \label{fig:boxplot_gt_clusters_per_model}
\end{figure*}

\Cref{fig:boxplot_gt_clusters_per_model} shows the angular error distribution for each ground-truth light direction cluster, as defined in \cref{fig:gt_clusters_spatial_dist}, for each model. This represents a more detailed version of \cref{fig:boxplot_gt_clusters_per_direction}, where each model is kept separate.  The same global trends can be observed as the model aggregated version  (\cref{fig:boxplot_gt_clusters_per_direction}), where the ``top back'' direction is considerably harder for all the models.  The ``front'' and ``top'' clusters have similar performances for most models and no significant bias for the ``left'' or ``right'' side.

\subsection{Colour metrics}
\label{suppsubsec:evaluation_colour}

\begin{figure*}[h]
  \centering
  \includegraphics[width=\linewidth]{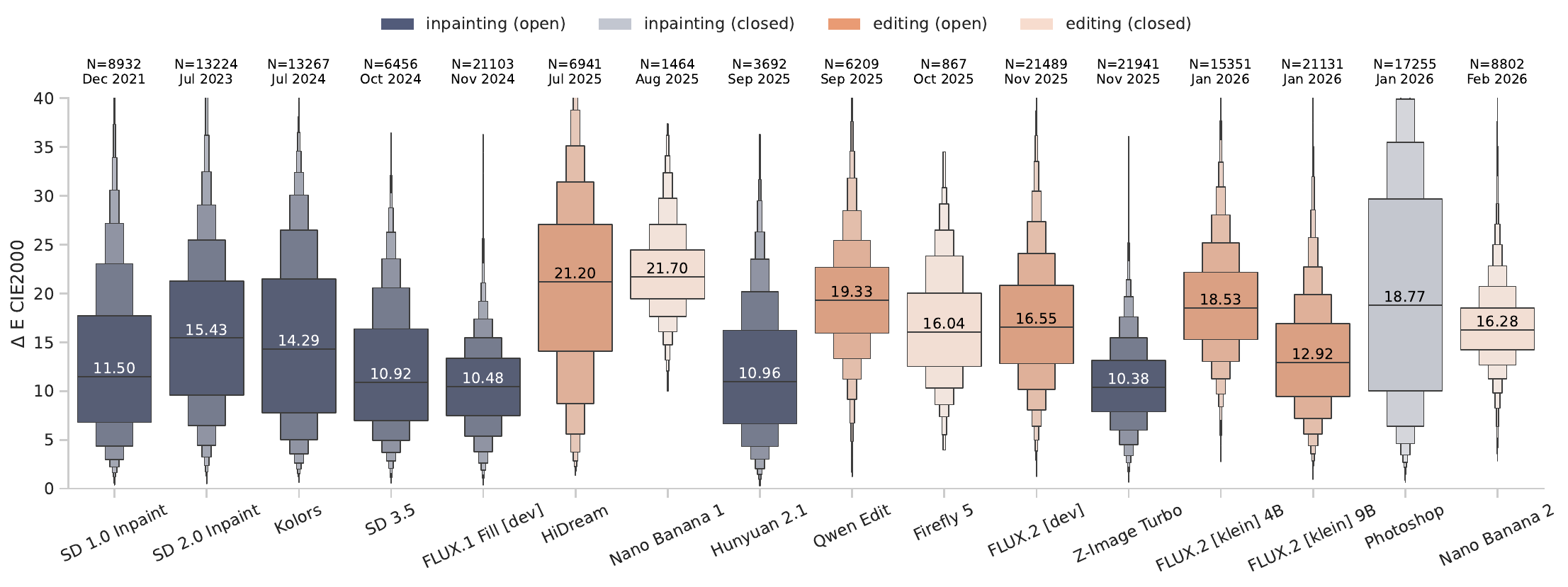}
  \caption{Distributions of the $\Delta\textrm{E CIE} 2000$ for each model, sorted by release date (displayed at the top of the distribution).  The range has been limited to $[0, 40]$ to better see the distribution.  The middle line and value displayed on the distribution corresponds to the median value.  Each step corresponds to the quantiles of the distribution.  The distributions are coloured by type (inpainting or editing) and availability (open or closed).  The number of probes ($N$) used to compute each distribution is displayed at the top of the distribution.  The purple line at $\Delta\textrm{E}=1$ corresponds to the Just Noticeable Difference (JND) of human perception, colour differences start to be noticeable between $\Delta\textrm{E} \in [3, 6]$ (pink regime) and appear as different colours for $\Delta\textrm{E}\geq6$ (light pink regime) \citep{fairchild_color_2013}.}
  \Description{Distributions of the $\Delta\textrm{E CIE} 2000$ for each model.}
  \label{fig:delta_E2000}
\end{figure*}

\begin{figure*}[h]
  \centering
  \includegraphics[width=\linewidth]{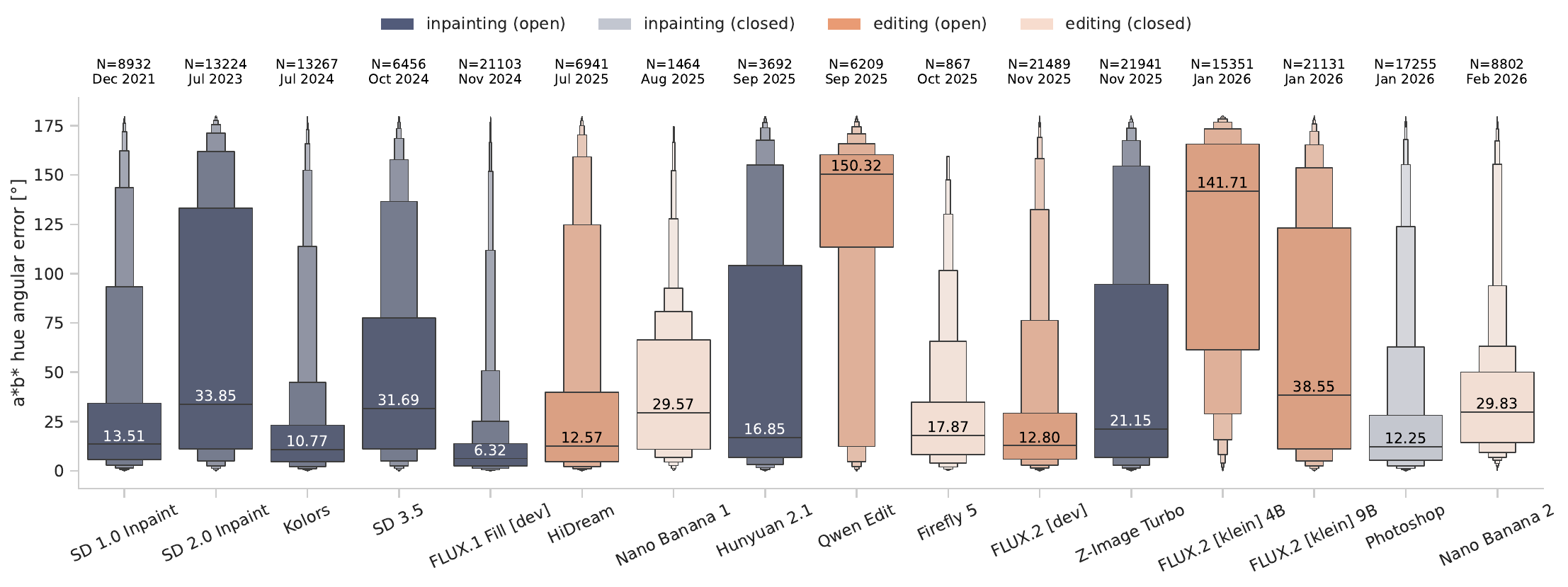}
  \caption{Distributions of the hue angular error for each model, sorted by release date (displayed at the top of the distribution).  The maximum possible error is \ang{180} given the circularity of hue values. The middle line and value displayed on the distribution corresponds to the median value.  Each step corresponds to the quantiles of the distribution.  The distributions are coloured by type (inpainting or editing) and availability (open or closed). The number of probes ($N$) used to compute each distribution is displayed at the top of the distribution.}
  \Description{Distributions of the hue angular error for each model.}
  \label{fig:hue_error_ab_boxenplot}
\end{figure*}

We further evaluate two other colour measures: $\Delta\textrm{E CIE} 2000$ (\cref{fig:delta_E2000}) and hue angular error (\cref{fig:hue_error_ab_boxenplot}). These measures are not included in the main sections because they are more prone to capturing co-occurring factors beyond colour alone.

$\Delta\textrm{E CIE} 2000$ (\cref{fig:delta_E2000}) considers errors not only in the chrominance $(a,b)$ channels of the Lab space, but also in the luminance $(L)$ channel. However, since our prompt only specifies a \textit{grey} sphere, errors in the $L$ channel may also arise from differences in albedo. Having said this, we can see that \fluxone and \SDone remain among the best methods, as was also the case for $\Delta E_{a,b}$, whilst \hidream and Nano Banana 1 become much more strongly penalised when luminance is included in the error measure.

Hue angular error (\cref{fig:hue_error_ab_boxenplot}) is commonly used in setups such as ours because it measures the chromatic direction of a colour independently of both luminance and saturation. However, hue estimation for colours close to the achromatic axis is highly unstable. Therefore, given the original grey colour of the sphere, scenes in which the illumination colour is only weakly pronounced can yield large hue errors despite negligible perceptual differences. With this caveat in mind, we nevertheless observe that \fluxone achieves the best performance, further highlighting the robustness of this model across multiple colour measures.

\begin{figure}[h]
  \centering
  \includegraphics[width=\linewidth]{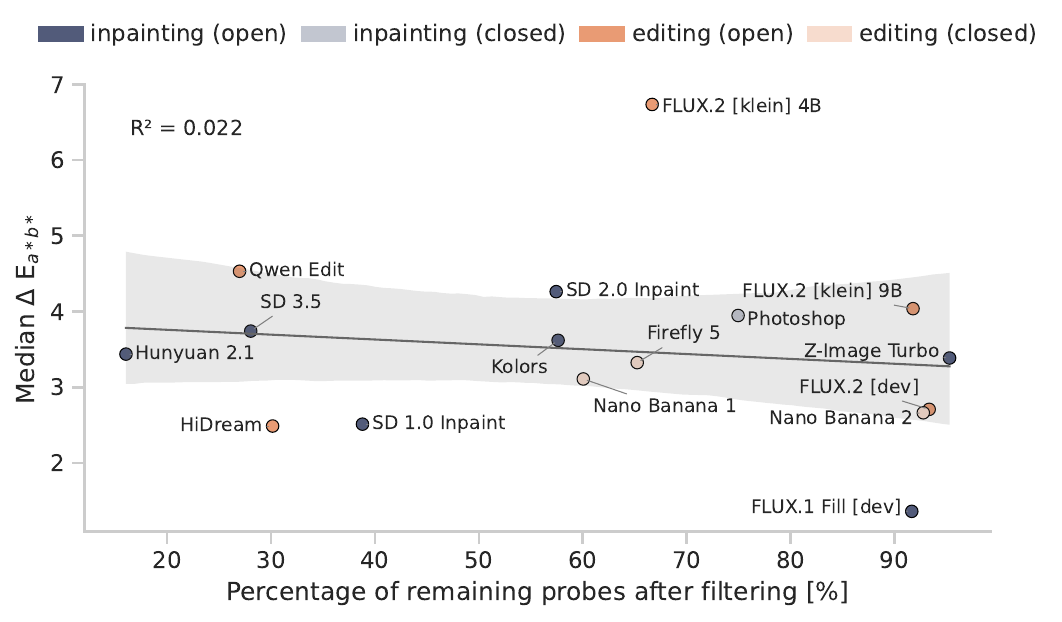}
    \caption{Median $\Delta\textrm{E}_\textrm{ab}$ as a function of the percentage of quality filtered probes. A linear regression (black line) is fitted on the data, with the $R^2$ displayed at the top, with the fit uncertainty around the line. Models are coloured by type (inpainting or editing) and availability (open or closed), as in \cref{fig:angular_error_direction_lumiere}.}
  \Description{Survivorship bias of the median $\Delta\textrm{E}_\textrm{ab}$.}
  \label{fig:discussion_survivorship_bias_colour}
\end{figure}

\Cref{fig:discussion_survivorship_bias_colour} is a complementary figure to \cref{fig:discussion_survivorship_bias} in the survivorship bias discussion (c.f. \cref{sec:discussion}), displaying the relationship between the median $\Delta\textrm{E}_\textrm{ab}$ and the number of probes remaining after the quality filtering.  The absence of trend (shown by the $R^2$ value) indicates that models with fewer valid probes do not have better colour accuracy than models with more valid probes.

\clearpage

\end{document}